\documentclass{article}
\usepackage[a4paper, margin=2.5cm]{geometry}
\usepackage{graphicx} 
\usepackage{amsmath}
\usepackage{amsthm}
\usepackage{amssymb}
\usepackage{algorithm}
\usepackage{algpseudocode}
\usepackage[backend=biber]{biblatex} 
\bibliography{references} 
\usepackage{hyperref}
\usepackage{comment}
\usepackage{dsfont}
\usepackage{graphicx}
\usepackage{subcaption}
\usepackage{rotating} 
\usepackage{multirow}
\usepackage[export]{adjustbox}
\usepackage{array}
\usepackage{float}
\usepackage{xurl}

\usepackage{booktabs}
\usepackage{makecell}
\usepackage{hyperref}
\usepackage[table]{xcolor}
\usepackage{booktabs}
\usepackage{multirow}

\usepackage{tcolorbox}

\newcommand{\heat}[2]{\cellcolor{red!#1!green!25!white}#2}

\newcommand{\graycell}[1]{\cellcolor{gray!15}#1}

\title{Stochastic Gradient Optimization with Model-Assisted Sampling}
\author{Jonne Pohjankukka and Jukka Heikkonen}
\date{University of Turku, Department of Computing}

\usepackage{xcolor}
\newenvironment{keywords}{\noindent\textbf{Keywords:} }{}
\newtheorem{assumption}{Assumption}
\newtheorem{definition}{Definition}
\DeclareMathOperator*{\argmin}{arg\,min}
\newtheorem{theorem}{Theorem}
\newtheorem*{theorem*}{Theorem}
\newcommand{\Qdiff}{Q_{\text{dif}}}

\newcommand{\datapdist}{\mathcal{P}}
\newcommand{\hypothesisSet}{\mathcal{H}}
\newcommand{\sample}{\mathcal{S}} 
\newcommand{\Es}[1]{\mathbb{E}_{\mathcal{D}}\left[#1\right]}
\newcommand{\bq}{\textbf{q}}

\newcommand{\subsampleIndexSymbol}{\mathcal{I}}
\newcommand{\subsample}{\sample_\subsampleIndexSymbol}
\newcommand{\lossSymbol}{\ell}

\newcommand{\boldmodelparams}{\boldsymbol{\theta}}
\newcommand{\indexsetall}{\mathcal{I}}

\newcommand{\datapoint}{\textbf{d}}

\newcommand{\sampleSize}{N}
\newcommand{\subsampleSize}{n}

\newcommand{\xdata}{\textbf{x}}
\newcommand{\ydata}{\textbf{y}}
\newcommand{\trueRiskSymbol}{R}
\newcommand{\empiricalRiskSymbol}{\hat{\trueRiskSymbol}}

\newcommand{\residual}{\textbf{e}}
\newcommand{\gradientPopulation}{\mathcal{G}}
\newcommand{\gradientSymbol}{\textbf{g}}
\newcommand{\gradientDataPointSymbol}{g}
\newcommand{\gradientModelSymbol}{q}
\newcommand{\populationSize}{N}
\newcommand{\subsampleIndexSet}{\mathcal{I}_J}
\newcommand{\firstSubsampleIndexSet}{\subsampleIndexSymbol^{1}}
\newcommand{\secondSubsampleIndexSet}{\subsampleIndexSymbol^{2}}
\newcommand{\firstSubsample}{\sample_{\firstSubsampleIndexSet}}
\newcommand{\secondSubsample}{\sample_{\secondSubsampleIndexSet}}
\newcommand{\firstGradsubsample}{\gradientPopulation_{\firstSubsampleIndexSet}}
\newcommand{\secondGradsubsample}{\gradientPopulation_{\secondSubsampleIndexSet}}
\newcommand{\firstSubsampleIndexSetSize}{m_1}
\newcommand{\sampleDistribution}{\mathcal{D}}

\begin{document}

\maketitle

\begin{abstract}
This work addresses the problem of variance in stochastic gradient estimation for machine learning optimization. Deep learning relies on mini-batch methods such as stochastic gradient descent, which approximate full gradients but introduce noise, creating trade-offs between convergence stability, speed, and generalization. Existing methods, including variance reduction techniques (e.g., SVRG and SAG) and adaptive optimizers, aim to mitigate gradient noise but may introduce additional computational overhead. We propose a model-assisted sampling framework that interprets mini-batch gradients through survey sampling theory, treating the dataset as a fixed finite population and gradients as sample-based estimates. Our aim is to bridge machine learning optimization and survey sampling theory by combining their perspectives on sample-based estimation and variance reduction. By incorporating auxiliary gradient-prediction models, we construct more efficient gradient estimators, with uniform sampling arising as a special case when no auxiliary information is used. Our approach integrates easily with existing optimizers, improving efficiency without altering their dynamics. Empirical results on synthetic and six benchmark datasets show performance gains in 71-86\% of the experiments, particularly for medium-sized input spaces in our benchmarks. Notably, with momentum-based optimizers such as AdamW, the proposed estimator achieves clearly better generalization in roughly half the training epochs compared to baseline estimator.
\end{abstract}


\section{Introduction}
In practical deep learning (DL) scenarios, one typically operates with fixed datasets, assumed to be realizations from some unknown data-generating distribution to which we do not have direct access. Given the substantial computational burden associated with training modern models, often comprising millions or billions of parameters, it is standard practice to use only subsets of the data, i.e., mini-batches, during training. This subsampling is not merely a convenience but a necessity: full-batch optimization would be computationally prohibitive and may even degrade generalization performance \cite{KeskarMNST16}. From a theoretical perspective, the dataset can be viewed as a finite population from which samples are repeatedly drawn. As a result, gradient estimates computed on mini-batches are inherently stochastic and subject to sampling variance, since they approximate the full population gradient using only partial information. This stochasticity can lead to oscillatory or unstable optimization trajectories and may slow convergence, particularly in smooth or ill-conditioned problems. On the other hand, in highly non-convex landscapes, gradient noise can provide beneficial exploratory behavior, helping the optimization process escape saddle points and sharp local minima. Consequently, there exists a fundamental trade-off between gradient variance, convergence stability, and generalization performance \cite{Smith2020,KeskarMNST16,zhu2019the}.

Within this stochastic setting, iterative optimization algorithms form the backbone of model training in machine learning (ML), particularly in DL. Due to computational constraints, gradient-based methods dominate, with much contemporary research focusing on extensions of classical gradient descent (GD), such as stochastic gradient descent (SGD), Adam \cite{Kingma2015Adam,AdamWCite}, and Nesterov’s accelerated momentum \cite{Nes83}, among others \cite{ruder2016overview}. These methods primarily aim to improve learning efficiency by adapting step sizes, incorporating momentum, or leveraging information from past updates. Despite their differences, they all rely on stochastic gradient estimates derived from mini-batches, and thus inherently contend with the variance introduced by subsampling. As a result, much of their effectiveness hinges on how well they balance fast progress toward minima with robustness to gradient noise, effectively determining the direction and magnitude of each update step to facilitate stable and efficient convergence \cite{Goodfellow-et-al-2016}.

Building on this, variance reduction of stochastic gradients has emerged in modern machine learning research. Standard stochastic optimizers exhibit a persistent variance floor due to noisy gradient estimates, which can slow convergence and limit solution accuracy, especially near optima. To address this, a rich body of work has developed methods that explicitly reduce gradient variance while preserving computational efficiency. Techniques such as stochastic variance reduced gradient (SVRG) and stochastic average gradient (SAG) employ control variates constructed from past gradients or periodically computed full gradients, yielding unbiased estimators with significantly lower variance and provably faster convergence in certain settings \cite{SVRGcite,SAGcite}. Extensions like SAGA further refine these ideas through memory-based corrections \cite{SAGAcite}, while mini-batch strategies trade additional computation for variance reduction via averaging. In the context of DL, adaptive methods such as Adam can also be interpreted as implicitly mitigating variance through moment estimation \cite{2015-kingma}. More recent research explores hybrid and data-dependent approaches, emphasizing that controlling both the magnitude and direction of gradient noise is crucial for achieving efficient, stable, and generalizable training \cite{SAGAcite,Bottou2010}.

While variance reduction has been extensively studied within ML, the broader problem of reducing noise in sample-based estimators has a long history in classical statistics, particularly in the field of survey sampling. The general problem of reducing the variance of noisy sample-based population estimates has been extensively studied in the classical context of statistical survey sampling, which focuses on efficiently estimating population parameters from limited samples when access to the full population is infeasible \cite{model2003sarndal}. The field encompasses a wide range of sampling-based methodologies, including design-based and model-based approaches, as well as hybrid model-assisted sampling methods that combine elements of both paradigms. In design-based sampling, inference is based on the randomization induced by the sampling design, and population parameters are treated as fixed but unknown quantities. These methods are generally robust and provide unbiased estimators under the sampling design, but may be statistically inefficient when strong auxiliary information is available but not fully utilized. In contrast, model-based approaches assume an explicit statistical model for the population generation process, allowing potentially more efficient estimation and prediction by leveraging structural assumptions about the data \cite{Dumelle2022}. However, their performance depends heavily on the validity of the assumed model and may become biased under model misspecification. Model-assisted approaches seek to combine the robustness of design-based inference with the efficiency gains of model-based methods. Common techniques in statistical survey sampling include simple random sampling, stratified sampling, systematic sampling, two-phase sampling or generalized regression estimator \cite{model2003sarndal,Cochran1977Sampling}.
    
In this work, we propose a novel model-assisted sampling framework to mitigate the variance-induced noise floor of stochastic gradient estimators. We show that the commonly used uniform sampling strategy employed in mini-batch SGD can be interpreted as a special case of our framework, corresponding to the use of a trivial prediction model for the unsampled loss gradients. Our approach formulates the empirical risk gradient field as an unknown and inaccessible population quantity that must be estimated from a stochastic sample drawn from a fixed population. This perspective enables the application of classical survey sampling methodology to stochastic optimization. Rather than proposing a new optimizer, our method improves existing optimization algorithms by replacing the standard mini-batch gradient estimator with a more statistically efficient estimator based on model-assisted sampling. To the best of our knowledge, the use of model-assisted survey sampling principles for mini-batch gradient estimation has received little attention in the ML optimization literature.

The remainder of this paper is organized as follows. In Section \ref{Section:ERM}, we briefly review the foundations of empirical risk minimization in ML, SGD, and classical survey sampling estimators, including the Horwitz–Thompson and difference estimators. We also link the gradient noise floor to the asymptotic convergence region of stochastic optimization, showing that lower-variance gradient estimators can reduce the residual error term and thereby improve convergence stability under standard smoothness and strong-convexity-type assumptions. In Section \ref{Section::MA-gradient-estimator}, we introduce the proposed model-assisted unbiased gradient estimator. Next in Section \ref{Section::empricial_analysis}, we empirically validate our estimator against the baseline uniform-sampling gradient estimator on benchmark datasets. Finally, in Sections \ref{Section::discussion} and \ref{Section::Conclusion} we cover the discussion and conclusions.



\section{Empirical risk minimization}\label{Section:ERM}

The general goal in ML, given a hypothesis set $\hypothesisSet$, a joint probability distribution $\datapdist(\xdata, \ydata)$, and a \emph{loss function} $\lossSymbol(h(\xdata), \ydata)$, is to find a hypothesis \(h\in \hypothesisSet\) such that the \emph{risk function}:
\begin{equation}
\label{eq:true_risk_definition}
    \trueRiskSymbol(h) = \mathbb{E}_{\datapdist}\left[\lossSymbol(h(\textbf{x}), \textbf{y})\right]=\int \lossSymbol(h(\textbf{x}), \textbf{y}) \; d\datapdist(\textbf{x}, \textbf{y}),
\end{equation}
is minimized, where \(\xdata, \ydata\) are the corresponding explanatory and response variables. The tuple \(\datapoint = (\xdata, \ydata)\) is generally called a \emph{data point}. Our task is thus to find a hypothesis $h^*$ which is defined as: 
\begin{equation}\label{Equation::optimal_hypothesis_definition}
    h^* := \argmin_{h\in\mathcal{H}} \trueRiskSymbol(h).
\end{equation}

Since \(\datapdist\) is practically always unknown to us, we need to rely on an estimator of $\trueRiskSymbol$, i.e., \(\empiricalRiskSymbol\) which relies on a sample \emph{population} $\sample=\{(\xdata_i, \ydata_i)\}$ for $i = 1, 2, ..., \sampleSize$, commonly defined as:
\begin{equation}\label{Equation::empirical_risk_definition}
\empiricalRiskSymbol_{\sample}(h) = \frac{1}{\sampleSize}\sum_{i=1}^\sampleSize \lossSymbol(h(\xdata_i), \ydata_i).   
\end{equation}
Due to the law of large numbers \(\empiricalRiskSymbol_{\sample}\) converges to the true risk $\trueRiskSymbol$ as \(\sampleSize\to\infty\). Thus, in practice, our target of optimization is \(\empiricalRiskSymbol_{\sample}\) rather than \(\trueRiskSymbol\), so the resulting optimization problem is to find a hypothesis $h^*$ such that: 
\begin{equation}\label{Equation::optimal_empirical_hypothesis_definition}
    h^* = \argmin_{h\in\hypothesisSet} \empiricalRiskSymbol(h).
\end{equation}

\subsection{Gradient descent optimization}\label{Section::Gradient_descent_optimization}

In many ML cases, analytical solutions of Eq. \ref{Equation::optimal_empirical_hypothesis_definition} are not possible, so we utilize iterative methods, most commonly using some variation of the \emph{gradient descent} (GD) algorithm \cite{Bottou1999,Boyd2004}. Our hypothesis \(h\) is typically parameterized by a vector \(\boldmodelparams\), i.e., \(h = h(\boldmodelparams)\). Thus, when searching for the optimal parameters \(\boldmodelparams^*\), the estimated empirical risk can be written directly as a function of \(\boldmodelparams\), i.e., \(
\empiricalRiskSymbol_{\sample}(h) = \empiricalRiskSymbol_{\sample}(\boldmodelparams)
\). In GD, we optimize the parameters with the general update rule as follows:

\begin{equation}\label{Equation::gradient_descent_definition}
\boldmodelparams_{t+1} = \boldmodelparams_t - \eta \nabla_{\boldmodelparams} \empiricalRiskSymbol_{\sample}(\boldmodelparams_t)
\end{equation}

where \( \boldmodelparams_t \) represents the parameter vector at iteration \( t \), \( \eta>0 \) is the learning rate, \( \nabla_{\boldmodelparams} \empiricalRiskSymbol_{\sample}(\boldmodelparams_t) \) is the gradient of the empirical risk with respect to \( \boldmodelparams \) at step \( t \). The goal is to iteratively update \( \boldmodelparams \) such that the value of \( \empiricalRiskSymbol_{\sample}(\boldmodelparams) \) decreases, ideally converging to a global minimum. The gradient of the empirical risk is clearly defined as:
\begin{equation}\label{Equation::empirical_gradient_definition}
\nabla_{\boldmodelparams} \empiricalRiskSymbol_{\sample}(\boldmodelparams) = \frac1\sampleSize\sum_{i=1}^\sampleSize \nabla_{\boldmodelparams} \lossSymbol(h(\xdata_i), \ydata_i) = \frac1\sampleSize\sum_{i=1}^\sampleSize \gradientSymbol_i,    
\end{equation}
where we have denoted \(\gradientSymbol_i =\nabla_{\boldmodelparams} \lossSymbol(h(\xdata_i), \ydata_i)\). In practice, especially in the deep learning setting where \(\boldmodelparams\) may consist of a very large number of parameters, computing the gradient of the empirical risk using the full sample population \(\sample\) is typically infeasible due to e.g. memory constraints. This is because large amounts of intermediate activations and gradients must be stored in memory in order to carry out the parameter update in Eq.~\ref{Equation::gradient_descent_definition}. For this reason, an unbiased \emph{mini-batch} solution is typically used, the \emph{stochastic gradient descent} (SGD) \cite{Bottou2010}, in which a subsample of the data set is taken $\subsample \subset \sample$, and the corresponding gradient is calculated using this subsample:
\begin{equation} \label{Equation::subsample_gradient_definition}
\nabla_{\boldmodelparams} \empiricalRiskSymbol_{\subsample}(\boldmodelparams) = \frac{1}{\subsampleSize}\sum_{i\in \subsampleIndexSymbol}  \gradientSymbol_i,     
\end{equation}
where $ \subsampleIndexSymbol \subset \left[\populationSize\right] =\{1, 2, ..., \sampleSize\}$ is a random subset of data indexes from the full sample set $\sample$ (called population) so that $|\subsampleIndexSymbol|=\subsampleSize$ for fixed \emph{batch size} $\subsampleSize$. Thus, the actual update steps become: 

\begin{equation}
\label{Equation::gradient_descent_empirical_step}
\boldmodelparams_{t+1} = \boldmodelparams_t - \eta \nabla_{\boldmodelparams} \empiricalRiskSymbol_{\subsample}(\boldmodelparams_t).
\end{equation}

In line with recent work on gradient estimation and sampling strategies \cite{Needell2014SGD,Zhao2015StochasticImportance, Alain2015VarianceReduction}, our focus is on improving the gradient itself, which determines the direction of movement in the parameter space. 


\subsection{Horwitz-Thompson and difference estimators} \label{Section::Horwitz-Thompson}

In the field of survey sampling \cite{Cochran1977Sampling,Lohr2009Sampling}, a common method for making an unbiased estimator, with data points having inclusion probabilities $\pi_i > 0$, i.e., probability of being included in a sample, is by using the \emph{Horwitz-Thompson} (also called the $\pi$-estimator) style estimators \cite{model2003sarndal}:
\begin{equation}
Q_{HT} = \frac{1}{\sampleSize}\sum_{i=1}^\sampleSize I_i \frac{\textbf{q}_i}{\pi_i} = \frac{1}{\sampleSize}\sum_{i\in \subsampleIndexSymbol} \frac{\textbf{q}_i}{\pi_i},     
\end{equation}
where $I_i$ is a random variable taking either value 0 or 1 if sample point is included in sample or not, \(\subsampleIndexSymbol\subset[\sampleSize]\), and $\textbf{q}_i$ is some fixed non-random data point from the full sample \(\sample \), also called the \emph{population}. The $\pi$-estimator is an unbiased estimator when averaged over the distribution \(\sampleDistribution\), i.e. \emph{sampling design}, of possible sample selections:
\begin{equation}
\mathbb{E}_{\sampleDistribution}\left[Q_{HT}\right] = \frac{1}{\sampleSize}\sum_{i=1}^\sampleSize \mathbb{E}_{\sampleDistribution}\left[I_i\right] \frac{\textbf{q}_i}{\pi_i}=\frac{1}{\sampleSize}\sum_{i=1}^\sampleSize \pi_i \frac{\textbf{q}_i}{\pi_i}=\frac{1}{\sampleSize}\sum_{i=1}^\sampleSize \textbf{q}_i,     
\end{equation}
and it has the covariance matrix \cite{model2003sarndal}: 
\begin{equation}\label{Equation::HT-variance_definition}
\mathbb{V}(Q_{HT}) = \sum_{i=1}^\sampleSize \sum_{j=1}^\sampleSize \frac{\pi_{ij} - \pi_i \pi_j}{\pi_i \pi_j}  \textbf{q}_i\textbf{q}_{j}^{\top},
\end{equation}
where $\pi_{ij}=\mathbb{E}_ {\mathcal{D}}\left[I_i I_j\right]$, i.e., the probability that both points $i$ and $j$ are included in the sample $\subsample$. 

There exists a potentially more efficient version of the $\pi$-estimator called \emph{difference estimator} (\(\pi_d\)-estimator), which is a \emph{model-assisted} estimator designed to incorporate model-learned estimation capability into correcting the error in the $\pi$-estimator. The difference estimator is defined as:
\begin{equation}\label{Equation::difference_estimator_definition}
\Qdiff = \frac{1}{\sampleSize}\left(\sum_{i=1}^\sampleSize \hat{\textbf{q}}_i + \sum_{i=1}^\sampleSize I_i\frac{\textbf{q}_i-\hat{\textbf{q}}_i}{\pi_i}\right)= \frac{1}{\sampleSize}\left(\sum_{i=1}^\sampleSize \hat{\textbf{q}}_i + \sum_{i\in \subsampleIndexSymbol}\frac{\textbf{q}_i-\hat{\textbf{q}}_i}{\pi_i}\right),     
\end{equation}
where \(\hat{\textbf{q}}_i\) is a model-based estimate of \(\textbf{q}_i\). For example, if \(\xdata_i\) is a vector of explanatory variables with corresponding response vector \(\textbf{q}_i\,\forall i\), then \(\hat{\textbf{q}}_i = \gradientModelSymbol(\xdata_i; \subsample)\) for some function \(\gradientModelSymbol\). The covariance matrix of the \(\pi_d\)-estimator is:

\begin{equation}\label{Equation::difference_estimator_variance}
\begin{aligned}
\mathbb{V}(\Qdiff)
&= \sum_{i,j}\frac{\pi_{ij}-\pi_i\pi_j}{\pi_i\pi_j}\,(\bq_i-\hat{\bq}_i)(\bq_j-\hat{\bq}_j)^{\!\top} \\
&= \sum_{i,j}p_{ij}\,\residual_i\residual_j^{\!\top},
\end{aligned}
\end{equation}
where we have denoted \(p_{ij}:= \frac{\pi_{ij}-\pi_i\pi_j}{\pi_i\pi_j}\) and \(\residual_i := \bq_i-\hat{\bq}_i \;\forall\,i,j\). Notice that in a simple random sampling without replacement (SRSWOR) case, i.e., \(\pi_i = \subsampleSize/\sampleSize\) where \(|\subsampleIndexSymbol|=\subsampleSize\) and \(\pi_{ij}=\subsampleSize(\subsampleSize-1)/\sampleSize(\sampleSize-1)\) with gradient model \(\gradientModelSymbol:=\textbf{0}\), the \(\pi_d\)-estimator reduces to the sample average, i.e., corresponding to the uniform mini-batch gradient in SGD. 

The trace of the variance is:
\begin{equation}\label{Equation::difference_estimator_variance_trace}
\begin{aligned}
\mathrm{Tr}\!\left\{\mathbb{V}(\Qdiff)\right\}
&= \sum_{i,j}p_{ij}\,
   \mathrm{Tr}\!\left\{\residual_i\residual_j^{\!\top}\right\} = \sum_{i,j}p_{ij}\,
   \left\langle \residual_i,\; \residual_j \right\rangle.
\end{aligned}
\end{equation}

It is clear that $\mathrm{Tr}\left\{{\mathbb{V}(\Qdiff)}\right\} \to 0$ as $\|\residual_i\| \to 0\; \forall i$. In other words, the better the model \(\gradientModelSymbol\), the more efficient the \(\pi_d\)-estimator is. This estimator induces the main motivation for our work; by producing a good model for the (population) gradient, we obtain a more efficient estimator for it.

\subsection{Theoretical motivation for model-assisted gradient estimation}

The convergence behavior of stochastic optimization methods is closely tied to the variance of the stochastic gradient estimator \cite{pmlr-v97-qian19b,garrigos2023handbook}. In particular, for strongly convex and smooth objective functions \(f\) (in our case, the loss function \(\lossSymbol\) in Equation \ref{eq:true_risk_definition}), the asymptotic convergence region of SGD-type parameter updates in Equation \ref{Equation::gradient_descent_empirical_step} is determined by the gradient noise variance. This provides theoretical motivation for variance-reduction strategies such as the proposed model-assisted gradient estimator.

We present our motivating theorem by following the work of Gower et al. \cite{pmlr-v97-qian19b} beginning from few key assumptions and definitions. Considering the general optimization problem (as in Equation \ref{Equation::empirical_risk_definition})

\begin{equation}
x^\ast = \arg \min_{x \in \mathbb{R}^d}
\left[
f(x) = \frac{1}{\populationSize}\sum_{i=1}^{\populationSize} f_i(x)\right],  
\end{equation}

where each \(f_i : \mathbb{R}^d \to \mathbb{R}\) is smooth, but not necessarily convex. Also, it is assumed that \(f\) has a unique global minimizer \(x^\ast\)
and is \(\mu\)-strongly (\(\mu>0\)) quasi-convex \cite{Karimi2016}, i.e.:
\begin{equation}
f(x^\ast)
\ge
f(x)
+
\langle \nabla f(x), x^\ast - x \rangle
+
\frac{\mu}{2}\|x^\ast - x\|^2
\end{equation}
for all \(x \in \mathbb{R}^d\).

\begin{definition}[Expected smoothness]
We say that \(f\) is \(L\)-smooth in expectation with respect to a sampling design \(\sampleDistribution\)
if there exists \(L = L(f,\sampleDistribution) > 0\) such that
\begin{equation}
\mathbb{E}_{\sampleDistribution}\left[
\left\|
\nabla f_v(x) - \nabla f_v(x^\ast)
\right\|^2
\right]
\leq
2L\bigl(f(x)-f(x^\ast)\bigr),
\end{equation}
for all \(x \in \mathbb{R}^d\), where \(f_v\) is an unbiased stochastic (mini-batch) estimator of \(f\). When the expected smoothness conditions hold for \(f\), we write \((f,\mathcal{D}) \sim ES(L)\).
\end{definition}

\begin{assumption}[Finite gradient noise]
The gradient noise is assumed to be finite at global minimum, i.e.:
\begin{equation}
\sigma^2 :=
\mathbb{E}_{\sampleDistribution}
\left[
\left\|
\nabla f_v(x^\ast)
\right\|^2
\right] < \infty.    
\end{equation}
\end{assumption}

Next, we present the theorem by Gower et al. using our analogous notation from section \ref{Section::Gradient_descent_optimization}:

\begin{tcolorbox}[colback=gray!15,colframe=gray!15]
\begin{theorem}
Assume \(\lossSymbol\) is \(\mu\)-quasi-strongly convex and that \((\lossSymbol,\sampleDistribution) \sim ES(L)\). Choose \(\eta_t = \eta \in \left(0,\frac{1}{2L}\right]\) for all \(t\). Then the iterates of \text{SGD} (Equation \ref{Equation::gradient_descent_empirical_step}) satisfy
\begin{equation}\label{Equation::convergence_theorem}
\mathbb{E}_{\sampleDistribution}\left[
\|\boldmodelparams_{t} - \boldmodelparams^\ast\|^2
\right]
\leq
(1-\eta\mu)^t
\|\boldmodelparams_0 - \boldmodelparams^\ast\|^2
+
\frac{2\eta\sigma^2}{\mu},
\end{equation}
where now \(\sigma^2 = \mathbb{E}_{\sampleDistribution}\left[\nabla_{\boldmodelparams} \empiricalRiskSymbol_{\subsample}(\boldmodelparams^\ast)\right]\).
\end{theorem}
\end{tcolorbox}

The first term in the bound decreases geometrically with the iteration index \(t\), while the second term defines the asymptotic noise floor induced by stochastic gradient variance \(\sigma^2\). Consequently, even when the optimization process converges, the iterates remain confined to a neighborhood around the optimum whose size is directly proportional to the variance of the gradient estimator.

This observation motivates the proposed \emph{model-assisted gradient estimator}. Since the asymptotic error floor scales linearly with \(\sigma^2\), reducing the variance of the stochastic gradient estimator directly tightens the convergence bound and decreases the limiting optimization error. In this sense, among unbiased gradient estimators, the estimator achieving the smallest variance is theoretically optimal with respect to this convergence guarantee in SGD optimization. For the full-batch gradient, it is clear that \(\sigma^2=0\), which sets the lower bound for stochastic gradient estimators. Using Theorem \ref{Equation::convergence_theorem} and the model-assisted estimator of Equation \ref{Equation::difference_estimator_variance_trace} it is obvious that the model-assisted gradient estimator \(\gradientModelSymbol^*\in\mathcal{Q}\) which minimizes the prediction residuals \(\residual_i\) in \ref{Equation::difference_estimator_variance_trace}, thus also minimizing gradient noise, is explicitly defined as:
\begin{equation}
\gradientModelSymbol^* = \arg\min_{\gradientModelSymbol\in\mathcal{Q}} \sigma_{\gradientModelSymbol}^2
=
\arg\min_{\gradientModelSymbol \in \mathcal{Q}}
\frac{1}{\populationSize}
\sum_{i=1}^{\populationSize}
\left\|
\gradientSymbol_i-\gradientModelSymbol(\xdata_i)
\right\|^2,
\end{equation}
where \(\mathcal{Q}\) is the set of gradient models and \(\sigma_{\gradientModelSymbol}^2\) is the variance of the model-assisted gradient estimator (using gradient model \(\gradientModelSymbol\)) at \(\boldmodelparams^\ast\).

\section{Model-assisted gradient estimator}\label{Section::MA-gradient-estimator}

\begin{figure}[t]
   \centering
    \includegraphics[width=\textwidth]{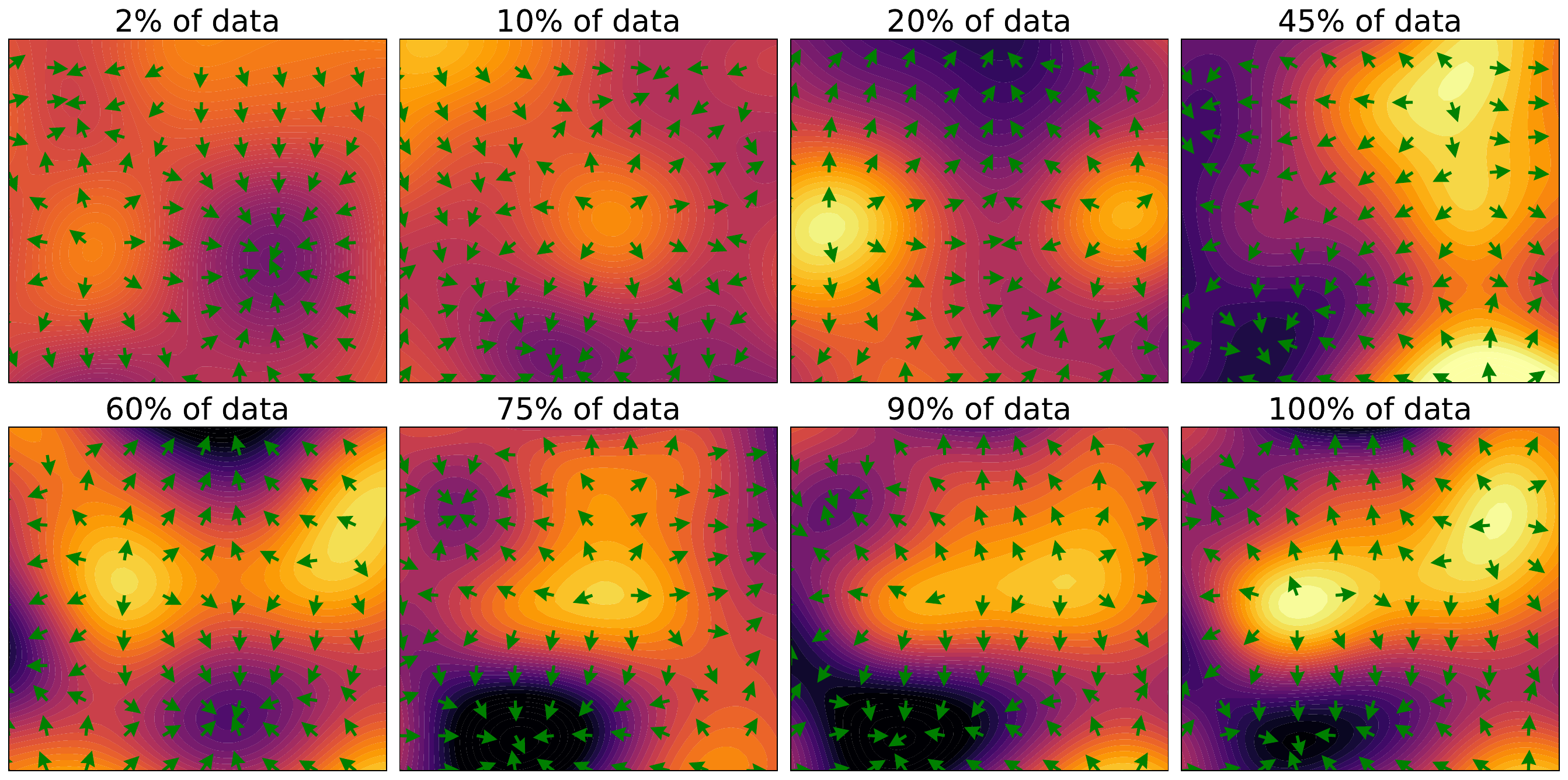} 
    \caption{Artificially generated subsampled risk surfaces in a two-dimensional parameter space. Normalized steepest descent negative gradient directions are visualized as green arrows. The figure illustrates how the estimated artificial risk surface develops as increasingly larger fractions of the data are used to compute the loss and its descent directions, ranging from 2\% of the data to the full dataset.}
    \label{Figure::artificial_risk_function_demo}
\end{figure}

The artificially generated risk surface in Figure \ref{Figure::artificial_risk_function_demo} gives visual motivation for our approach. Given a dataset defining a population risk function through a finite-sum objective \cite{garrigos2023handbook}, each individual sample contributes to the geometry of the overall loss surface in parameter space. As the number of sampled data points increases, the empirical risk surface converges toward the population risk surface. In the small-sample regime, however, the empirical risk may differ substantially from the true population risk, leading to inaccurate gradient estimates. Consequently, the optimization trajectory may exhibit noisy oscillations or move in suboptimal directions. More accurate risk estimates yield descent directions that better approximate the true population gradients, resulting in more stable optimization.

\begin{figure}[t]
    \centering
    \includegraphics[width=\textwidth]{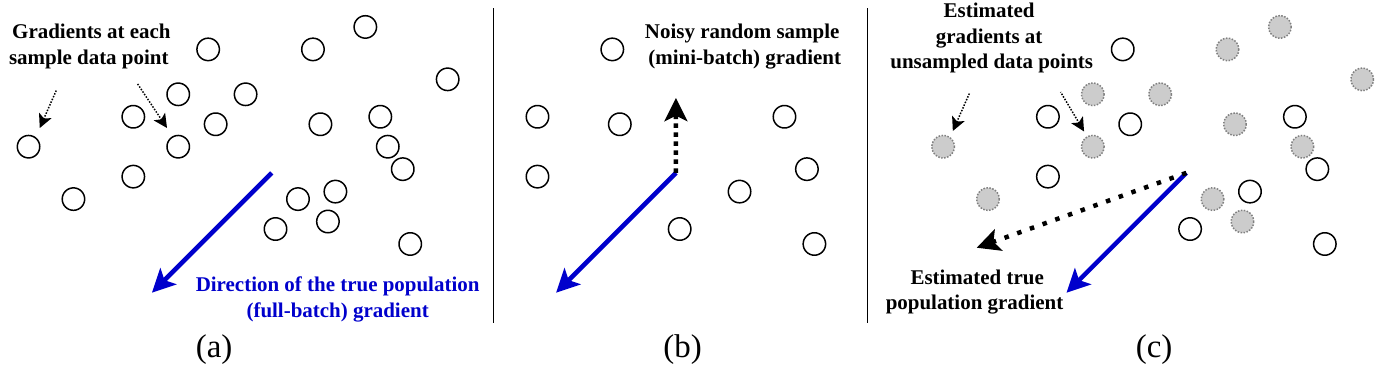}
    \caption{(a) The true population loss gradient calculated using all samples (full-batch). (b) Estimated gradient using a sample (i.e. mini-batch) of the population data. (c) Model-assisted gradient estimation. A mini-batch of data is sampled and used to estimate unsampled loss gradients.}
    \label{fig:noisy_gradient}
\end{figure}

The model-assisted gradient estimation is illustrated in Figure \ref{fig:noisy_gradient}. As the (full-batch) population gradient is generally inaccessible in practical optimization settings, the true population risk surface and its exact gradients cannot be easily evaluated. Therefore, we rely on mini-batch gradient estimators constructed from small random subsets of the data. The proposed gradient estimator combines observed sample gradients with predictions from a gradient model estimating the unsampled components.

The most similar study to ours was presented by Wang et al. \cite{NIPS2013_9766527f} who proposed a control-variate method to reduce the variance of stochastic gradients. The key idea is to reduce stochastic-gradient variance by adding a control-variate correction based on data statistics, such as low-order moments of the entire data. Their estimator replaces the noisy stochastic gradient with an unbiased corrected gradient by subtracting a control-variate term, which requires computing the control variate expectation \(h_d(w)\) \cite{NIPS2013_9766527f} and estimating the scaling matrix or coefficient \(A\) \cite{NIPS2013_9766527f} so that the correction reduces variance. 

In our model-assisted approach from survey sampling theory, the estimators reduce the gradient variance by explicitly modeling the target gradient. Let $\sample$ denote the data population as before.  
We define the corresponding \emph{gradient population} as
\begin{equation}\label{Equation::gradient_population_definition}
\gradientPopulation = \left\{\gradientDataPointSymbol_i=(\xdata_i, \gradientSymbol_i) \,|\, \datapoint_i \in \sample, i\in[\populationSize] \right\},    
\end{equation}
where \(\gradientSymbol_i =\nabla_{\boldmodelparams} \lossSymbol(h(\xdata_i), \ydata_i)\), \(\xdata_i\in\mathbb{R}^m\) and \(\gradientSymbol_i \in \mathbb{R}^d\;\forall i\) where \(m, d\in \mathbb{N}\) is the number of input data dimensions and trainable parameters of hypothesis \(h\) respectively. In other words, \(\gradientPopulation\) contains tuples pairing each input feature vector with its corresponding risk/loss gradient. Given these, we present the model-based gradient estimator in Algorithm \ref{Algorithm::gradient_difference_estimator}. 
\begin{algorithm}[t]
\caption{Model-assisted gradient estimator}
\label{Algorithm::gradient_difference_estimator}
\begin{algorithmic}[1] 
\Require Fixed gradient population \(\gradientPopulation\).  
\Ensure Unbiased gradient estimate $\gradientSymbol_{\mathrm{dif}}$

\State Draw uniformly a random index set $\firstSubsampleIndexSet \subset \left[\populationSize\right]$
\For{each $i \in \firstSubsampleIndexSet$}
    \State Set inclusion probability $\pi_i = 1$
\EndFor
\For{each $i \notin \firstSubsampleIndexSet$}
    \State Choose $0 < \pi_i < 1$
\EndFor
\State Define a sampling design $\mathcal{D}$ induced by the inclusion probabilities \(\pi_i, \,i\in \left[\populationSize\right]\)
\State Given \(\gradientDataPointSymbol \sim \mathcal{D}\), obtain a random sample \(\gradientPopulation_{\indexsetall} = \firstGradsubsample \cup \secondGradsubsample \subset \gradientPopulation\), where index sets \(\firstSubsampleIndexSet\) and \(\secondSubsampleIndexSet\) correspond to \(\pi_i=1\) and \(0<\pi_i<1\) cases respectively, with \(|\firstSubsampleIndexSet| + |\secondSubsampleIndexSet| = n_1 + n_2=\subsampleSize\)
\State Fit gradient model \(\gradientModelSymbol:\mathbb{R}^m \to \mathbb{R}^d,\;\xdata \mapsto \gradientSymbol\) using $\firstGradsubsample$
\For{$i = 1,\dots,N$}
    \State $\hat{\gradientSymbol}_i \gets \gradientModelSymbol(\xdata_i)$
\EndFor

\State Compute difference estimator
\[
\gradientSymbol_{\mathrm{dif}} =
\frac{1}{N} \left(
\sum_{i=1}^{N} \hat{\gradientSymbol}_i
+ \sum_{j \in \firstSubsampleIndexSet} (\gradientSymbol_j - \hat{\gradientSymbol}_j)
+ \sum_{k \in \secondSubsampleIndexSet} \frac{\gradientSymbol_k - \hat{\gradientSymbol}_k}{\pi_k}
\right)
\]

\State \Return $\gradientSymbol_{\mathrm{dif}}$

\end{algorithmic}
\end{algorithm}
First, an initial subset of indices \(\firstSubsampleIndexSet \subset \left[\populationSize\right]\) is drawn uniformly at random. In the second stage, inclusion probabilities are assigned: elements in \(\firstSubsampleIndexSet \) are always selected \(\pi_i=1\), while the remaining elements receive probabilities \(0<\pi_i<1\). A random sample \(\gradientPopulation_{\indexsetall} = \firstGradsubsample \cup \secondGradsubsample \subset \gradientPopulation\) is then drawn according to sampling design \(\mathcal{D}\) induced by these probabilities, corresponding to index sets \(\firstSubsampleIndexSet\) and \(\secondSubsampleIndexSet \subset \left[\populationSize\right] \setminus \firstSubsampleIndexSet\), with \(|\firstSubsampleIndexSet| + |\secondSubsampleIndexSet| = n_1 + n_2=\subsampleSize\). Next, we train the  gradient model \(\gradientModelSymbol:\mathbb{R}^m \to \mathbb{R}^d,\;\xdata \mapsto \gradientSymbol\) using $\firstGradsubsample$. Thus, the gradient model with closed-form solution is deterministic w.r.t \(\mathcal{D}\). Finally, an unbiased difference estimator \(\gradientSymbol_{\mathrm{dif}}\) of the population gradient is formed using Equation \ref{Equation::difference_estimator_definition}.

The performance of the proposed estimator is fundamentally linked to the predictive accuracy of the gradient model \(\gradientModelSymbol\). Consequently, its effectiveness may deteriorate in high-dimensional settings due to curse-of-dimensionality (COD) effects, highly complex gradient landscapes, or insufficient sample sizes for reliable model construction. Nevertheless, the conventional uniform mini-batch gradient estimator arises as a special case (\(\gradientModelSymbol:=\boldsymbol{0}, \pi_i=\frac{\subsampleSize}{\sampleSize}\,\forall i\)) of the proposed formulation, indicating that the proposed approach generalizes standard stochastic gradient estimation.

\section{Empirical analysis}\label{Section::empricial_analysis}

We empirically evaluate the behavior of our model-assisted gradient estimator against a baseline uniformly sampled gradient (i.e. mini-batch) over four different optimizers: SGD, SGD with momentum (SGD-M, momentum coefficient \(\beta=0.9\)), Adam and AdamW. The experiments are tested over seven datasets: synthetic (a random 1D sinusoidal function with a downward-opening parabolic trend), Airfoil self-noise \cite{airfoil_self-noise_291}, Appliances energy \cite{appliances_energy_prediction_374}, MNIST \cite{lecun-mnisthandwrittendigit-2010}, Fashion-MNIST \cite{xiao2017fashionmnist}, CIFAR-10 and CIFAR-100 \cite{CIFAR10_and_100_Cite}. Our primary goal is to understand how the proposed variance-reduced gradient estimator compares to standard mini-batch estimator and full-batch optimization in terms of convergence speed, stability, and final performance across different multilayer perceptron (MLP) and convolutional neural networks (CNN) with varying input space and model parameter sizes. The tested datasets and corresponding models are listed in Table \ref{tab:model_architectures}. As the proposed model-assisted estimator in Algorithm \ref{Algorithm::gradient_difference_estimator} is based on a prediction model \(\gradientModelSymbol\), we will utilize kernel ridge regression (KRR, regularized kernel least squares) \cite{2012Alvarez, pahikkala2008phdthesis} to model the gradient as we can solve the optimal solution analytically and cheaply (with small batch size). For the KRR-model, computations were performed using Scikit-learn library \cite{SKLEARNcite} with Gaussian kernel and default parameters (\(\alpha=0.1, \gamma=1\)). 

\begin{table*}[th]
\centering
\setlength{\tabcolsep}{4pt}
\renewcommand{\arraystretch}{1.15}

\caption{Model architectures used for each dataset in the experiments. CIFAR-10 and CIFAR-100 use the same CNN architecture, differing only in the output layer dimensionality.}
\label{tab:model_architectures}

\resizebox{\textwidth}{!}{%
\begin{tabular}{p{3cm} p{1.8cm} p{2.3cm} p{2.3cm} p{9cm}}
\toprule
\textbf{Dataset(s)} &
\textbf{Model} &
\textbf{Input size} &
\textbf{\# Params} &
\textbf{Architecture} \\
\midrule

Synthetic
& MLP
& 1
& 321
& Linear$(1 \rightarrow 16)$ + ReLU, Linear$(16 \rightarrow 16)$ + ReLU, Linear$(16 \rightarrow 1)$. Regression output without final activation. \\

Airfoil self-noise
& MLP
& 5
& 385
& Linear$(5 \rightarrow 16)$ + ReLU, Linear$(16 \rightarrow 16)$ + ReLU, Linear$(16 \rightarrow 1)$. Regression output without final activation. \\

Appliances energy
& MLP
& 27
& 737
& Linear$(27 \rightarrow 16)$ + ReLU, Linear$(16 \rightarrow 16)$ + ReLU, Linear$(16 \rightarrow 1)$. Regression output without final activation. \\

MNIST, FashionMNIST
& CNN
& 784
& 13\,978
& Conv2d$(1 \rightarrow 8, k=3, p=1)$ + ReLU + MaxPool$(2)$, Conv2d$(8 \rightarrow 16, k=3, p=1)$ + ReLU + MaxPool$(2)$, Flatten$(16 \times 7 \times 7)$, Linear$(784 \rightarrow 16)$ + ReLU, Linear$(16 \rightarrow 10)$. \\

CIFAR-10
& CNN
& 3072
& 17\,962
& Conv2d$(3 \rightarrow 8, k=3, p=1)$ + ReLU + MaxPool$(2)$, Conv2d$(8 \rightarrow 16, k=3, p=1)$ + ReLU + MaxPool$(2)$, Flatten$(16 \times 8 \times 8)$, Linear$(1024 \rightarrow 16)$ + ReLU, Linear$(16 \rightarrow 10)$. \\

CIFAR-100
& CNN
& 3072
& 19\,492
& Same architecture as CIFAR-10, except the final classification layer is Linear$(16 \rightarrow 100)$ instead of Linear$(16 \rightarrow 10)$. \\

\bottomrule
\end{tabular}
}
\end{table*}

We focus on three complementary metrics: (i) test loss, capturing generalization quality, (ii) loss standard deviation across runs, reflecting stability, and (iii) the epoch at which the loss reaches its minimum, indicating efficiency. These metrics allow us to jointly assess not only how well an optimizer performs but also how reliably and how quickly it reaches a good solution. The learning rate is fixed to \(\eta=5\times10^{
-3}\) and run for 100 epochs in all experiments with batch sizes of 10 (\(n_1=8, n_2=2\)), 50 (\(n_1=30, n_2=20\)) and 100 (\(n_1=80, n_2=20\)). Reported results correspond to averaged performance trajectories over 400 distinct runs with random model initialization and data seeds (visualizations are shown up to 50 epochs when curve trajectories have mostly stabilized). To make repeated experiments computationally feasible while still allowing variability across runs to be estimated reliably, each experiment was performed on a randomly sampled subset of 1,000 examples from the corresponding dataset. Of these, 800 examples were used for training and 200 were held out to evaluate generalization performance using the test loss.

\begin{figure*}[t]
\centering
\setlength{\tabcolsep}{3pt}
\renewcommand{\arraystretch}{0.8}

\begin{tabular}{>{\centering\arraybackslash}m{0.04\textwidth}
                >{\centering\arraybackslash}m{0.29\textwidth}
                >{\centering\arraybackslash}m{0.29\textwidth}
                >{\centering\arraybackslash}m{0.29\textwidth}}

& \textbf{Batch size 10} & \textbf{Batch size 50} & \textbf{Batch size 100} \\

\rotatebox[origin=c]{90}{\footnotesize\textbf{Synthetic}} &
\includegraphics[valign=m,width=0.29\textwidth]{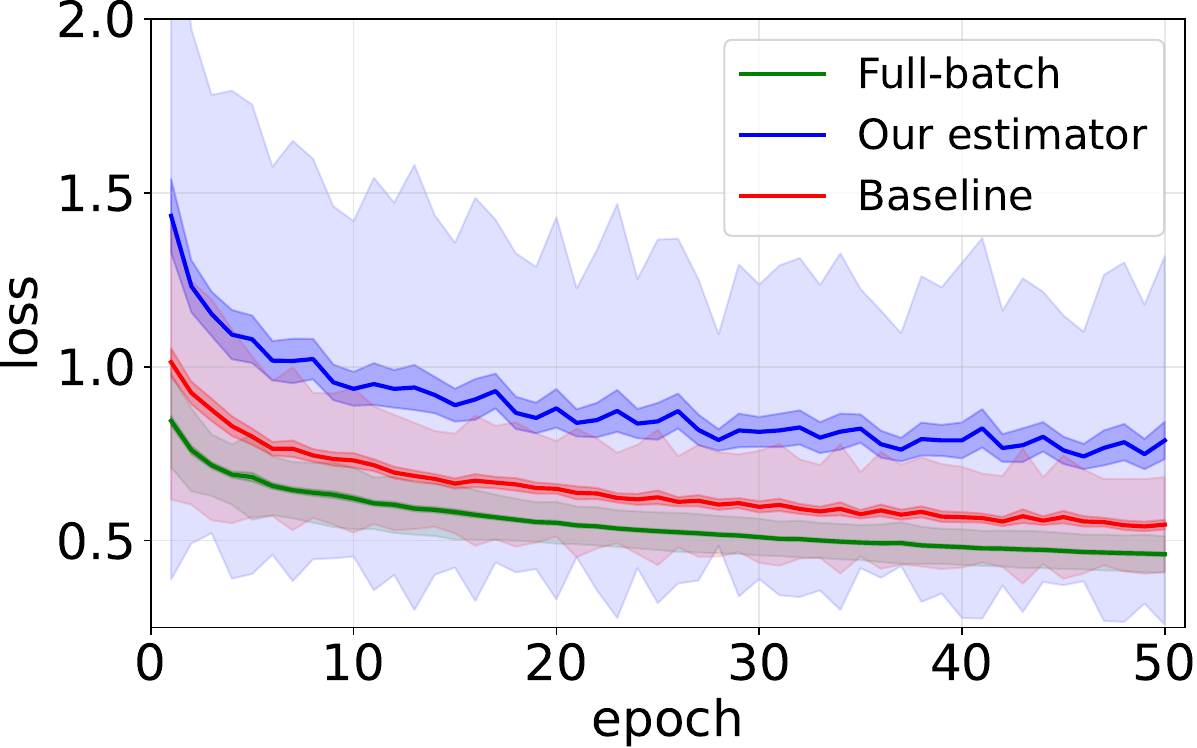} &
\includegraphics[valign=m,width=0.29\textwidth]{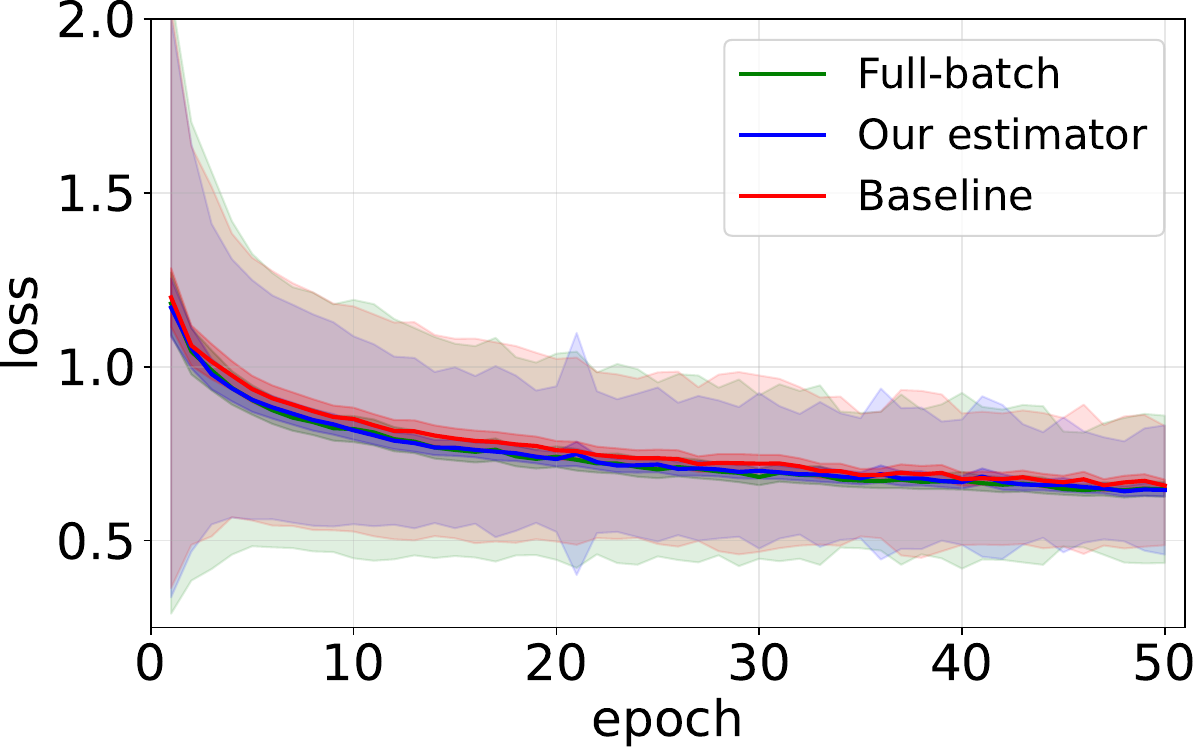} &
\includegraphics[valign=m,width=0.29\textwidth]{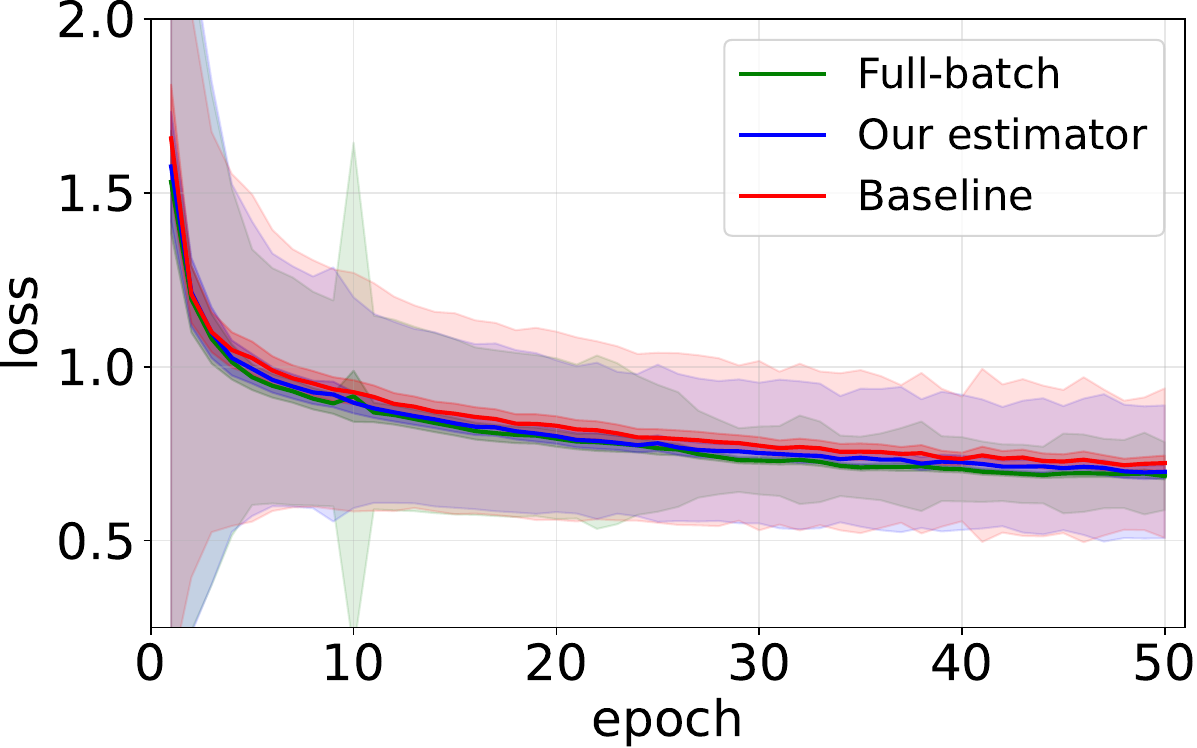} \\

\rotatebox[origin=c]{90}{\footnotesize\textbf{Airfoil self-noise}} &
\includegraphics[valign=m,width=0.29\textwidth]{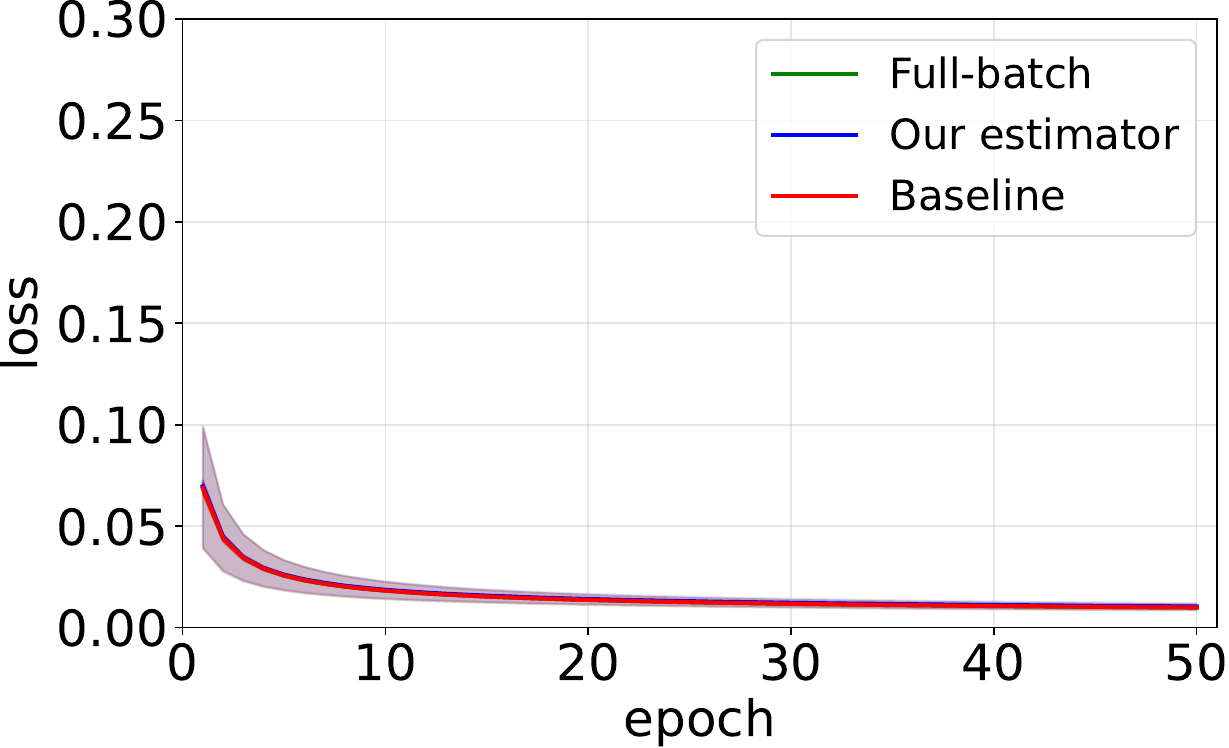} &
\includegraphics[valign=m,width=0.29\textwidth]{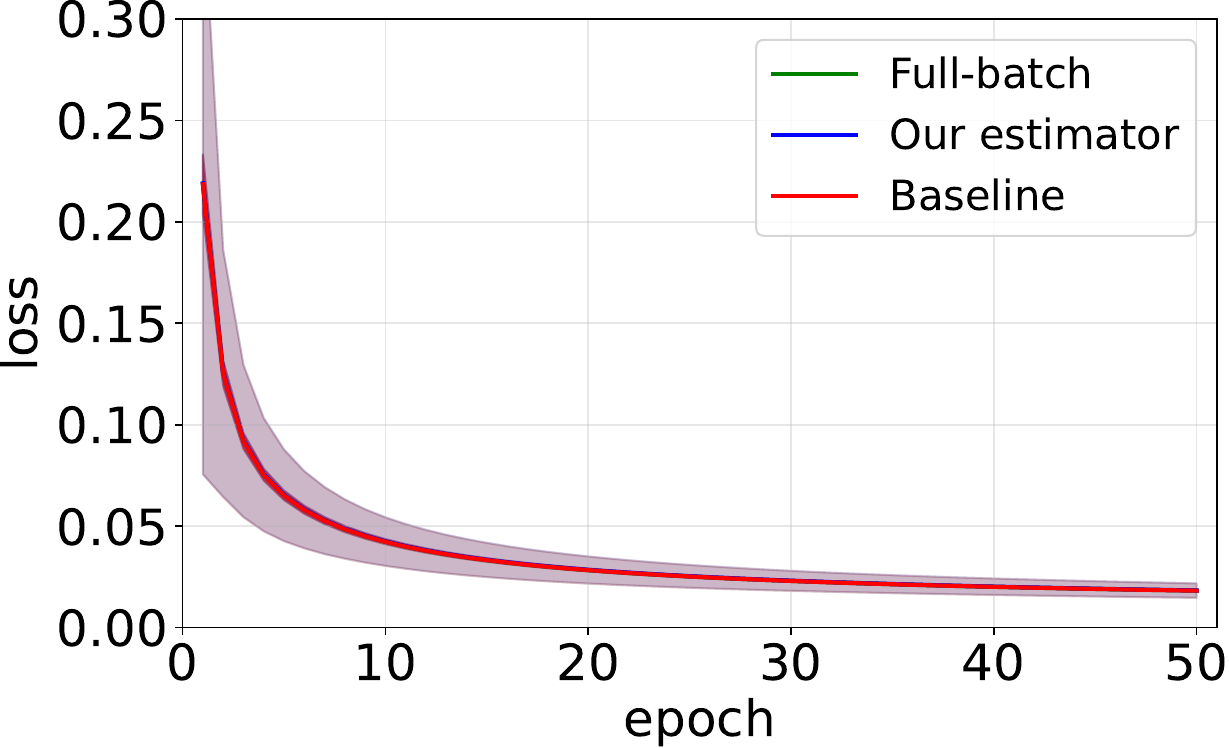} &
\includegraphics[valign=m,width=0.29\textwidth]{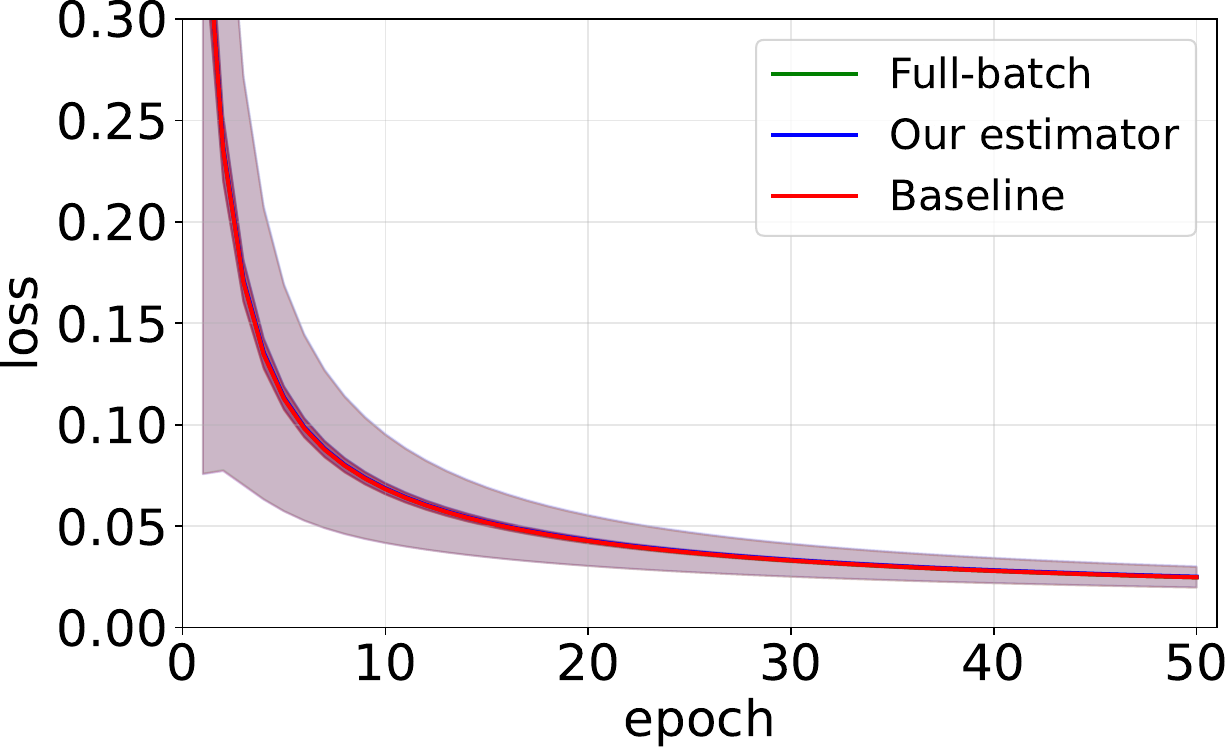} \\

\rotatebox[origin=c]{90}{\footnotesize\textbf{Appliances energy}} &
\includegraphics[valign=m,width=0.29\textwidth]{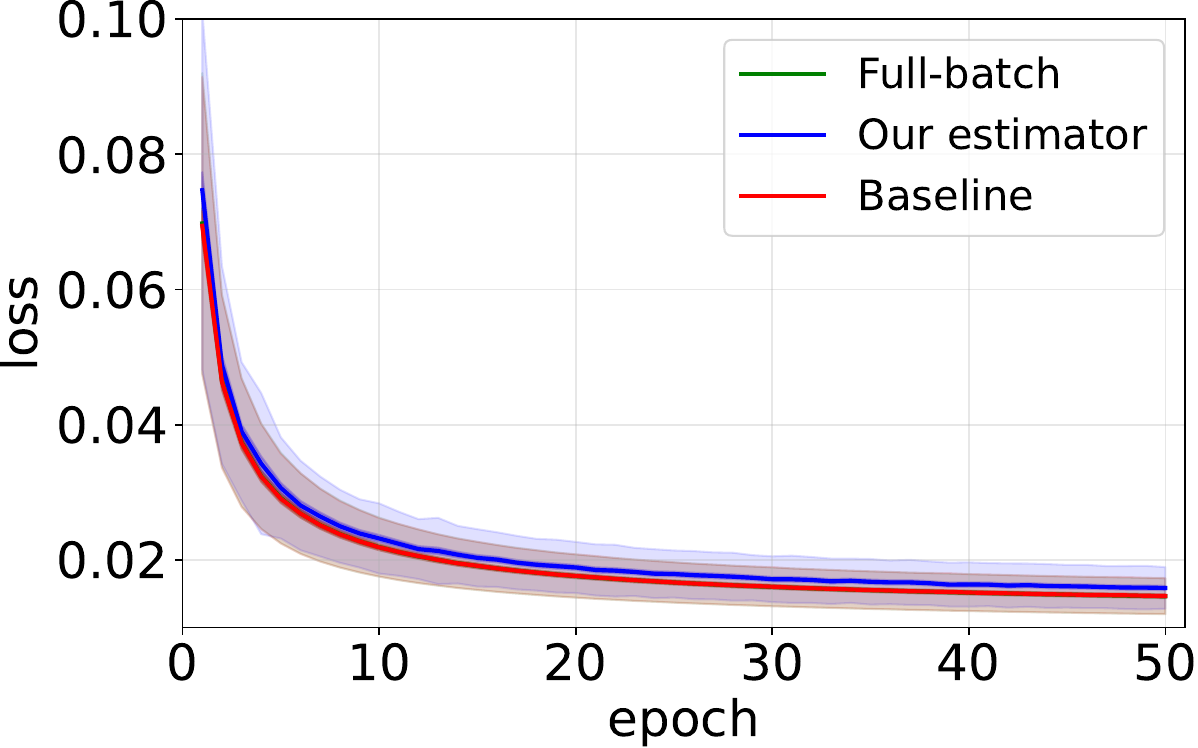} &
\includegraphics[valign=m,width=0.29\textwidth]{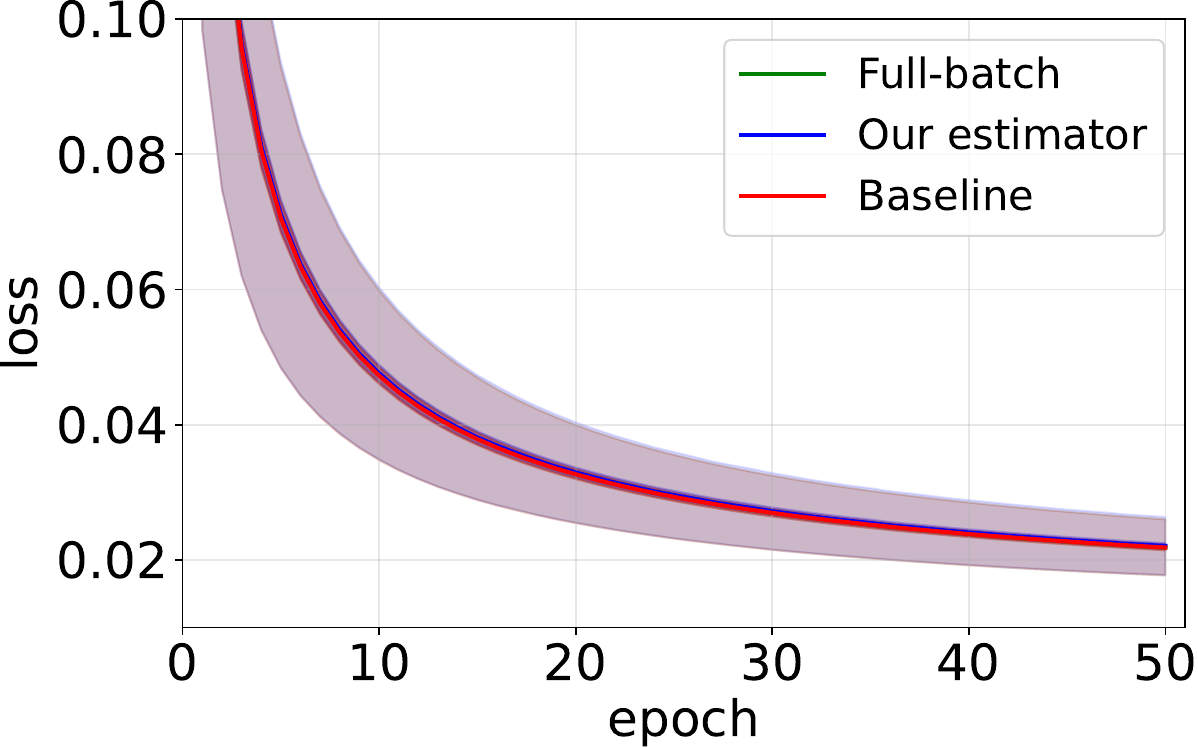} &
\includegraphics[valign=m,width=0.29\textwidth]{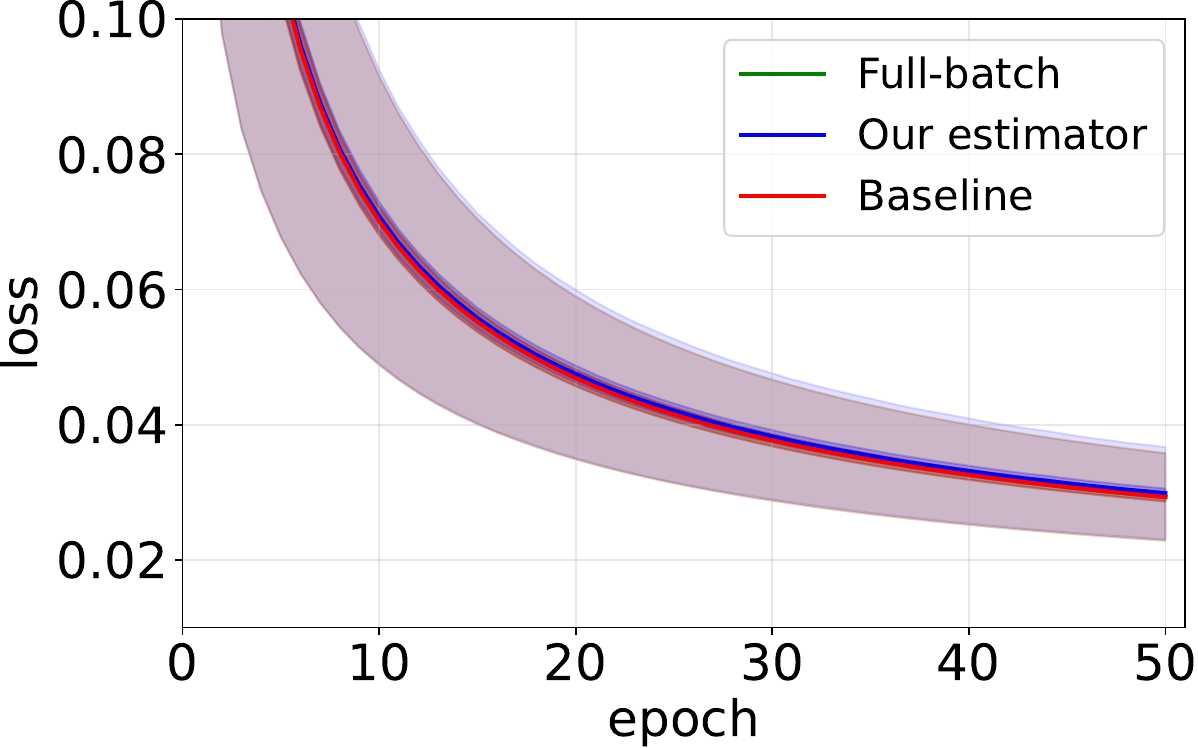} \\

\rotatebox[origin=c]{90}{\footnotesize\textbf{MNIST}} &
\includegraphics[valign=m,width=0.29\textwidth]{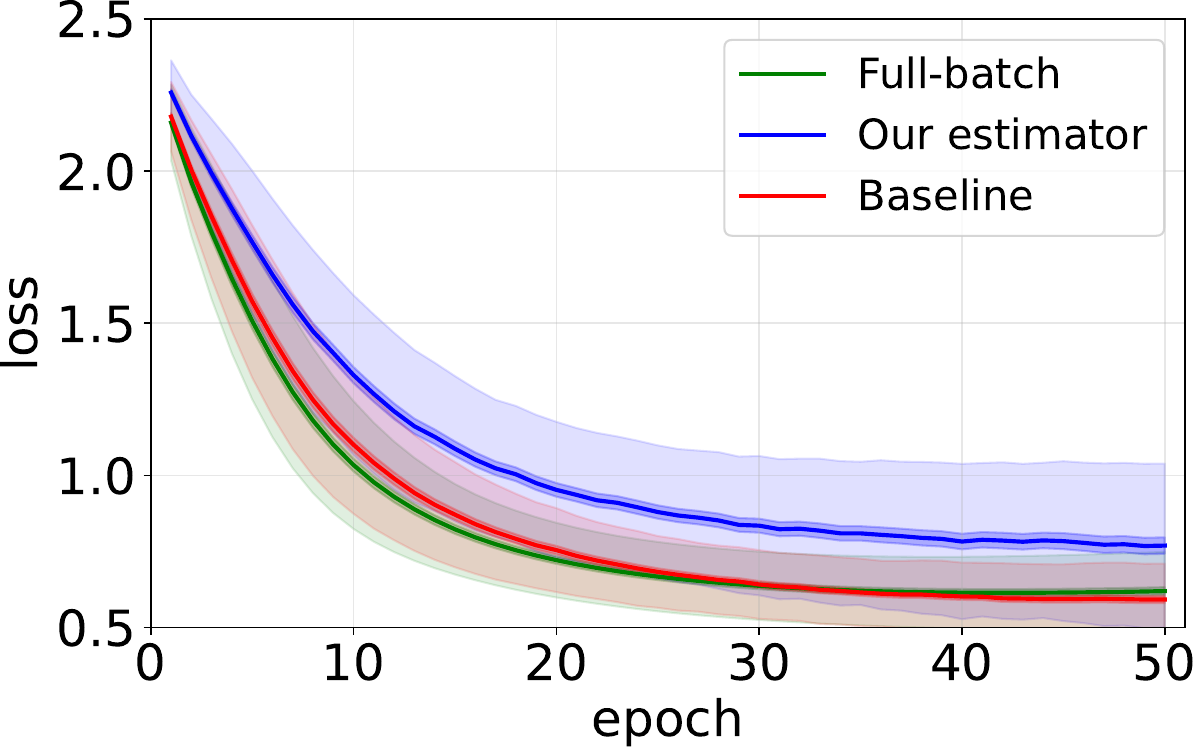} &
\includegraphics[valign=m,width=0.29\textwidth]{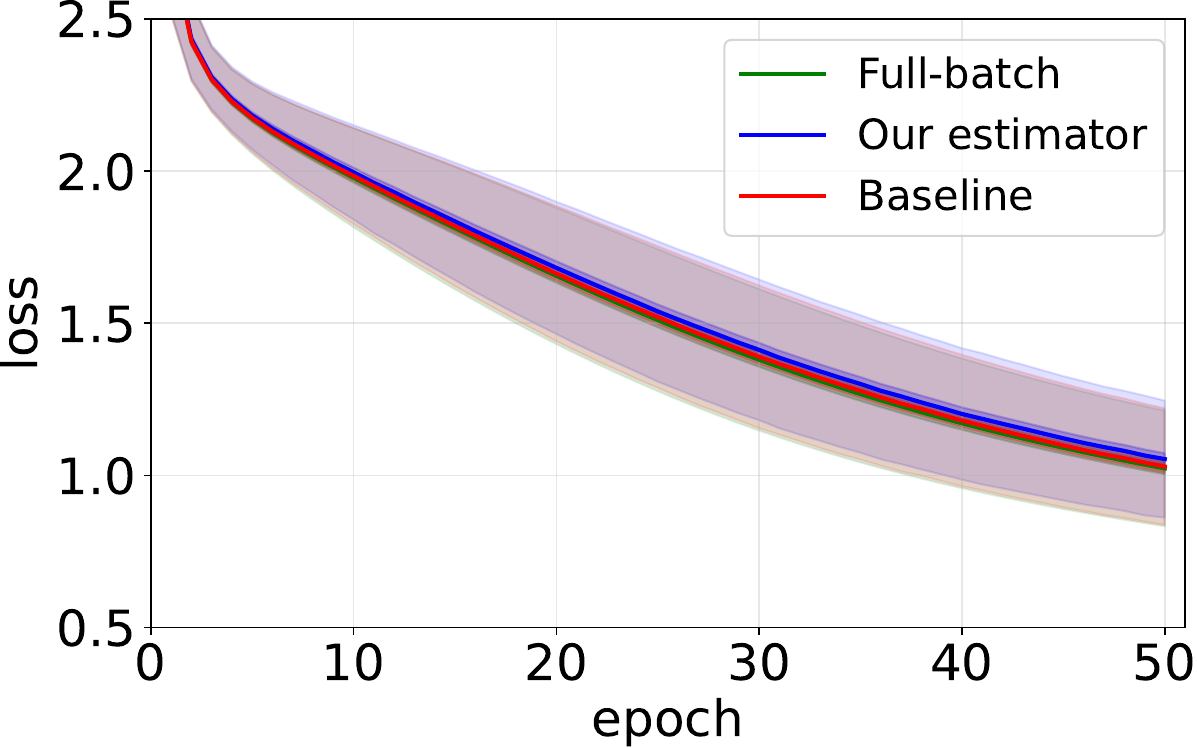} &
\includegraphics[valign=m,width=0.29\textwidth]{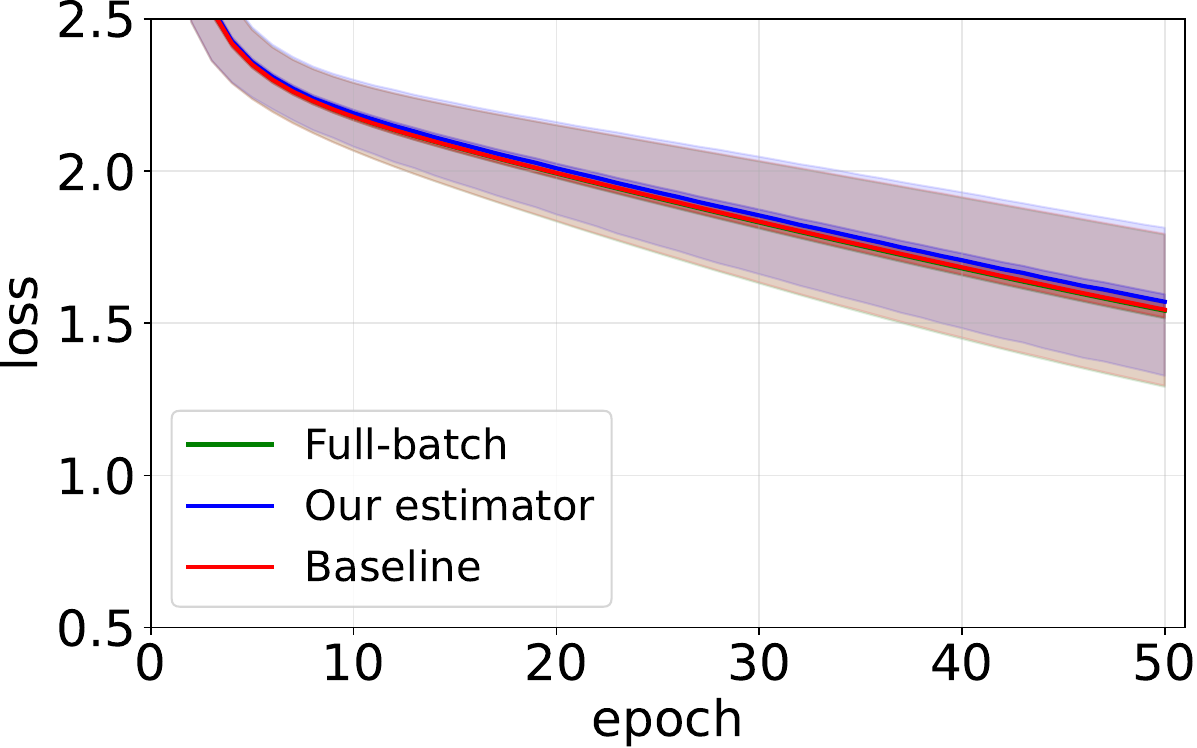} \\

\rotatebox[origin=c]{90}{\footnotesize\textbf{Fashion-MNIST}} &
\includegraphics[valign=m,width=0.29\textwidth]{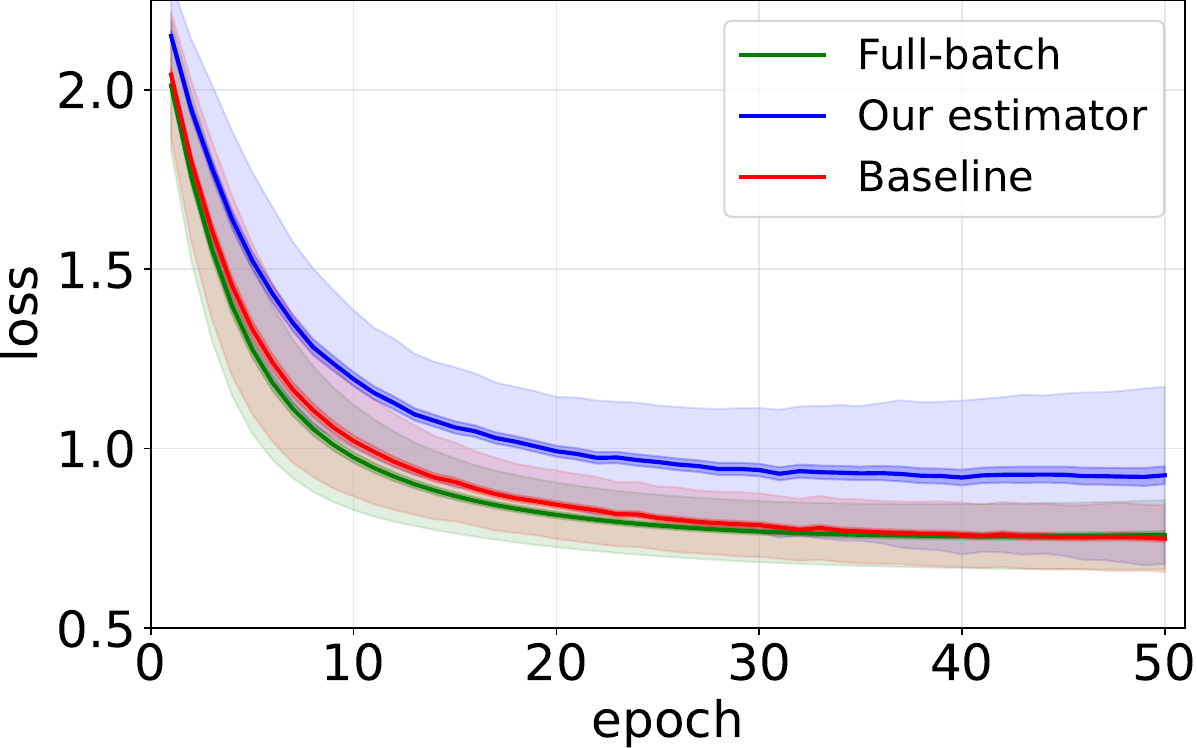} &
\includegraphics[valign=m,width=0.29\textwidth]{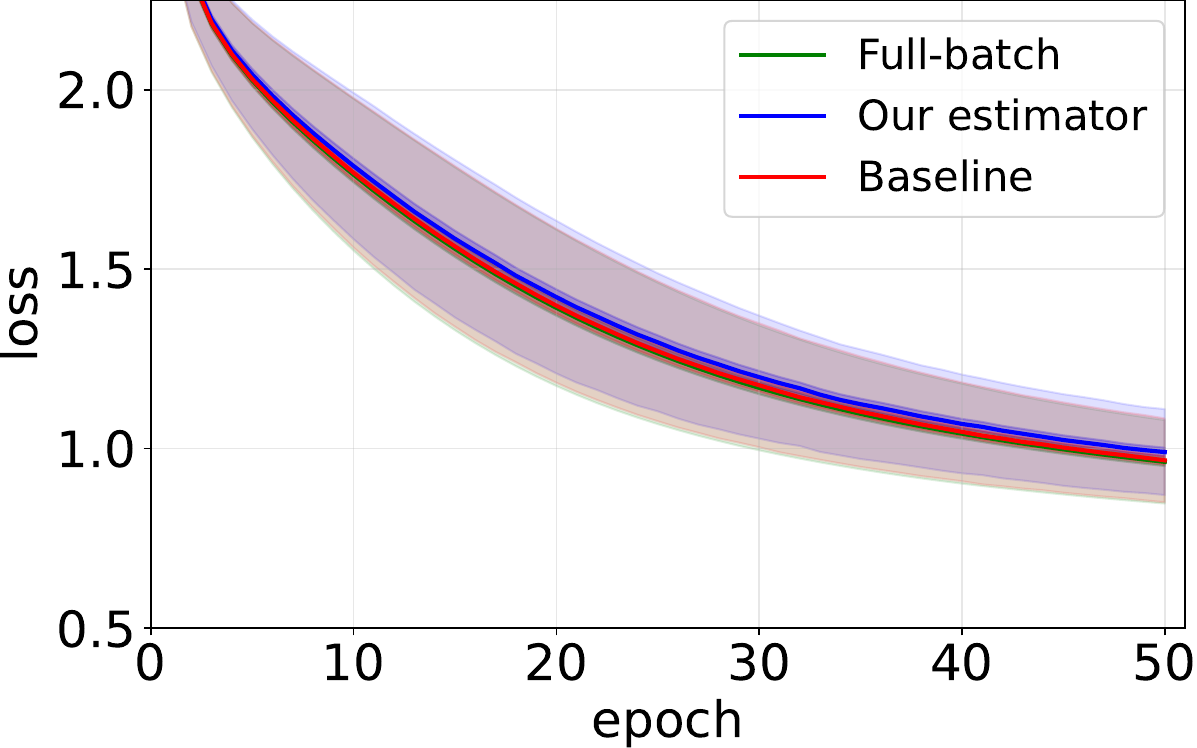} &
\includegraphics[valign=m,width=0.29\textwidth]{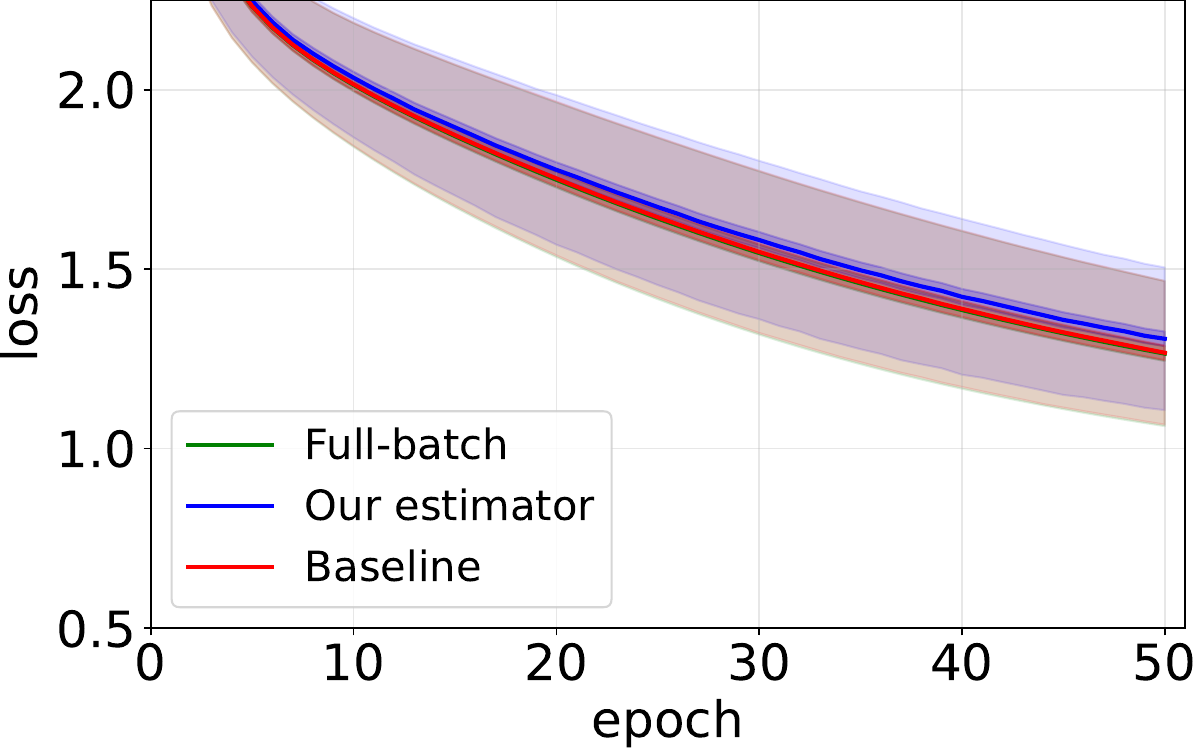} \\

\rotatebox[origin=c]{90}{\footnotesize\textbf{CIFAR-10}} &
\includegraphics[valign=m,width=0.29\textwidth]{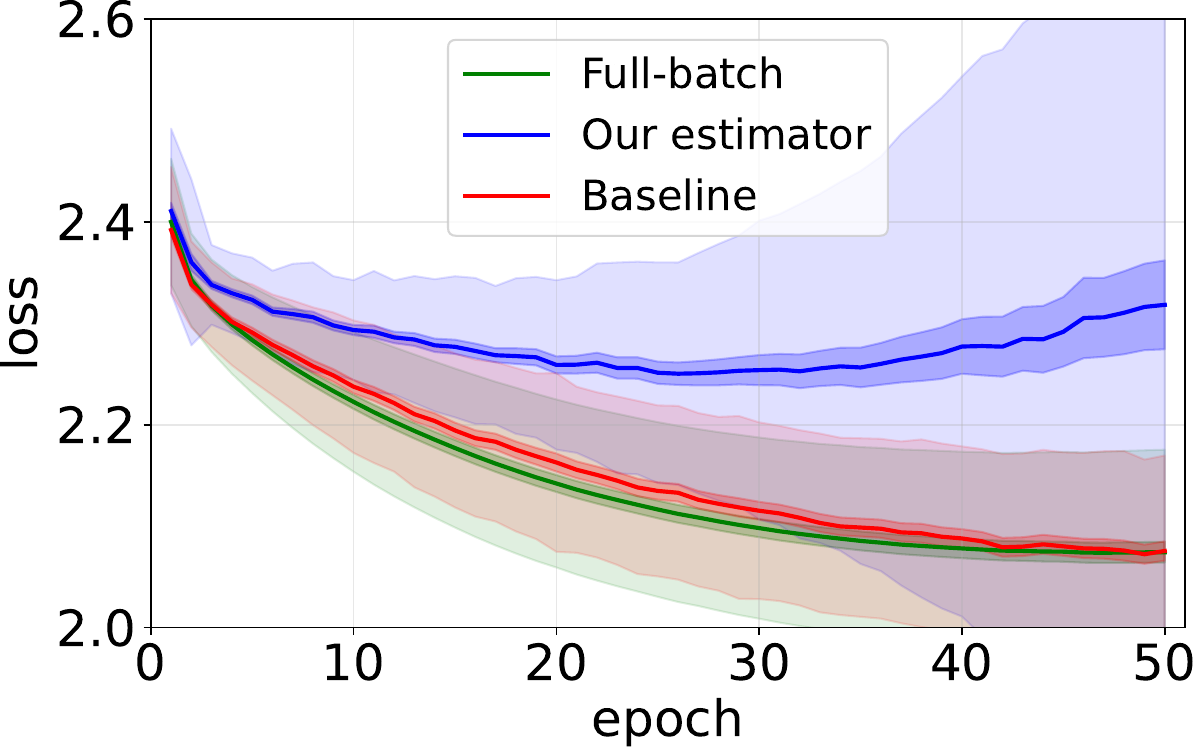} &
\includegraphics[valign=m,width=0.29\textwidth]{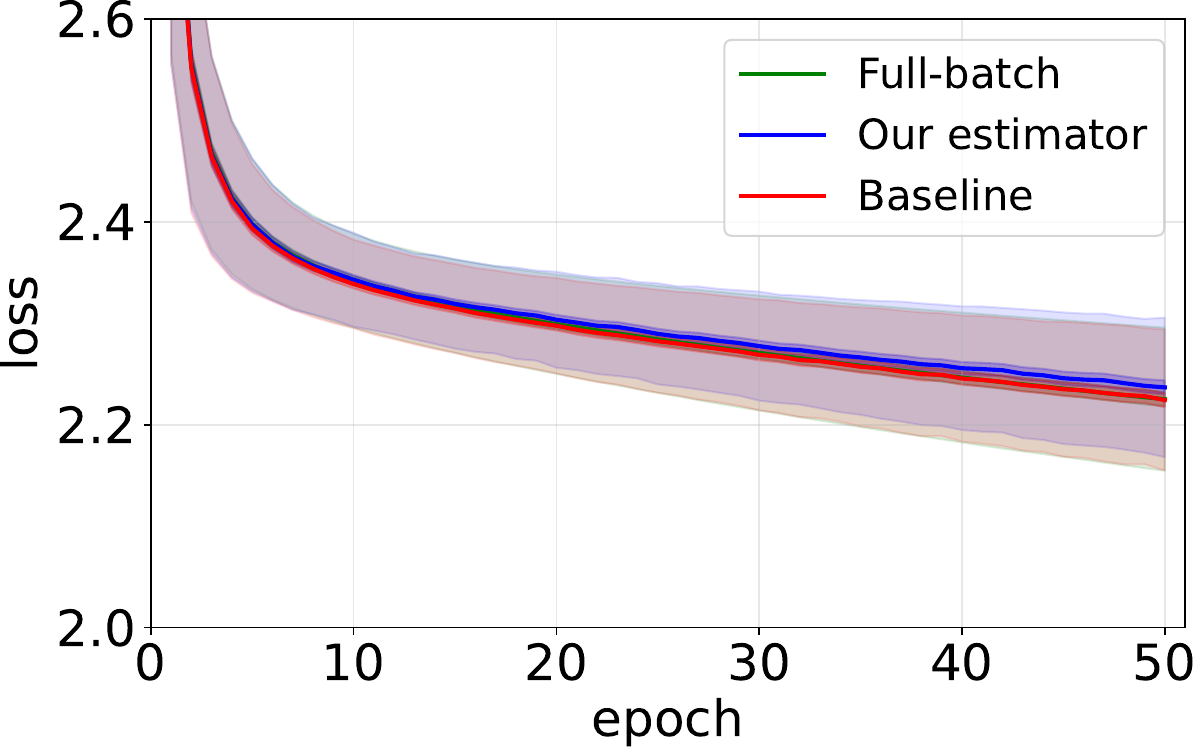} &
\includegraphics[valign=m,width=0.29\textwidth]{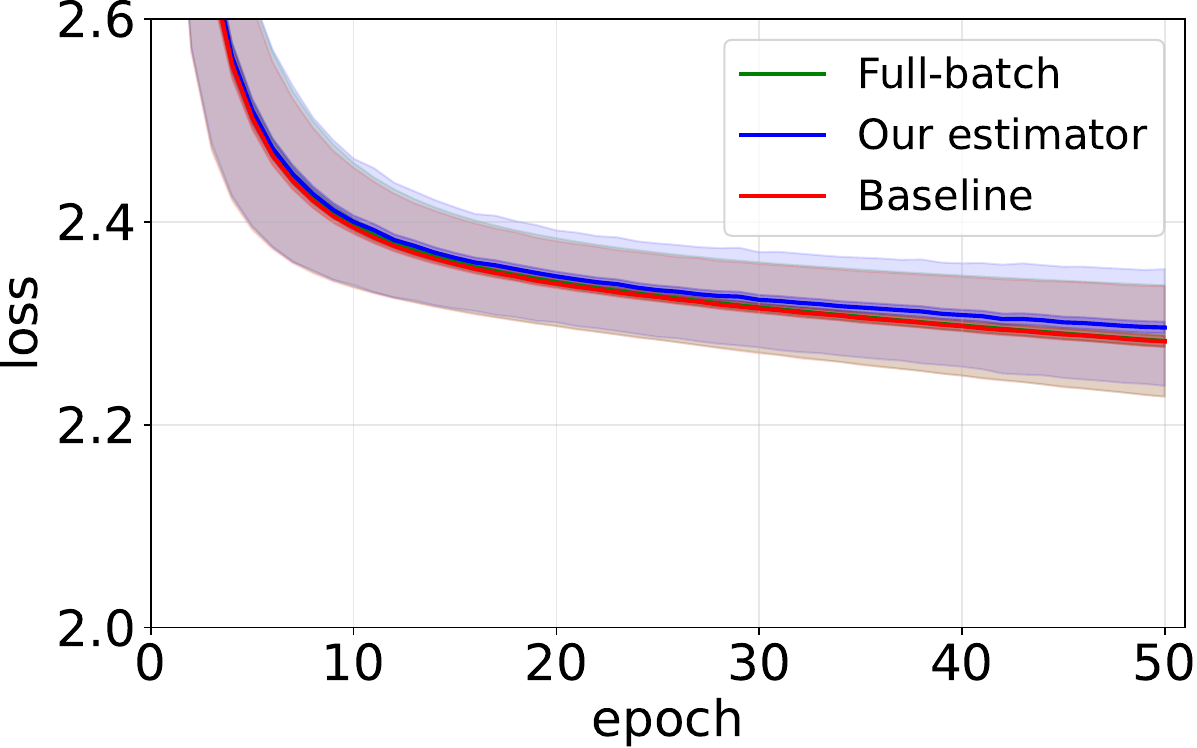} \\

\rotatebox[origin=c]{90}{\footnotesize\textbf{CIFAR-100}} &
\includegraphics[valign=m,width=0.29\textwidth]{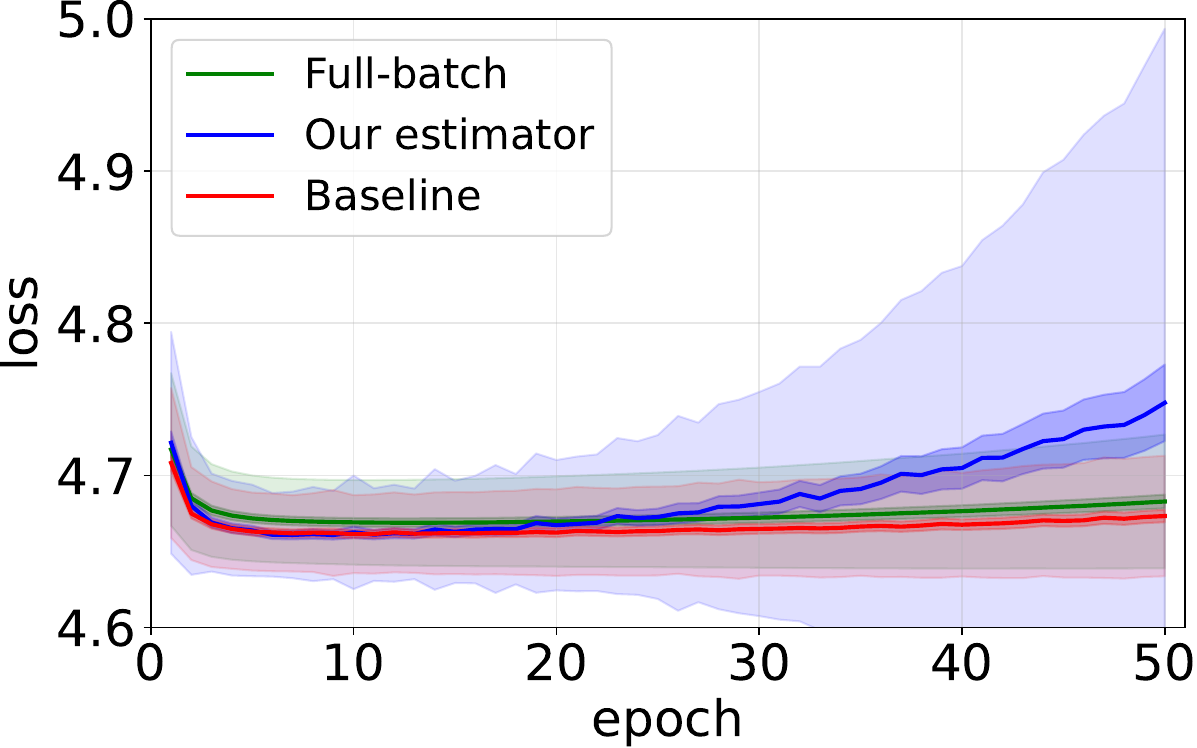} &
\includegraphics[valign=m,width=0.29\textwidth]{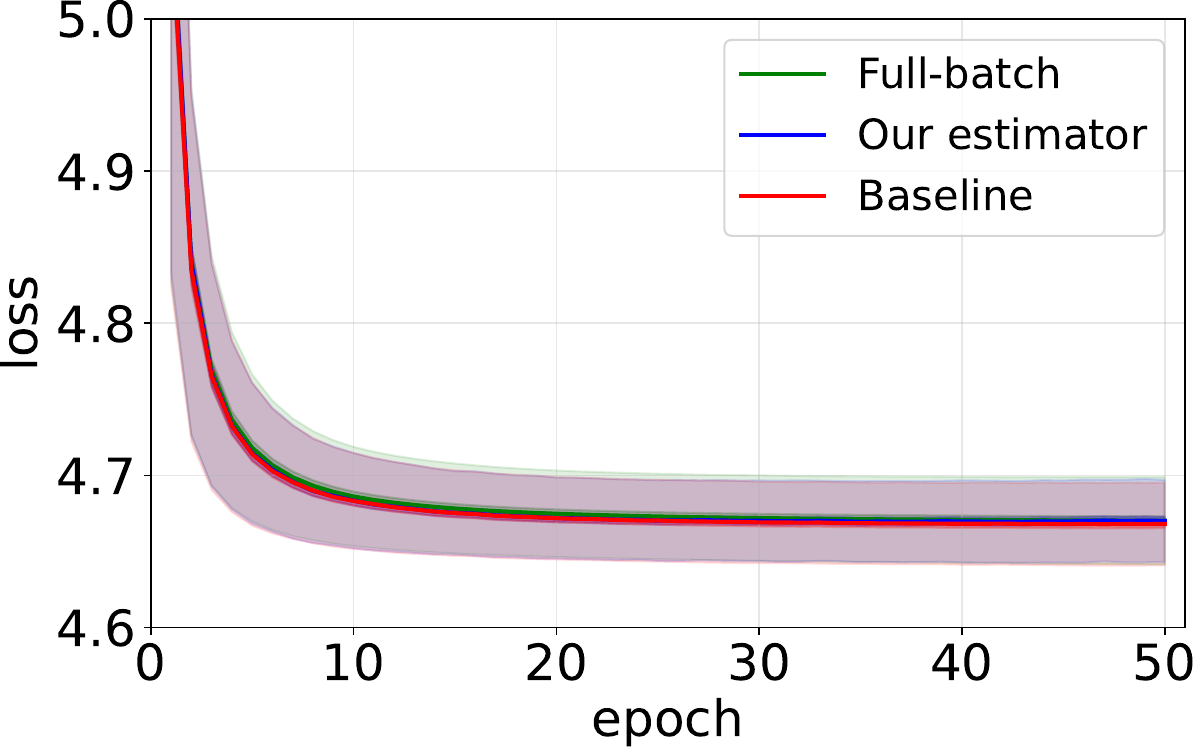} &
\includegraphics[valign=m,width=0.29\textwidth]{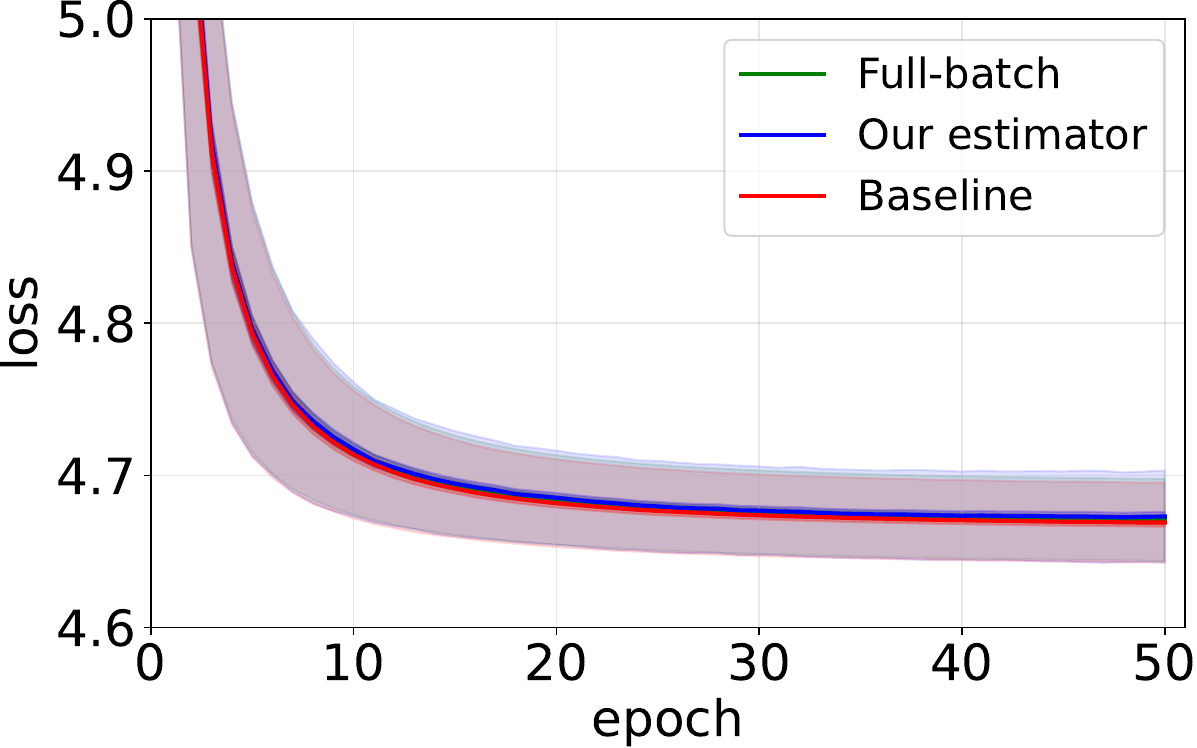} \\

\end{tabular}

\caption{The generalization performance for SGD optimizer with baseline (uniform mini-batch), model-assisted (our estimator) and full-batch gradients. Full-batch gradient case is listed for comparison purposes. The darker curve represents the average test loss of 400 runs, the dark shading is the 95\% confidence interval and lighter shading shows the standard deviation. Columns correspond to batch sizes and rows to datasets.}
\label{fig:sgd_loss_grid}
\end{figure*}

\begin{figure*}[t]
\centering
\setlength{\tabcolsep}{3pt}
\renewcommand{\arraystretch}{0.8}

\begin{tabular}{>{\centering\arraybackslash}m{0.04\textwidth}
                >{\centering\arraybackslash}m{0.29\textwidth}
                >{\centering\arraybackslash}m{0.29\textwidth}
                >{\centering\arraybackslash}m{0.29\textwidth}}

& \textbf{Batch size 10} & \textbf{Batch size 50} & \textbf{Batch size 100} \\

\rotatebox[origin=c]{90}{\footnotesize\textbf{Synthetic}} &
\includegraphics[valign=m,width=0.29\textwidth]{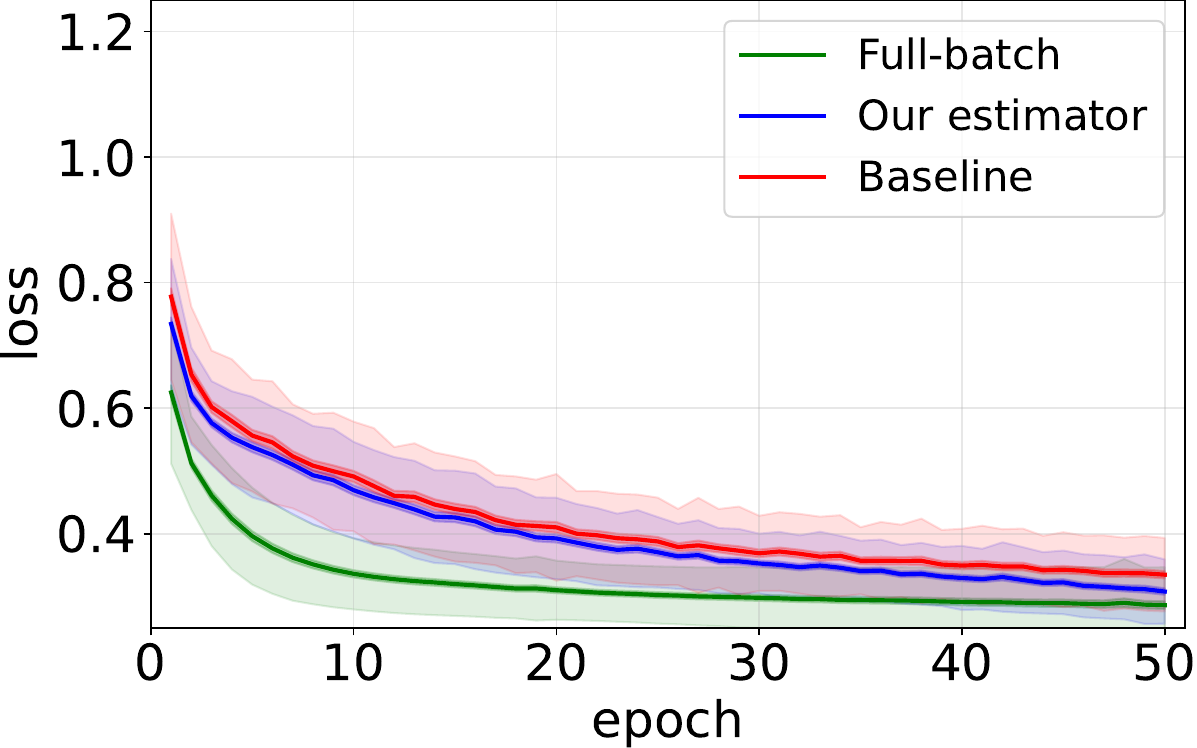} &
\includegraphics[valign=m,width=0.29\textwidth]{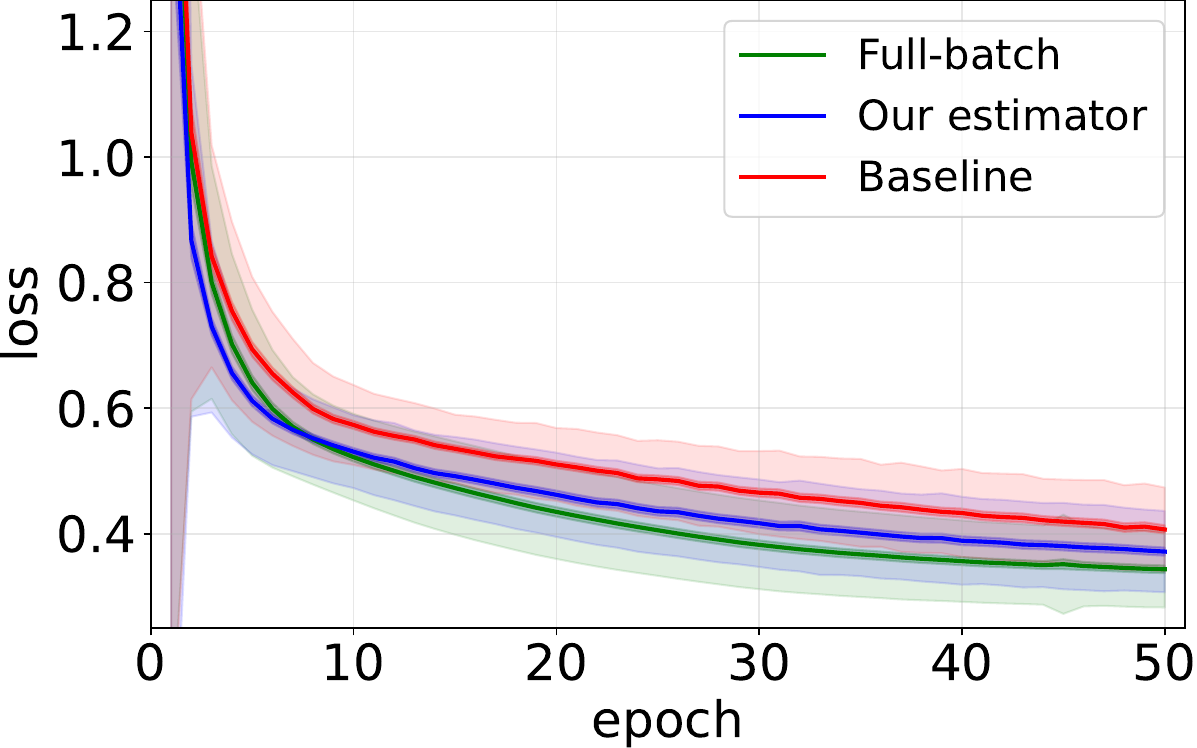} &
\includegraphics[valign=m,width=0.29\textwidth]{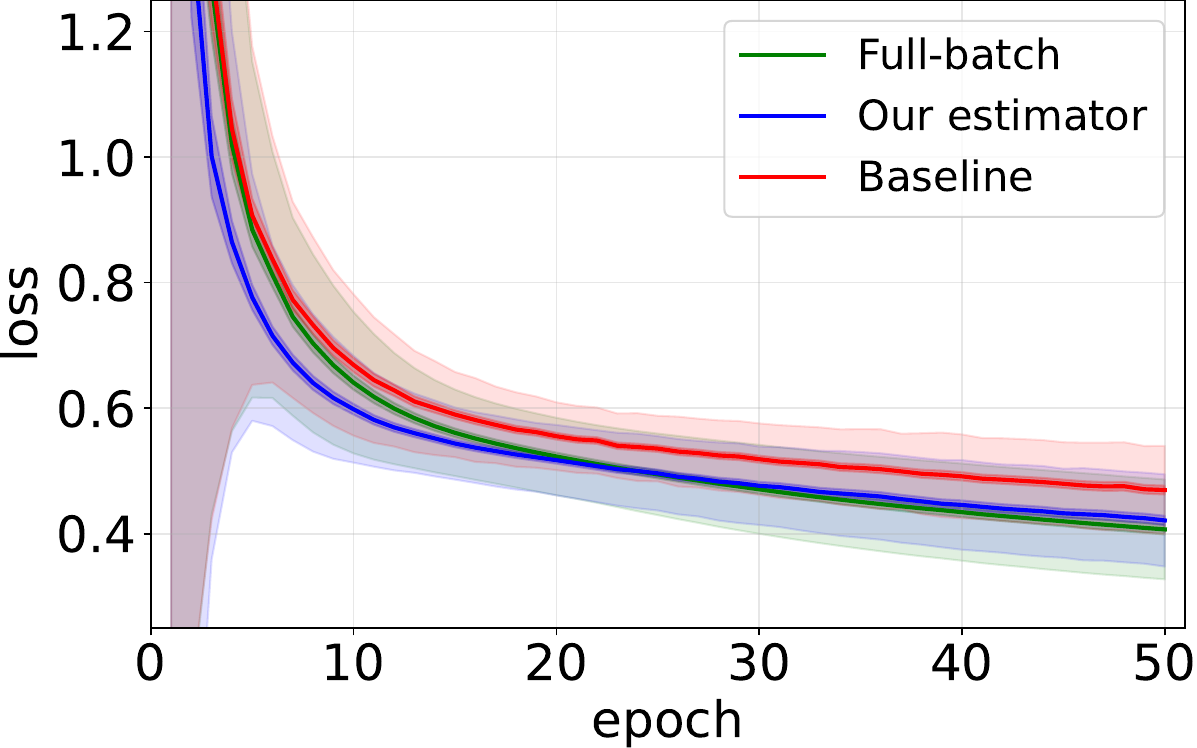} \\

\rotatebox[origin=c]{90}{\footnotesize\textbf{Airfoil self-noise}} &
\includegraphics[valign=m,width=0.29\textwidth]{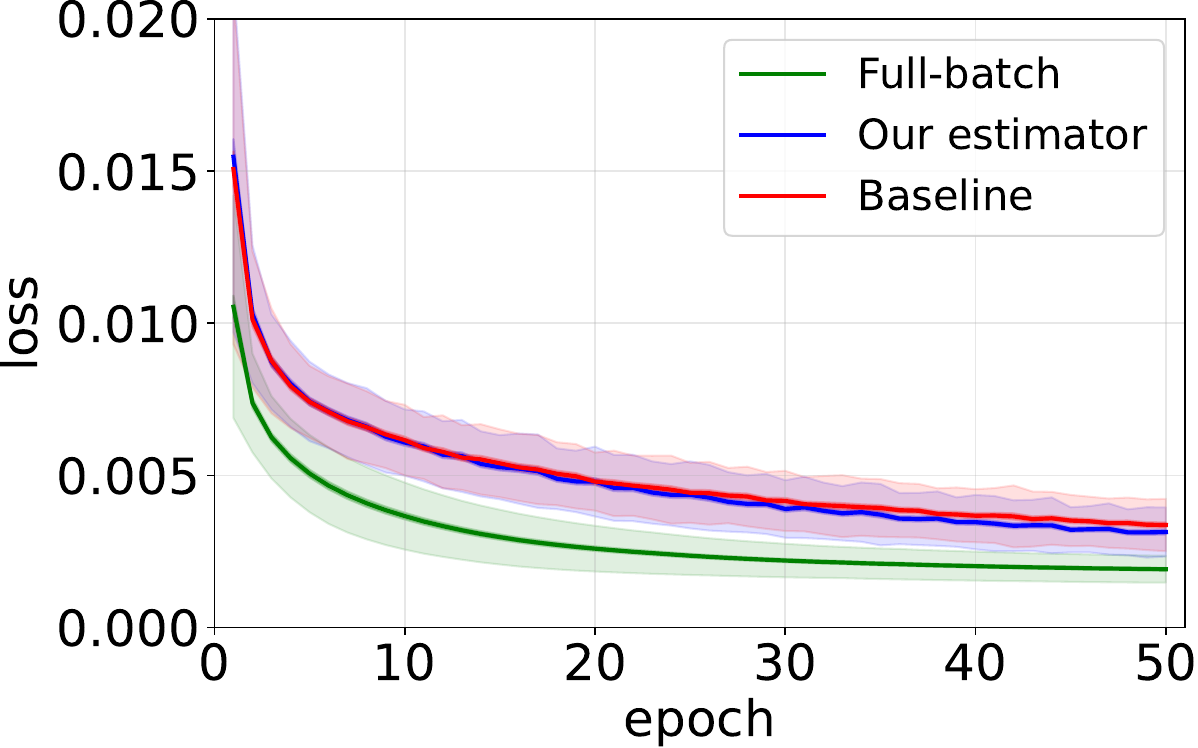} &
\includegraphics[valign=m,width=0.29\textwidth]{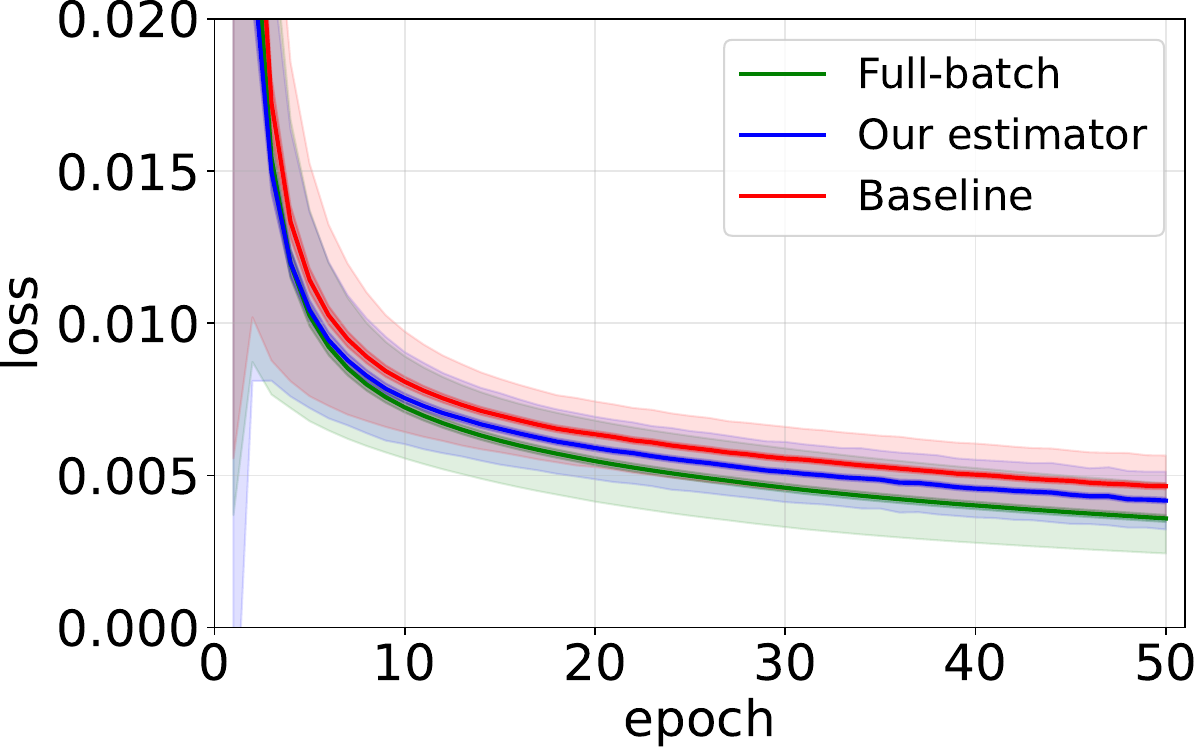} &
\includegraphics[valign=m,width=0.29\textwidth]{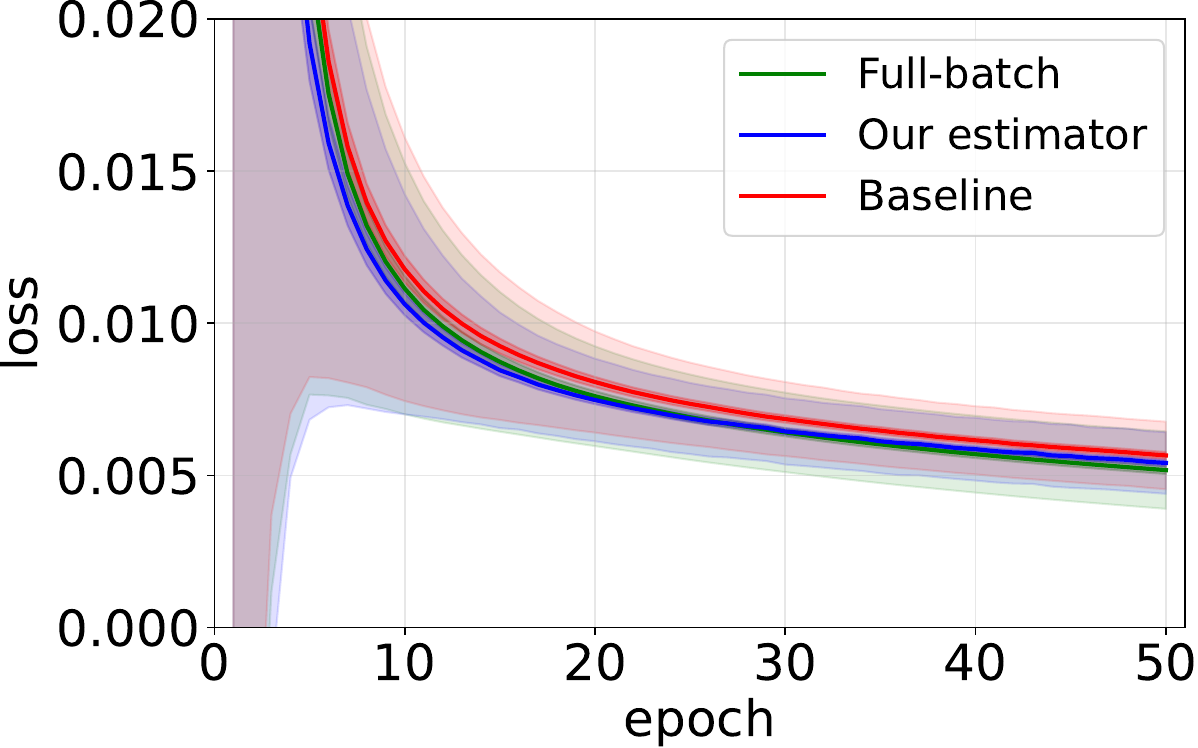} \\

\rotatebox[origin=c]{90}{\footnotesize\textbf{Appliances energy}} &
\includegraphics[valign=m,width=0.29\textwidth]{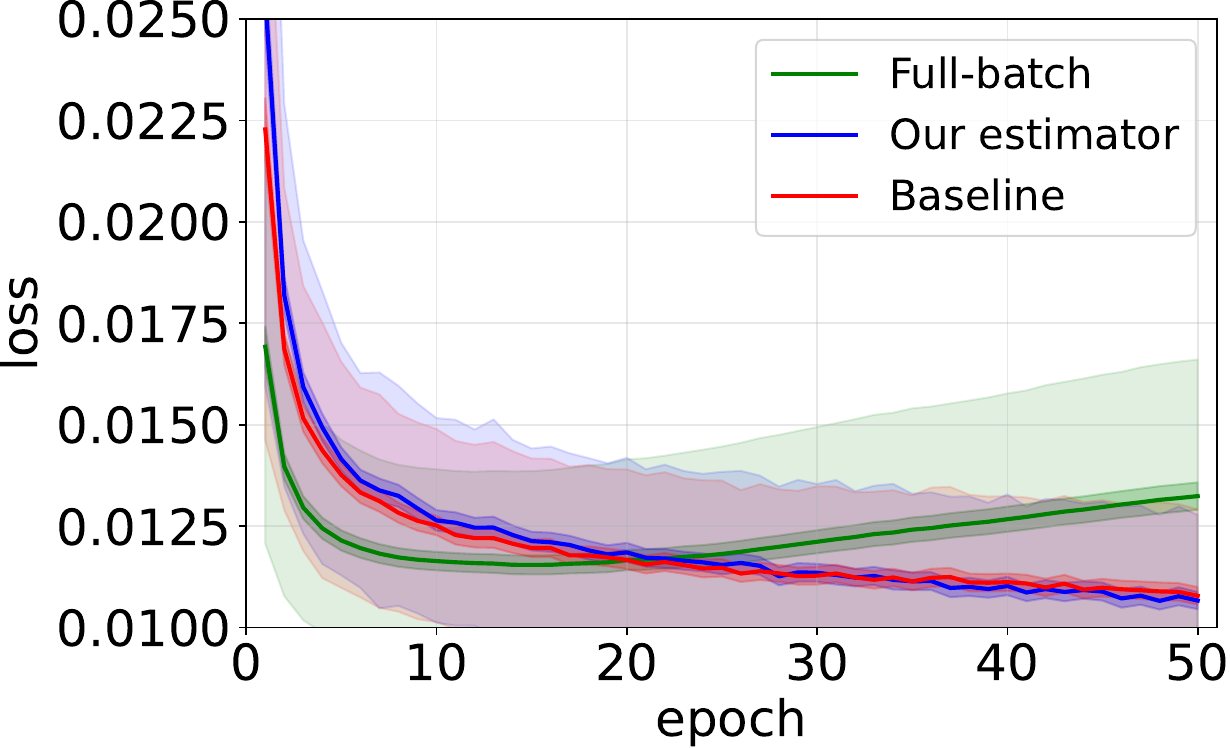} &
\includegraphics[valign=m,width=0.29\textwidth]{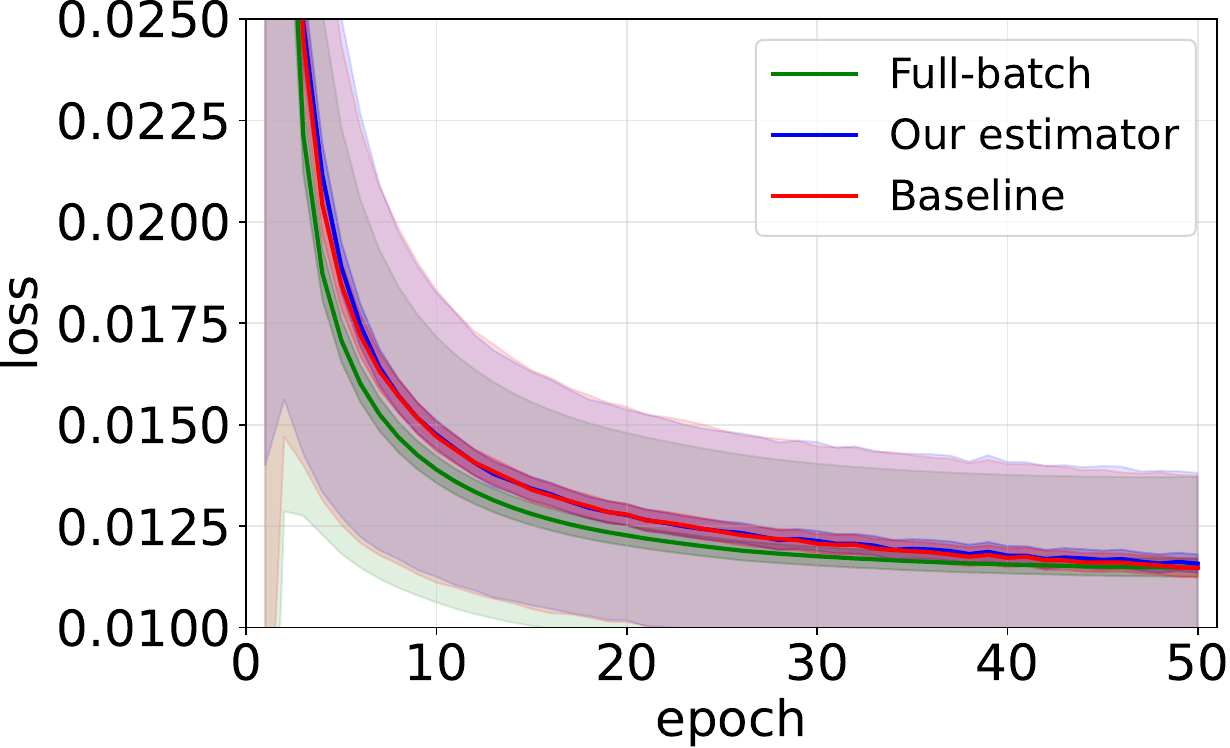} &
\includegraphics[valign=m,width=0.29\textwidth]{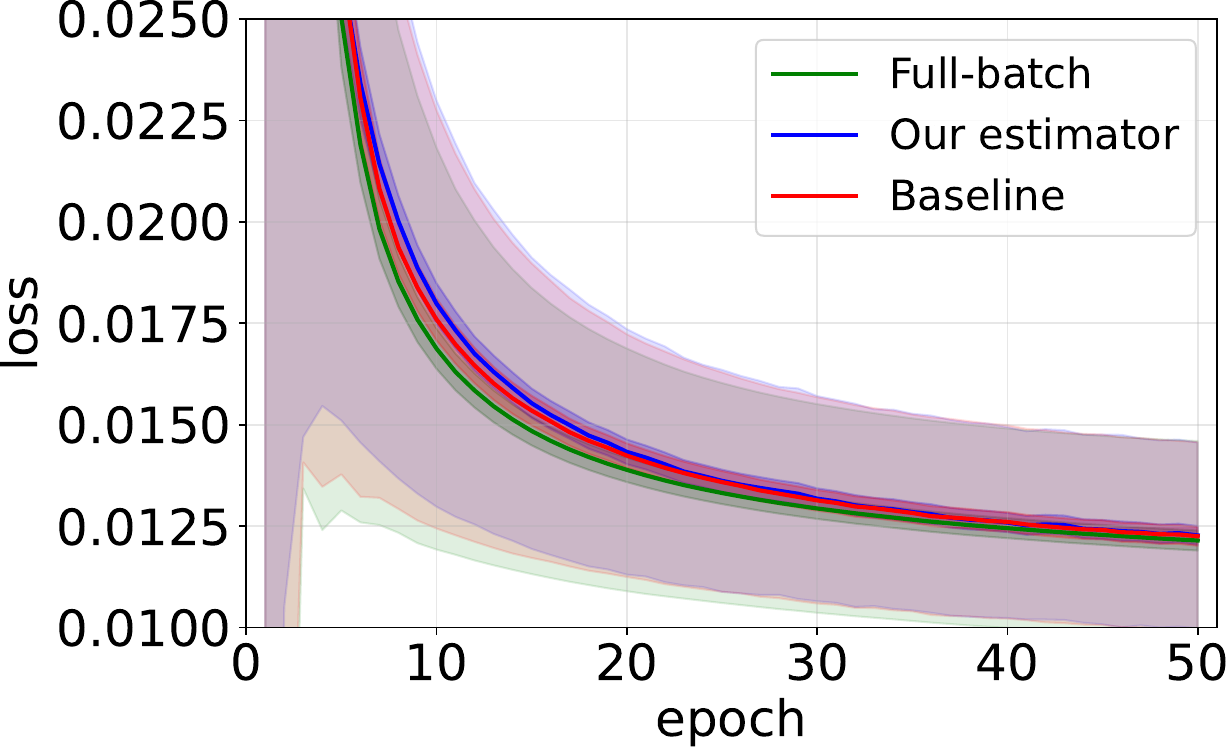} \\

\rotatebox[origin=c]{90}{\footnotesize\textbf{MNIST}} &
\includegraphics[valign=m,width=0.29\textwidth]{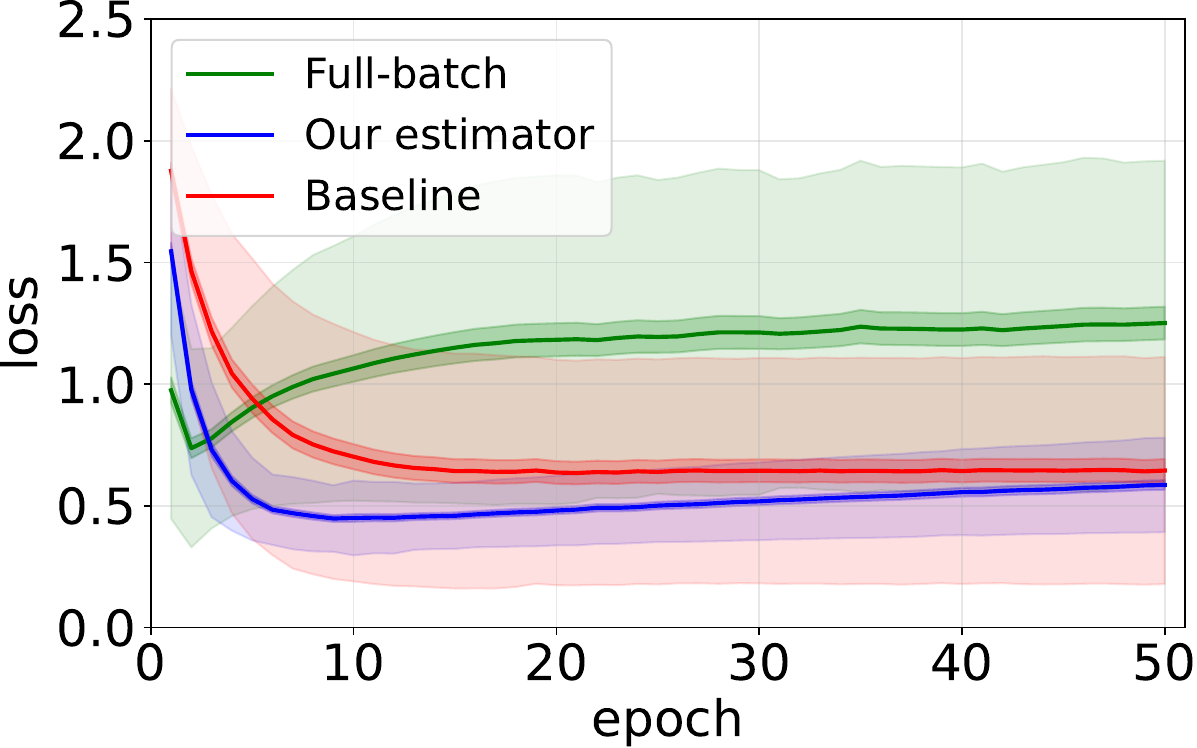} &
\includegraphics[valign=m,width=0.29\textwidth]{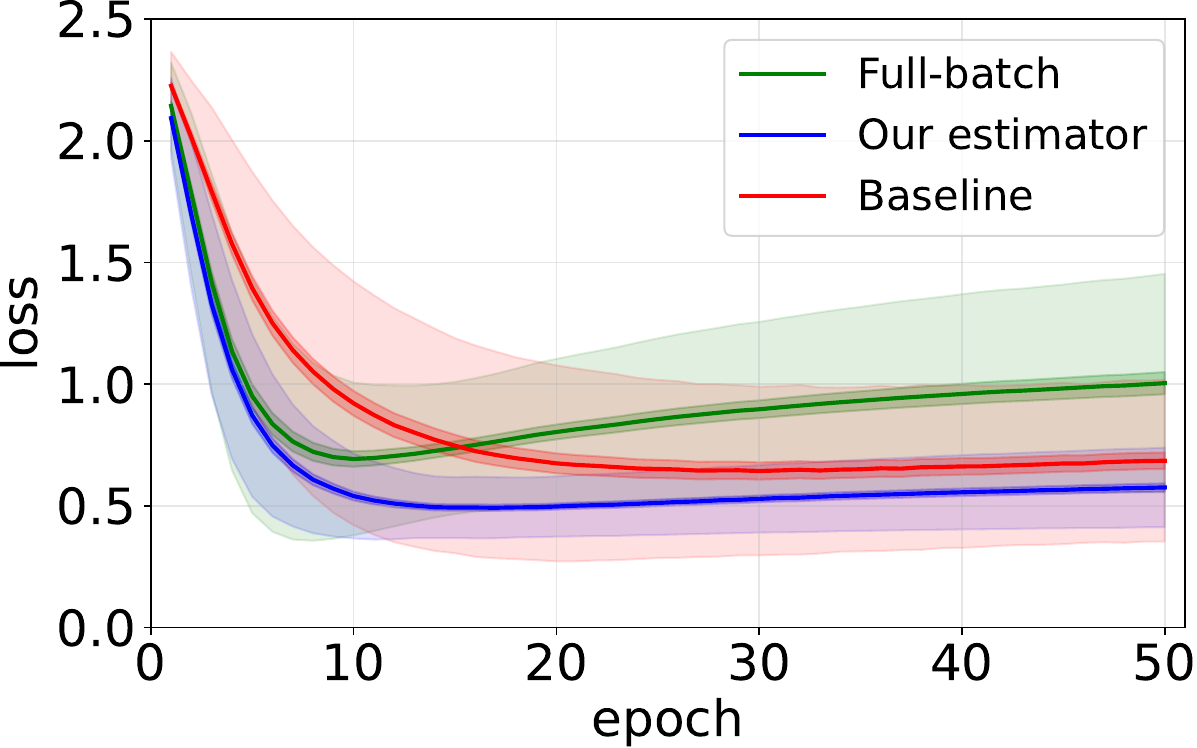} &
\includegraphics[valign=m,width=0.29\textwidth]{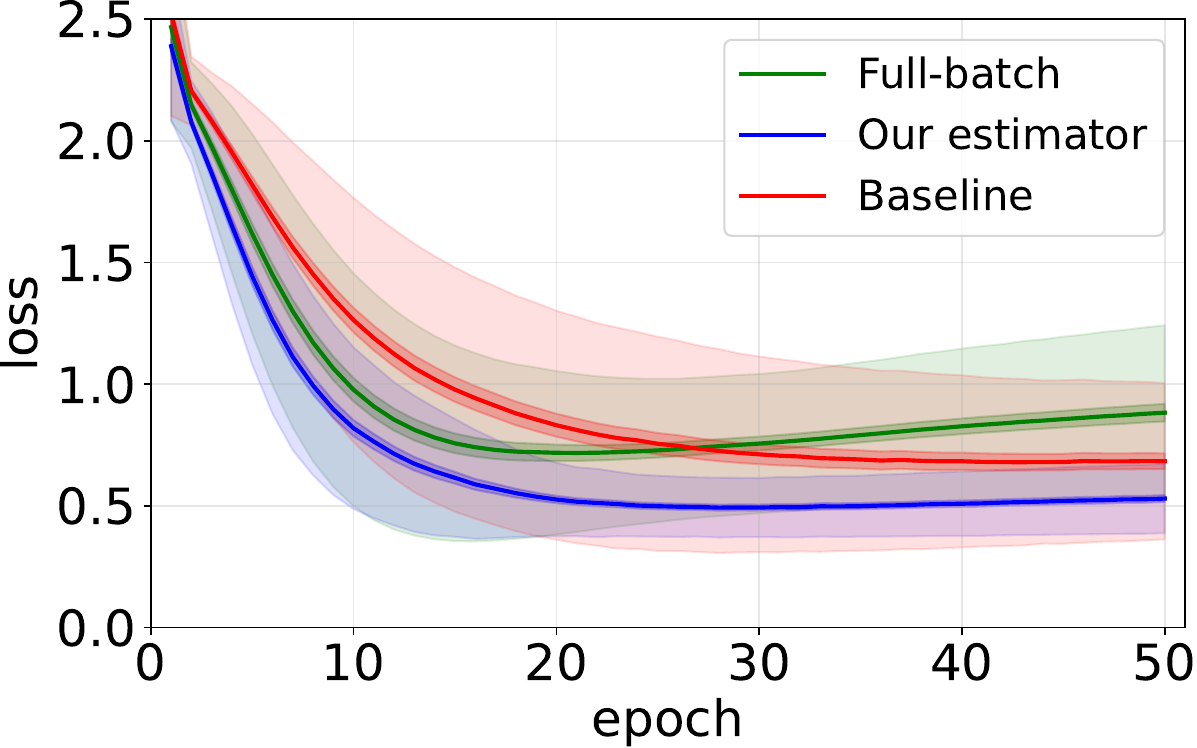} \\

\rotatebox[origin=c]{90}{\footnotesize\textbf{Fashion-MNIST}} &
\includegraphics[valign=m,width=0.29\textwidth]{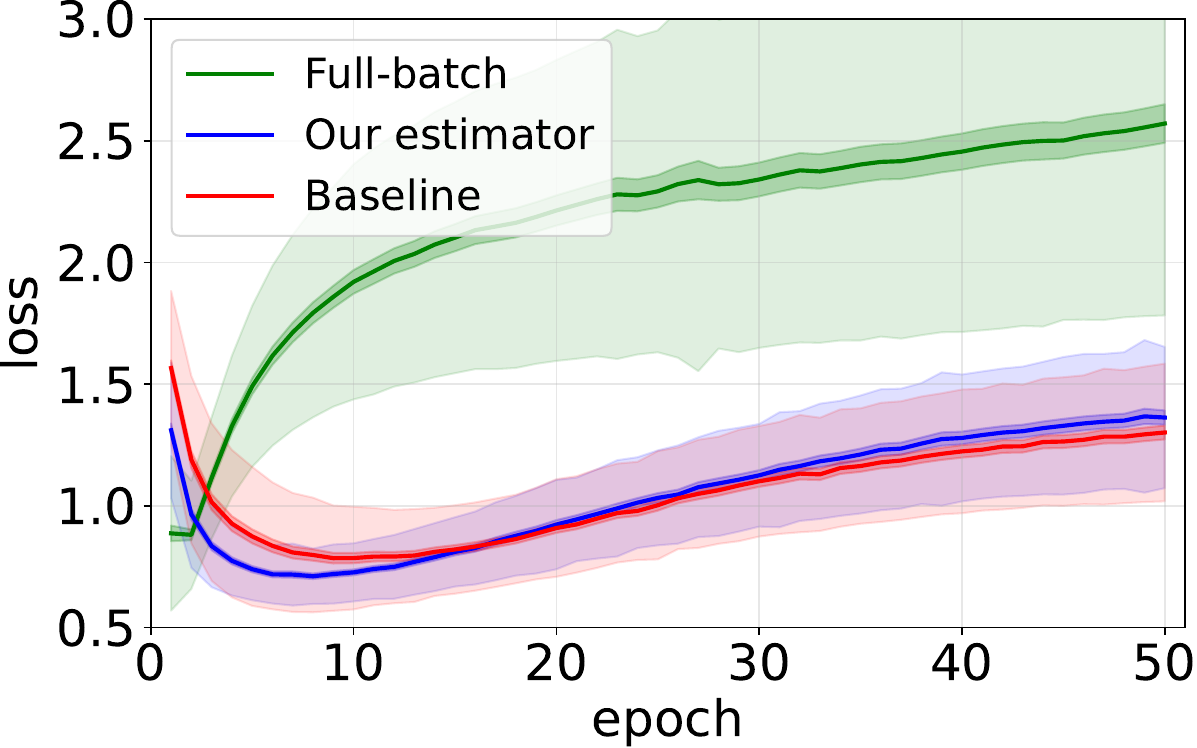} &
\includegraphics[valign=m,width=0.29\textwidth]{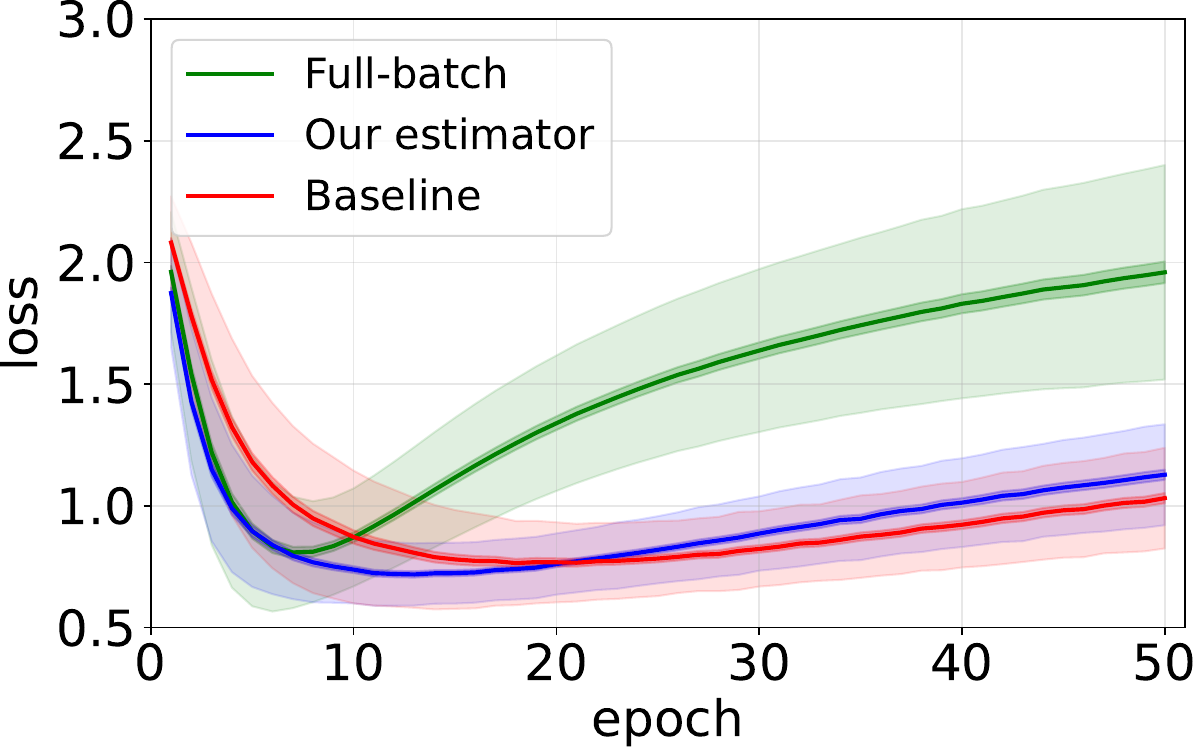} &
\includegraphics[valign=m,width=0.29\textwidth]{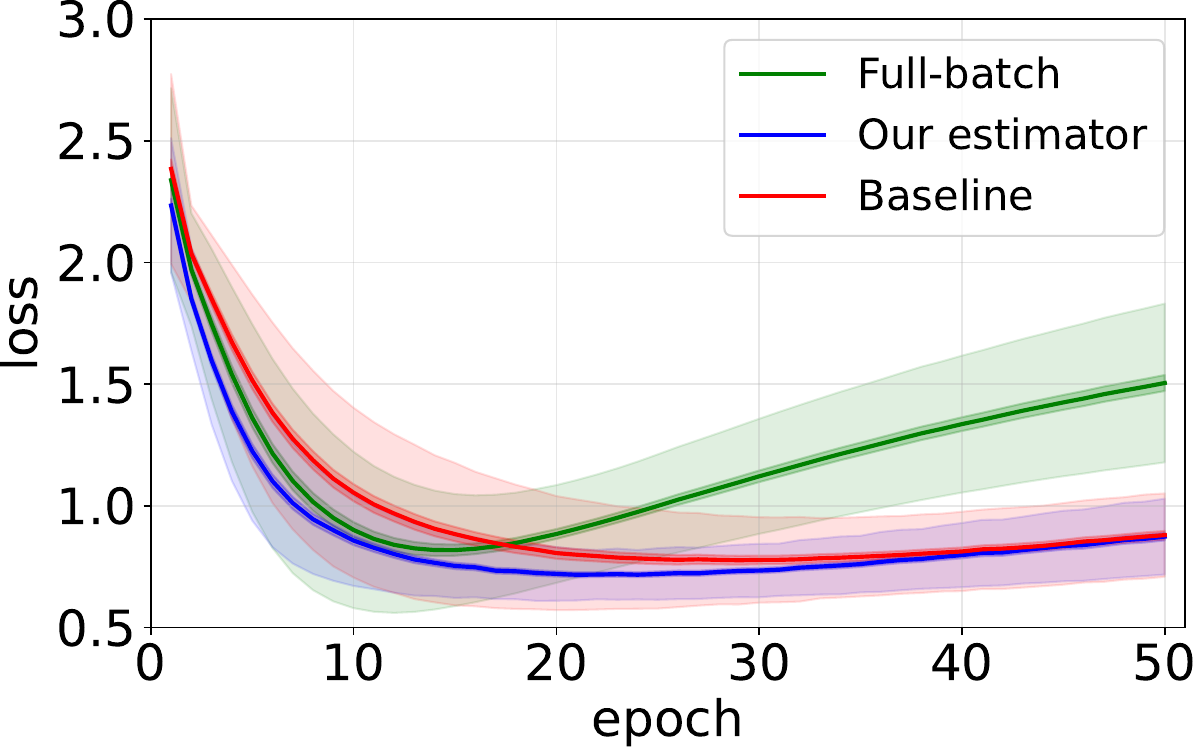} \\

\rotatebox[origin=c]{90}{\footnotesize\textbf{CIFAR-10}} &
\includegraphics[valign=m,width=0.29\textwidth]{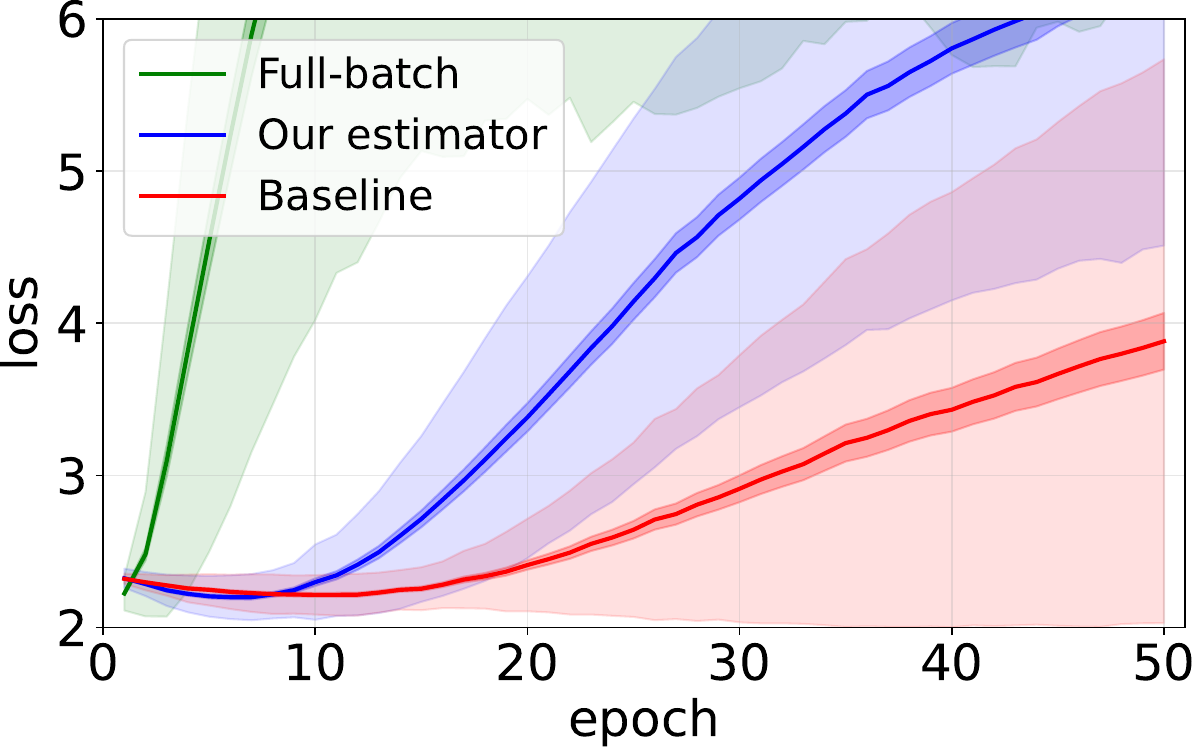} &
\includegraphics[valign=m,width=0.29\textwidth]{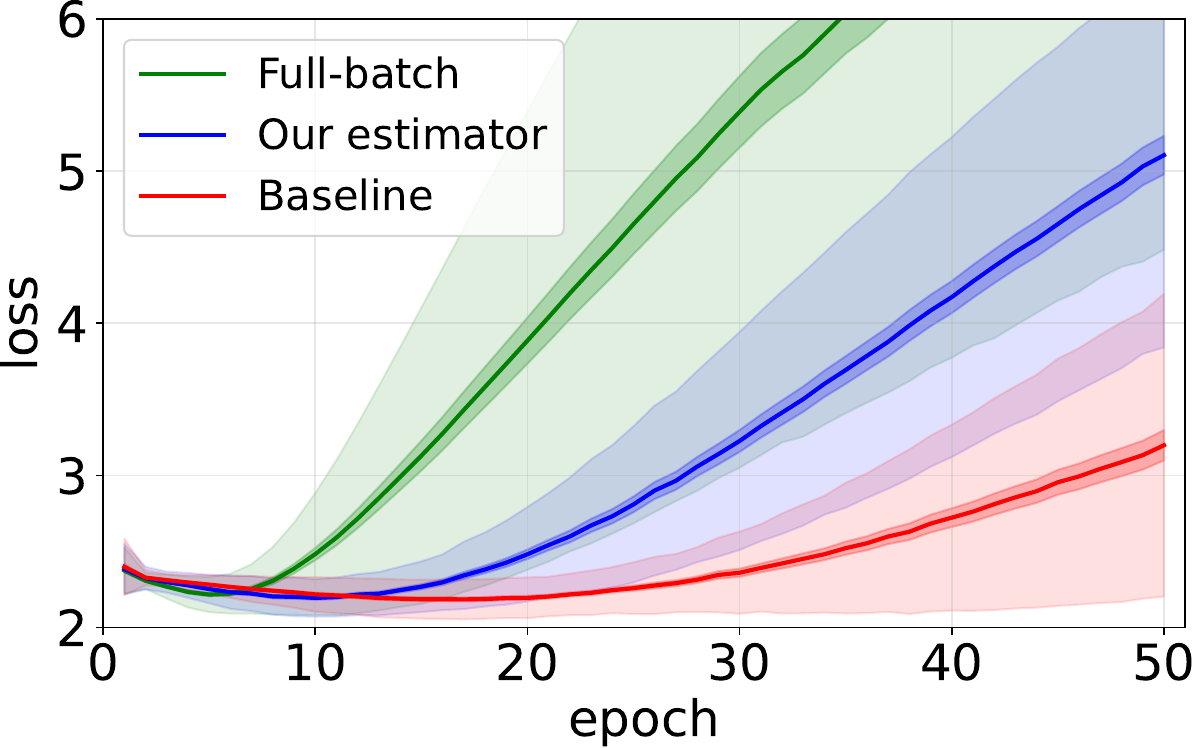} &
\includegraphics[valign=m,width=0.29\textwidth]{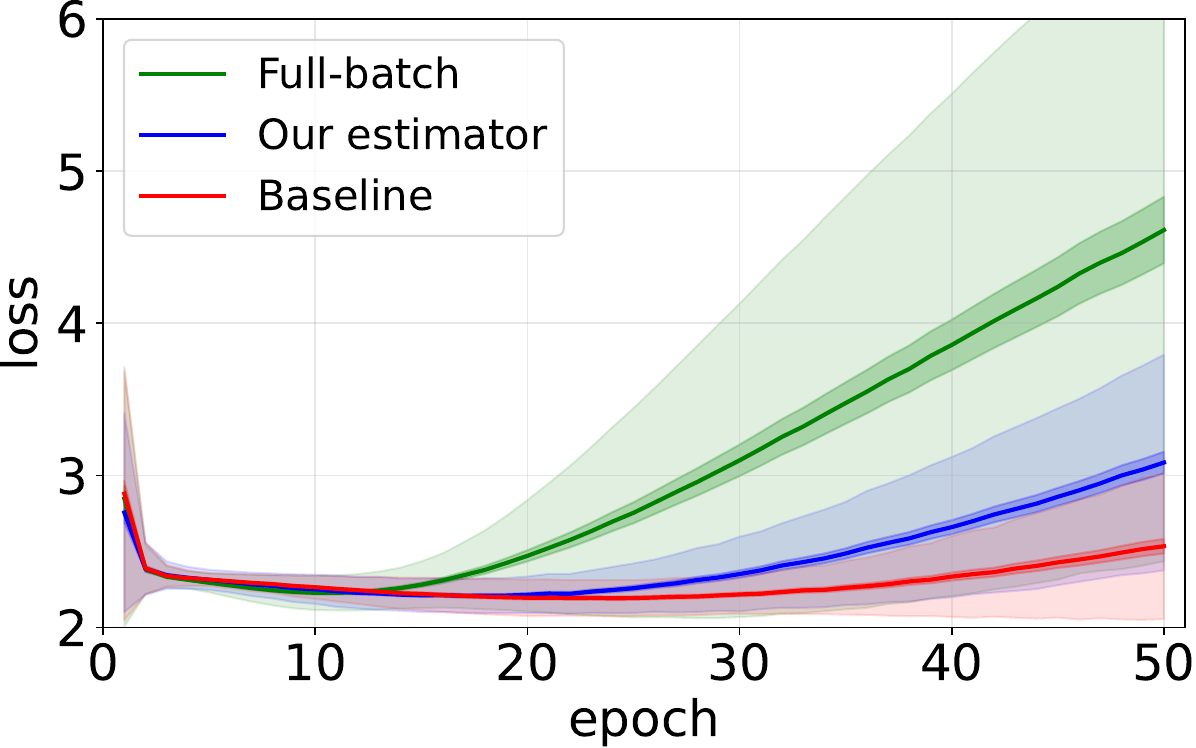} \\

\rotatebox[origin=c]{90}{\footnotesize\textbf{CIFAR-100}} &
\includegraphics[valign=m,width=0.29\textwidth]{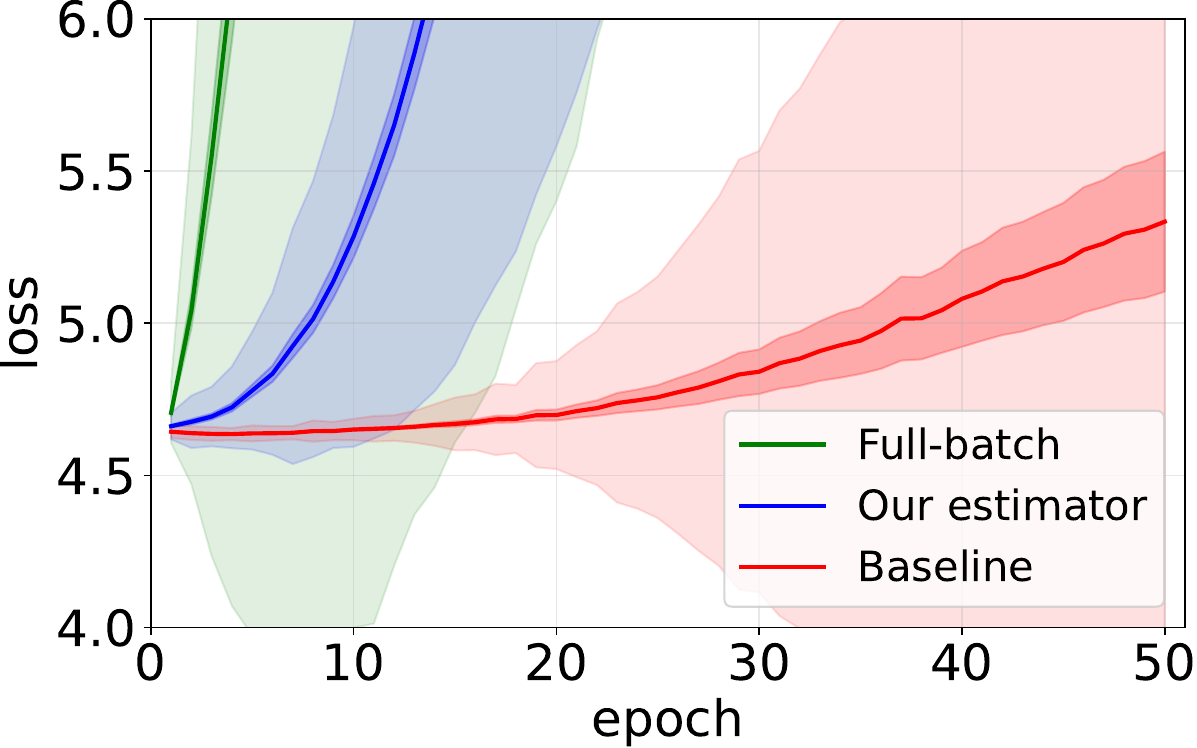} &
\includegraphics[valign=m,width=0.29\textwidth]{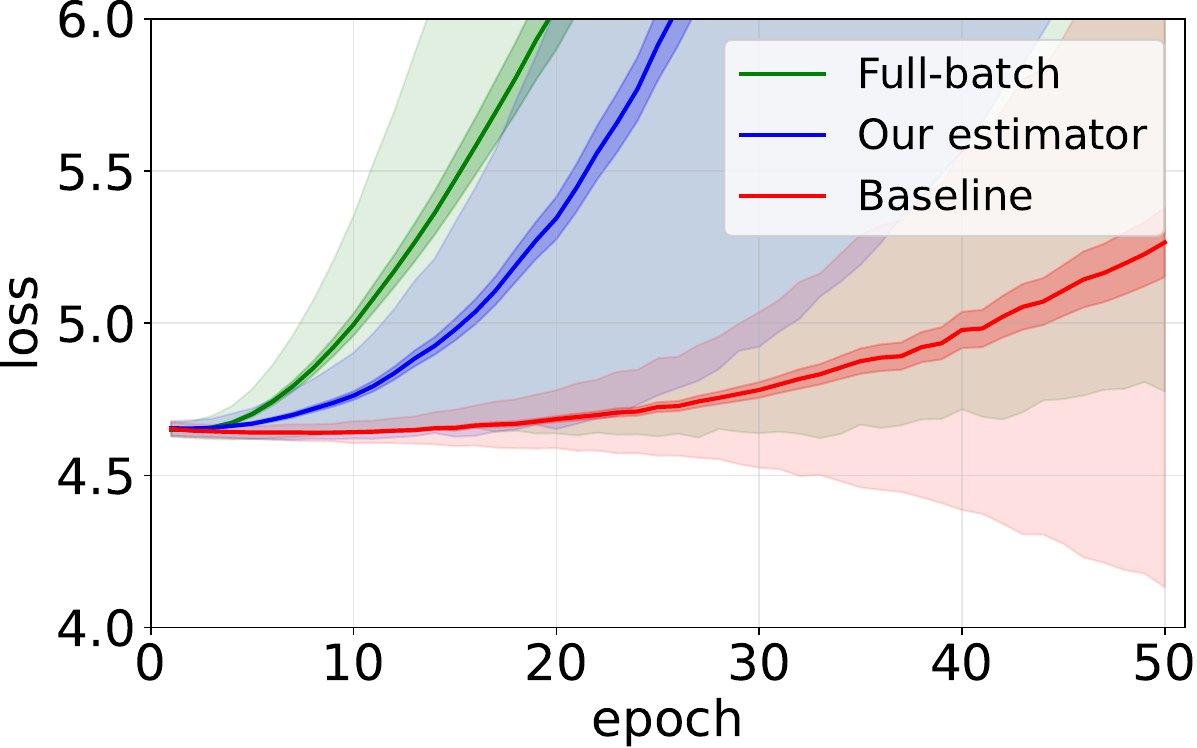} &
\includegraphics[valign=m,width=0.29\textwidth]{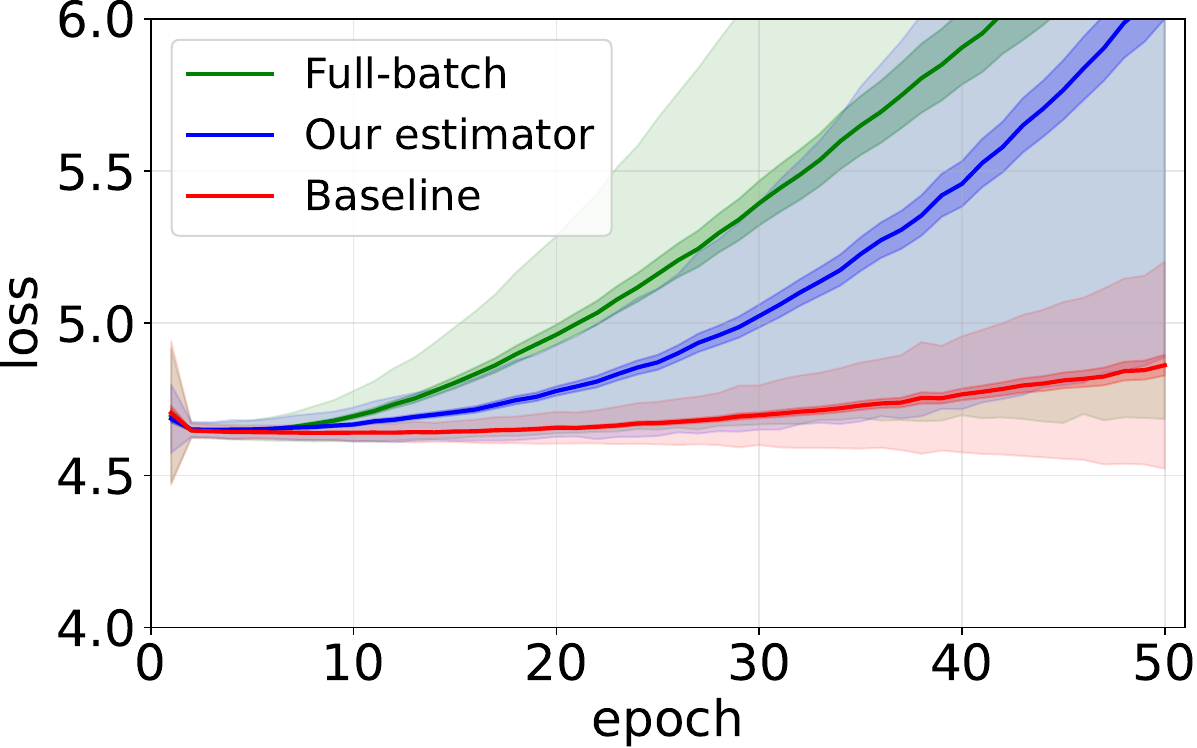} \\

\end{tabular}

\caption{The generalization performance for AdamW optimizer with baseline (uniform mini-batch), model-assisted (our estimator) and full-batch gradients. Full-batch gradient case is listed for comparison purposes. The darker curve represents the average test loss of 400 runs, the dark shading is the 95\% confidence interval and lighter shading shows the standard deviation. Columns correspond to batch sizes and rows to datasets.}
\label{fig:AdamW_loss_grid}
\end{figure*}

\begin{figure*}[t]
\centering
\setlength{\tabcolsep}{3pt}
\renewcommand{\arraystretch}{0.8}

\begin{tabular}{>{\centering\arraybackslash}m{0.04\textwidth}
                >{\centering\arraybackslash}m{0.29\textwidth}
                >{\centering\arraybackslash}m{0.29\textwidth}
                >{\centering\arraybackslash}m{0.29\textwidth}}

& \textbf{Batch size 10} & \textbf{Batch size 50} & \textbf{Batch size 100} \\

\rotatebox[origin=c]{90}{\footnotesize\textbf{Synthetic}} &
\includegraphics[valign=m,width=0.29\textwidth]{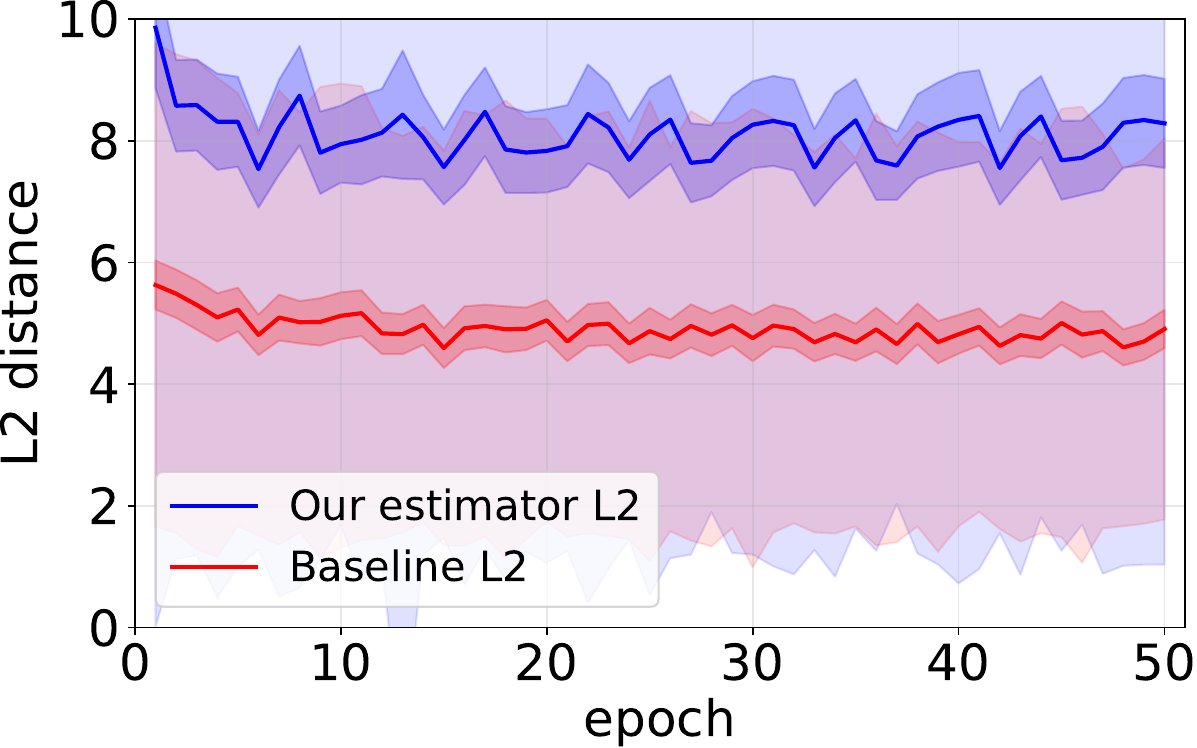} &
\includegraphics[valign=m,width=0.29\textwidth]{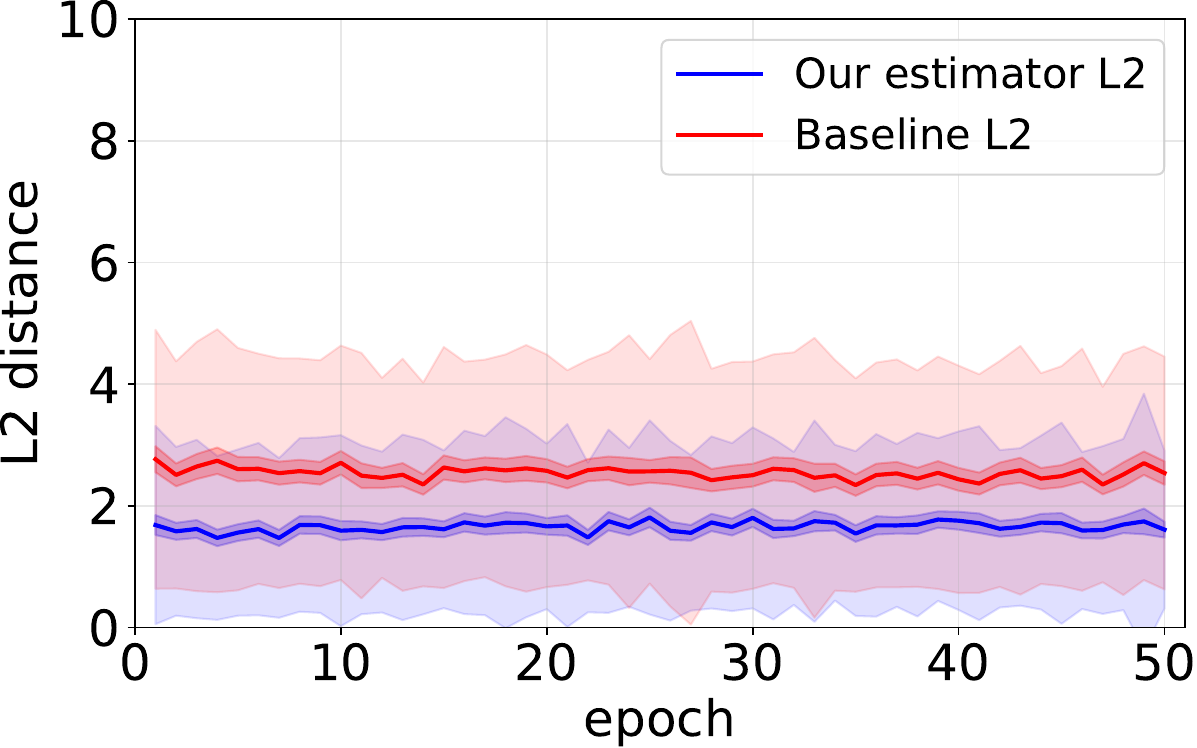} &
\includegraphics[valign=m,width=0.29\textwidth]{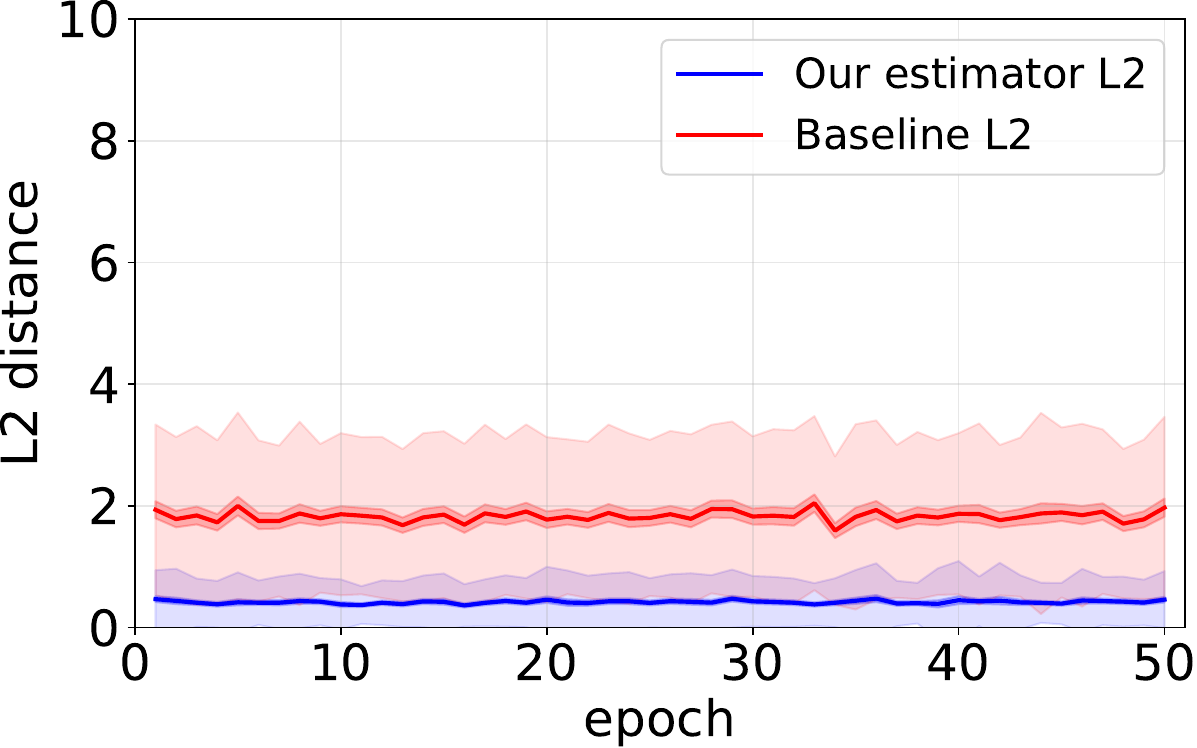} \\

\rotatebox[origin=c]{90}{\footnotesize\textbf{Airfoil self-noise}} &
\includegraphics[valign=m,width=0.29\textwidth]{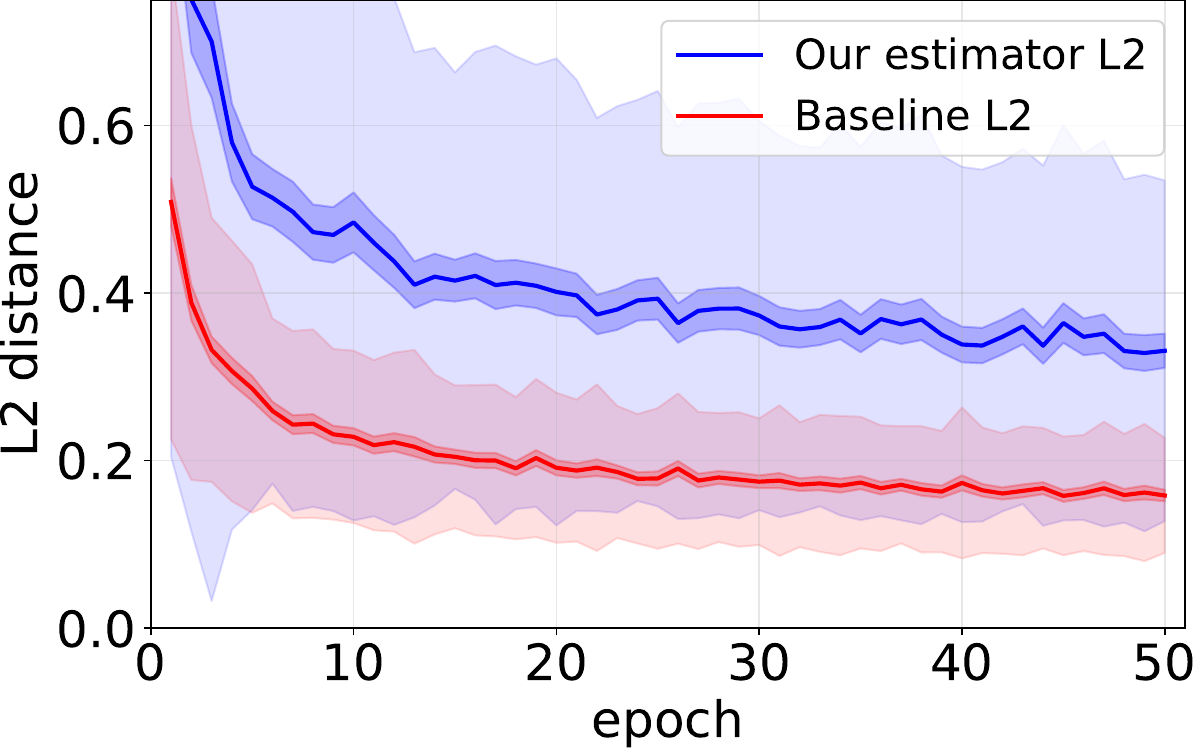} &
\includegraphics[valign=m,width=0.29\textwidth]{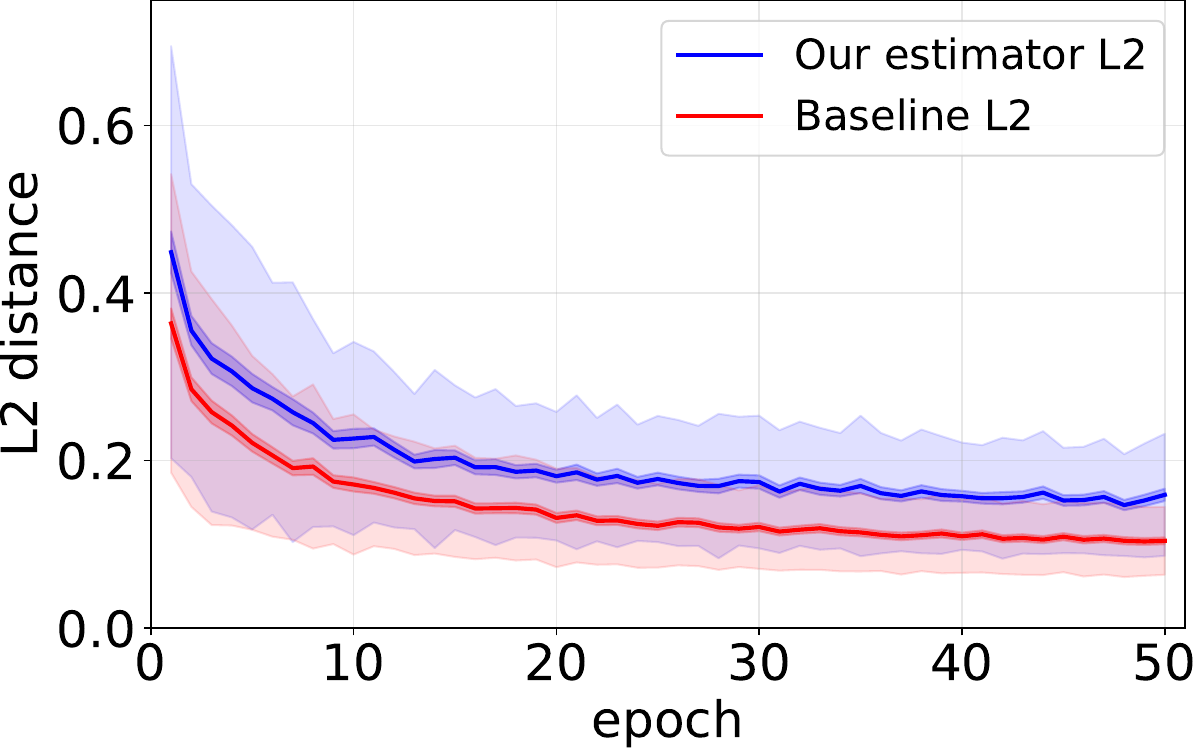} &
\includegraphics[valign=m,width=0.29\textwidth]{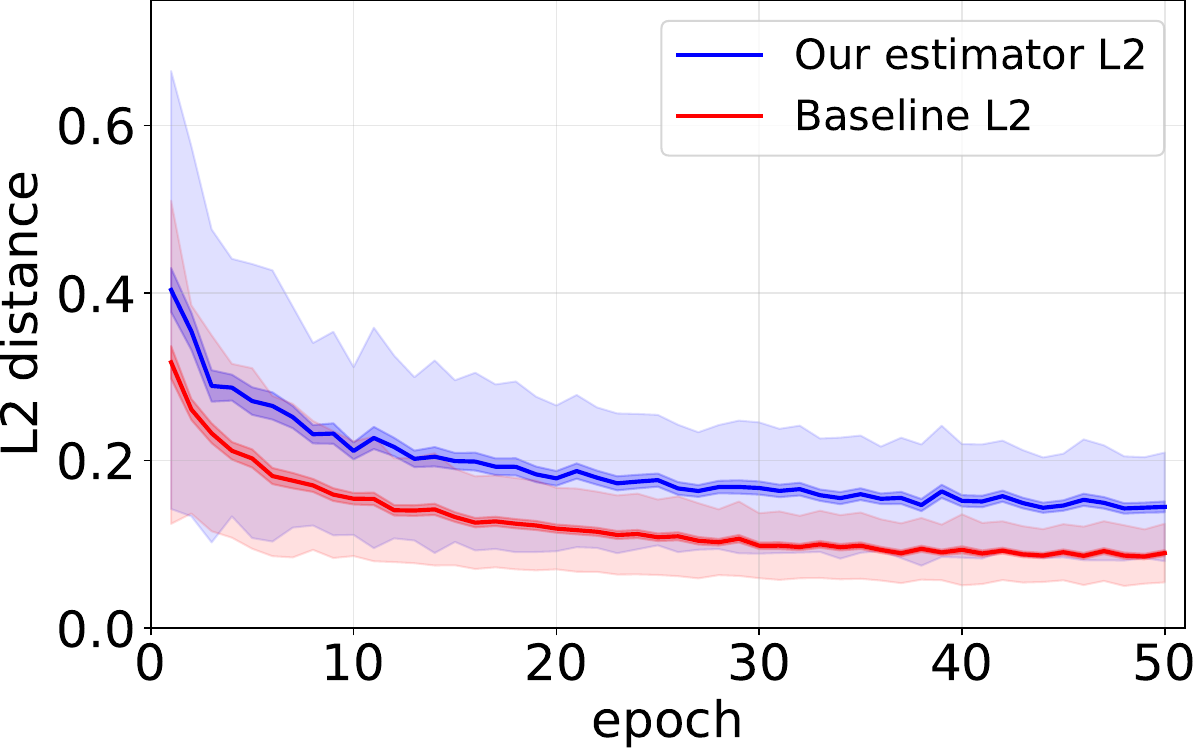} \\

\rotatebox[origin=c]{90}{\footnotesize\textbf{Appliances energy}} &
\includegraphics[valign=m,width=0.29\textwidth]{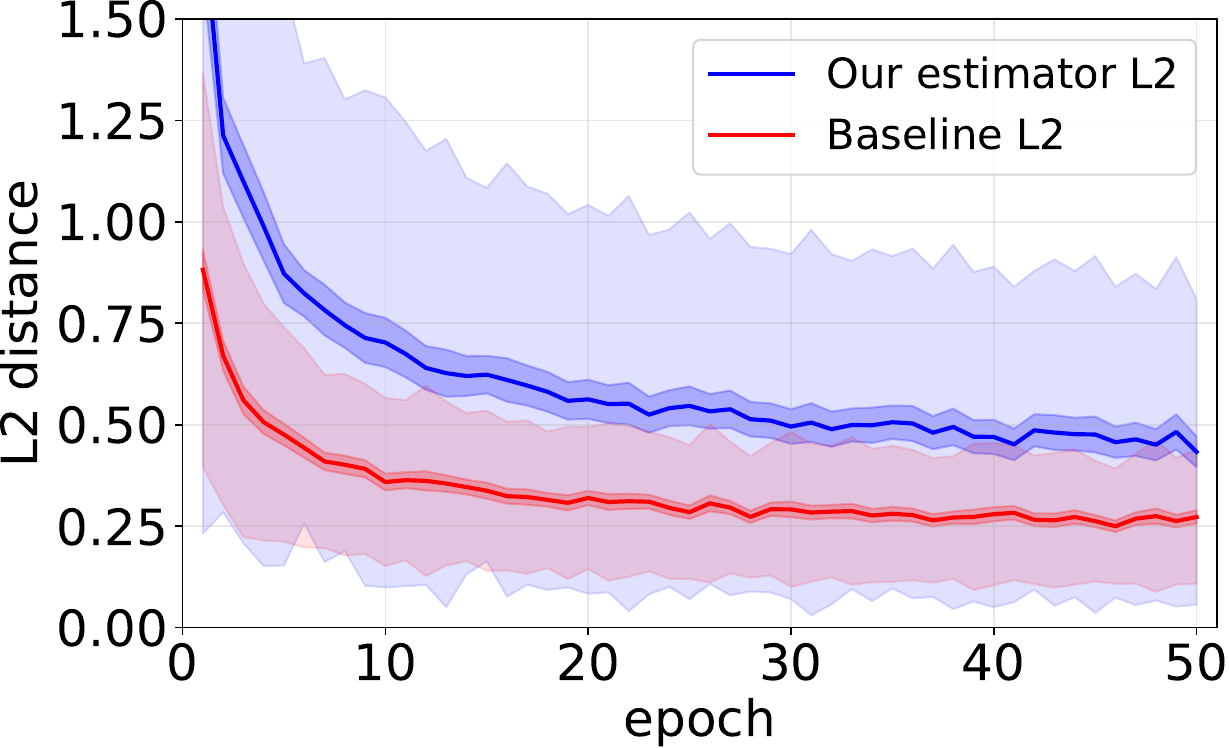} &
\includegraphics[valign=m,width=0.29\textwidth]{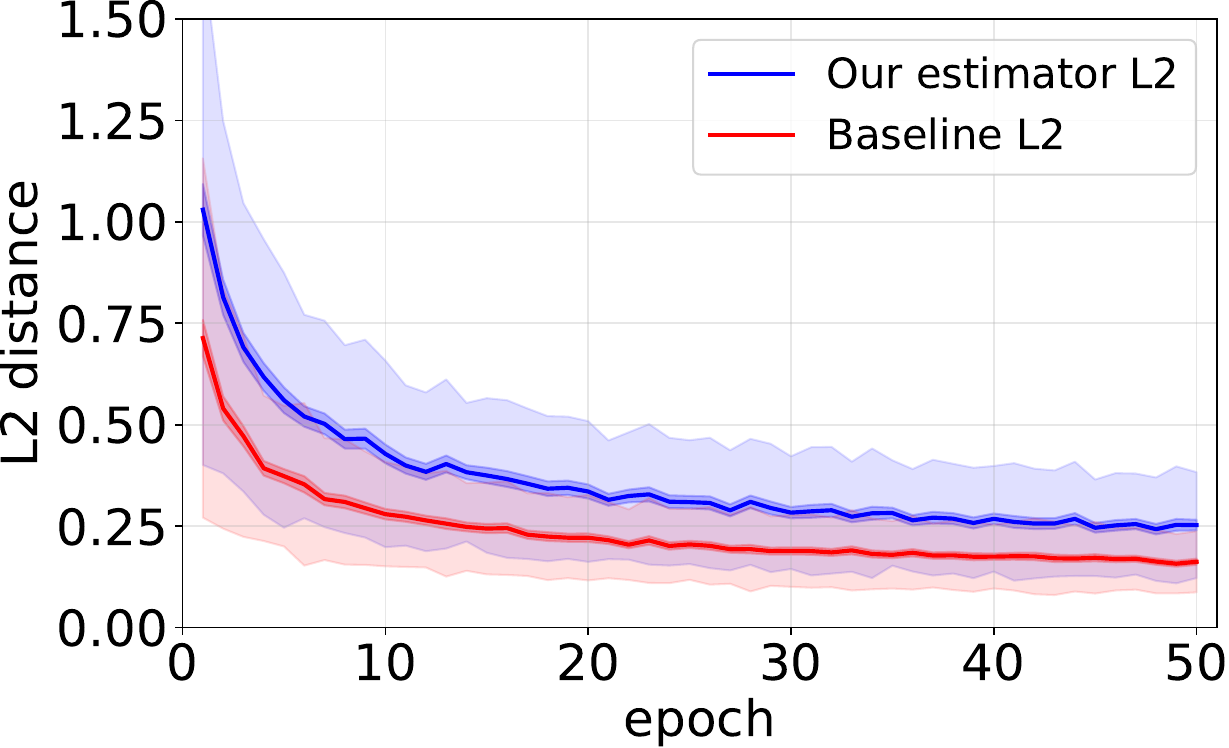} &
\includegraphics[valign=m,width=0.29\textwidth]{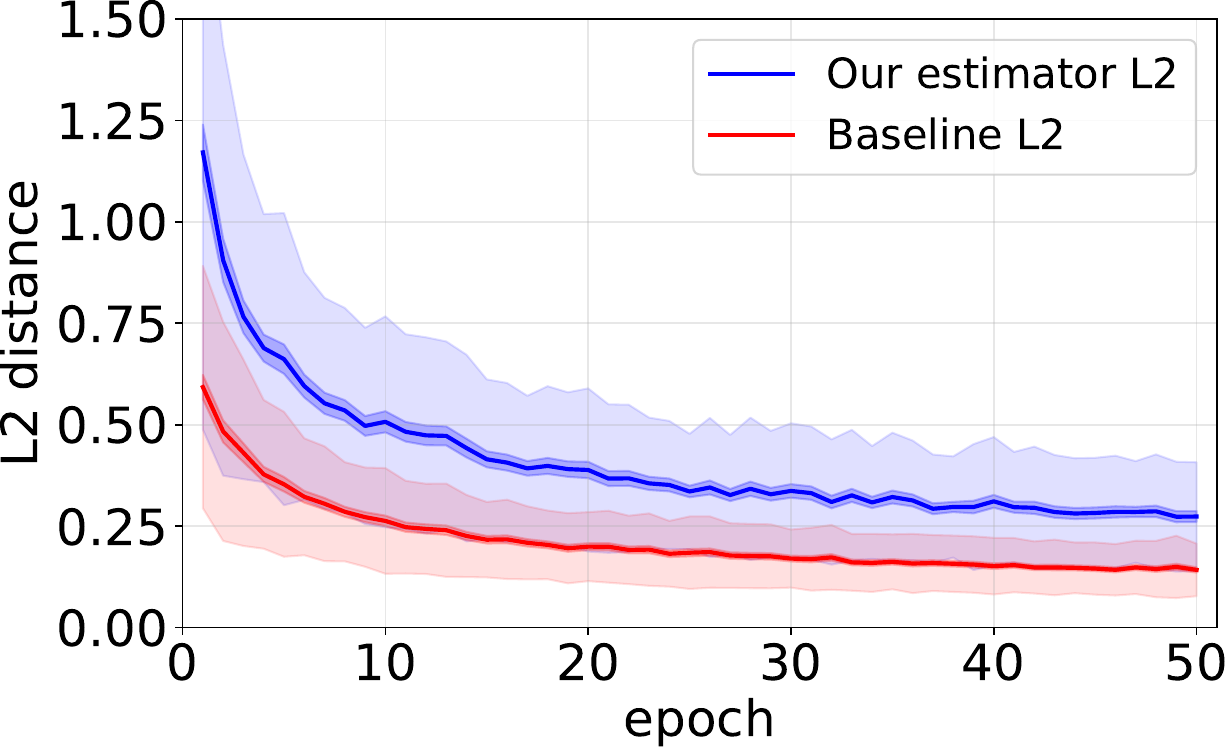} \\

\rotatebox[origin=c]{90}{\footnotesize\textbf{MNIST}} &
\includegraphics[valign=m,width=0.29\textwidth]{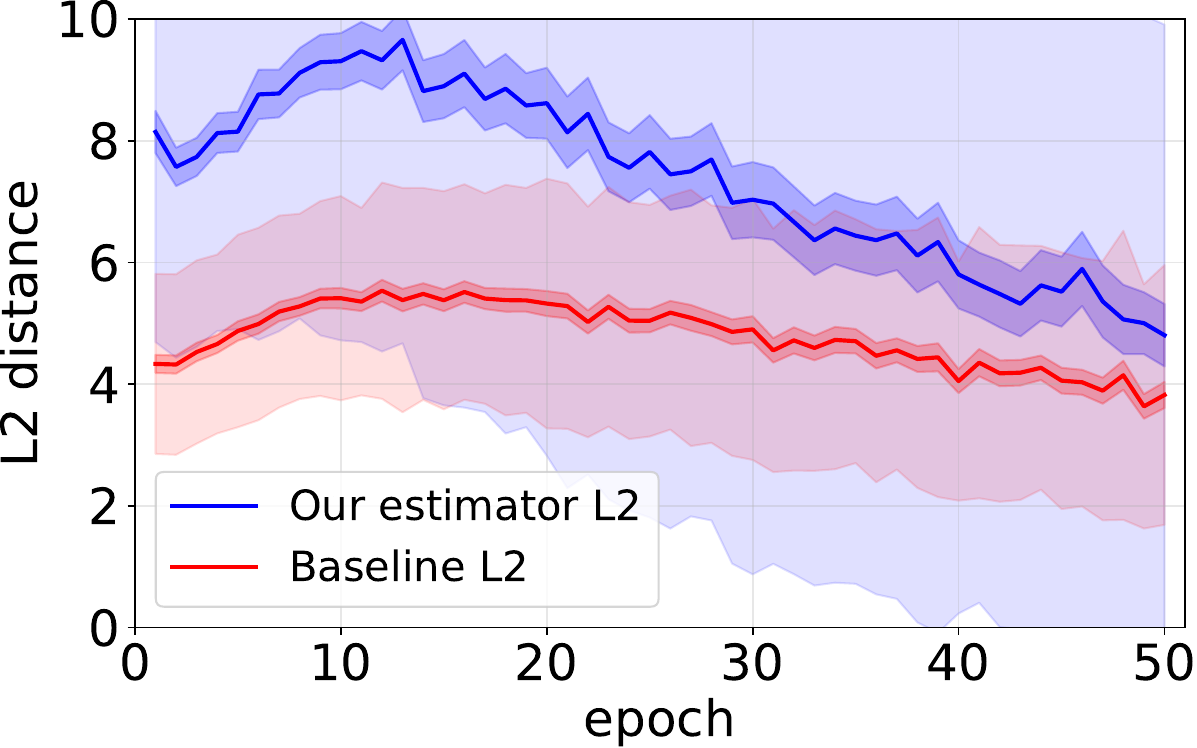} &
\includegraphics[valign=m,width=0.29\textwidth]{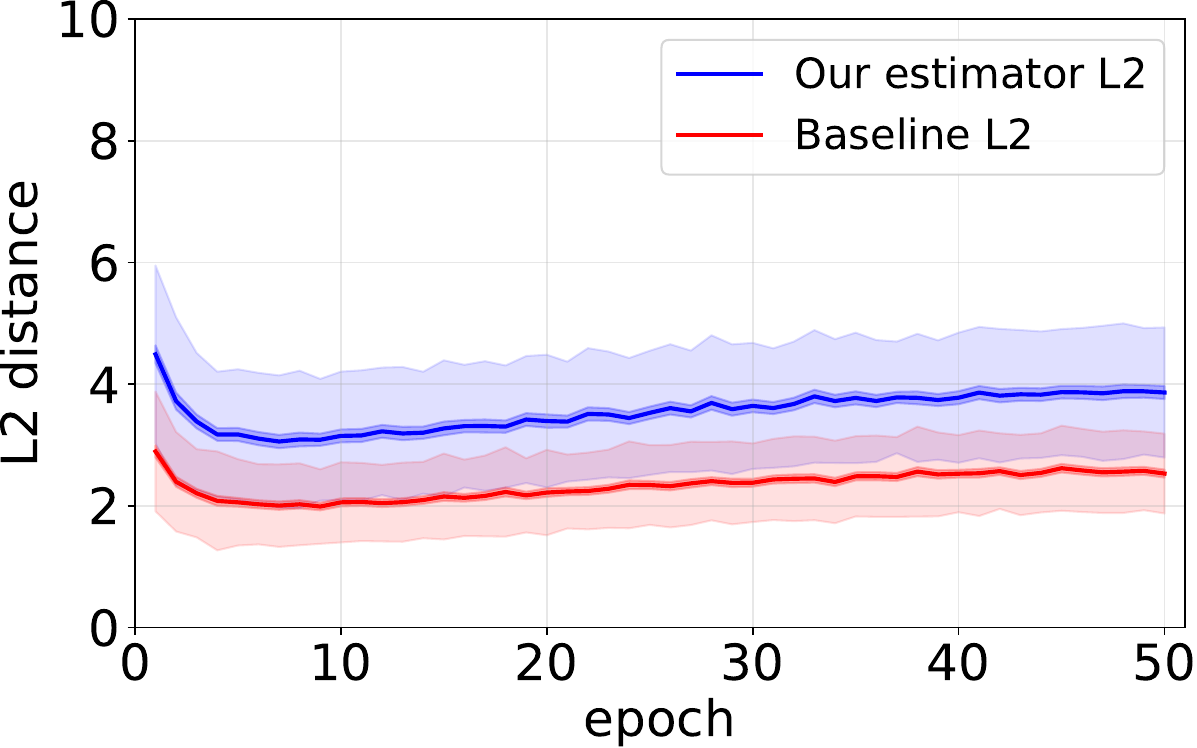} &
\includegraphics[valign=m,width=0.29\textwidth]{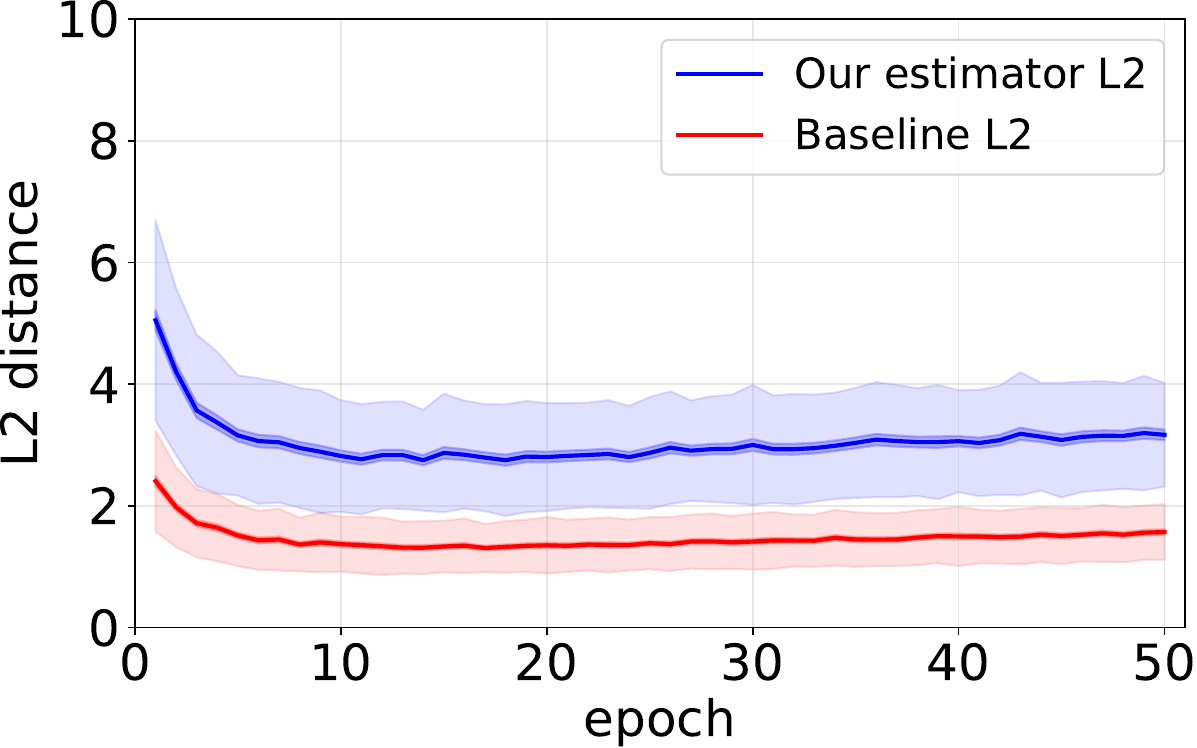} \\

\rotatebox[origin=c]{90}{\footnotesize\textbf{Fashion-MNIST}} &
\includegraphics[valign=m,width=0.29\textwidth]{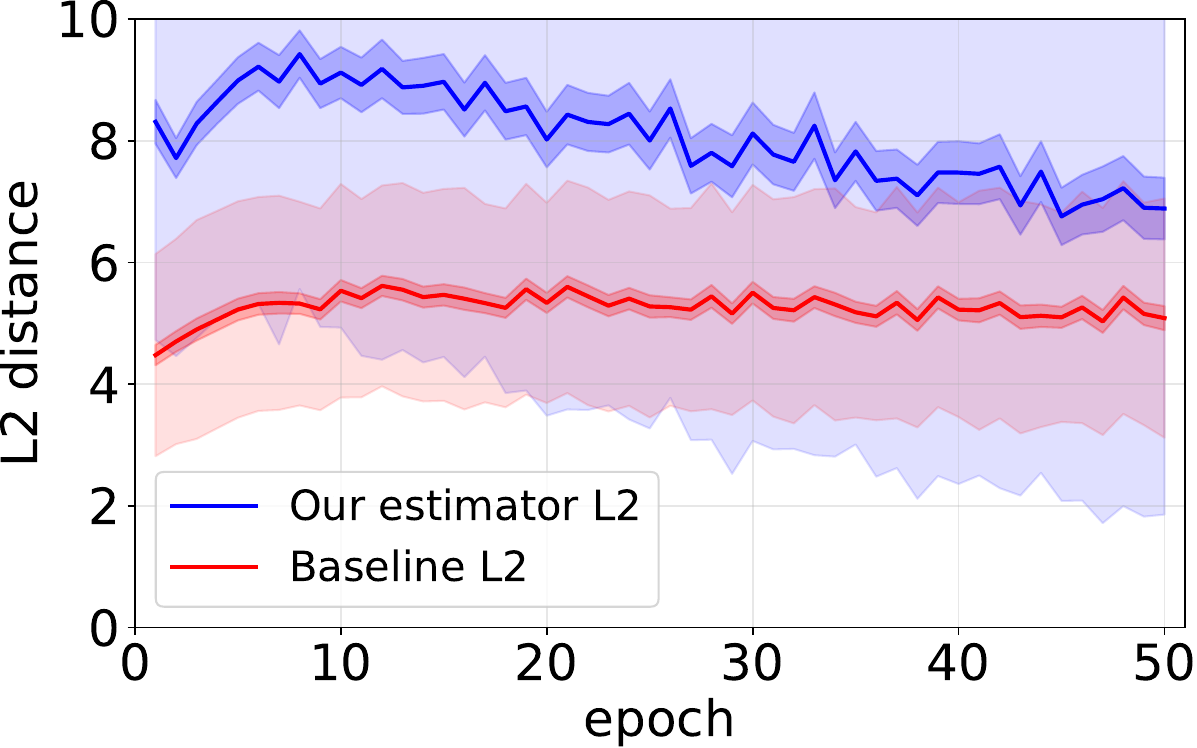} &
\includegraphics[valign=m,width=0.29\textwidth]{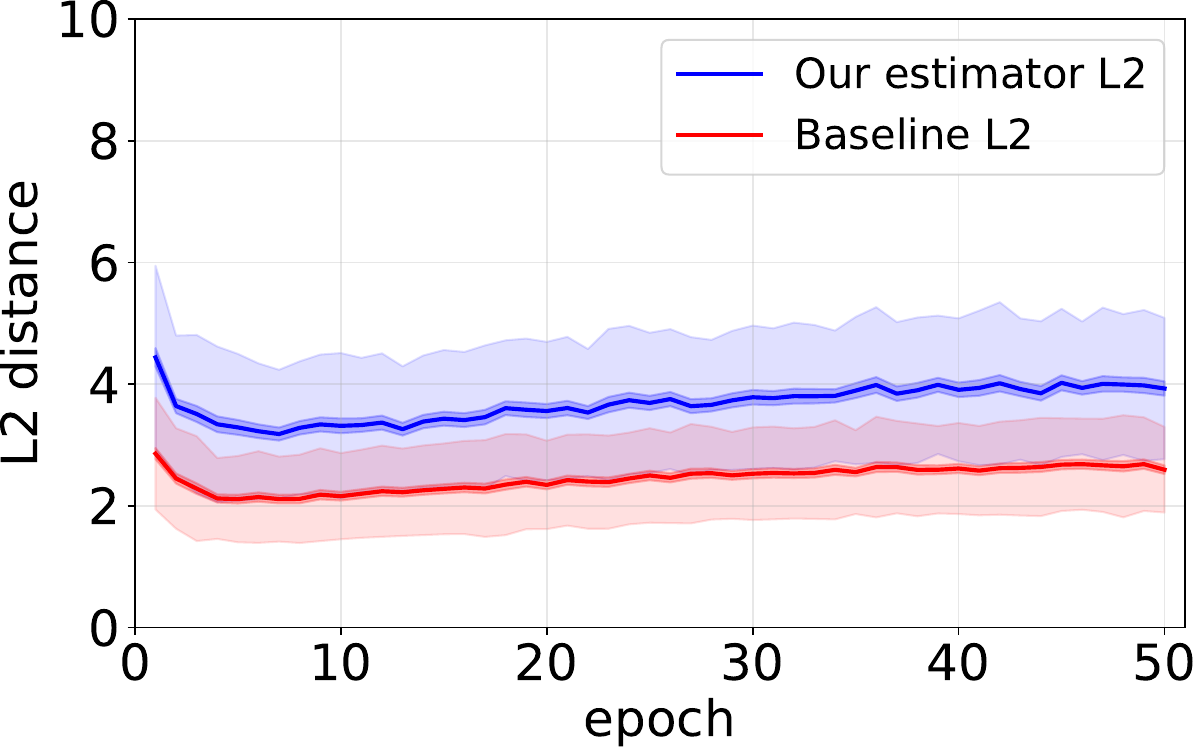} &
\includegraphics[valign=m,width=0.29\textwidth]{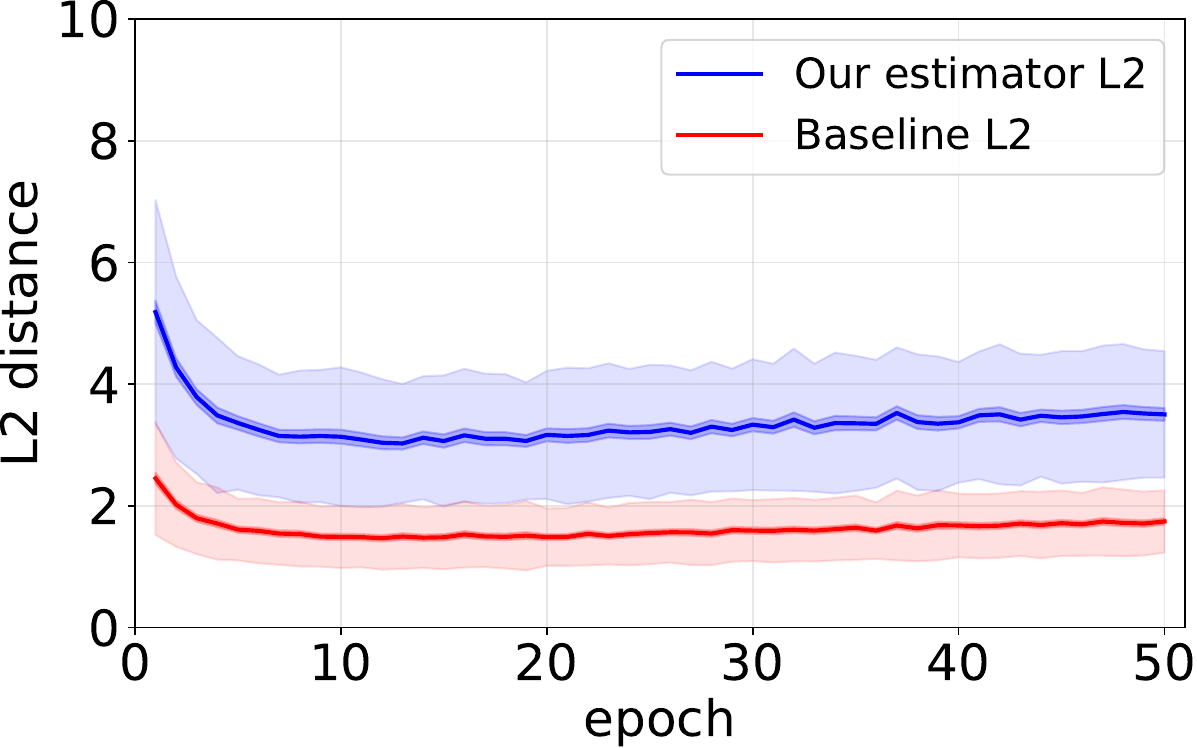} \\

\rotatebox[origin=c]{90}{\footnotesize\textbf{CIFAR-10}} &
\includegraphics[valign=m,width=0.29\textwidth]{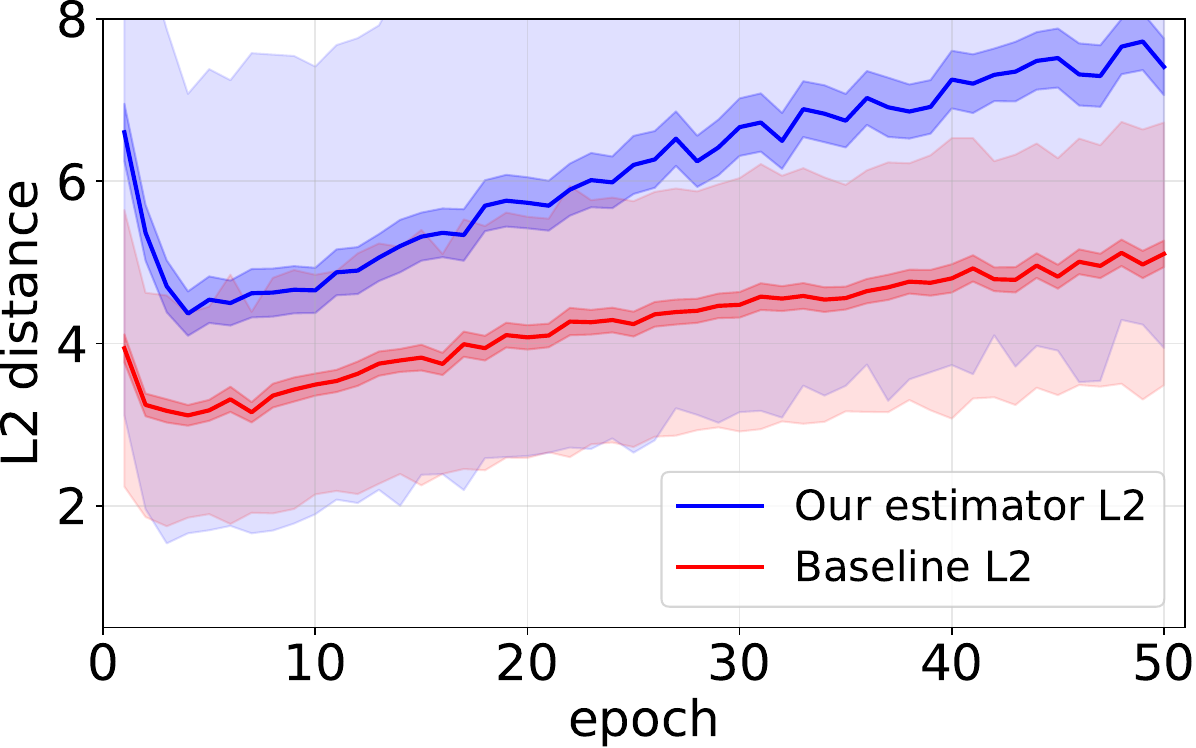} &
\includegraphics[valign=m,width=0.29\textwidth]{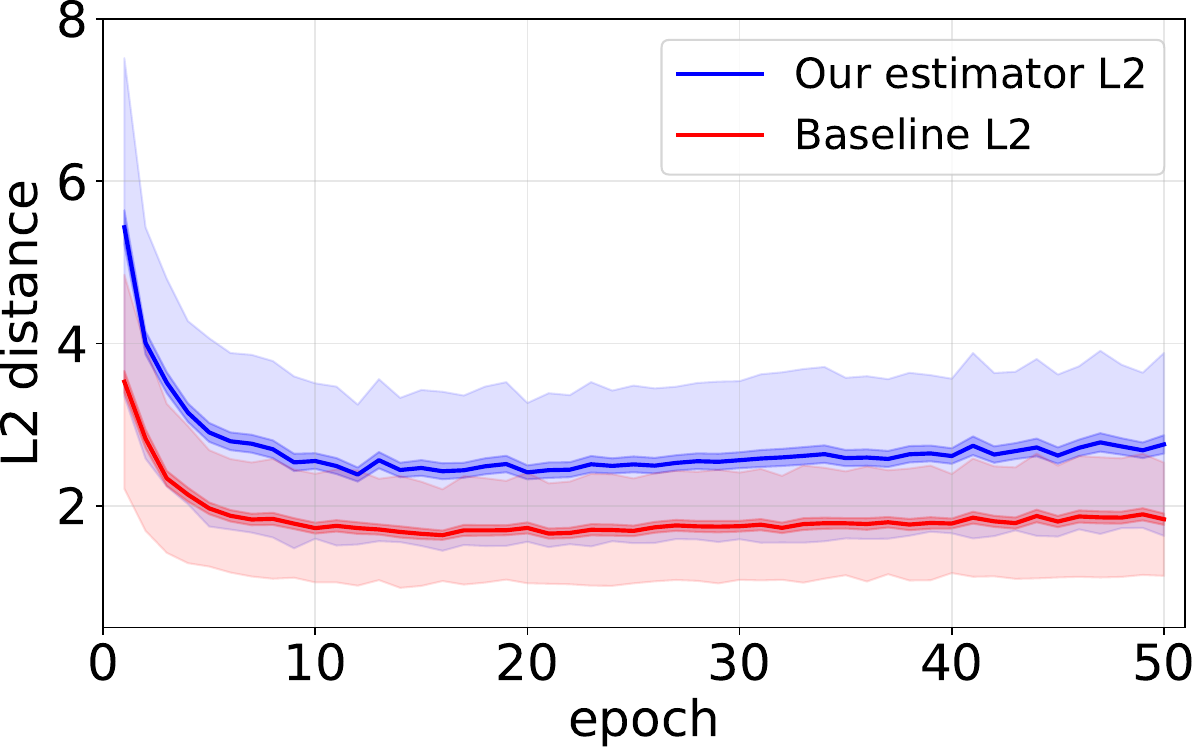} &
\includegraphics[valign=m,width=0.29\textwidth]{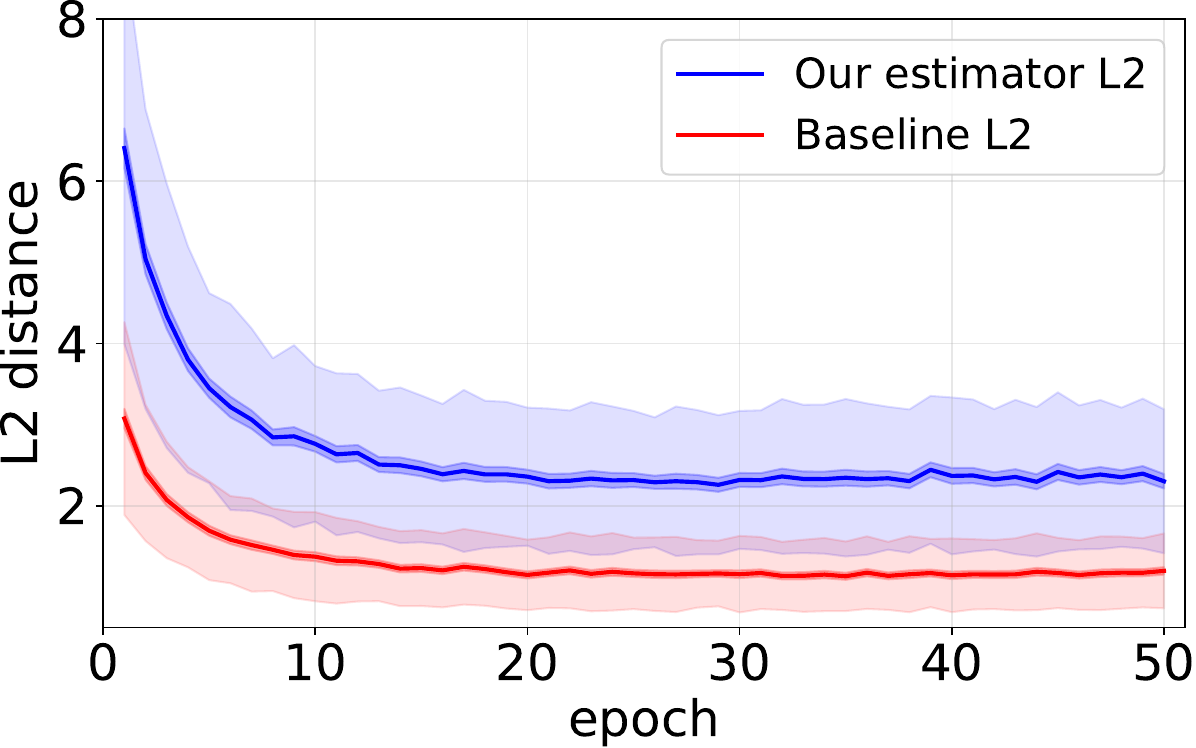} \\

\rotatebox[origin=c]{90}{\footnotesize\textbf{CIFAR-100}} &
\includegraphics[valign=m,width=0.29\textwidth]{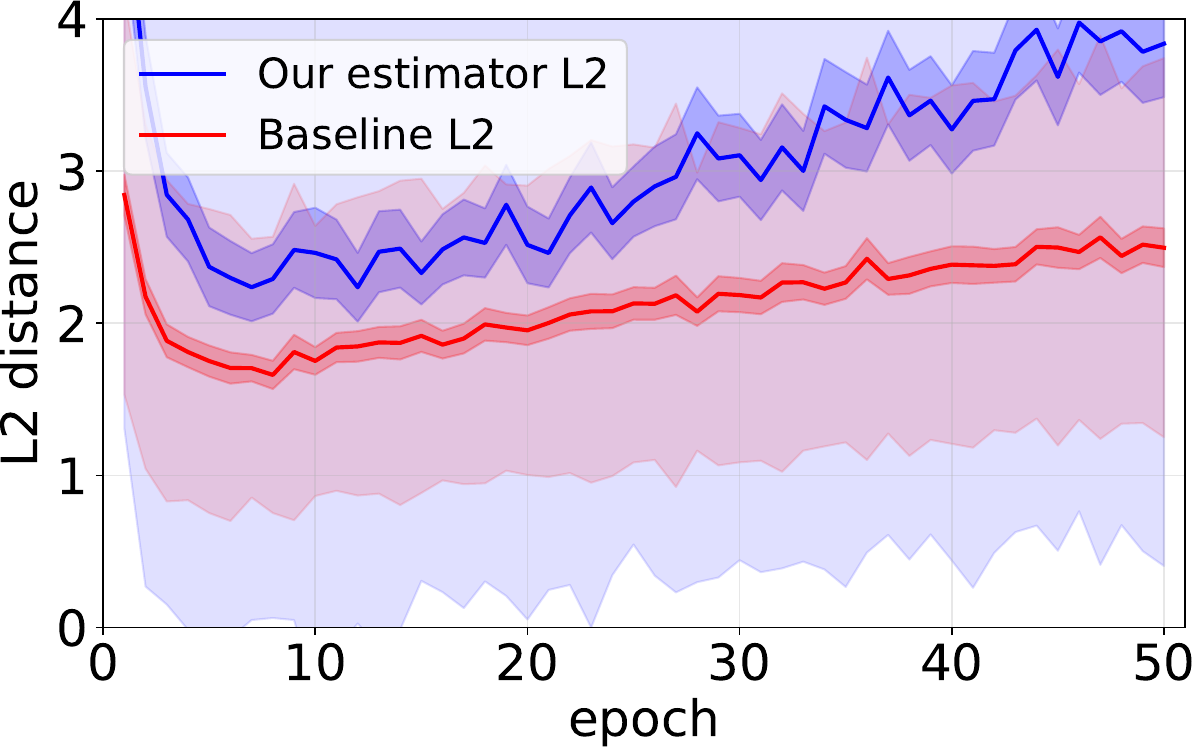} &
\includegraphics[valign=m,width=0.29\textwidth]{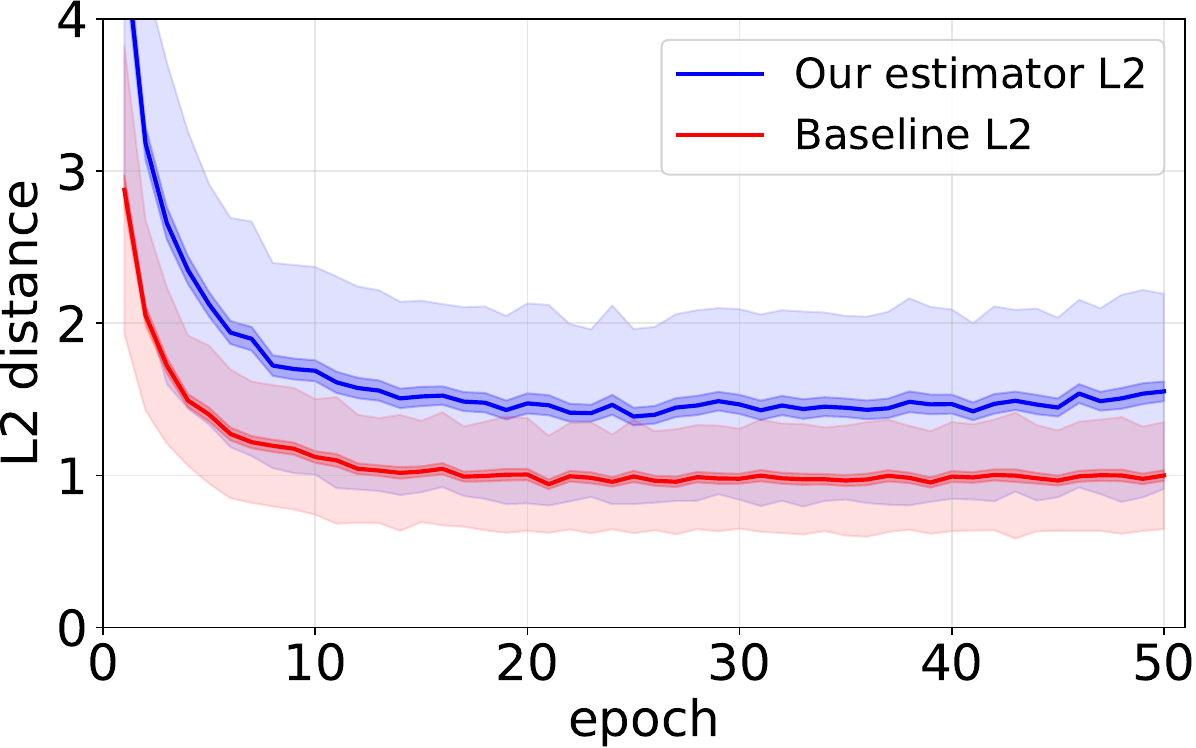} &
\includegraphics[valign=m,width=0.29\textwidth]{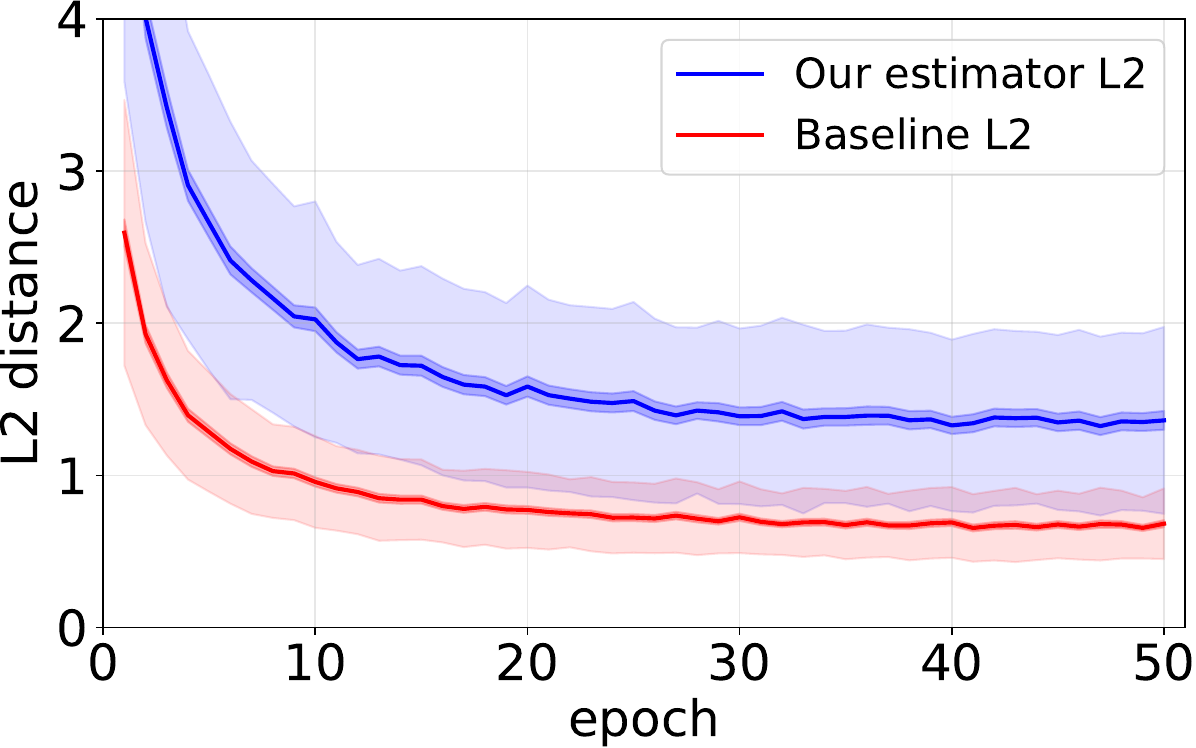} \\

\end{tabular}

\caption{L2 distance to full-batch gradient with SGD optimizer. The darker curve represents the average of 400 runs, the dark shading is the 95\% confidence interval and lighter shading shows the standard deviation. Columns correspond to batch sizes and rows to datasets.}
\label{fig:sgd_l2_grid}
\end{figure*}

\begin{figure*}[t]
\centering
\setlength{\tabcolsep}{3pt}
\renewcommand{\arraystretch}{0.8}

\begin{tabular}{>{\centering\arraybackslash}m{0.04\textwidth}
                >{\centering\arraybackslash}m{0.29\textwidth}
                >{\centering\arraybackslash}m{0.29\textwidth}
                >{\centering\arraybackslash}m{0.29\textwidth}}

& \textbf{Batch size 10} & \textbf{Batch size 50} & \textbf{Batch size 100} \\

\rotatebox[origin=c]{90}{\footnotesize\textbf{Synthetic}} &
\includegraphics[valign=m,width=0.29\textwidth]{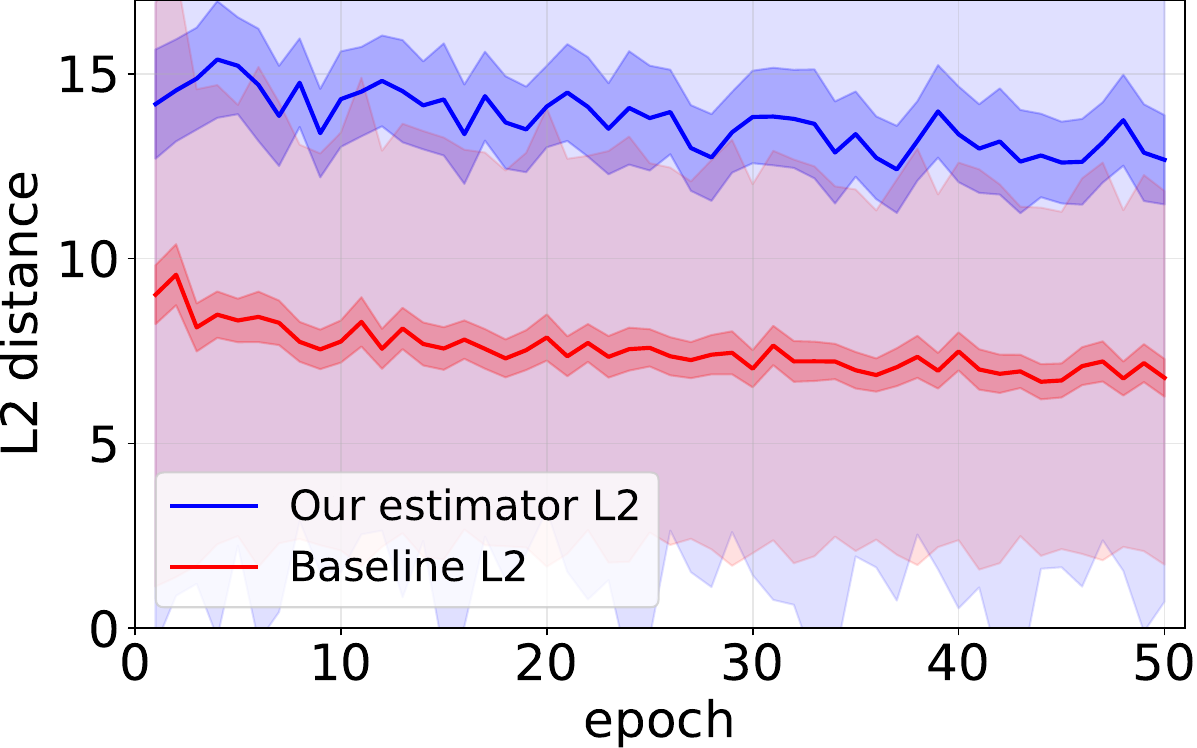} &
\includegraphics[valign=m,width=0.29\textwidth]{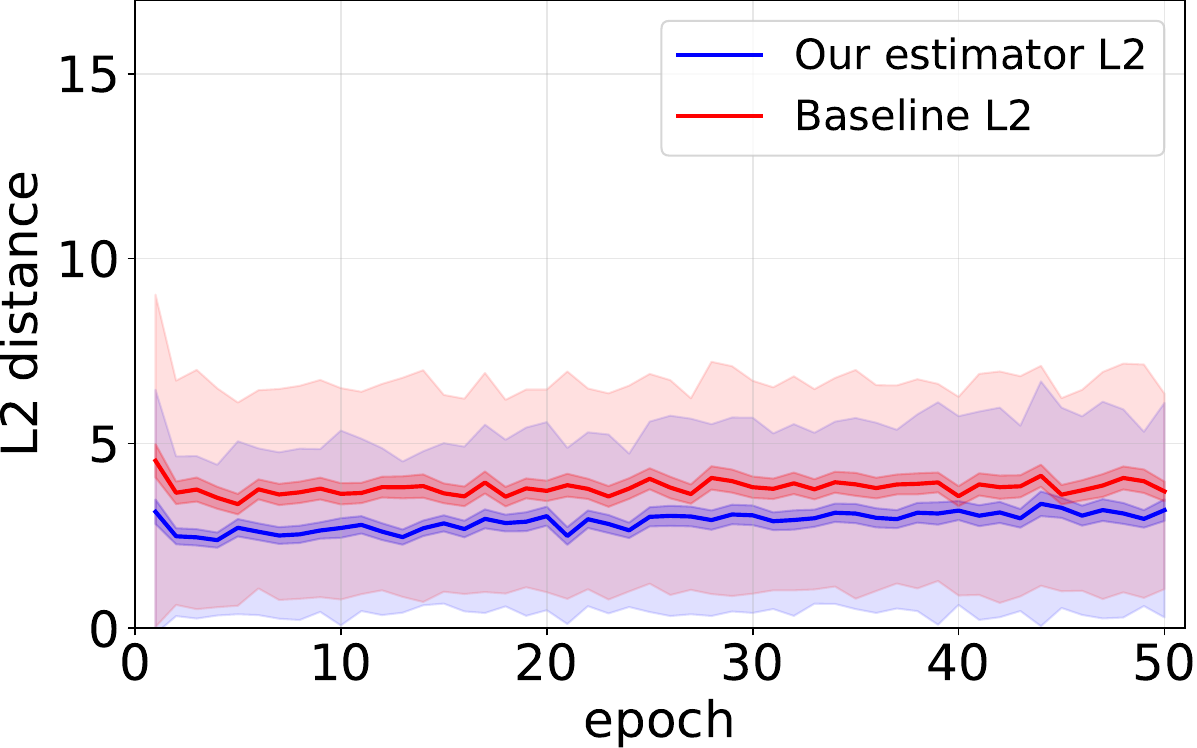} &
\includegraphics[valign=m,width=0.29\textwidth]{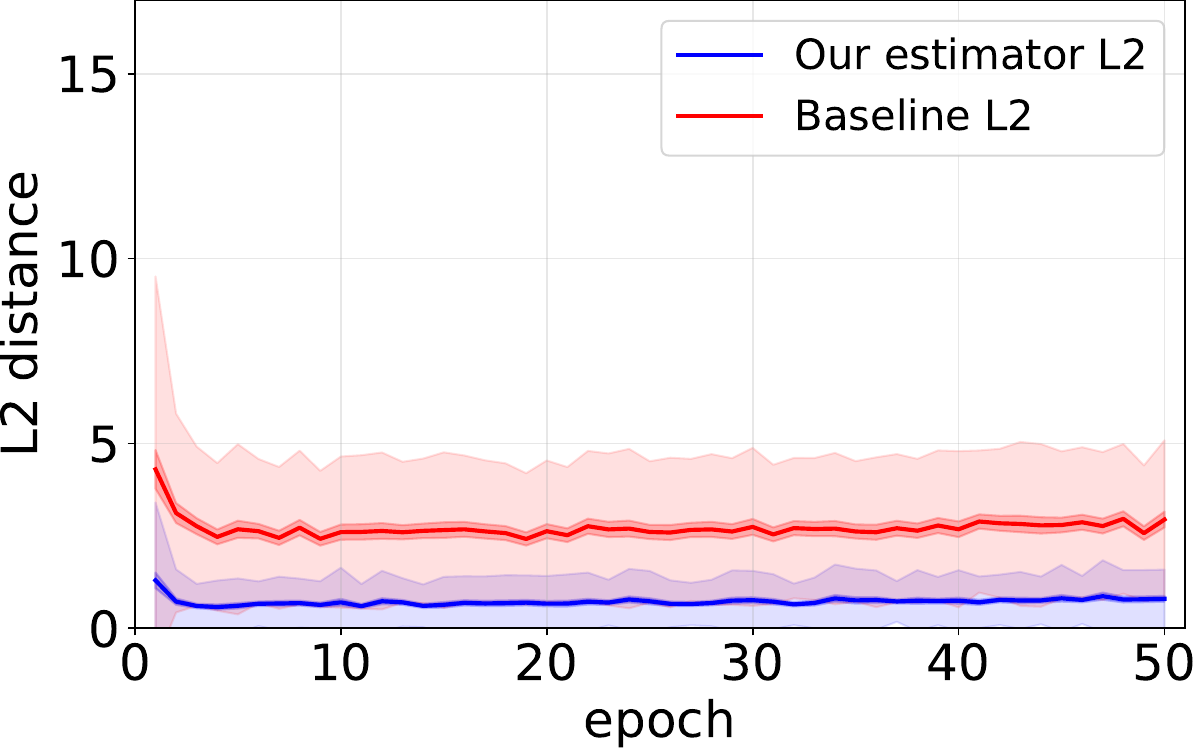} \\

\rotatebox[origin=c]{90}{\footnotesize\textbf{Airfoil self-noise}} &
\includegraphics[valign=m,width=0.29\textwidth]{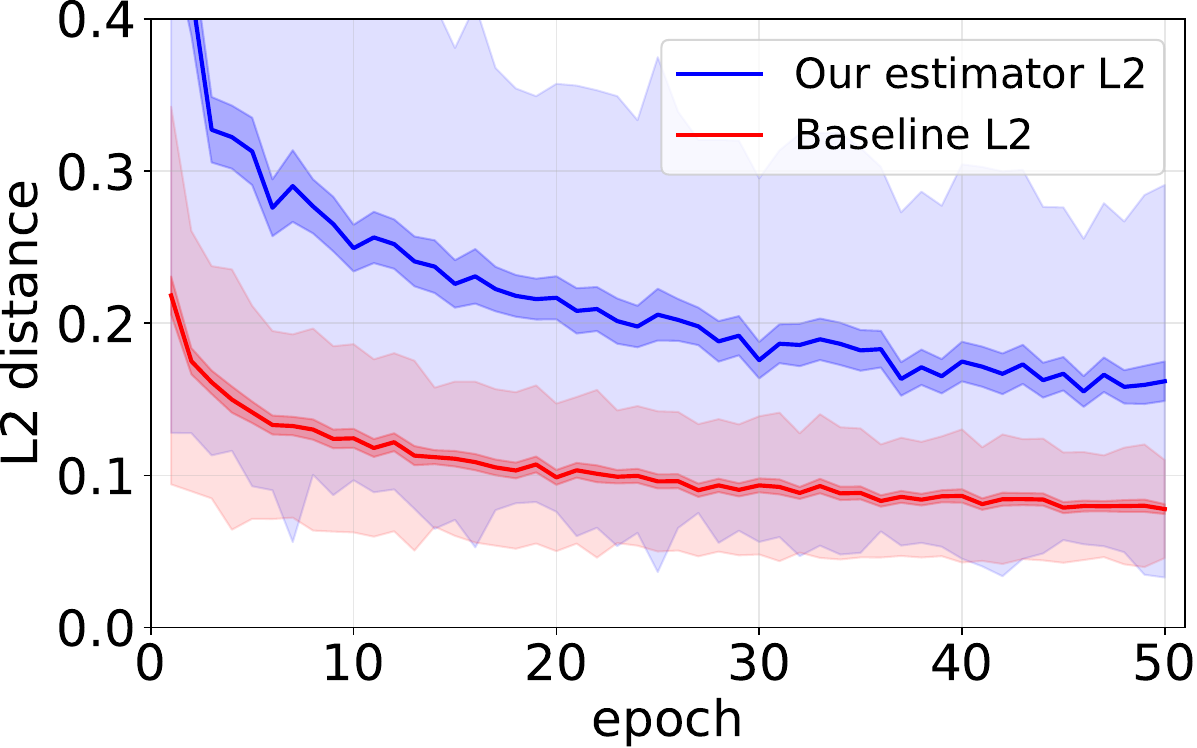} &
\includegraphics[valign=m,width=0.29\textwidth]{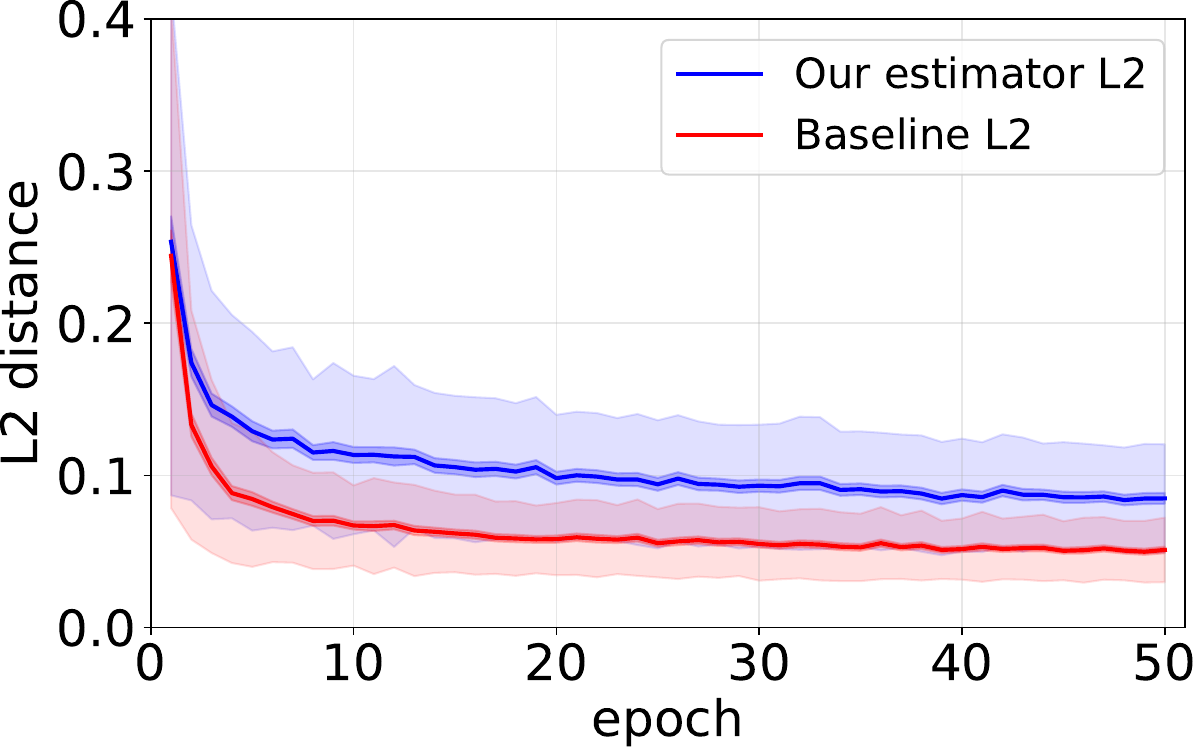} &
\includegraphics[valign=m,width=0.29\textwidth]{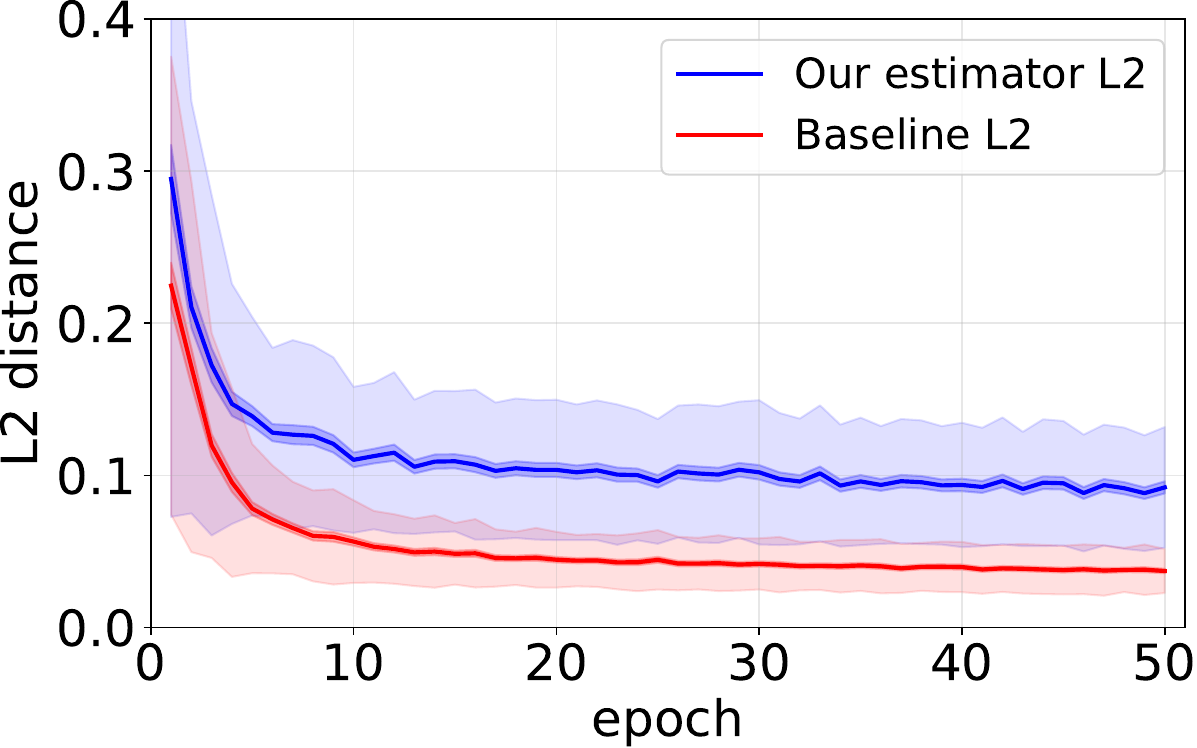} \\

\rotatebox[origin=c]{90}{\footnotesize\textbf{Appliances energy}} &
\includegraphics[valign=m,width=0.29\textwidth]{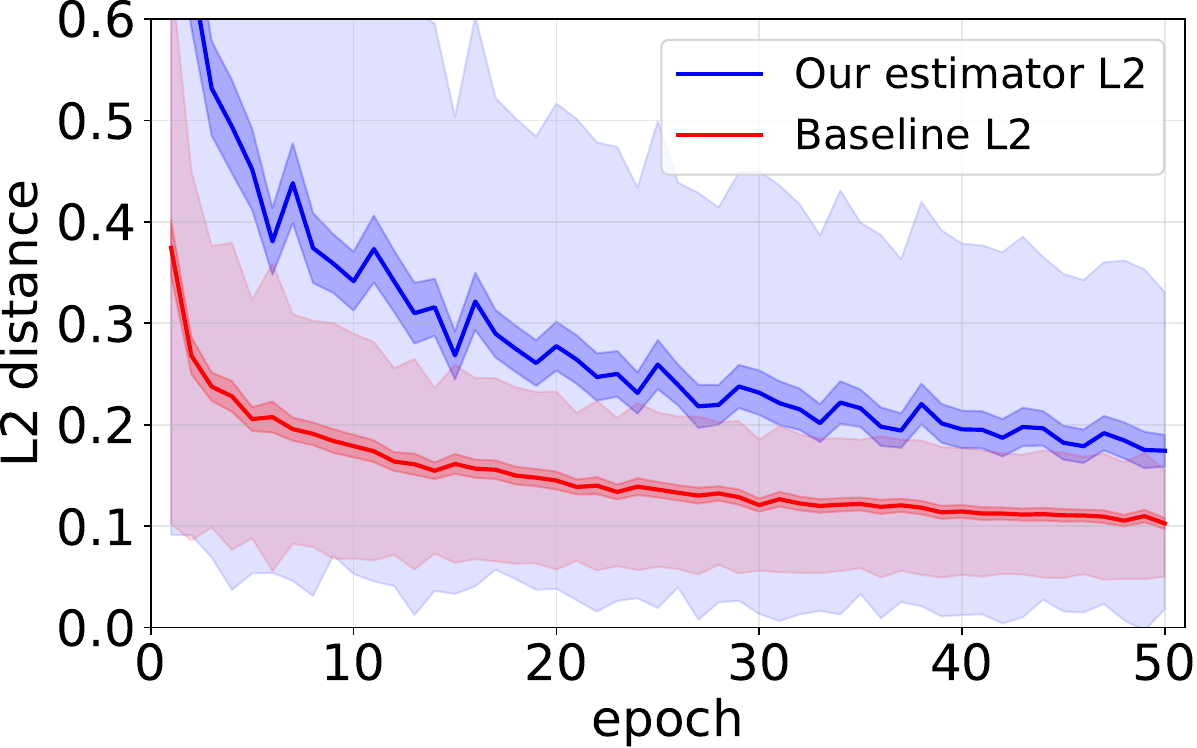} &
\includegraphics[valign=m,width=0.29\textwidth]{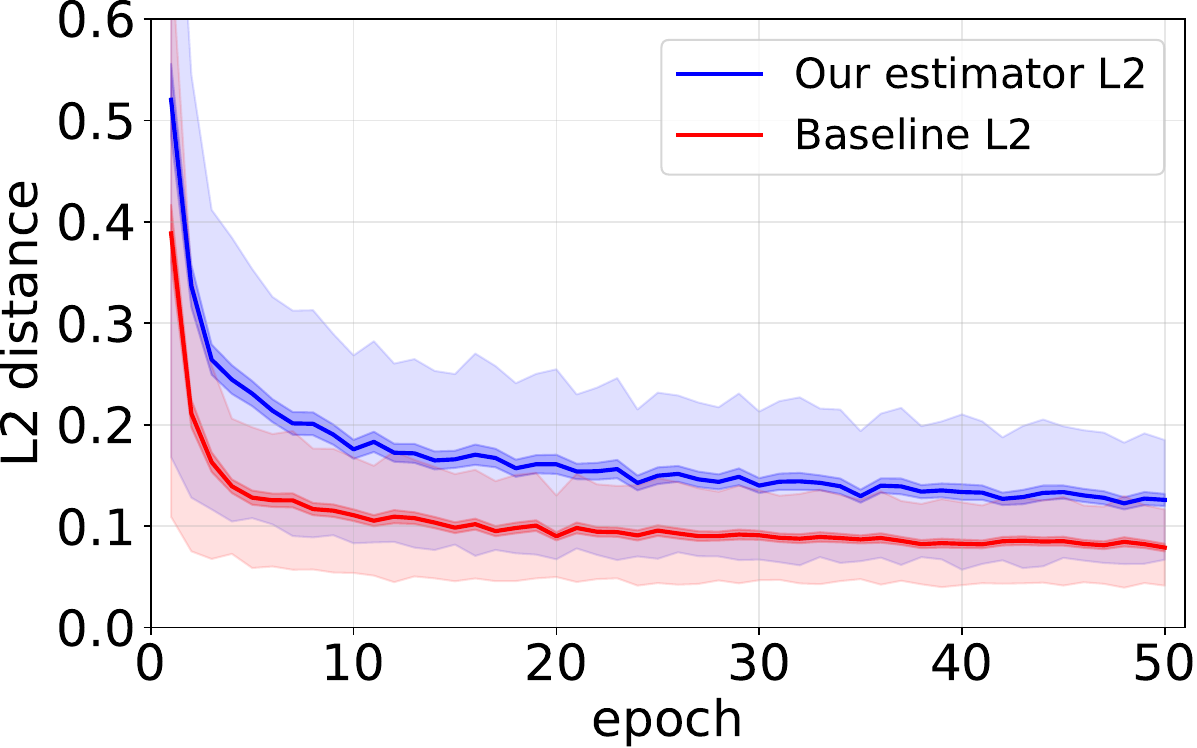} &
\includegraphics[valign=m,width=0.29\textwidth]{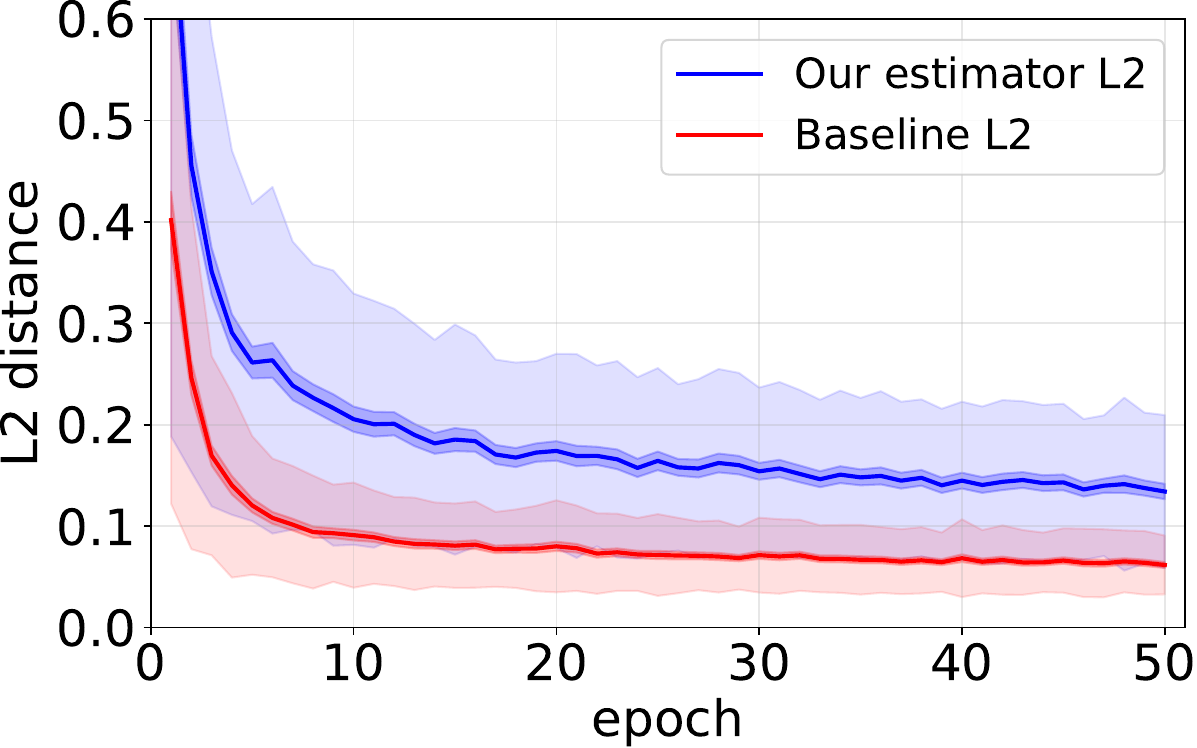} \\

\rotatebox[origin=c]{90}{\footnotesize\textbf{MNIST}} &
\includegraphics[valign=m,width=0.29\textwidth]{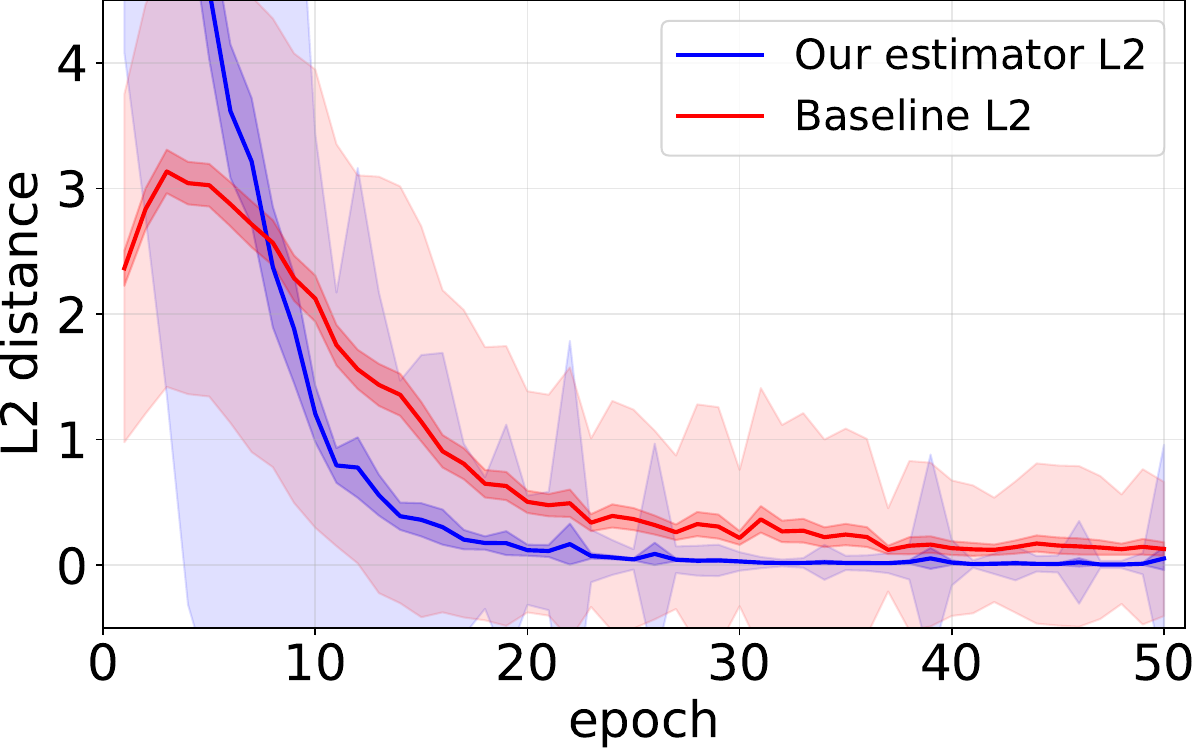} &
\includegraphics[valign=m,width=0.29\textwidth]{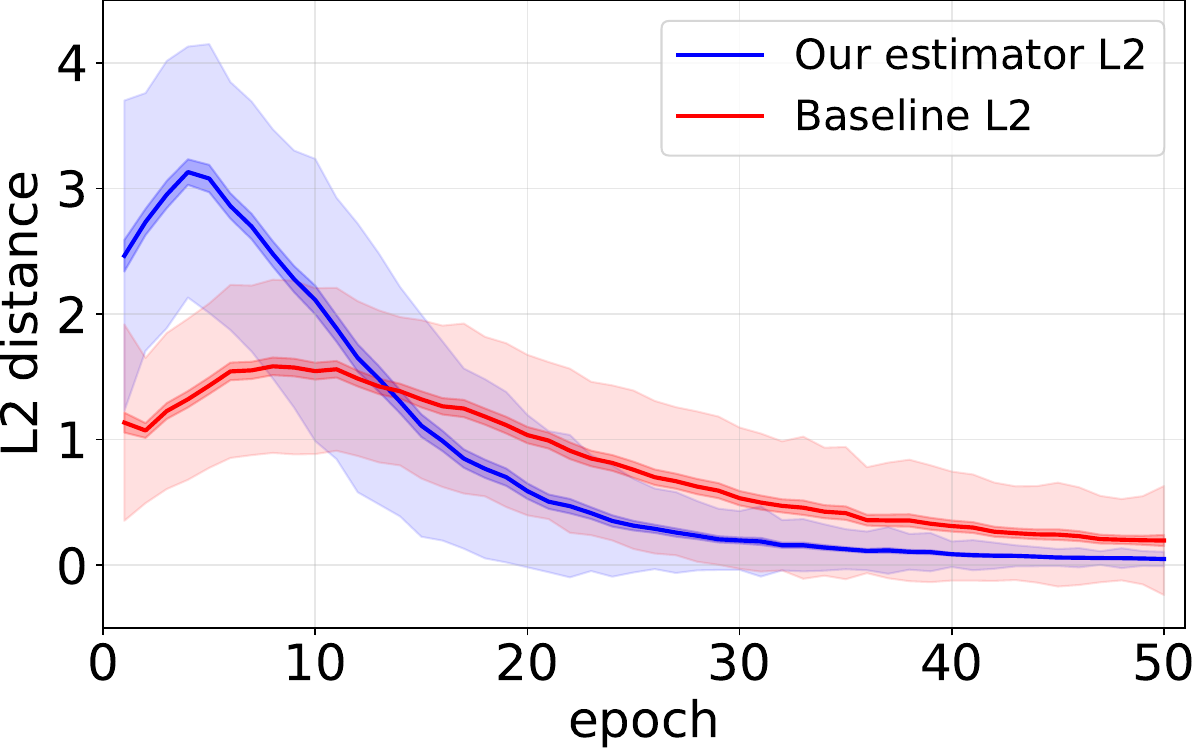} &
\includegraphics[valign=m,width=0.29\textwidth]{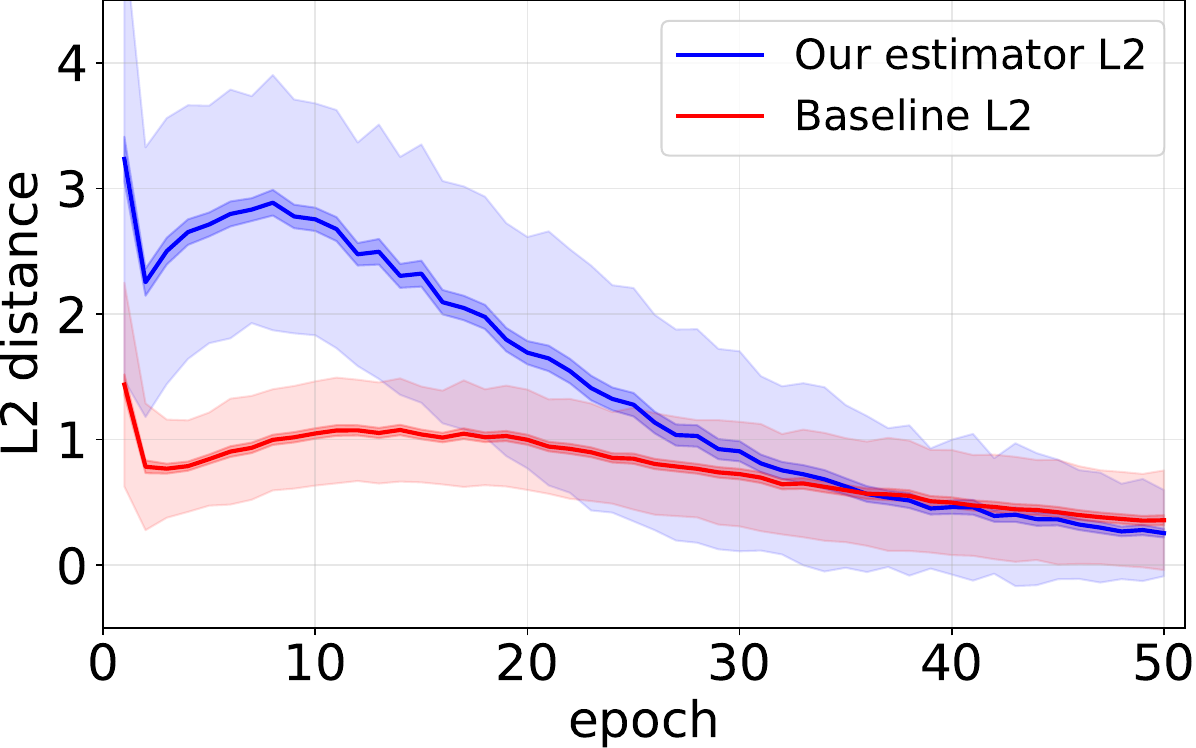} \\

\rotatebox[origin=c]{90}{\footnotesize\textbf{Fashion-MNIST}} &
\includegraphics[valign=m,width=0.29\textwidth]{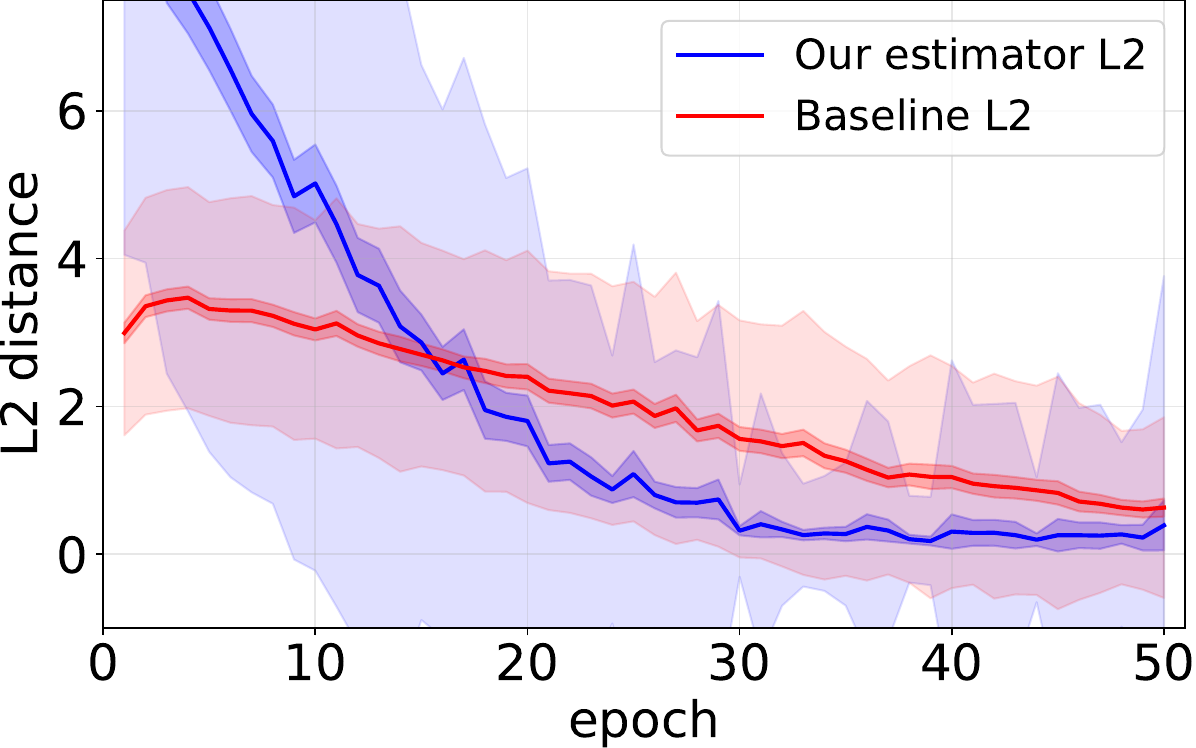} &
\includegraphics[valign=m,width=0.29\textwidth]{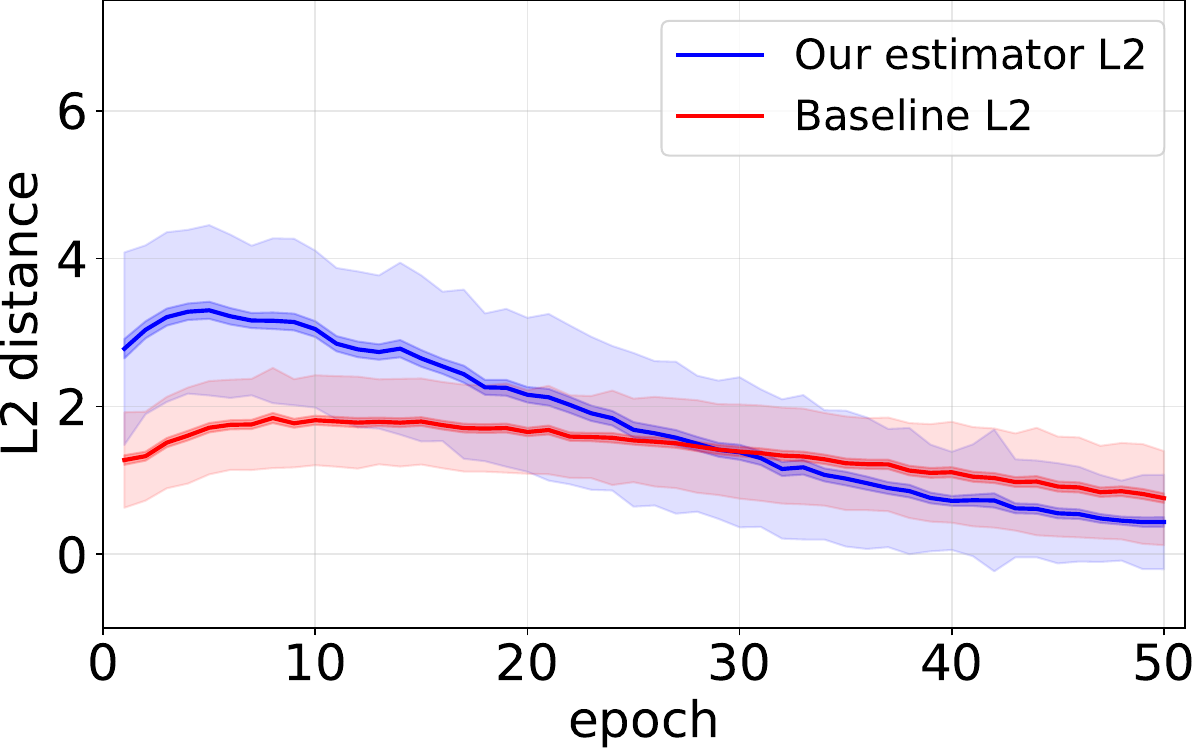} &
\includegraphics[valign=m,width=0.29\textwidth]{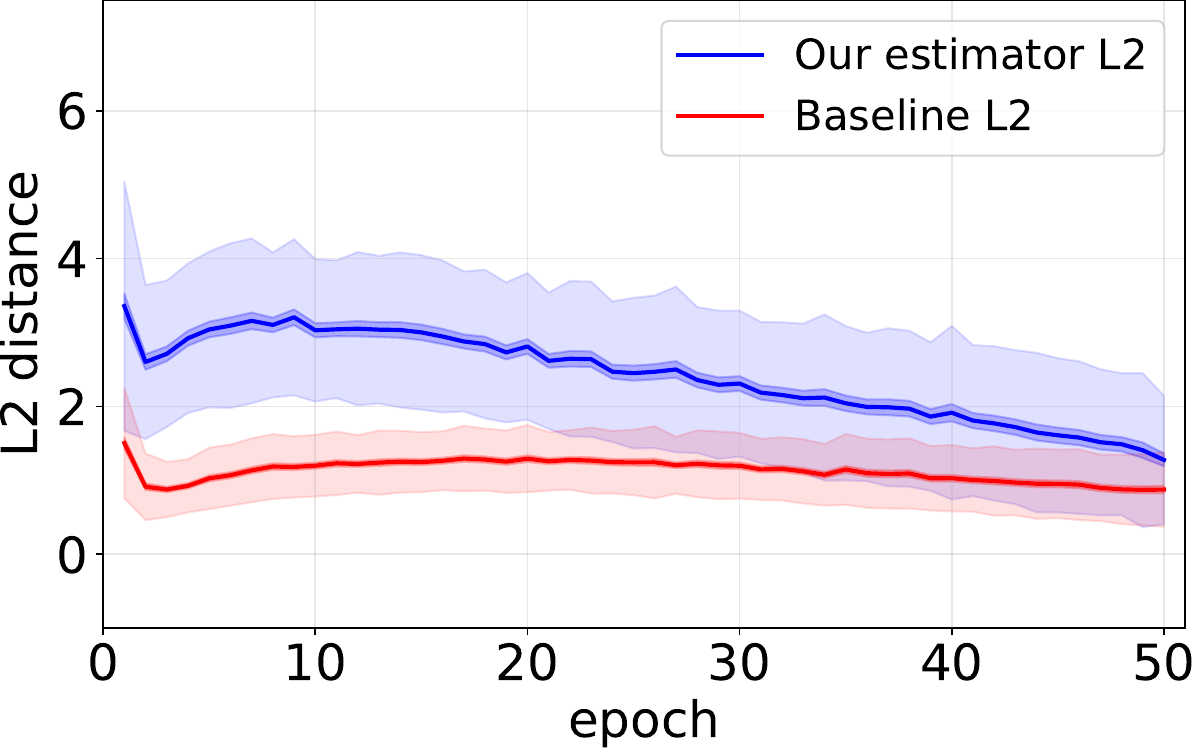} \\

\rotatebox[origin=c]{90}{\footnotesize\textbf{CIFAR-10}} &
\includegraphics[valign=m,width=0.29\textwidth]{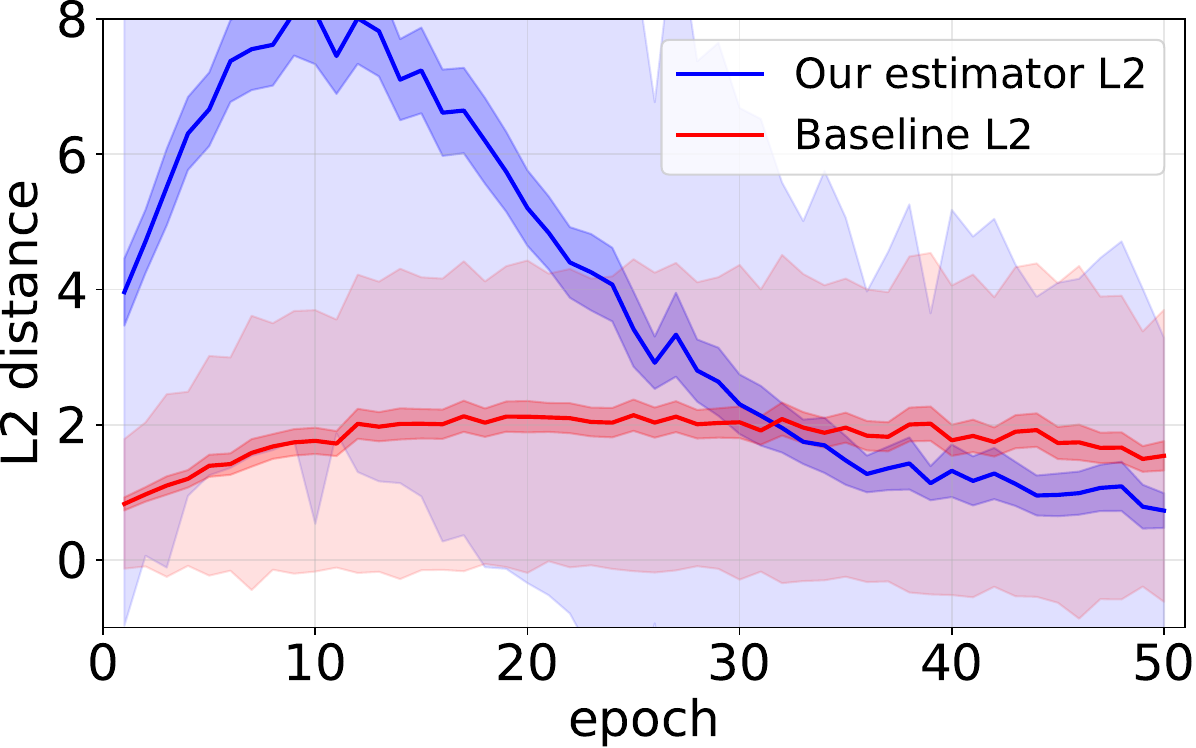} &
\includegraphics[valign=m,width=0.29\textwidth]{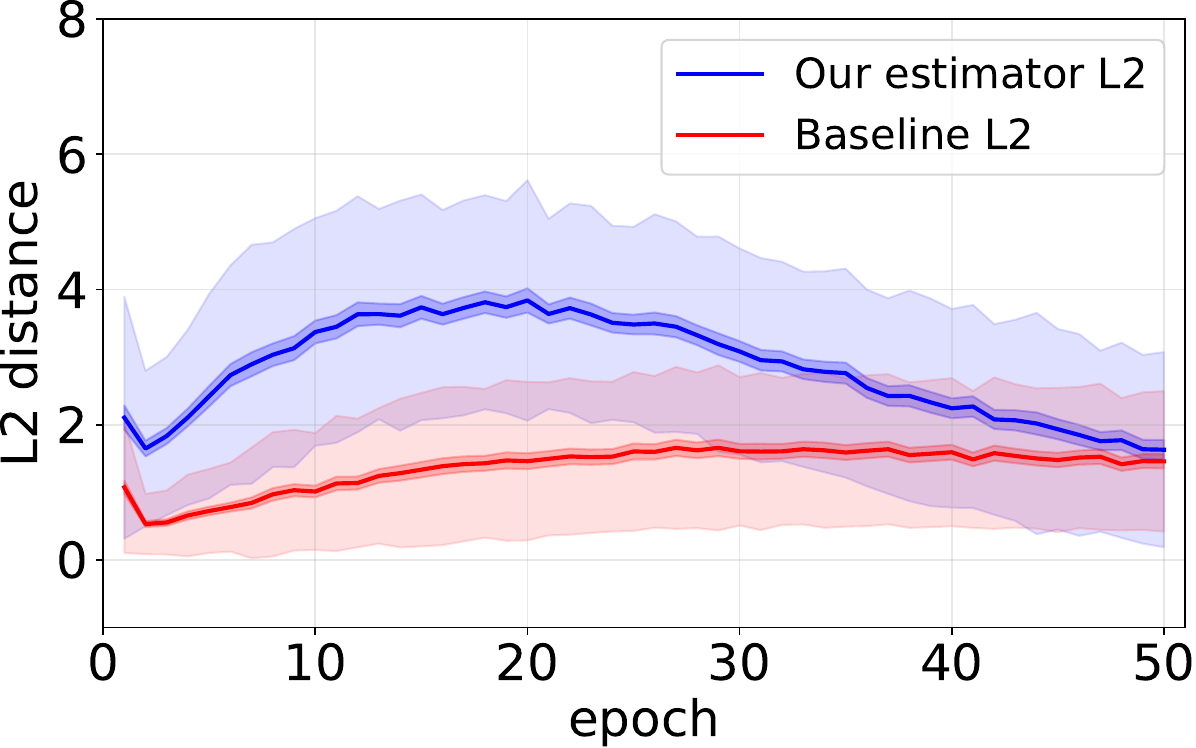} &
\includegraphics[valign=m,width=0.29\textwidth]{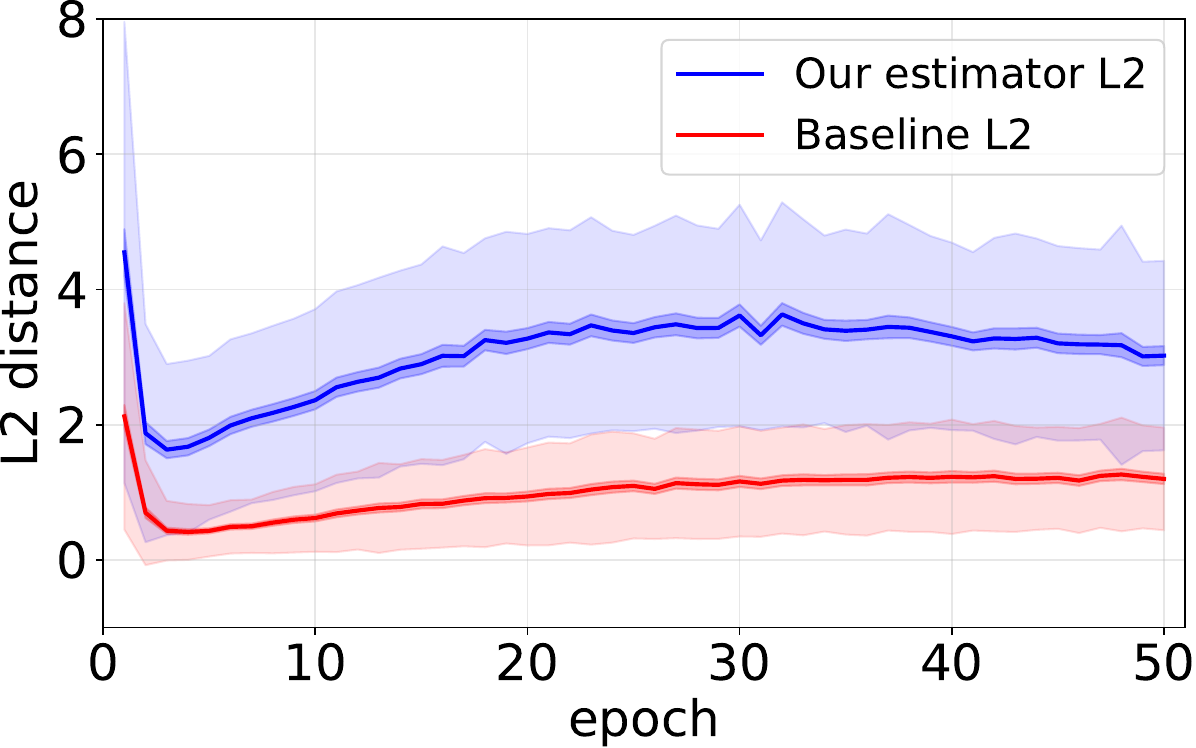} \\

\rotatebox[origin=c]{90}{\footnotesize\textbf{CIFAR-100}} &
\includegraphics[valign=m,width=0.29\textwidth]{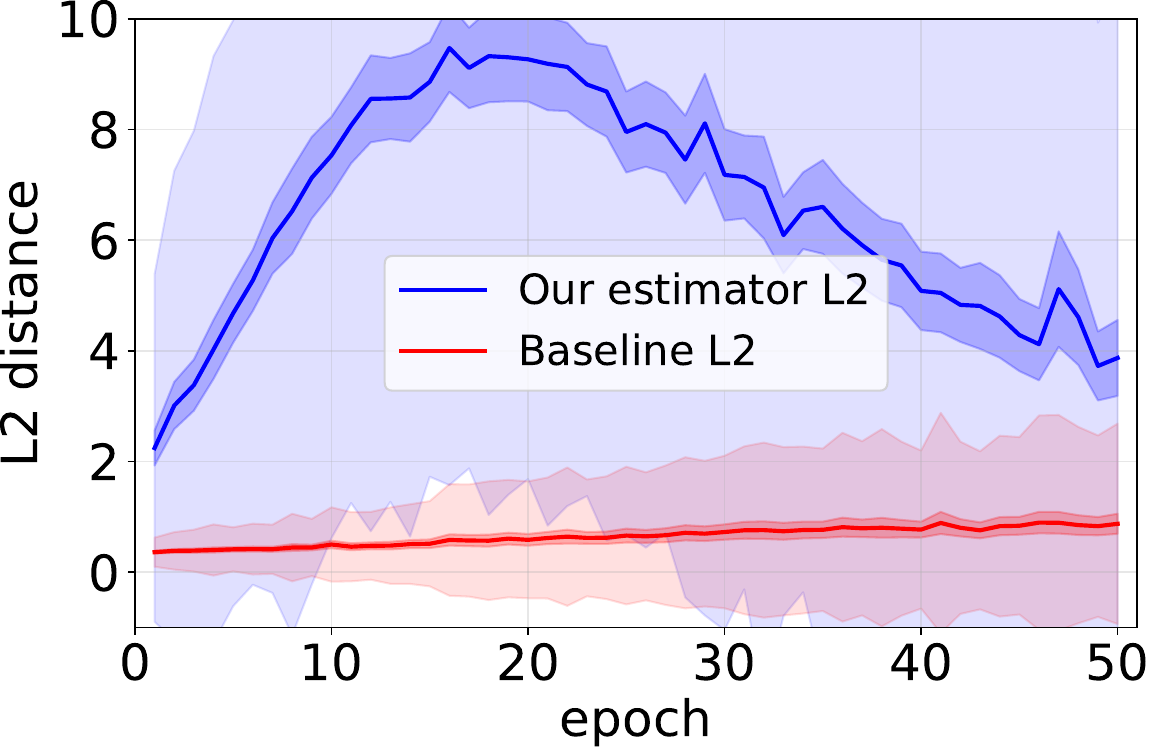} &
\includegraphics[valign=m,width=0.29\textwidth]{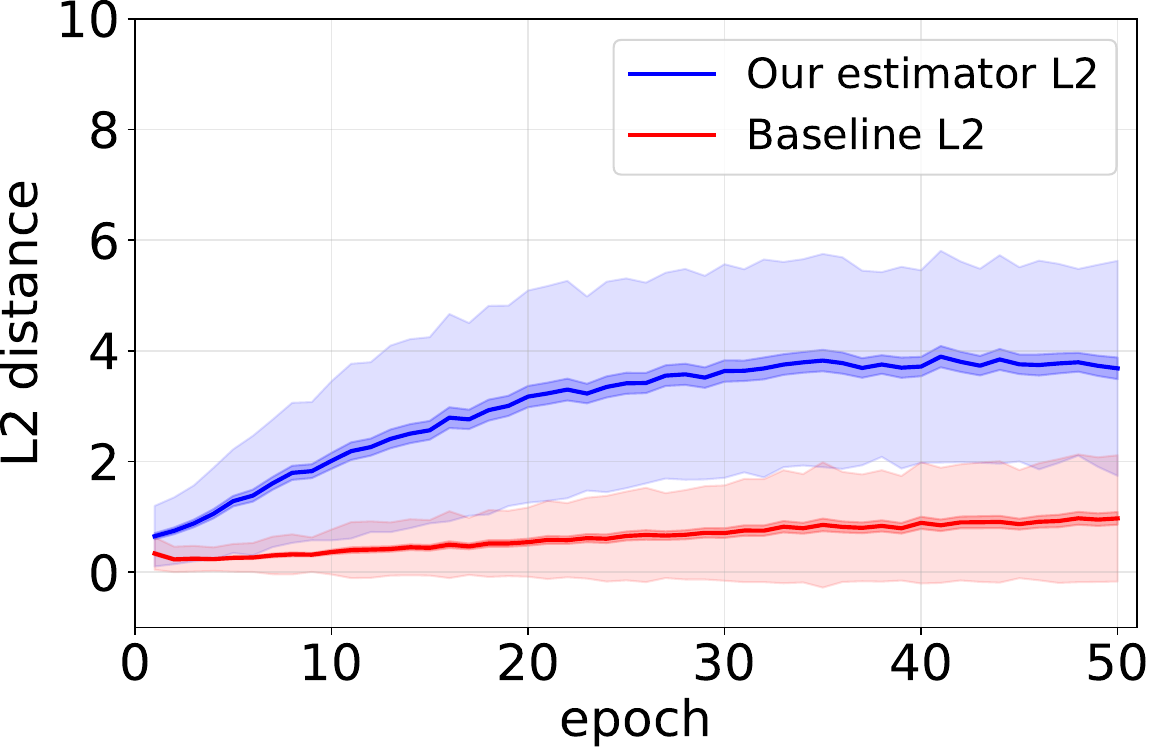} &
\includegraphics[valign=m,width=0.29\textwidth]{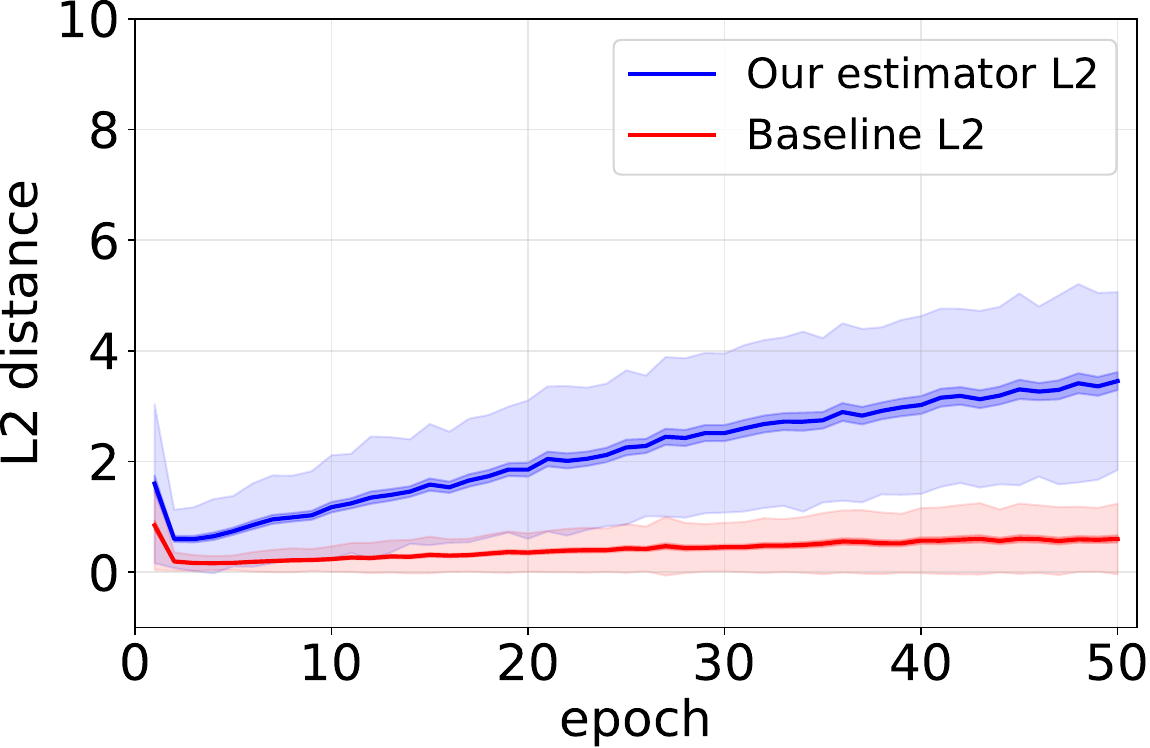} \\

\end{tabular}

\caption{L2 distance to full-batch gradient with AdamW optimizer. The darker curve represents the average of 400 runs, the dark shading is the 95\% confidence interval and lighter shading shows the standard deviation. Columns correspond to batch sizes and rows to datasets.}
\label{fig:AdamW_l2_grid}
\end{figure*}

The results presented are intended to reveal consistent patterns across datasets rather than optimize performance for any single task. In particular, we are interested in whether model-assisted gradient estimators provide systematic improvements over standard baseline approach and how these improvements interact with optimizer choice and dataset characteristics. The Figures \ref{fig:sgd_loss_grid} and \ref{fig:AdamW_loss_grid} represent the mean test loss curves for the SGD and AdamW optimizers, corresponding to the worst and best cases for our estimator respectively. For regression datasets, the loss is measured in terms of \emph{mean squared error} and for image classification datasets, the loss is measured in terms of \emph{cross-entropy loss}. The corresponding SGD-M and Adam figures are presented in the Appendix. Similarly, we illustrate the corresponding evolution of L2 distance to the full-batch gradient in Figures \ref{fig:sgd_l2_grid} and \ref{fig:AdamW_l2_grid}, which highlight how closely the two competing estimators follow the true population gradient.

With vanilla SGD, our estimator does not show a clear overall advantage over the baseline, except for the synthetic sinusoid dataset. In this dataset case, we can see that our estimator achieves best test loss performance, as well as L2-distance to full-batch gradient, especially with higher batch sizes. The worst performance for our estimator with SGD occurs with higher dimensional input data cases (CIFAR). With momentum and past gradient utilizing optimizers (SGD-M, Adam, AdamW) we can notice a drastic change in the performance for our estimator. Although the model-assisted estimator shows a slightly larger average deviation from the full-batch gradient, it generally outperforms the baseline in terms of generalization performance, with the clearest and most statistically significant improvements observed on the MNIST datasets.

In Tables \ref{tab:batch10_results_colored}, \ref{tab:batch50_results_colored} and \ref{tab:batch100_results_colored} we have listed detailed best results over all batch size cases respectively, which support the visual inspection of Figures \ref{fig:sgd_loss_grid}, \ref{fig:AdamW_loss_grid}, \ref{fig:sgdm_loss_grid} and \ref{fig:Adam_loss_grid}. The tables' cell colors are based on an equally weighted normalized score computed within each dataset from test loss, standard deviation, and epoch. Thus, the coloring highlights overall estimator performance in terms of accuracy, stability, and efficiency. With vanilla SGD, our estimator outperforms the baseline in approximately \(14\text{--}29\%\) of dataset cases. In contrast, when using momentum-based optimizers, the win rate increases substantially to \(57\text{--}86\%\). The best observed performance is achieved with AdamW at batch size \(100\), where our estimator surpasses the baseline in \(86\%\) of cases. Across datasets, performance varies: CIFAR-100 represents the most challenging setting (win rate \(25\text{--}50\%\)), whereas MNIST yields the strongest results (\(75\%\)), along with consistently lower and more stable test loss, achieved in nearly half the number of epochs. When results are aggregated over optimizers and datasets, the proposed estimator outperforms the baseline in \(54\%\) of the comparisons for batch sizes \(10\) and \(50\), and in \(61\%\) of the comparisons for batch size \(100\). When the optimizer is treated as a tunable hyperparameter on a per-dataset basis, the win rate further improves to \(71\text{--}86\%\).



\begin{table}[tbh]
\centering
\caption{Results for batch size 10. Values are reported as mean minimum test loss $\pm$ standard deviation (epoch) over 400 distinct runs. Cell color is based on an equally weighted normalized combined score within each dataset, computed from loss, standard deviation, and epoch. Greener indicates better combined performance, while redder indicates worse combined performance. Full-batch rows are shown in gray as references. Bold indicates the lowest (best) score between our estimator and baseline within each dataset. The superscript $^*$ indicates better score between our estimator and the baseline within the same dataset and optimizer. The final column reports dataset-wise win rates, and the bottom row reports optimizer-wise win rates for our estimator. The bottom-right gray cell reports the overall win rate of our estimator across all dataset--optimizer comparisons in the table.}
\label{tab:batch10_results_colored}
\setlength{\tabcolsep}{4pt}
\renewcommand{\arraystretch}{1.1}

\resizebox{\textwidth}{!}{%
\begin{tabular}{llcccc>{\centering\arraybackslash}p{1.5cm}}
\toprule
\textbf{Dataset} & \textbf{Case} & \textbf{SGD} & \textbf{SGD-M} & \textbf{Adam} & \textbf{AdamW} & \textbf{Row win rate} \\
\midrule

\multirow{3}{*}{Synthetic}
& Our estimator 
& \heat{43}{0.62 $\pm$ 0.29 (100)} 
& \heat{67}{3.06 $\pm$ 1.49 (1)} 
& \heat{33}{\textbf{0.22 $\pm$ 0.067 (98)}$^*$} 
& \heat{33}{0.24 $\pm$ 0.062 (100)$^*$} 
& \multirow{3}{*}{\textbf{50.0\%}} \\

& Baseline      
& \heat{37}{0.46 $\pm$ 0.11 (98)$^*$} 
& \heat{51}{1.91 $\pm$ 1.38 (4)$^*$} 
& \heat{34}{0.24 $\pm$ 0.067 (100)} 
& \heat{34}{0.26 $\pm$ 0.063 (100)}
& \\

& Full-batch    
& \graycell{0.39 $\pm$ 0.058 (100)} 
& \graycell{0.44 $\pm$ 0.59 (97)} 
& \graycell{0.26 $\pm$ 0.075 (100)} 
& \graycell{0.26 $\pm$ 0.078 (100)}
& \\

\midrule

\multirow{3}{*}{Airfoil self-noise}
& Our estimator 
& \heat{100}{0.0087 $\pm$ 0.0012 (100)} 
& \heat{61}{0.0050 $\pm$ 0.0008 (100)} 
& \heat{33}{0.0023 $\pm$ 0.0005 (100)$^*$} 
& \heat{17}{0.0024 $\pm$ 0.0006 (99)} 
& \multirow{3}{*}{\textbf{25.0\%}} \\

& Baseline      
& \heat{88}{0.0082 $\pm$ 0.0010 (100)$^*$} 
& \heat{61}{0.0050 $\pm$ 0.0008 (100)} 
& \heat{35}{0.0025 $\pm$ 0.0006 (100)} 
& \heat{2}{\textbf{0.0026 $\pm$ 0.0006 (98)}$^*$}
& \\

& Full-batch    
& \graycell{0.0082 $\pm$ 0.0010 (100)} 
& \graycell{0.0049 $\pm$ 0.0008 (100)} 
& \graycell{0.0018 $\pm$ 0.0004 (98)} 
& \graycell{0.0017 $\pm$ 0.0004 (97)}
& \\

\midrule

\multirow{3}{*}{Appliances energy}
& Our estimator 
& \heat{85}{0.015 $\pm$ 0.0030 (96)} 
& \heat{23}{0.012 $\pm$ 0.0022 (92)$^*$} 
& \heat{6}{0.010 $\pm$ 0.0022 (91)$^*$} 
& \heat{4}{\textbf{0.010 $\pm$ 0.0021 (92)}$^*$} 
& \multirow{3}{*}{\textbf{75.0\%}} \\

& Baseline      
& \heat{73}{0.014 $\pm$ 0.0025 (100)$^*$} 
& \heat{46}{0.012 $\pm$ 0.0021 (99)} 
& \heat{23}{0.010 $\pm$ 0.0021 (97)} 
& \heat{39}{0.010 $\pm$ 0.0022 (100)}
& \\

& Full-batch    
& \graycell{0.013 $\pm$ 0.0025 (100)} 
& \graycell{0.012 $\pm$ 0.0021 (100)} 
& \graycell{0.012 $\pm$ 0.0023 (10)} 
& \graycell{0.012 $\pm$ 0.0024 (15)}
& \\

\midrule

\multirow{3}{*}{MNIST}
& Our estimator 
& \heat{81}{0.76 $\pm$ 0.27 (67)} 
& \heat{21}{0.58 $\pm$ 0.18 (11)$^*$} 
& \heat{3}{0.46 $\pm$ 0.13 (9)$^*$} 
& \heat{2}{\textbf{0.45 $\pm$ 0.14 (9)}$^*$} 
& \multirow{3}{*}{\textbf{75.0\%}} \\

& Baseline      
& \heat{39}{0.59 $\pm$ 0.11 (51)$^*$} 
& \heat{27}{0.60 $\pm$ 0.18 (15)} 
& \heat{40}{0.58 $\pm$ 0.34 (16)} 
& \heat{60}{0.63 $\pm$ 0.46 (21)}
& \\

& Full-batch    
& \graycell{0.61 $\pm$ 0.12 (43)} 
& \graycell{0.66 $\pm$ 0.17 (7)} 
& \graycell{0.69 $\pm$ 0.33 (2)} 
& \graycell{0.73 $\pm$ 0.39 (2)}
& \\

\midrule

\multirow{3}{*}{Fashion-MNIST}
& Our estimator 
& \heat{86}{0.92 $\pm$ 0.24 (49)} 
& \heat{17}{0.78 $\pm$ 0.11 (10)$^*$} 
& \heat{11}{0.74 $\pm$ 0.14 (7)$^*$} 
& \heat{4}{\textbf{0.71 $\pm$ 0.11 (8)}$^*$} 
& \multirow{3}{*}{\textbf{75.0\%}} \\

& Baseline      
& \heat{39}{0.75 $\pm$ 0.091 (50)$^*$} 
& \heat{21}{0.79 $\pm$ 0.12 (14)} 
& \heat{60}{0.87 $\pm$ 0.34 (10)} 
& \heat{30}{0.78 $\pm$ 0.22 (9)}
& \\

& Full-batch    
& \graycell{0.76 $\pm$ 0.089 (43)} 
& \graycell{0.78 $\pm$ 0.13 (6)} 
& \graycell{0.91 $\pm$ 0.25 (2)} 
& \graycell{0.88 $\pm$ 0.22 (2)}
& \\

\midrule

\multirow{3}{*}{CIFAR-10}
& Our estimator 
& \heat{64}{2.25 $\pm$ 0.12 (26)} 
& \heat{33}{\textbf{2.14 $\pm$ 0.11 (14)}$^*$} 
& \heat{55}{2.22 $\pm$ 0.13 (6)} 
& \heat{57}{2.20 $\pm$ 0.14 (6)} 
& \multirow{3}{*}{\textbf{25.0\%}} \\

& Baseline      
& \heat{33}{2.07 $\pm$ 0.096 (60)$^*$} 
& \heat{45}{2.18 $\pm$ 0.12 (19)} 
& \heat{53}{2.22 $\pm$ 0.13 (9)$^*$} 
& \heat{55}{2.21 $\pm$ 0.13 (10)$^*$}
& \\

& Full-batch    
& \graycell{2.07 $\pm$ 0.096 (47)} 
& \graycell{2.10 $\pm$ 0.11 (10)} 
& \graycell{2.24 $\pm$ 0.11 (1)} 
& \graycell{2.22 $\pm$ 0.11 (1)}
& \\

\midrule

\multirow{3}{*}{CIFAR-100}
& Our estimator 
& \heat{69}{4.66 $\pm$ 0.031 (7)$^*$} 
& \heat{28}{4.64 $\pm$ 0.022 (5)$^*$} 
& \heat{49}{4.66 $\pm$ 0.034 (1)} 
& \heat{67}{4.66 $\pm$ 0.044 (1)} 
& \multirow{3}{*}{\textbf{50.0\%}} \\

& Baseline      
& \heat{74}{4.66 $\pm$ 0.027 (10)} 
& \heat{37}{4.63 $\pm$ 0.021 (11)} 
& \heat{11}{\textbf{4.63 $\pm$ 0.018 (4)}$^*$} 
& \heat{15}{4.64 $\pm$ 0.020 (4)$^*$}
& \\

& Full-batch    
& \graycell{4.67 $\pm$ 0.029 (14)} 
& \graycell{4.64 $\pm$ 0.027 (7)} 
& \graycell{4.70 $\pm$ 0.087 (1)} 
& \graycell{4.71 $\pm$ 0.11 (1)}
& \\

\midrule

\multicolumn{2}{l}{\textbf{Column win rate}}
& \textbf{14.3\%}
& \textbf{71.4\%}
& \textbf{71.4\%}
& \textbf{57.1\%}
& \cellcolor{gray!20} \textbf{53.6\%} \\

\bottomrule
\end{tabular}
}
\end{table}
\begin{table}[H]
\centering
\caption{Results for batch size 50. Notation, coloring, bolding, and superscript markers follow Table~\ref{tab:batch10_results_colored}.}
\label{tab:batch50_results_colored}
\setlength{\tabcolsep}{4pt}
\renewcommand{\arraystretch}{1.1}

\resizebox{\textwidth}{!}{%
\begin{tabular}{llcccc>{\centering\arraybackslash}p{1.5cm}}
\toprule
\textbf{Dataset} & \textbf{Case} & \textbf{SGD} & \textbf{SGD-M} & \textbf{Adam} & \textbf{AdamW} & \textbf{Row win rate} \\
\midrule

\multirow{3}{*}{Synthetic}
& Our estimator 
& \heat{21}{0.57 $\pm$ 0.097 (95)$^*$} 
& \heat{100}{1.15 $\pm$ 1.28 (100)} 
& \heat{24}{\textbf{0.33 $\pm$ 0.048 (98)}$^*$} 
& \heat{34}{0.33 $\pm$ 0.049 (100)} 
& \multirow{3}{*}{\textbf{50.0\%}} \\

& Baseline 
& \heat{47}{0.59 $\pm$ 0.17 (100)} 
& \heat{27}{0.60 $\pm$ 0.63 (93)$^*$} 
& \heat{25}{0.35 $\pm$ 0.048 (98)} 
& \heat{30}{0.35 $\pm$ 0.052 (99)$^*$}
& \\

& Full-batch 
& \graycell{0.56 $\pm$ 0.11 (99)} 
& \graycell{0.50 $\pm$ 0.65 (99)} 
& \graycell{0.31 $\pm$ 0.056 (100)} 
& \graycell{0.31 $\pm$ 0.051 (100)}
& \\

\midrule

\multirow{3}{*}{Airfoil self-noise}
& Our estimator 
& \heat{100}{0.014 $\pm$ 0.0024 (100)} 
& \heat{50}{0.0070 $\pm$ 0.00091 (100)} 
& \heat{33}{0.0030 $\pm$ 0.00071 (100)} 
& \heat{34}{0.0030 $\pm$ 0.00073 (100)$^*$} 
& \multirow{3}{*}{\textbf{25.0\%}} \\

& Baseline 
& \heat{100}{0.014 $\pm$ 0.0024 (100)} 
& \heat{50}{0.0071 $\pm$ 0.00090 (100)} 
& \heat{3}{\textbf{0.0033 $\pm$ 0.00083 (99)}$^*$} 
& \heat{37}{0.0034 $\pm$ 0.00082 (100)}
& \\

& Full-batch 
& \graycell{0.014 $\pm$ 0.0024 (100)} 
& \graycell{0.0071 $\pm$ 0.00091 (100)} 
& \graycell{0.0026 $\pm$ 0.00074 (100)} 
& \graycell{0.0025 $\pm$ 0.00076 (100)}
& \\

\midrule

\multirow{3}{*}{Appliances energy}
& Our estimator 
& \heat{100}{0.018 $\pm$ 0.0033 (100)} 
& \heat{47}{0.013 $\pm$ 0.0022 (100)$^*$} 
& \heat{3}{\textbf{0.011 $\pm$ 0.0021 (96)}$^*$} 
& \heat{36}{0.011 $\pm$ 0.0021 (100)} 
& \multirow{3}{*}{\textbf{50.0\%}} \\

& Baseline 
& \heat{97}{0.017 $\pm$ 0.0032 (100)$^*$} 
& \heat{49}{0.013 $\pm$ 0.0023 (100)} 
& \heat{27}{0.011 $\pm$ 0.0020 (99)} 
& \heat{11}{0.011 $\pm$ 0.0021 (97)$^*$}
& \\

& Full-batch 
& \graycell{0.018 $\pm$ 0.0032 (100)} 
& \graycell{0.013 $\pm$ 0.0023 (100)} 
& \graycell{0.012 $\pm$ 0.0021 (52)} 
& \graycell{0.011 $\pm$ 0.0022 (65)}
& \\

\midrule

\multirow{3}{*}{MNIST}
& Our estimator 
& \heat{68}{0.75 $\pm$ 0.13 (100)} 
& \heat{19}{0.59 $\pm$ 0.13 (30)$^*$} 
& \heat{6}{0.50 $\pm$ 0.16 (17)$^*$} 
& \heat{0}{\textbf{0.49 $\pm$ 0.13 (17)}$^*$} 
& \multirow{3}{*}{\textbf{75.0\%}} \\

& Baseline 
& \heat{62}{0.71 $\pm$ 0.13 (100)$^*$} 
& \heat{32}{0.62 $\pm$ 0.17 (39)} 
& \heat{59}{0.66 $\pm$ 0.35 (26)} 
& \heat{57}{0.64 $\pm$ 0.35 (30)}
& \\

& Full-batch 
& \graycell{0.71 $\pm$ 0.12 (100)} 
& \graycell{0.64 $\pm$ 0.17 (34)} 
& \graycell{0.73 $\pm$ 0.45 (9)} 
& \graycell{0.70 $\pm$ 0.34 (10)}
& \\

\midrule

\multirow{3}{*}{Fashion-MNIST}
& Our estimator 
& \heat{70}{0.84 $\pm$ 0.095 (100)} 
& \heat{24}{0.76 $\pm$ 0.11 (28)$^*$} 
& \heat{7}{\textbf{0.72 $\pm$ 0.11 (13)}$^*$} 
& \heat{16}{0.72 $\pm$ 0.14 (13)$^*$} 
& \multirow{3}{*}{\textbf{75.0\%}} \\

& Baseline 
& \heat{60}{0.81 $\pm$ 0.084 (100)$^*$} 
& \heat{31}{0.76 $\pm$ 0.11 (43)} 
& \heat{40}{0.77 $\pm$ 0.16 (20)} 
& \heat{50}{0.77 $\pm$ 0.20 (18)}
& \\

& Full-batch 
& \graycell{0.81 $\pm$ 0.085 (100)} 
& \graycell{0.77 $\pm$ 0.12 (31)} 
& \graycell{0.81 $\pm$ 0.23 (7)} 
& \graycell{0.82 $\pm$ 0.28 (7)}
& \\

\midrule

\multirow{3}{*}{CIFAR-10}
& Our estimator 
& \heat{55}{2.17 $\pm$ 0.083 (100)} 
& \heat{23}{\textbf{2.10 $\pm$ 0.10 (39)}$^*$} 
& \heat{58}{2.20 $\pm$ 0.13 (10)$^*$} 
& \heat{60}{2.19 $\pm$ 0.13 (10)$^*$} 
& \multirow{3}{*}{\textbf{75.0\%}} \\

& Baseline 
& \heat{49}{2.15 $\pm$ 0.085 (100)$^*$} 
& \heat{39}{2.12 $\pm$ 0.11 (57)} 
& \heat{64}{2.21 $\pm$ 0.13 (17)} 
& \heat{62}{2.18 $\pm$ 0.13 (17)}
& \\

& Full-batch 
& \graycell{2.15 $\pm$ 0.086 (100)} 
& \graycell{2.11 $\pm$ 0.11 (49)} 
& \graycell{2.22 $\pm$ 0.11 (5)} 
& \graycell{2.22 $\pm$ 0.12 (5)}
& \\

\midrule

\multirow{3}{*}{CIFAR-100}
& Our estimator 
& \heat{85}{4.67 $\pm$ 0.027 (43)$^*$} 
& \heat{44}{4.65 $\pm$ 0.026 (18)} 
& \heat{42}{4.65 $\pm$ 0.026 (2)} 
& \heat{45}{4.65 $\pm$ 0.027 (2)} 
& \multirow{3}{*}{\textbf{25.0\%}} \\

& Baseline 
& \heat{98}{4.67 $\pm$ 0.028 (62)} 
& \heat{16}{4.64 $\pm$ 0.021 (30)$^*$} 
& \heat{14}{\textbf{4.64 $\pm$ 0.024 (7)}$^*$} 
& \heat{38}{4.64 $\pm$ 0.027 (8)$^*$}
& \\

& Full-batch 
& \graycell{4.67 $\pm$ 0.030 (72)} 
& \graycell{4.64 $\pm$ 0.023 (33)} 
& \graycell{4.65 $\pm$ 0.027 (2)} 
& \graycell{4.65 $\pm$ 0.027 (2)}
& \\

\midrule

\multicolumn{2}{l}{\textbf{Column win rate}}
& \textbf{28.6\%}
& \textbf{57.1\%}
& \textbf{71.4\%}
& \textbf{57.1\%}
& \cellcolor{gray!20} \textbf{53.6\%} \\

\bottomrule
\end{tabular}
}
\end{table}
\begin{table}[tbh]
\centering
\caption{Results for batch size 100. Notation, coloring, bolding, and superscript markers follow Table~\ref{tab:batch10_results_colored}.}
\label{tab:batch100_results_colored}
\setlength{\tabcolsep}{4pt}
\renewcommand{\arraystretch}{1.1}

\resizebox{\textwidth}{!}{%
\begin{tabular}{llcccc>{\centering\arraybackslash}p{1.5cm}}
\toprule
\textbf{Dataset} & \textbf{Case} & \textbf{SGD} & \textbf{SGD-M} & \textbf{Adam} & \textbf{AdamW} & \textbf{Row win rate} \\
\midrule

\multirow{3}{*}{Synthetic}
& Our estimator 
& \heat{44}{0.63 $\pm$ 0.085 (99)$^*$} 
& \heat{92}{1.02 $\pm$ 1.15 (98)} 
& \heat{34}{0.36 $\pm$ 0.070 (100)$^*$} 
& \heat{33}{\textbf{0.36 $\pm$ 0.065 (100)}$^*$} 
& \multirow{3}{*}{\textbf{75.0\%}} \\

& Baseline      
& \heat{49}{0.64 $\pm$ 0.13 (100)} 
& \heat{22}{0.55 $\pm$ 0.47 (91)$^*$} 
& \heat{35}{0.39 $\pm$ 0.068 (100)} 
& \heat{35}{0.40 $\pm$ 0.065 (100)}
& \\

& Full-batch    
& \graycell{0.62 $\pm$ 0.076 (99)} 
& \graycell{0.54 $\pm$ 0.59 (100)} 
& \graycell{0.34 $\pm$ 0.067 (100)} 
& \graycell{0.34 $\pm$ 0.065 (100)}
& \\

\midrule

\multirow{3}{*}{Airfoil self-noise}
& Our estimator 
& \heat{67}{0.018 $\pm$ 0.0033 (100)} 
& \heat{12}{0.0081 $\pm$ 0.0011 (100)$^*$} 
& \heat{0}{\textbf{0.0041 $\pm$ 0.0009 (100)}$^*$} 
& \heat{0}{0.0040 $\pm$ 0.0009 (100)$^*$} 
& \multirow{3}{*}{\textbf{75.0\%}} \\

& Baseline      
& \heat{65}{0.018 $\pm$ 0.0032 (100)$^*$} 
& \heat{13}{0.0082 $\pm$ 0.0012 (100)} 
& \heat{1}{0.0042 $\pm$ 0.0010 (100)} 
& \heat{2}{0.0042 $\pm$ 0.0010 (100)}
& \\

& Full-batch    
& \graycell{0.018 $\pm$ 0.0032 (100)} 
& \graycell{0.0082 $\pm$ 0.0012 (100)} 
& \graycell{0.0036 $\pm$ 0.0011 (100)} 
& \graycell{0.0037 $\pm$ 0.0011 (100)}
& \\

\midrule

\multirow{3}{*}{Appliances energy}
& Our estimator 
& \heat{100}{0.022 $\pm$ 0.0046 (100)} 
& \heat{44}{0.013 $\pm$ 0.0024 (100)$^*$} 
& \heat{36}{0.012 $\pm$ 0.0022 (100)} 
& \heat{1}{\textbf{0.011 $\pm$ 0.0021 (95)}$^*$} 
& \multirow{3}{*}{\textbf{50.0\%}} \\

& Baseline      
& \heat{94}{0.022 $\pm$ 0.0043 (100)$^*$} 
& \heat{45}{0.013 $\pm$ 0.0025 (100)} 
& \heat{35}{0.012 $\pm$ 0.0021 (100)$^*$} 
& \heat{33}{0.011 $\pm$ 0.0020 (100)}
& \\

& Full-batch    
& \graycell{0.022 $\pm$ 0.0042 (100)} 
& \graycell{0.013 $\pm$ 0.0025 (100)} 
& \graycell{0.012 $\pm$ 0.0022 (99)} 
& \graycell{0.012 $\pm$ 0.0024 (99)}
& \\

\midrule

\multirow{3}{*}{MNIST}
& Our estimator 
& \heat{80}{1.11 $\pm$ 0.22 (100)} 
& \heat{18}{0.61 $\pm$ 0.13 (56)$^*$} 
& \heat{2}{0.51 $\pm$ 0.14 (30)$^*$} 
& \heat{1}{\textbf{0.50 $\pm$ 0.13 (30)}$^*$} 
& \multirow{3}{*}{\textbf{75.0\%}} \\

& Baseline      
& \heat{79}{1.07 $\pm$ 0.23 (100)$^*$} 
& \heat{36}{0.65 $\pm$ 0.16 (77)} 
& \heat{48}{0.68 $\pm$ 0.36 (39)} 
& \heat{48}{0.68 $\pm$ 0.34 (43)}
& \\

& Full-batch    
& \graycell{1.07 $\pm$ 0.22 (100)} 
& \graycell{0.66 $\pm$ 0.17 (68)} 
& \graycell{0.75 $\pm$ 0.38 (20)} 
& \graycell{0.72 $\pm$ 0.33 (21)}
& \\

\midrule

\multirow{3}{*}{Fashion-MNIST}
& Our estimator 
& \heat{76}{1.02 $\pm$ 0.13 (100)} 
& \heat{22}{0.76 $\pm$ 0.094 (61)$^*$} 
& \heat{18}{0.74 $\pm$ 0.15 (24)$^*$} 
& \heat{4}{\textbf{0.72 $\pm$ 0.11 (21)}$^*$} 
& \multirow{3}{*}{\textbf{75.0\%}} \\

& Baseline      
& \heat{70}{0.97 $\pm$ 0.13 (100)$^*$} 
& \heat{29}{0.76 $\pm$ 0.098 (75)} 
& \heat{46}{0.80 $\pm$ 0.22 (29)} 
& \heat{40}{0.78 $\pm$ 0.21 (29)}
& \\

& Full-batch    
& \graycell{0.97 $\pm$ 0.12 (100)} 
& \graycell{0.77 $\pm$ 0.10 (64)} 
& \graycell{0.87 $\pm$ 0.32 (15)} 
& \graycell{0.82 $\pm$ 0.24 (14)}
& \\

\midrule

\multirow{3}{*}{CIFAR-10}
& Our estimator 
& \heat{73}{2.24 $\pm$ 0.084 (100)} 
& \heat{39}{\textbf{2.10 $\pm$ 0.099 (74)}$^*$} 
& \heat{59}{2.21 $\pm$ 0.13 (18)} 
& \heat{50}{2.21 $\pm$ 0.11 (17)$^*$} 
& \multirow{3}{*}{\textbf{50.0\%}} \\

& Baseline      
& \heat{62}{2.22 $\pm$ 0.074 (100)$^*$} 
& \heat{55}{2.12 $\pm$ 0.10 (99)} 
& \heat{55}{2.19 $\pm$ 0.12 (23)$^*$} 
& \heat{56}{2.19 $\pm$ 0.13 (24)}
& \\

& Full-batch    
& \graycell{2.22 $\pm$ 0.075 (100)} 
& \graycell{2.10 $\pm$ 0.10 (100)} 
& \graycell{2.24 $\pm$ 0.12 (11)} 
& \graycell{2.23 $\pm$ 0.11 (10)}
& \\

\midrule

\multirow{3}{*}{CIFAR-100}
& Our estimator 
& \heat{93}{4.67 $\pm$ 0.031 (78)} 
& \heat{26}{4.65 $\pm$ 0.024 (35)$^*$} 
& \heat{28}{4.65 $\pm$ 0.027 (3)} 
& \heat{18}{4.65 $\pm$ 0.025 (3)} 
& \multirow{3}{*}{\textbf{25.0\%}} \\

& Baseline      
& \heat{76}{4.67 $\pm$ 0.027 (99)$^*$} 
& \heat{31}{4.64 $\pm$ 0.025 (72)} 
& \heat{13}{4.64 $\pm$ 0.026 (9)$^*$} 
& \heat{2}{\textbf{4.64 $\pm$ 0.023 (8)}$^*$}
& \\

& Full-batch    
& \graycell{4.67 $\pm$ 0.028 (100)} 
& \graycell{4.64 $\pm$ 0.023 (66)} 
& \graycell{4.65 $\pm$ 0.026 (3)} 
& \graycell{4.65 $\pm$ 0.026 (3)}
& \\

\midrule

\multicolumn{2}{l}{\textbf{Column win rate}}
& \textbf{14.3\%}
& \textbf{85.7\%}
& \textbf{57.1\%}
& \textbf{85.7\%}
& \cellcolor{gray!20} \textbf{60.7\%} \\

\bottomrule
\end{tabular}
}
\end{table}

\section{Discussion}\label{Section::discussion}

The empirical results reveal dataset-dependent behavior for the proposed estimator, with performance patterns varying across datasets and optimizers. The model-assisted estimator generally improves generalization performance and reach their minimum loss in fewer epochs than standard mini-batch estimator. The proposed estimator also tend to reduce variance across runs, although this improvement is not observed in all settings. In controlled synthetic settings, both generalization performance and the L2-distance to the full-batch gradient behaved as intuitively expected, providing a validation of the model-assisted model optimization under well-specified conditions.

Notably, optimizer improvements do not always seem correlate with smaller L2-distance to the full-batch gradient. While L2-distance measures error magnitude to the population (full-batch) gradient, it does not seem to capture overall directional quality in the longer run. Estimators with larger norm errors relative to the full-batch gradient can still produce more reliable descent directions. This effect is especially relevant for adaptive methods such as Adam and AdamW, where gradients are rescaled using moment estimates. In these cases, the temporal consistency and directional structure of the gradient estimates may matter more than the magnitude of the gradient estimate at any single iteration.

On smaller and larger datasets (e.g., Airfoil Self-Noise, CIFAR-10/100), the benefits are more limited, though still generally positive. In the CIFAR settings in particular, model-assisted estimators tend to more closely track the performance of full-batch optimization, including its eventual degradation as full-batch optimization does not uniformly outperform stochastic methods, highlighting the beneficial role of noise for exploration and generalization.

Finally, the improvements using model-assisted estimator comes with additional computational overhead from constructing model-assisted estimators, introducing a trade-off between efficiency and accuracy. While the auxiliary model is typically cheap to evaluate, its quality directly affects the accuracy of the resulting gradient estimates. Given that a single gradient computation can be expensive, especially in large models, the effectiveness of the approach depends on whether the low-cost model can provide sufficiently accurate guidance to justify its use in practice.

\section{Conclusions}\label{Section::Conclusion}
    
In this work, we examined gradient variance reduction using a model-assisted estimator inspired by survey sampling theory. Across multiple datasets and optimizers, overall, model-assisted gradient estimation improved training stability, convergence speed, and generalization, with the strongest gains observed for momentum-based methods such as Adam, while vanilla SGD benefited less. At the same time, our results show that variance reduction alone is not sufficient to fully explain generalization, highlighting the need to balance variance control with beneficial stochasticity. We hope this work opens new directions for applying model-assisted estimator frameworks in ML optimization.

\printbibliography

@article{Nes83,
  author    = {Yu. E. Nesterov},
  title     = {A method of solving a convex programming problem with convergence rate $O\bigl(\frac{1}{k^2}\bigr)$},
  journal   = {Dokl. Akad. Nauk SSSR},
  year      = {1983},
  volume    = {269},
  number    = {3},
  pages     = {543--547},
  url       = {http://mi.mathnet.ru/dan46009},
}

@inproceedings{Kingma2015Adam,
  author    = {Diederik P. Kingma and Jimmy Ba},
  title     = {Adam: A Method for Stochastic Optimization},
  booktitle = {Proceedings of the 3rd International Conference on Learning Representations (ICLR)},
  year      = {2015},
  url       = {https://arxiv.org/abs/1412.6980}
}

@book{Boyd2004,
  author = {Boyd, Stephen and Vandenberghe, Lieven},
  description = {CiteULike},
  howpublished = {Hardcover},
  isbn = {0521833787},
  publisher = {{Cambridge University Press}},
  title = {Convex Optimization},
  year = {2004}
}

@InProceedings{Bottou2010,
author="Bottou, L{\'e}on",
editor="Lechevallier, Yves
and Saporta, Gilbert",
title="Large-Scale Machine Learning with Stochastic Gradient Descent",
booktitle="Proceedings of COMPSTAT'2010",
year="2010",
publisher="Physica-Verlag HD",
address="Heidelberg",
pages="177--186",
isbn="978-3-7908-2604-3"
}

@article{garrigos2023handbook,
  title={Handbook of convergence theorems for (stochastic) gradient methods},
  author={Garrigos, Guillaume and Gower, Robert Michael},
  journal={arXiv preprint arXiv:2301.11235},
  year={2023}
}

@article{Needell2014SGD,
  title={Stochastic gradient descent, weighted sampling, and the randomized Kaczmarz algorithm},
  author={Needell, Deanna and De Sa, Christopher and Tropp, Joel},
  journal={Mathematical Programming},
  volume={155},
  pages={549--573},
  year={2016}
}

@inproceedings{Zhao2015StochasticImportance,
  title={Stochastic Optimization with Importance Sampling for Regularized Loss Minimization},
  author={Zhao, Peilin and Zhang, Tong},
  booktitle={Proceedings of the 32nd International Conference on Machine Learning (ICML)},
  pages={1--9},
  year={2015}
}

@article{Alain2015VarianceReduction,
  title={Variance Reduction in SGD by Distributed Importance Sampling},
  author={Alain, Guillaume and Bengio, Yoshua and others},
  journal={arXiv preprint arXiv:1511.06481},
  year={2015}
}

@book{Cochran1977Sampling,
  title={Sampling Techniques},
  author={Cochran, William G.},
  edition={3rd},
  publisher={John Wiley \& Sons},
  year={1977}
}

@book{Lohr2009Sampling,
  title={Sampling: Design and Analysis},
  author={Lohr, Sharon L.},
  edition={2nd},
  publisher={Brooks/Cole},
  year={2009}
}

@inbook{Bottou1999,
author = {Bottou, L\'{e}on},
title = {On-line learning and stochastic approximations},
year = {1999},
isbn = {0521652634},
publisher = {Cambridge University Press},
address = {USA},
booktitle = {On-Line Learning in Neural Networks},
pages = {9–42},
numpages = {34}
}

@book{2012Alvarez,
author = {\'{A}lvarez, Mauricio A. and Rosasco, Lorenzo and Lawrence, Neil D.},
title = {Kernels for Vector-Valued Functions},
year = {2012},
isbn = {1601985584},
publisher = {Now Publishers Inc.},
address = {Hanover, MA, USA},
}

@PhdThesis{pahikkala2008phdthesis,
  Title                    = {New Kernel Functions and Learning Methods for Text and Data Mining},
  Author                   = {Tapio Pahikkala},
  School                   = {Turku Centre for Computer Science (TUCS)},
  Year                     = {2008},
  Address                  = {Turku, Finland}
}

@InProceedings{pmlr-v97-qian19b,
  title = 	 {{SGD}: General Analysis and Improved Rates},
  author =       {Gower, Robert Mansel and Loizou, Nicolas and Qian, Xun and Sailanbayev, Alibek and Shulgin, Egor and Richt{\'a}rik, Peter},
  booktitle = 	 {Proceedings of the 36th International Conference on Machine Learning},
  pages = 	 {5200--5209},
  year = 	 {2019},
  editor = 	 {Chaudhuri, Kamalika and Salakhutdinov, Ruslan},
  volume = 	 {97},
  series = 	 {Proceedings of Machine Learning Research},
  publisher =    {PMLR},
}

@InProceedings{Karimi2016,
author="Karimi, Hamed
and Nutini, Julie
and Schmidt, Mark",
editor="Frasconi, Paolo
and Landwehr, Niels
and Manco, Giuseppe
and Vreeken, Jilles",
title="Linear Convergence of Gradient and Proximal-Gradient Methods Under the Polyak-{\L}ojasiewicz Condition",
booktitle="Machine Learning and Knowledge Discovery in Databases",
year="2016",
publisher="Springer International Publishing",
address="Cham",
pages="795--811",
}

@article{Masters2018RevisitingSB,
  title={Revisiting Small Batch Training for Deep Neural Networks},
  author={Dominic Masters and Carlo Luschi},
  journal={ArXiv},
  year={2018},
  volume={abs/1804.07612},
  url={https://api.semanticscholar.org/CorpusID:5032969}
}

@misc{ruder2016overview,
  author = {Ruder, Sebastian},
  description = {An overview of gradient descent optimization algorithms},
  title = {An overview of gradient descent optimization algorithms.},
  url = {http://arxiv.org/abs/1609.04747},
  year = 2016
}

@inproceedings{2015-kingma,
  author = {Kingma, Diederik P. and Ba, Jimmy},
  booktitle = {ICLR (Poster)},
  editor = {Bengio, Yoshua and LeCun, Yann},
  ee = {http://arxiv.org/abs/1412.6980},
  title = {Adam: A Method for Stochastic Optimization.},
  url = {http://dblp.uni-trier.de/db/conf/iclr/iclr2015.html#KingmaB14},
  year = 2015
}

@inproceedings{AdamWCite,
  author       = {Ilya Loshchilov and
                  Frank Hutter},
  title        = {Decoupled Weight Decay Regularization},
  booktitle    = {7th International Conference on Learning Representations, {ICLR} 2019,
                  New Orleans, LA, USA, May 6-9, 2019},
  publisher    = {OpenReview.net},
  year         = {2019},
  url          = {https://openreview.net/forum?id=Bkg6RiCqY7},
  bibsource    = {dblp computer science bibliography, https://dblp.org}
}

@article{KeskarMNST16,
  author = {Keskar, Nitish Shirish and Mudigere, Dheevatsa and Nocedal, Jorge and Smelyanskiy, Mikhail and Tang, Ping Tak Peter},
  bibsource = {dblp computer science bibliography, http://dblp.org},
  journal = {CoRR},
  title = {On Large-Batch Training for Deep Learning: Generalization Gap and Sharp Minima.},
  url = {http://arxiv.org/abs/1609.04836},
  volume = {abs/1609.04836},
  year = 2016
}

@article{lecun-mnisthandwrittendigit-2010,
  author = {LeCun, Yann and Cortes, Corinna},
  howpublished = {http://yann.lecun.com/exdb/mnist/},
  lastchecked = {2016-01-14 14:24:11},
  title = {{MNIST} handwritten digit database},
  url = {http://yann.lecun.com/exdb/mnist/},
  username = {mhwombat},
  year = 2010
}

@misc{airfoil_self-noise_291,
  author       = {Brooks, Thomas and Pope, D. and Marcolini, Michael},
  title        = {{Airfoil Self-Noise}},
  year         = {1989},
  howpublished = {UCI Machine Learning Repository},
  note         = {{DOI}: https://doi.org/10.24432/C5VW2C}
}

@misc{appliances_energy_prediction_374,
  author       = {Candanedo, Luis},
  title        = {{Appliances Energy Prediction}},
  year         = {2017},
  howpublished = {UCI Machine Learning Repository},
  doi          = {10.24432/C5VC8G}
}

@misc{xiao2017fashionmnist,
  author = {Xiao, Han and Rasul, Kashif and Vollgraf, Roland},
  description = {Fashion-MNIST: a Novel Image Dataset for Benchmarking Machine Learning Algorithms},
  note = {cite arxiv:1708.07747Comment: Dataset is freely available at  https://github.com/zalandoresearch/fashion-mnist Benchmark is available at  http://fashion-mnist.s3-website.eu-central-1.amazonaws.com/},
  title = {Fashion-MNIST: a Novel Image Dataset for Benchmarking Machine Learning
  Algorithms},
  url = {http://arxiv.org/abs/1708.07747},
  year = 2017
}

@techreport{CIFAR10_and_100_Cite,
  title={Learning multiple layers of features from tiny images},
  author={Krizhevsky, Alex},
  year={2009},
  institution={University of Toronto}
}

@inproceedings{SVRGcite,
author = {Johnson, Rie and Zhang, Tong},
title = {Accelerating stochastic gradient descent using predictive variance reduction},
year = {2013},
publisher = {Curran Associates Inc.},
address = {Red Hook, NY, USA},
pages = {315–323},
numpages = {9},
series = {NIPS'13}
}

@article{model2003sarndal,
  asin = {0387406204},
  author = {Särndal, Carl-Erik and Swensson, Bengt and Wretman, Jan},
  description = {Amazon.com: Model Assisted Survey Sampling (Springer Series in Statistics): Books: Carl-Erik Särndal,Bengt Swensson,Jan Wretman},
  isbn = {0387406204},
  title = {Model Assisted Survey Sampling (Springer Series in Statistics)},
  typesource = {Simple CitationSource},
  url = {http://www.amazon.com/Assisted-Survey-Sampling-Springer-Statistics/dp/0387406204/sr=8-1/qid=1172587067/ref=pd_bbs_sr_1/103-2111122-6886251?ie=UTF8&s=books},
  year = 2003
}

@article{SAGcite,
author = {Schmidt, Mark and Le Roux, Nicolas and Bach, Francis},
title = {Minimizing finite sums with the stochastic average gradient},
year = {2017},
issue_date = {March     2017},
publisher = {Springer-Verlag},
address = {Berlin, Heidelberg},
volume = {162},
number = {1–2},
issn = {0025-5610},
url = {https://doi.org/10.1007/s10107-016-1030-6},
doi = {10.1007/s10107-016-1030-6},
journal = {Math. Program.},
month = mar,
pages = {83–112},
numpages = {30}
}

@inproceedings{SAGAcite,
author = {Defazio, Aaron and Bach, Francis and Lacoste-Julien, Simon},
title = {SAGA: a fast incremental gradient method with support for non-strongly convex composite objectives},
year = {2014},
publisher = {MIT Press},
booktitle = {Proceedings of the 28th International Conference on Neural Information Processing Systems - Volume 1},
pages = {1646–1654},
numpages = {9},
series = {NIPS'14}
}

@article{Dumelle2022,
author = {Dumelle, Michael and Higham, Matt and Ver Hoef, Jay M. and Olsen, Anthony R. and Madsen, Lisa},
title = {A comparison of design-based and model-based approaches for finite population spatial sampling and inference},
journal = {Methods in Ecology and Evolution},
volume = {13},
number = {9},
pages = {2018-2029},
doi = {https://doi.org/10.1111/2041-210X.13919},
year = {2022}
}

@inproceedings{NIPS2013_9766527f,
 author = {Wang, Chong and Chen, Xi and Smola, Alexander and Xing, Eric},
 booktitle = {Advances in Neural Information Processing Systems},
 editor = {C.J. Burges and L. Bottou and M. Welling and Z. Ghahramani and K. Weinberger},
 pages = {},
 publisher = {Curran Associates, Inc.},
 title = {Variance Reduction for Stochastic Gradient Optimization},
 url = {https://proceedings.neurips.cc/paper_files/paper/2013/file/9766527f2b5d3e95d4a733fcfb77bd7e-Paper.pdf},
 volume = {26},
 year = {2013}
}

@book{Goodfellow-et-al-2016,
  author = {Goodfellow, Ian and Bengio, Yoshua and Courville, Aaron},
  note = {Book in preparation for MIT Press},
  publisher = {MIT Press},
  title = {Deep Learning},
  url = {http://www.deeplearningbook.org},
  year = 2016
}

@inproceedings{Smith2020,
author = {Smith, Samuel L. and Elsen, Erich and De, Soham},
title = {On the generalization benefit of noise in stochastic gradient descent},
year = {2020},
publisher = {JMLR.org},
booktitle = {Proceedings of the 37th International Conference on Machine Learning},
articleno = {840},
numpages = {10},
series = {ICML'20}
}

@misc{
zhu2019the,
title={The Anisotropic Noise in Stochastic Gradient Descent: Its Behavior of Escaping from Minima and Regularization Effects},
author={Zhanxing Zhu and Jingfeng Wu and Bing Yu and Lei Wu and Jinwen Ma},
year={2019},
url={https://openreview.net/forum?id=H1M7soActX},
}

@article{SKLEARNcite,
author = {Pedregosa, Fabian and Varoquaux, Ga\"{e}l and Gramfort, Alexandre and Michel, Vincent and Thirion, Bertrand and Grisel, Olivier and Blondel, Mathieu and Prettenhofer, Peter and Weiss, Ron and Dubourg, Vincent and Vanderplas, Jake and Passos, Alexandre and Cournapeau, David and Brucher, Matthieu and Perrot, Matthieu and Duchesnay, \'{E}douard},
title = {Scikit-learn: Machine Learning in Python},
year = {2011},
issue_date = {2/1/2011},
publisher = {JMLR.org},
volume = {12},
number = {null},
issn = {1532-4435},
journal = {J. Mach. Learn. Res.},
month = nov,
pages = {2825–2830},
numpages = {6}
}

\newpage

\section*{Appendix}

\subsection*{Kernel ridge regression in gradient modeling}
Given the set \(\gradientPopulation_{\firstSubsampleIndexSet
} := \{\gradientDataPointSymbol_i: i \in \firstSubsampleIndexSet\}\subset\gradientPopulation\), the regularized least squares problem in a reproducing kernel Hilbert space (RKHS) is defined as:
\begin{equation}\label{equation:RLS_problem_definition}
\sum_{j=1}^{d} \frac{1}{\firstSubsampleIndexSetSize} \sum_{i\in\firstSubsampleIndexSet} \left( \gradientModelSymbol_j(\xdata_i) - \gradientSymbol_{ji} \right)^2
+ \lambda \|\gradientModelSymbol\|_{\mathbf{K}}^2,    
\end{equation}
where \(\firstSubsampleIndexSetSize = |\firstSubsampleIndexSet|\), \(\gradientSymbol_{ji}\) is the \(j\)th component of the \(i\)th gradient, and \(\gradientModelSymbol_j(\xdata_i)\) is the corresponding model estimate evaluated at \(\xdata_i\), and \(\lambda > 0\) is the regularization coefficient. By the representer theorem, the solution admits the form
\begin{equation}
\gradientModelSymbol(\xdata) \;=\; \sum_{i\in\firstSubsampleIndexSet} \mathbf{K}(\xdata_i, \xdata)\,\mathbf{c}_i,
\end{equation}
where $\mathbf{K}: \mathcal{X}\times \mathcal{X} \to \mathbb{R}^{d \times d}$ is a matrix-valued kernel and $\mathbf{c}_i \in \mathbb{R}^d$ are the coefficient vectors associated with sample $i$. The function \(\mathbf{K}\) corresponds to a positive semi-definite matrix with each component defined as
\begin{equation}
(\mathbf{K}(\xdata,\xdata'))_{d,d'}=k(\xdata, \xdata')\,k_T(d,d'),
\end{equation}
where \( k(\xdata, \xdata')\) is a scalar kernel acting on the input space \(\mathcal{X}\), 
and \( k_T(d,d') \) is a scalar kernel acting on the task (or output) indices that captures correlations between output dimensions. 
For tractability, we make the simplifying assumption that the output dimensions are uncorrelated, 
which corresponds to \( k_T(d,d') = \delta_{d,d'} \), where \(\delta_{d,d'}\) denotes the Kronecker delta. 
Under this assumption, each block of the operator-valued kernel reduces to a scaled identity,
\[
\mathbf{K}(\xdata,\xdata') = k(\xdata,\xdata')\,\mathbf{I}_d,
\]
and the full Gram matrix over \(\firstSubsampleIndexSetSize\) samples can be expressed compactly as
\[
\mathbf{K}(\mathbf{X},\mathbf{X}) = \mathbf{I}_d \otimes \mathbf{K}_X,
\]
where \((\mathbf{K}_X)_{ij} = k(\xdata_i, \xdata_j),i,j\in\firstSubsampleIndexSet\), 
\(\mathbf{K}_X\in\mathbb{R}^{\firstSubsampleIndexSetSize\times \firstSubsampleIndexSetSize}\), 
and consequently \(\mathbf{K}(\mathbf{X},\mathbf{X}) \in \mathbb{R}^{(\firstSubsampleIndexSetSize d)\times(\firstSubsampleIndexSetSize d)}\). It is clear that with the assumption of uncorrelated outputs, \(\mathbf{K}(\mathbf{X},\mathbf{X})\) is block diagonal. The solution that minimizes~\ref{equation:RLS_problem_definition} is then given by
\begin{equation}\label{Equation::krr_solution}
    \overline{\mathbf{c}} = \left(\mathbf{K}(\mathbf{X}, \mathbf{X}) + \lambda \firstSubsampleIndexSetSize\,\mathbf{I}\right)^{-1}\,\overline{\ydata},
\end{equation}
where the stacked vectors 
\(\overline{\mathbf{c}}\) and \(\overline{\mathbf{y}}\) are obtained by vectorizing the 
coefficient and output matrices 
\(\mathbf{C} = [\,\mathbf{c}_i\,]_{i\in\firstSubsampleIndexSet}^\top\) and 
\(\mathbf{Y} = [\,\mathbf{y}_i\,]_{i\in\firstSubsampleIndexSet}^\top\), respectively, 
with \(\mathbf{c}_i, \mathbf{y}_i \in \mathbb{R}^{d\times 1}\) denoting the coefficient and output vectors for sample \(i\in\firstSubsampleIndexSet\). In other words, \(\overline{\mathbf{c}} = \operatorname{vec}(\mathbf{C})\) and 
\(\overline{\mathbf{y}} = \operatorname{vec}(\mathbf{Y})\),
where \(\operatorname{vec}(\cdot)\) stacks the columns of a matrix into a single column vector,
so that all elements corresponding to the first output dimension come first, followed by the second, and so on.

Thanks to the Kronecker structure, the inverse in Equation \ref{Equation::krr_solution} can be computed efficiently without explicitly forming the 
\((\firstSubsampleIndexSetSize d)\times(\firstSubsampleIndexSetSize d)\) matrix. 
Using the property
\[
(\mathbf{I}_d \otimes \left(\mathbf{K}_X + \lambda \firstSubsampleIndexSetSize\,\mathbf{I}_{\firstSubsampleIndexSetSize}\right))^{-1}
= \mathbf{I}_d \otimes (\mathbf{K}_X + \lambda \firstSubsampleIndexSetSize\,\mathbf{I}_{\firstSubsampleIndexSetSize})^{-1},
\]
we obtain
\[
\overline{\mathbf{c}} 
= \left(\mathbf{I}_d \otimes (\mathbf{K}_X + \lambda \firstSubsampleIndexSetSize\,\mathbf{I}_{\firstSubsampleIndexSetSize})^{-1}\right)\,\overline{\ydata},
\]

Thus, only the \(\firstSubsampleIndexSetSize\times \firstSubsampleIndexSetSize\) matrix 
\((\mathbf{K}_X + \lambda \firstSubsampleIndexSetSize\,\mathbf{I}_{\firstSubsampleIndexSetSize})\) needs to be inverted,
which greatly reduces the computational cost. Furthermore, the output dimensions coefficients can be solved in parallel due to our assumption of uncorrelated outputs. Let us denote \(\mathbf{A} = (\mathbf{K}_X + \lambda \firstSubsampleIndexSetSize\,\mathbf{I}_{\firstSubsampleIndexSetSize})^{-1}\) and then the model coefficients are solved by \(\mathbf{C} = \mathbf{A}\mathbf{Y}\). The prediction model, evaluated at new point \(\xdata^*\), can then be written compactly as
\begin{equation}\label{Equation::unweighted_krr}
\gradientModelSymbol(\xdata^*) = \mathbf{C}^\top \mathbf{k}_{\xdata^*} 
= \mathbf{Y}^\top \mathbf{A}\,\mathbf{k}_{\xdata^*}= \sum_{i\in\firstSubsampleIndexSet} k(\xdata_i, \xdata^*)\,\mathbf{c}_i,
\end{equation}
where \(\mathbf{k}_{\xdata^*}=\left[k(\xdata_i, \xdata^*)\right]^\top, \; i\in\firstSubsampleIndexSet\).

\subsection*{Input space effect on model-assisted gradient estimation}

\begin{figure}[tbh]
\centering

\begin{subfigure}{0.32\textwidth}
  \centering
  \includegraphics[width=\linewidth]{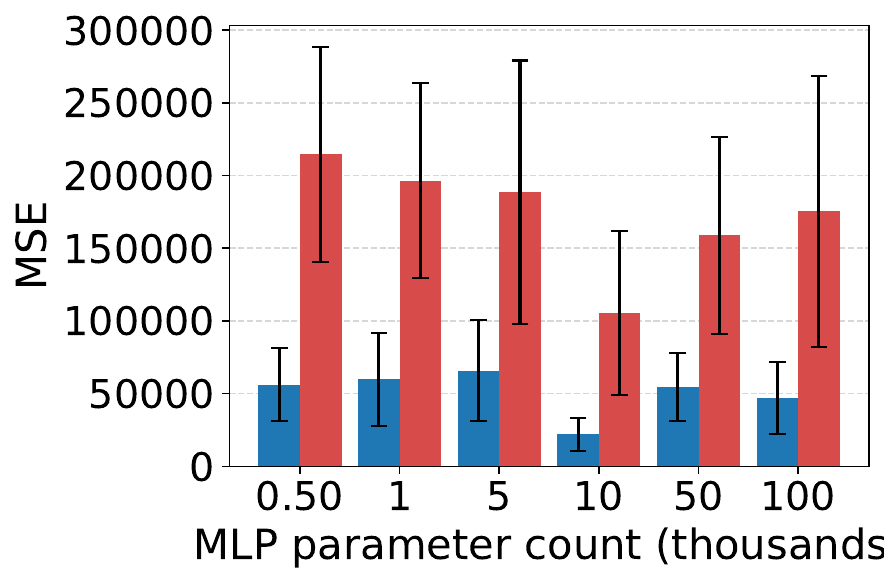}
  \caption{$m=5$, $\rho=0.7$}
\end{subfigure}
\hfill
\begin{subfigure}{0.32\textwidth}
  \centering
  \includegraphics[width=\linewidth]{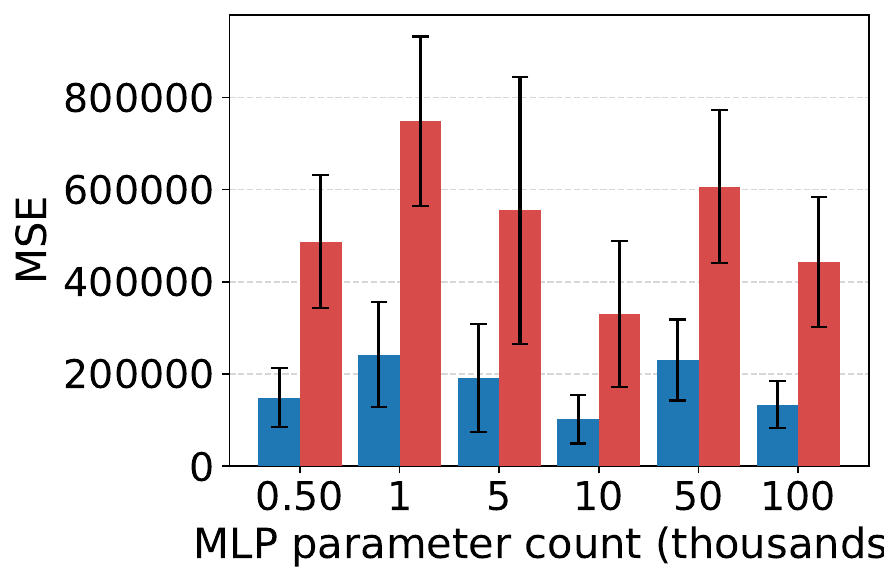}
  \caption{$m=5$, $\rho=0.5$}
\end{subfigure}
\hfill
\begin{subfigure}{0.32\textwidth}
  \centering
  \includegraphics[width=\linewidth]{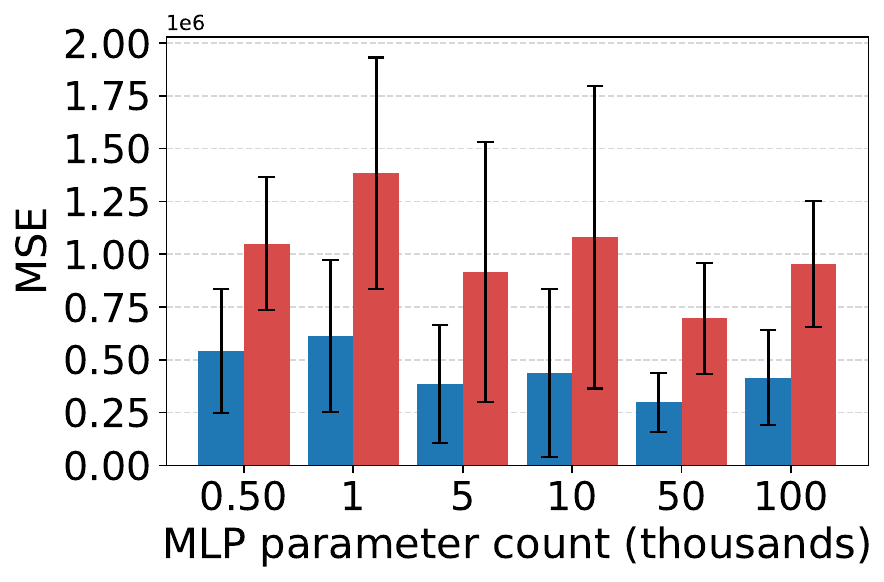}
  \caption{$m=5$, $\rho=0.3$}
\end{subfigure}

\vspace{0.4cm}

\begin{subfigure}{0.32\textwidth}
  \centering
  \includegraphics[width=\linewidth]{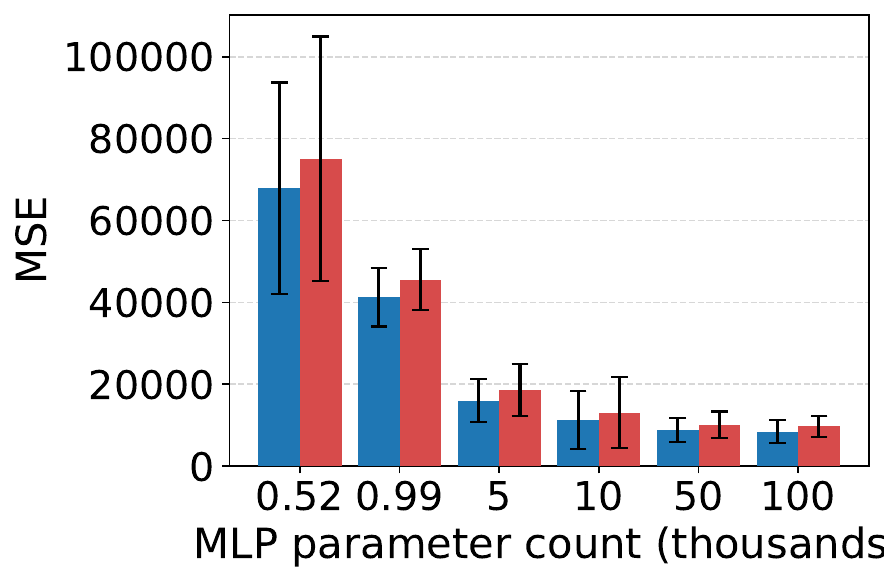}
  \caption{$m=50$, $\rho=0.7$}
\end{subfigure}
\hfill
\begin{subfigure}{0.32\textwidth}
  \centering
  \includegraphics[width=\linewidth]{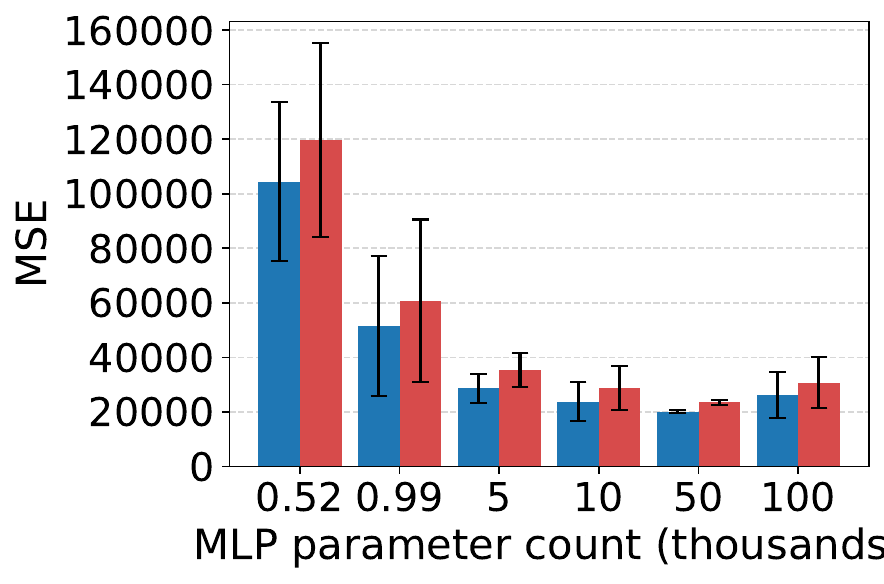}
  \caption{$m=50$, $\rho=0.5$}
\end{subfigure}
\hfill
\begin{subfigure}{0.32\textwidth}
  \centering
  \includegraphics[width=\linewidth]{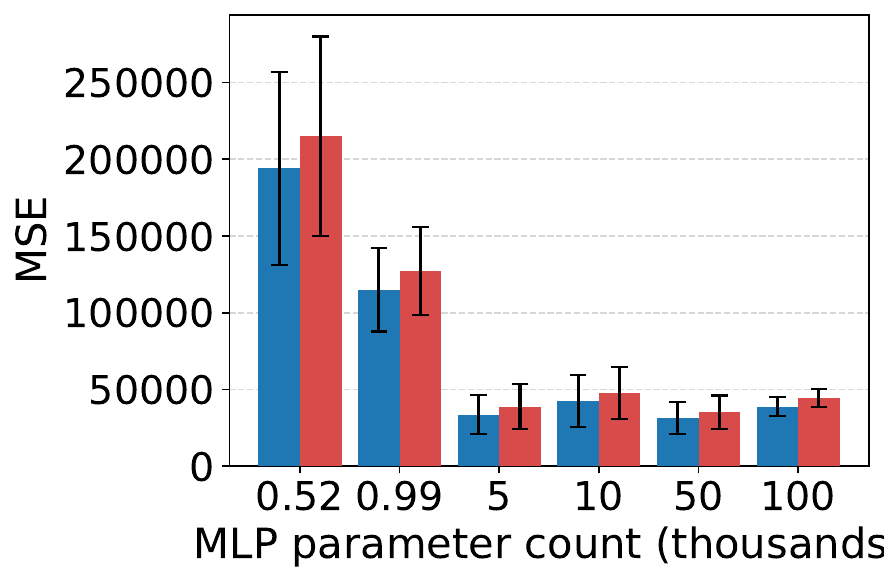}
  \caption{$m=50$, $\rho=0.3$}
\end{subfigure}

\vspace{0.4cm}

\begin{subfigure}{0.32\textwidth}
  \centering
  \includegraphics[width=\linewidth]{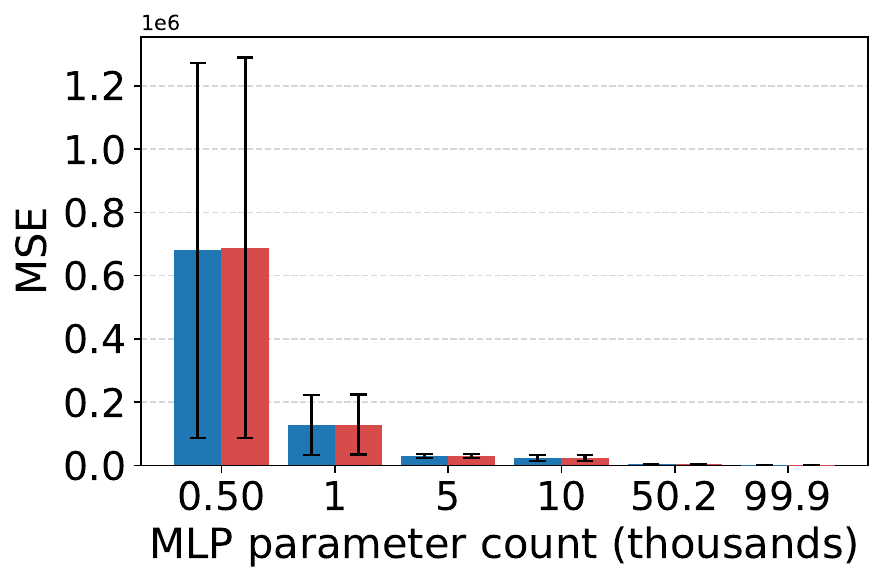}
  \caption{$m=500$, $\rho=0.7$}
\end{subfigure}
\hfill
\begin{subfigure}{0.32\textwidth}
  \centering
  \includegraphics[width=\linewidth]{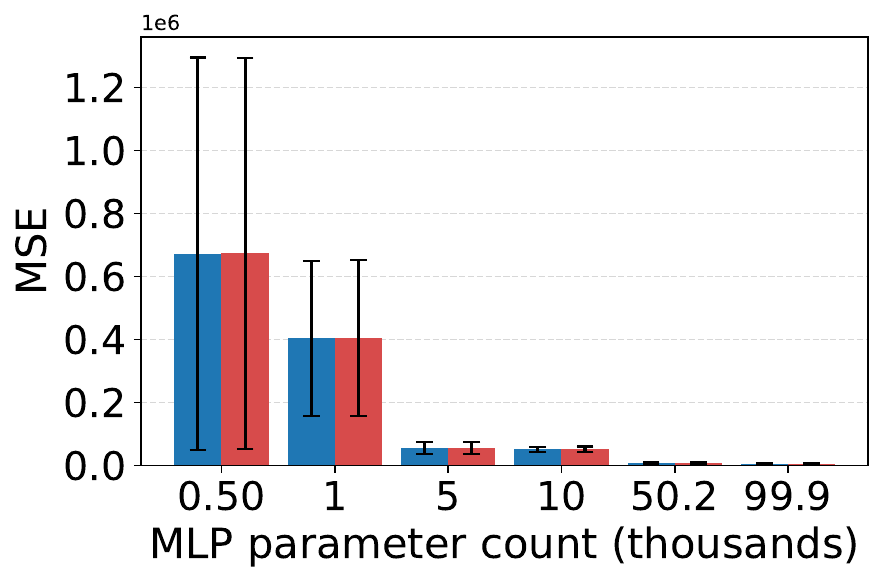}
  \caption{$m=500$, $\rho=0.5$}
\end{subfigure}
\hfill
\begin{subfigure}{0.32\textwidth}
  \centering
  \includegraphics[width=\linewidth]{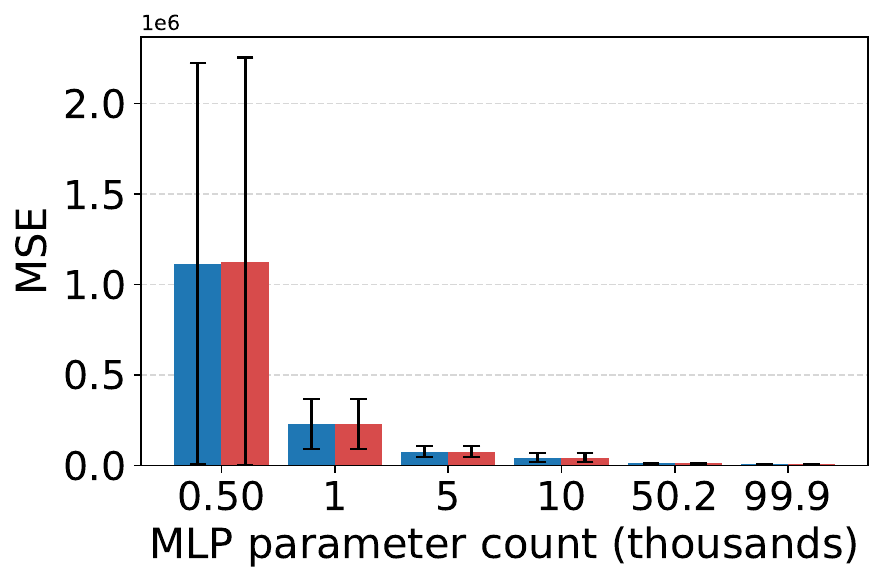}
  \caption{$m=500$, $\rho=0.3$}
\end{subfigure}

\caption{
Comparison of the proposed model-assisted gradient estimator and the standard uniform mini-batch gradient estimator in estimating the true population gradient over 100 randomly generated synthetic regression datasets. The plots report the mean squared error (MSE) between the estimated gradients and the true population gradient for the model-assisted estimator (blue) and the uniform mini-batch estimator (red). Experiments are conducted using multilayer perceptron models with varying parameter counts. The input dimensionality is varied as $m \in \{5,50,500\}$, while the sampling fraction used for gradient estimation is varied as $\rho \in \{0.7,0.5,0.3\}$, where $\rho$ denotes the fraction of the population sample used for gradient estimation.
}
\label{fig:input_space_effect_on_magd}
\end{figure}

In Figure \ref{fig:input_space_effect_on_magd} we compare the average performance of the proposed model-assisted gradient estimator against the standard uniform mini-batch gradient estimator in estimating the true population gradient over 100 randomly generated synthetic regression datasets. The plots show the mean squared error (MSE) between the estimated gradients and the true population gradient for the model-assisted estimator (blue) and the uniform mini-batch estimator (red). The experiments are performed using multilayer perceptron models with varying parameter counts. Furthermore, the input dimensionality is varied across 5, 50, and 500 features, and the fraction of sampled data (of population) used for gradient estimation is set to 0.7, 0.5, and 0.3. As the gradient prediction model \(q\), we utilize kernel ridge regression (KRR) with an RBF kernel. To mitigate unstable extrapolation effects, the model prediction is constrained to decay toward zero whenever the queried parameter point lies sufficiently far from the gradient model training samples in the input space.

The plots illustrate a clear benefit of using model-assisted gradient estimation especially when the input space size is smaller, but decreases clearly as the input space size decreases. The number of MLP seems to have relatively small effect in the difference in performance between model-assisted and mini-batch gradients. The synthetic results suggest, as expected by the COD effect, that the input space has the clearest effect on the performance of the model-assisted estimator.

\subsection*{Computational complexity analysis with SGD and KRR model}
In order for the optimization of ML models to be efficient using model-assisted approach, the added computational complexity needs to be outweighted in terms of generalization error and/or required epochs. The computational overhead of the proposed model-assisted gradient estimator arises mainly from the gradient model training used to predict unsampled gradient contributions. In a kernel ridge regression (KRR, e.g. \cite{pahikkala2008phdthesis}) case, which we apply in this study, requires solving a dense linear system involving the kernel matrix, resulting in \(\mathcal{O}(\subsampleSize^3)\) for training the model, memory complexity of \(\mathcal{O}(\subsampleSize^2)\) and predicting the population gradients \(\mathcal{O}(\subsampleSize\populationSize)\). However, in our setting the number of sampled gradients used for constructing the predictor is usually much smaller than the whole population size, i.e. \(\subsampleSize \ll \populationSize\), making the computational overhead manageable. Common used batch sizes are in powers of two such as \(n\in \{16, 32, 64, 128\}\) which are significantly lower than usual full population (i.e., dataset) sizes. Smaller batch sizes are also recommended for better generalization performance \cite{Masters2018RevisitingSB}.

Let \(E_{\mathrm{SGD}}\) denote the total number of epochs required by standard SGD to reach a given optimization criterion for the loss function \(\lossSymbol\). Let \(E_{\mathrm{MA}}\) denote the corresponding number of epochs required by the model-assisted optimizer. Furthermore, let \(\populationSize\) be the population dataset size, \(\subsampleSize\) the minibatch size, and \(d\) the computational cost of processing one sample gradient.

The total computational complexity of standard SGD is
\begin{equation}
C_{\mathrm{SGD}}
=
\mathcal{O}\!\left(E_{\mathrm{SGD}}\populationSize d\right),
\end{equation}
since each epoch processes all \(\populationSize\) samples and each individual data sample incurs computational cost \(\mathcal{O}(d)\). For the proposed model-assisted gradient estimator, additional computational overhead is introduced through KRR training and prediction. The added worst-case (\(n_1=\subsampleSize\)) per-iteration computational complexity is approximately
\begin{equation}
\mathcal{O}\!\left(\subsampleSize^3+\subsampleSize\populationSize\right),
\end{equation}
where \(\mathcal{O}(\subsampleSize^3)\) corresponds to inversion/training of the kernel system and \(\mathcal{O}(\subsampleSize\populationSize)\) corresponds to predicting gradients for the remaining dataset points.

Thus, the total computational complexity of the KRR-assisted optimizer is
\begin{equation}
C_{\mathrm{MA}}
=
\mathcal{O}\!\left(
E_{\mathrm{MA}}
\left[
\populationSize d
+
\frac{\populationSize}{\subsampleSize}
\left(
\subsampleSize^3+\subsampleSize\populationSize
\right)
\right]
\right).
\end{equation}

Equivalently, this can be simplified as
\begin{equation}
C_{\mathrm{MA}}
=
\mathcal{O}\!\left(
E_{\mathrm{MA}}\populationSize
\left[
d+\subsampleSize^2+\populationSize
\right]
\right).
\end{equation}

The KRR-assisted optimizer is computationally advantageous when \(C_{\mathrm{MA}}<C_{\mathrm{SGD}}\) which yields straightforwardly a condition:
\begin{equation}
\frac{E_{\mathrm{MA}}}{E_{\mathrm{SGD}}}
<
\frac{d}{d+\subsampleSize^2+\populationSize}.
\end{equation}

The left-hand side represents the fraction of SGD epochs required by the
model-assisted method, while the right-hand side gives the maximum
allowable epoch ratio for the KRR-assisted optimizer to remain
computationally cheaper than SGD. When \(d\) is large relative to \(\subsampleSize^2+\populationSize\),
the cost of ordinary gradient computation dominates the comparison. In
this case, the additional KRR overhead is relatively small compared with
the baseline cost of SGD, so a moderate reduction in epochs is enough to provide computational advantage. In contrast, when either
\(\subsampleSize\) or \(\populationSize\) is large relative to \(d\), the overhead caused by KRR becomes more significant: the \(\subsampleSize^2\) term reflects
the cost associated with training the KRR model, while the
\(\populationSize\) term reflects the cost of predicting gradients over
the population. Consequently, the KRR-assisted optimizer must achieve a
larger reduction in epochs to offset these added training and prediction
costs.

\subsection*{Result figures for SGD-M and Adam}

\begin{figure*}[t]
\centering
\setlength{\tabcolsep}{3pt}
\renewcommand{\arraystretch}{0.8}

\begin{tabular}{>{\centering\arraybackslash}m{0.04\textwidth}
                >{\centering\arraybackslash}m{0.29\textwidth}
                >{\centering\arraybackslash}m{0.29\textwidth}
                >{\centering\arraybackslash}m{0.29\textwidth}}

& \textbf{Batch size 10} & \textbf{Batch size 50} & \textbf{Batch size 100} \\

\rotatebox[origin=c]{90}{\footnotesize\textbf{Synthetic}} &
\includegraphics[valign=m,width=0.29\textwidth]{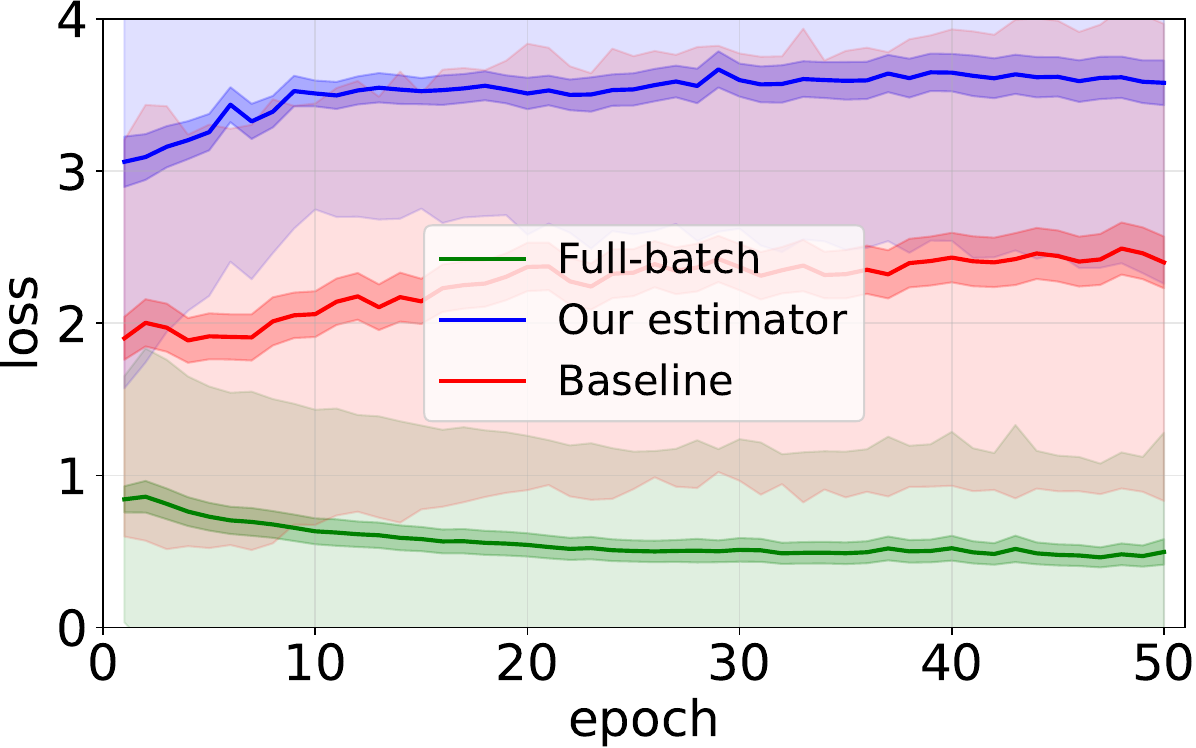} &
\includegraphics[valign=m,width=0.29\textwidth]{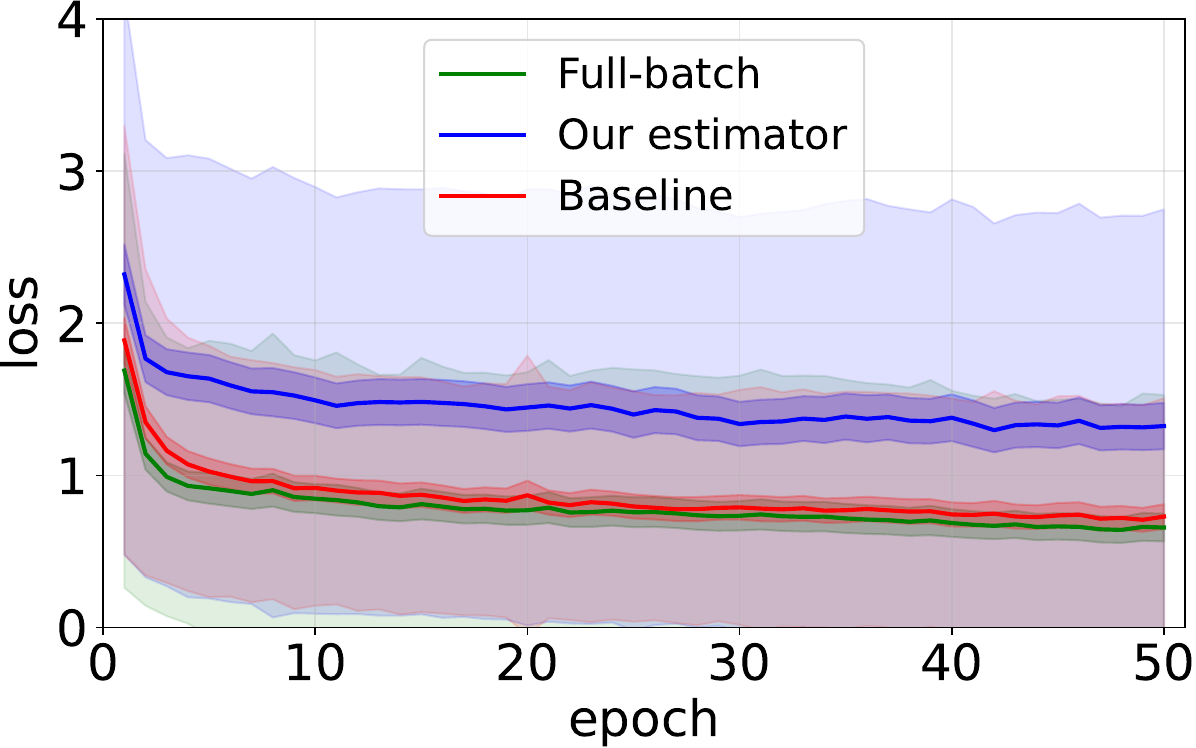} &
\includegraphics[valign=m,width=0.29\textwidth]{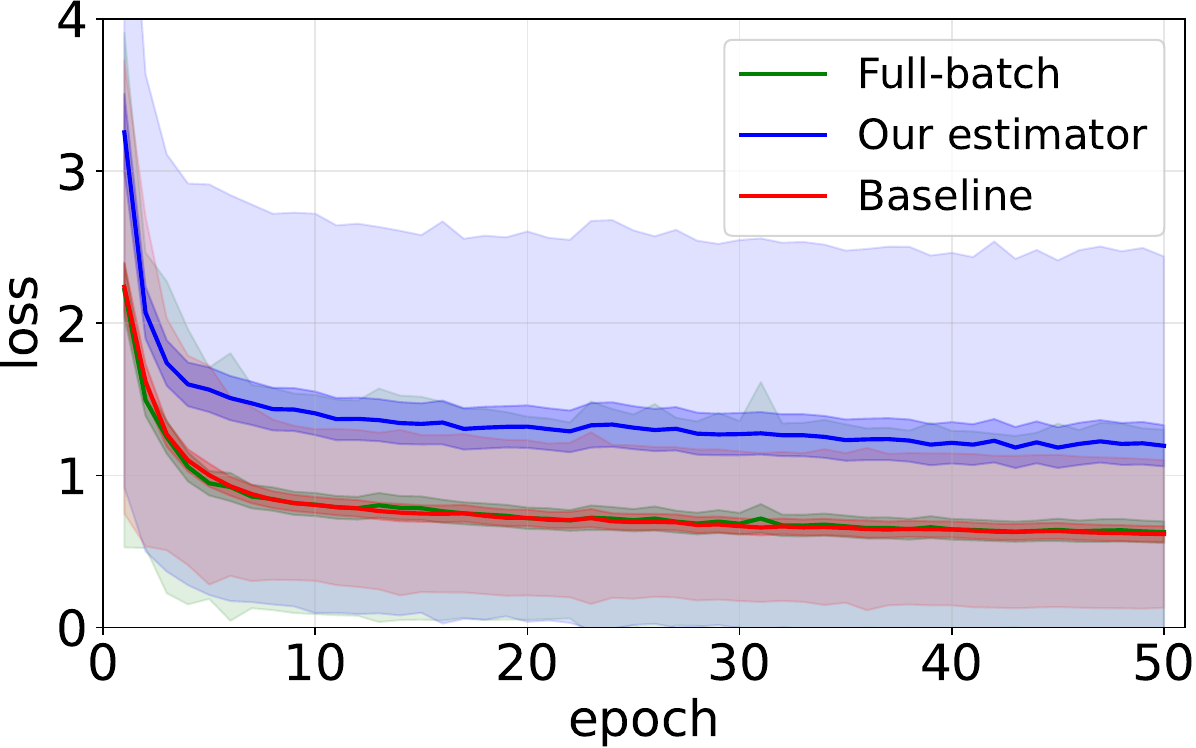} \\

\rotatebox[origin=c]{90}{\footnotesize\textbf{Airfoil self-noise}} &
\includegraphics[valign=m,width=0.29\textwidth]{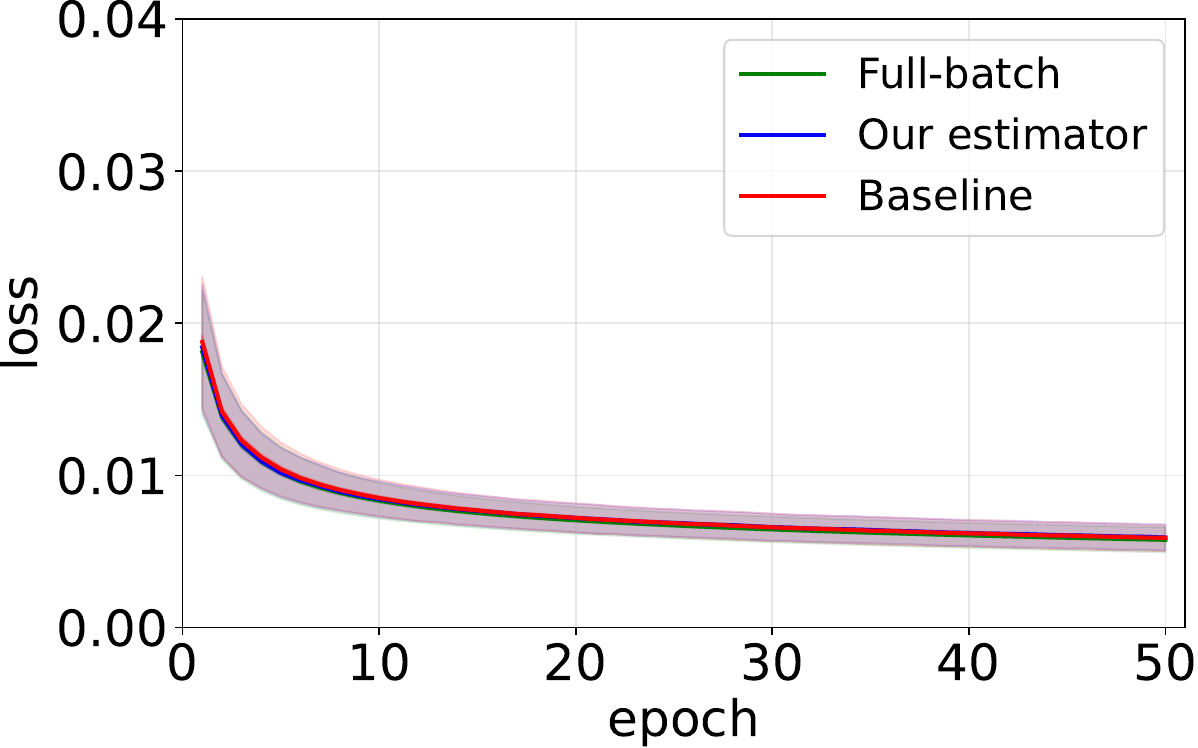} &
\includegraphics[valign=m,width=0.29\textwidth]{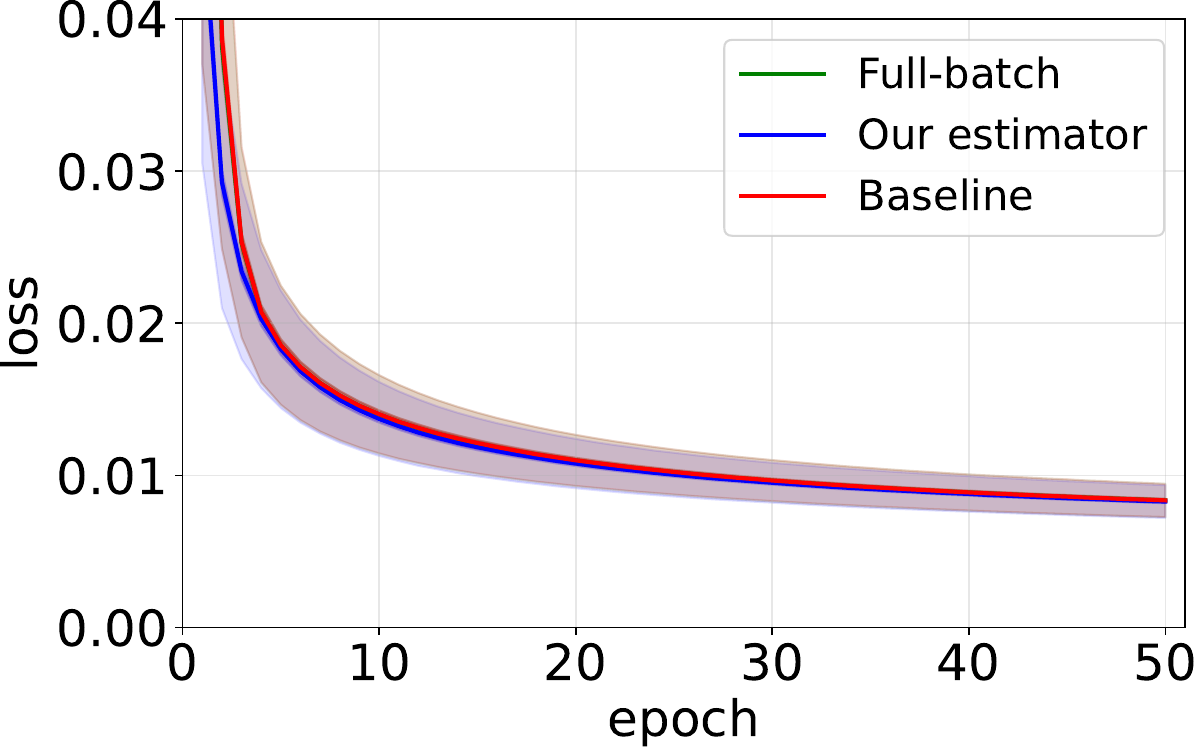} &
\includegraphics[valign=m,width=0.29\textwidth]{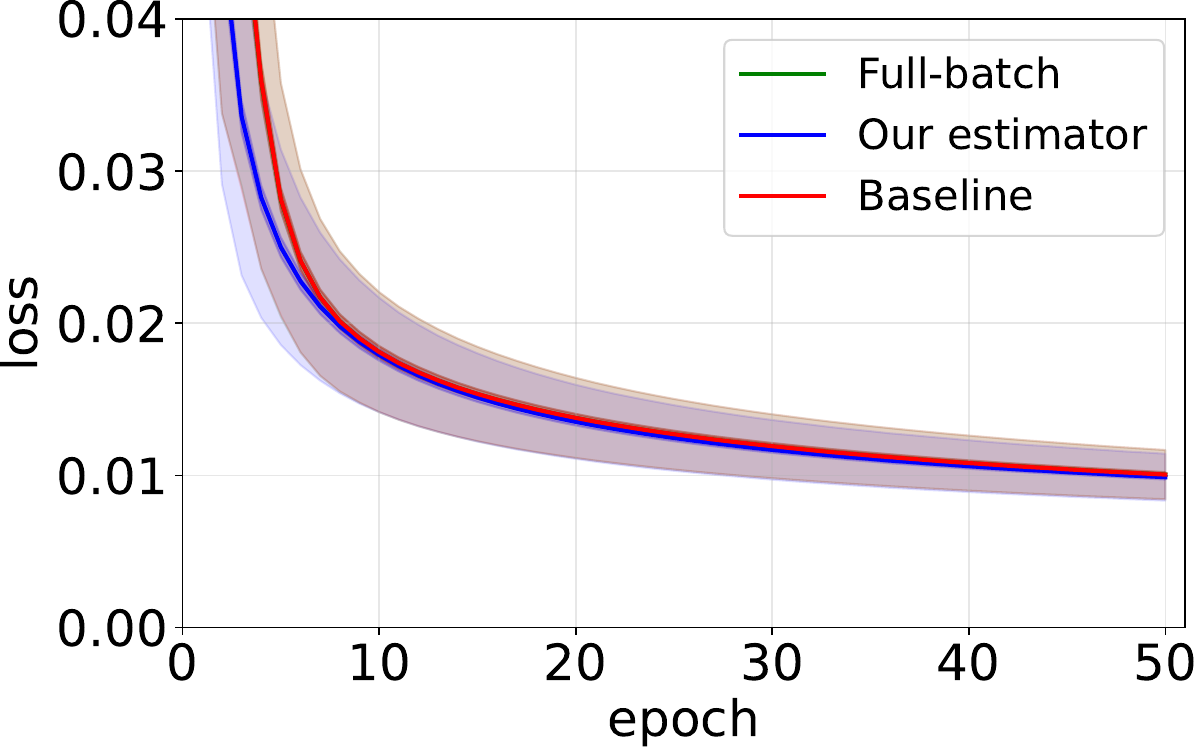} \\

\rotatebox[origin=c]{90}{\footnotesize\textbf{Appliances energy}} &
\includegraphics[valign=m,width=0.29\textwidth]{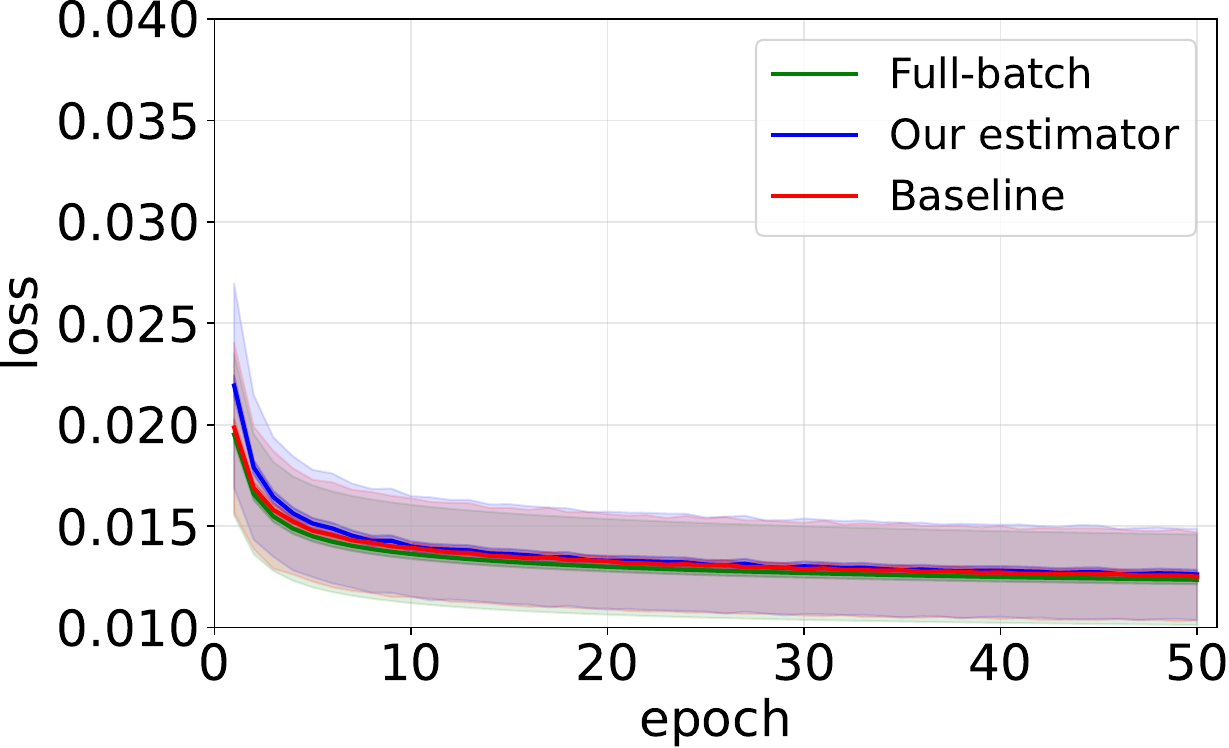} &
\includegraphics[valign=m,width=0.29\textwidth]{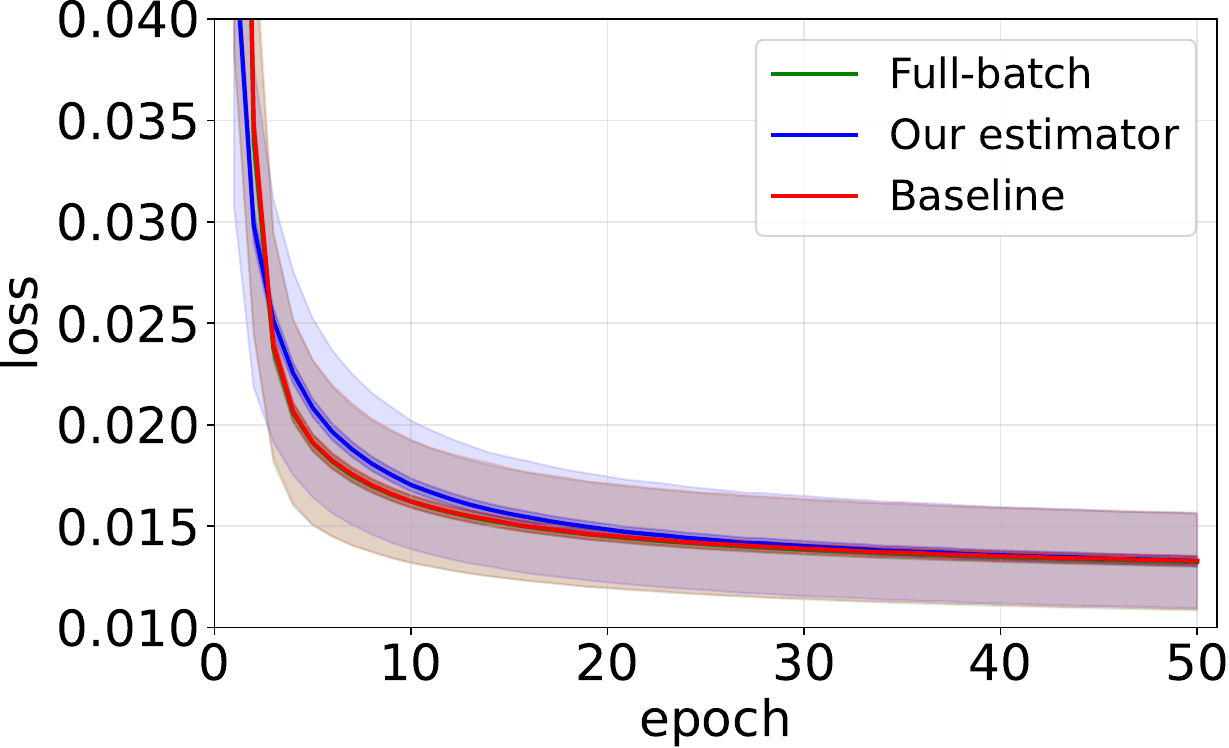} &
\includegraphics[valign=m,width=0.29\textwidth]{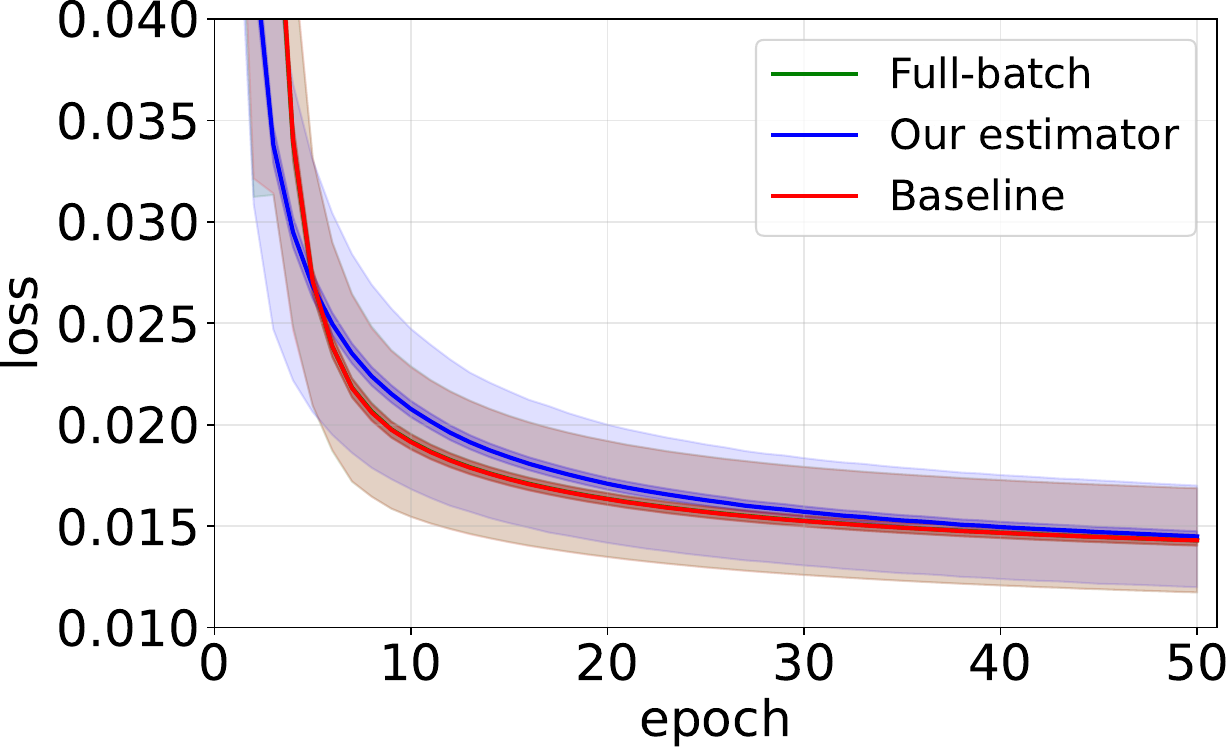} \\

\rotatebox[origin=c]{90}{\footnotesize\textbf{MNIST}} &
\includegraphics[valign=m,width=0.29\textwidth]{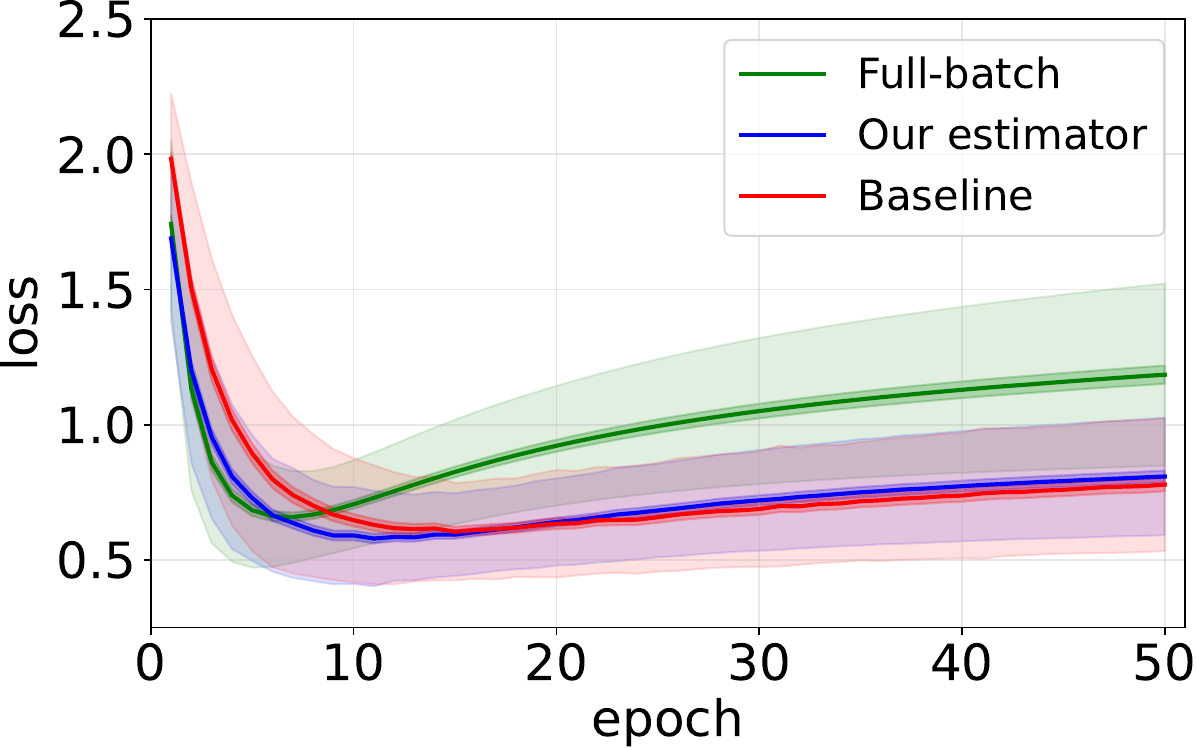} &
\includegraphics[valign=m,width=0.29\textwidth]{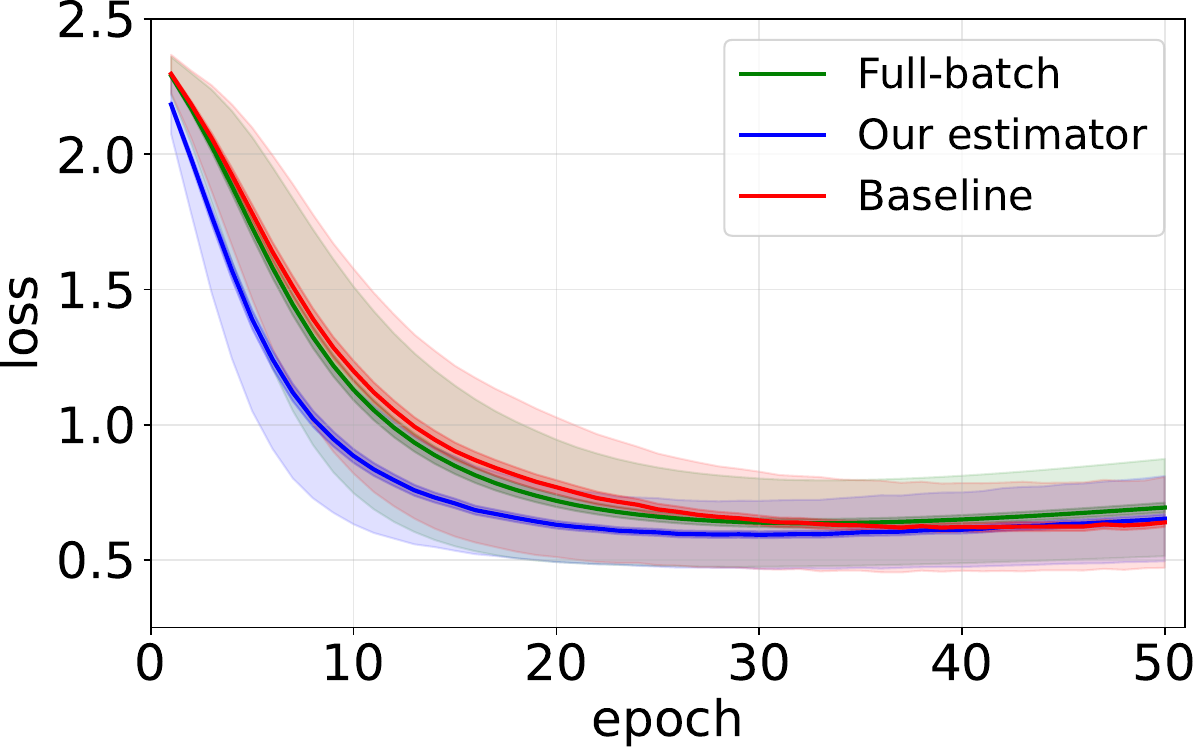} &
\includegraphics[valign=m,width=0.29\textwidth]{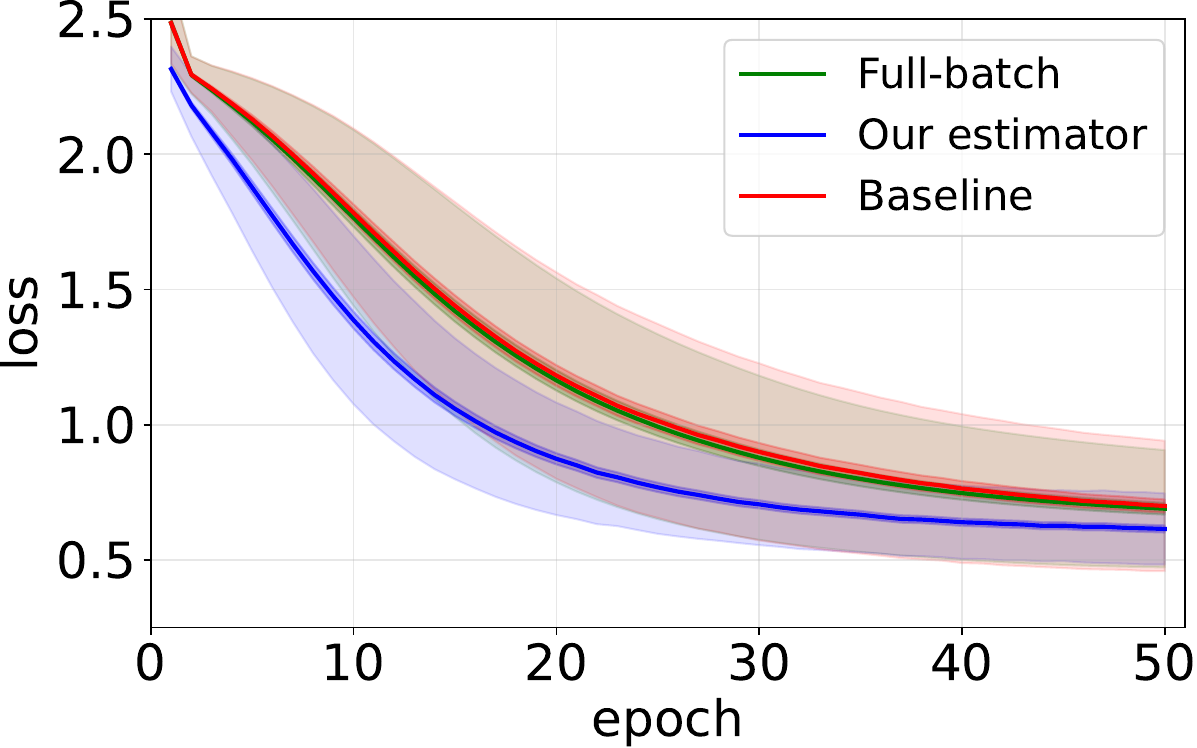} \\

\rotatebox[origin=c]{90}{\footnotesize\textbf{Fashion-MNIST}} &
\includegraphics[valign=m,width=0.29\textwidth]{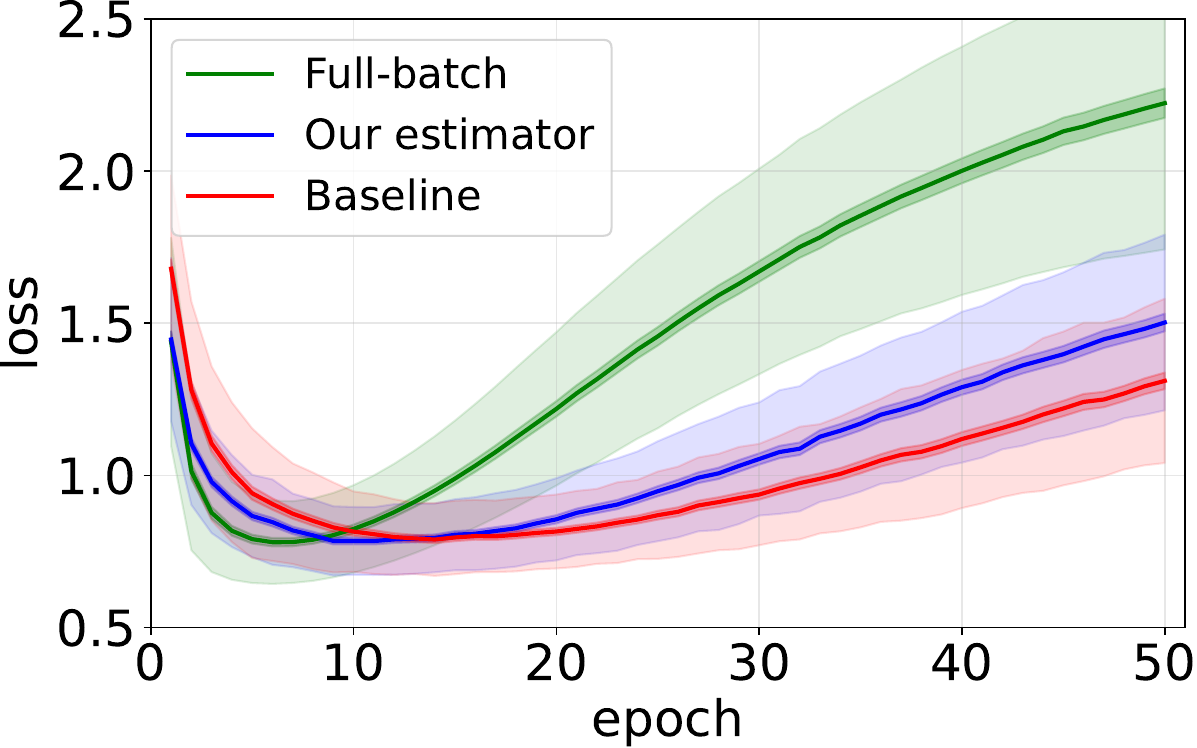} &
\includegraphics[valign=m,width=0.29\textwidth]{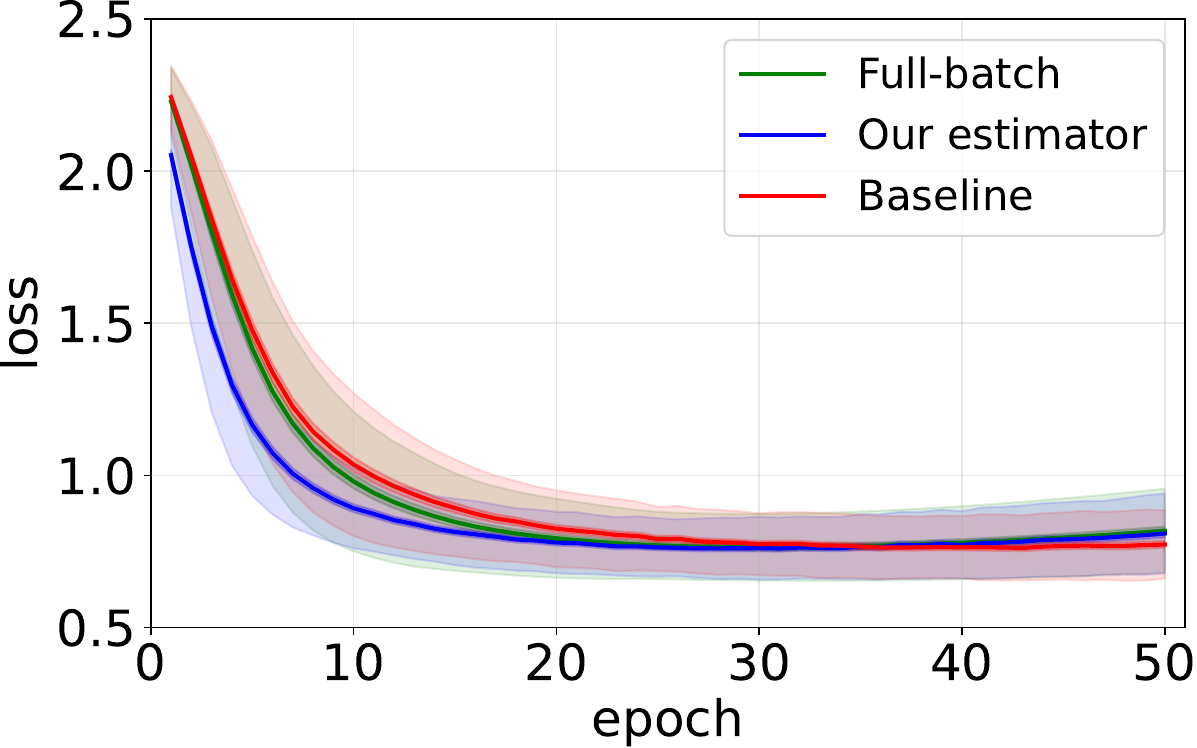} &
\includegraphics[valign=m,width=0.29\textwidth]{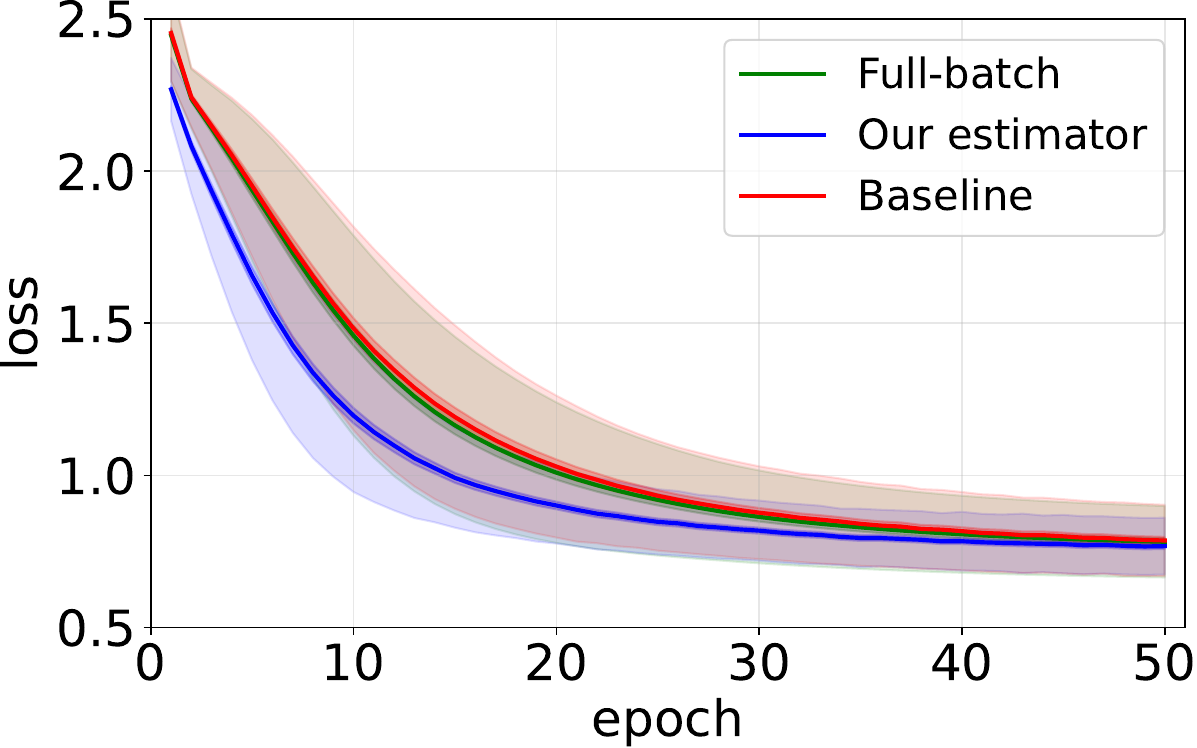} \\

\rotatebox[origin=c]{90}{\footnotesize\textbf{CIFAR-10}} &
\includegraphics[valign=m,width=0.29\textwidth]{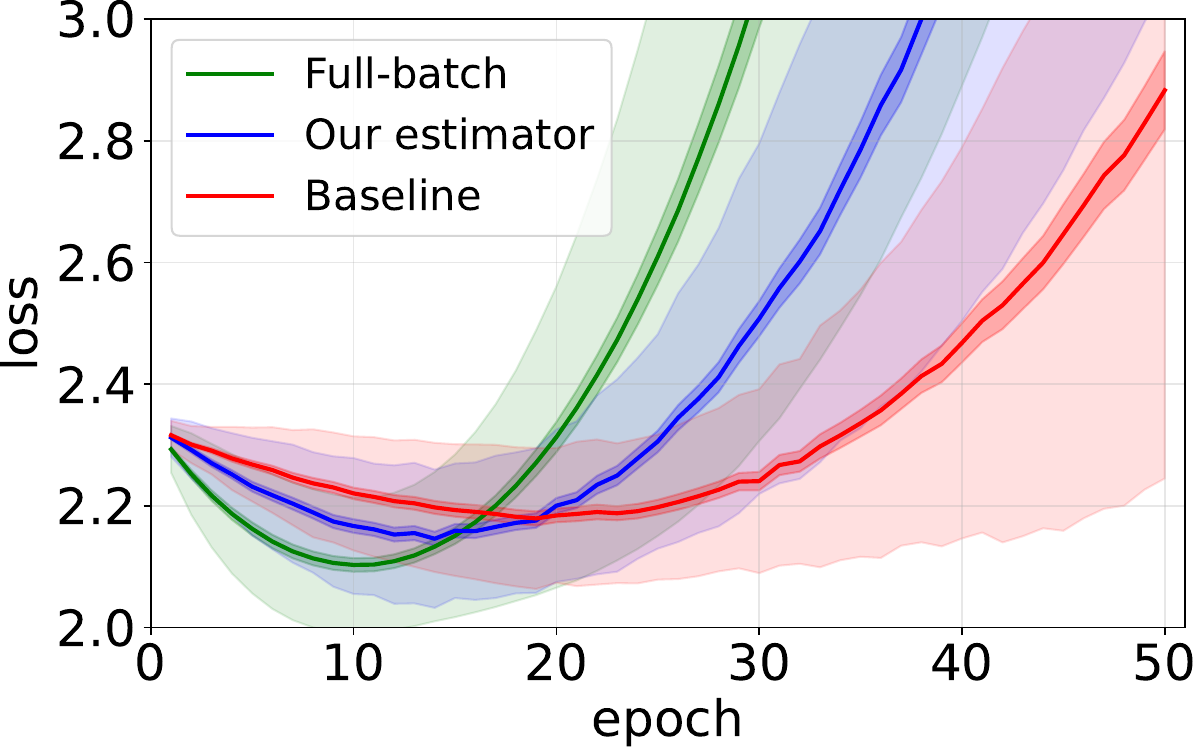} &
\includegraphics[valign=m,width=0.29\textwidth]{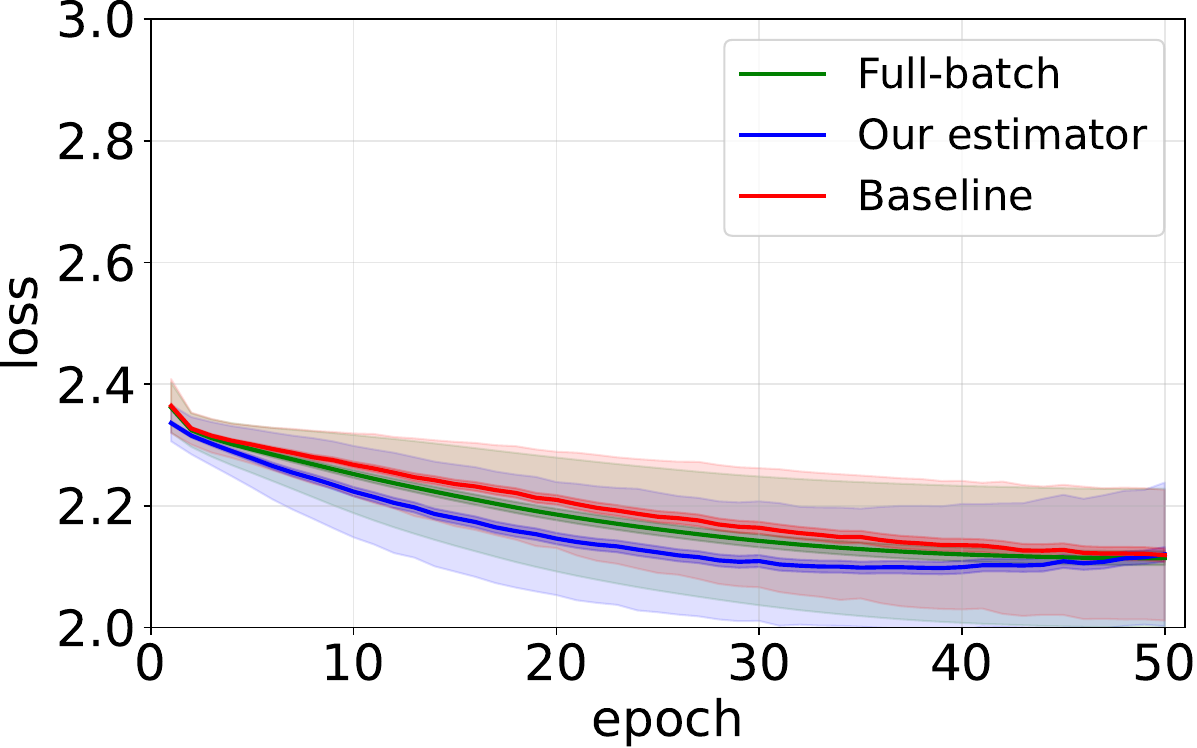} &
\includegraphics[valign=m,width=0.29\textwidth]{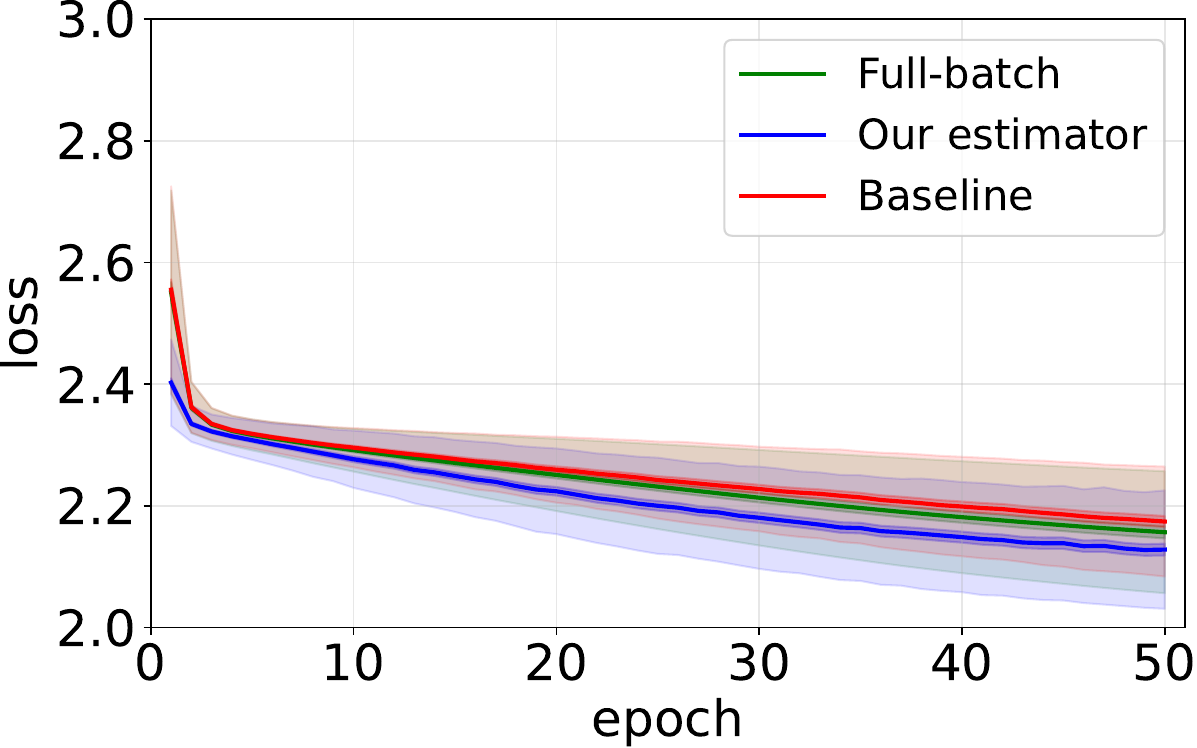} \\

\rotatebox[origin=c]{90}{\footnotesize\textbf{CIFAR-100}} &
\includegraphics[valign=m,width=0.29\textwidth]{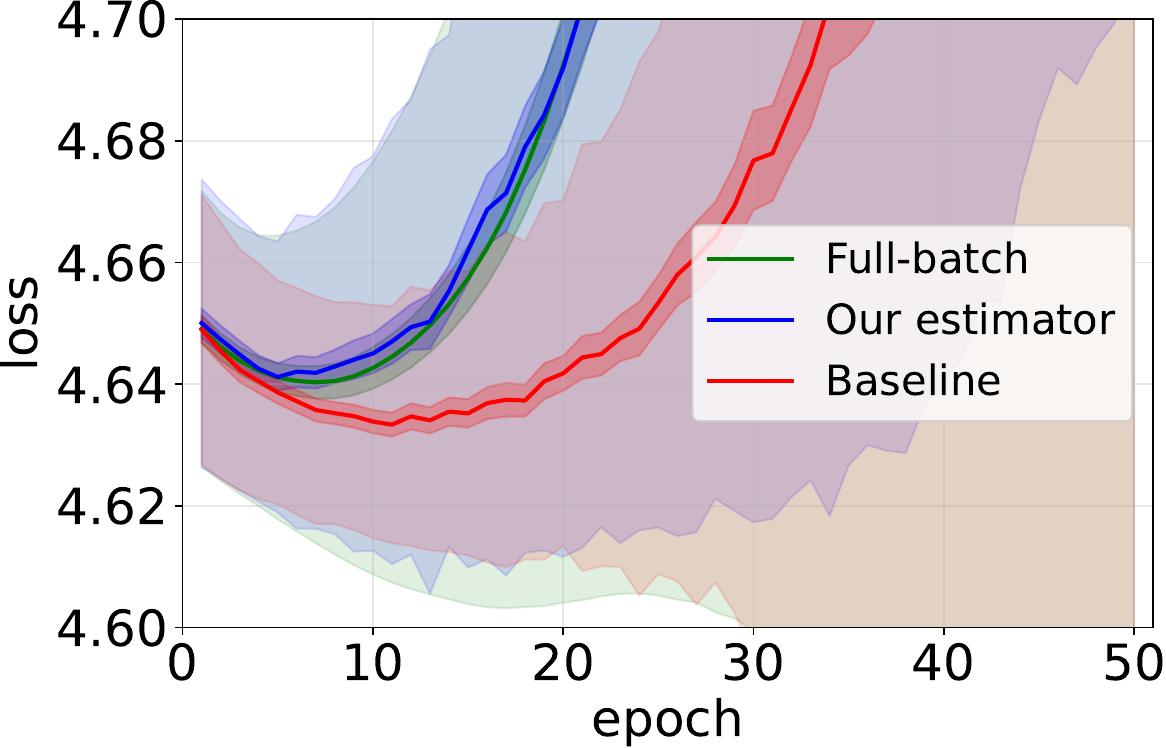} &
\includegraphics[valign=m,width=0.29\textwidth]{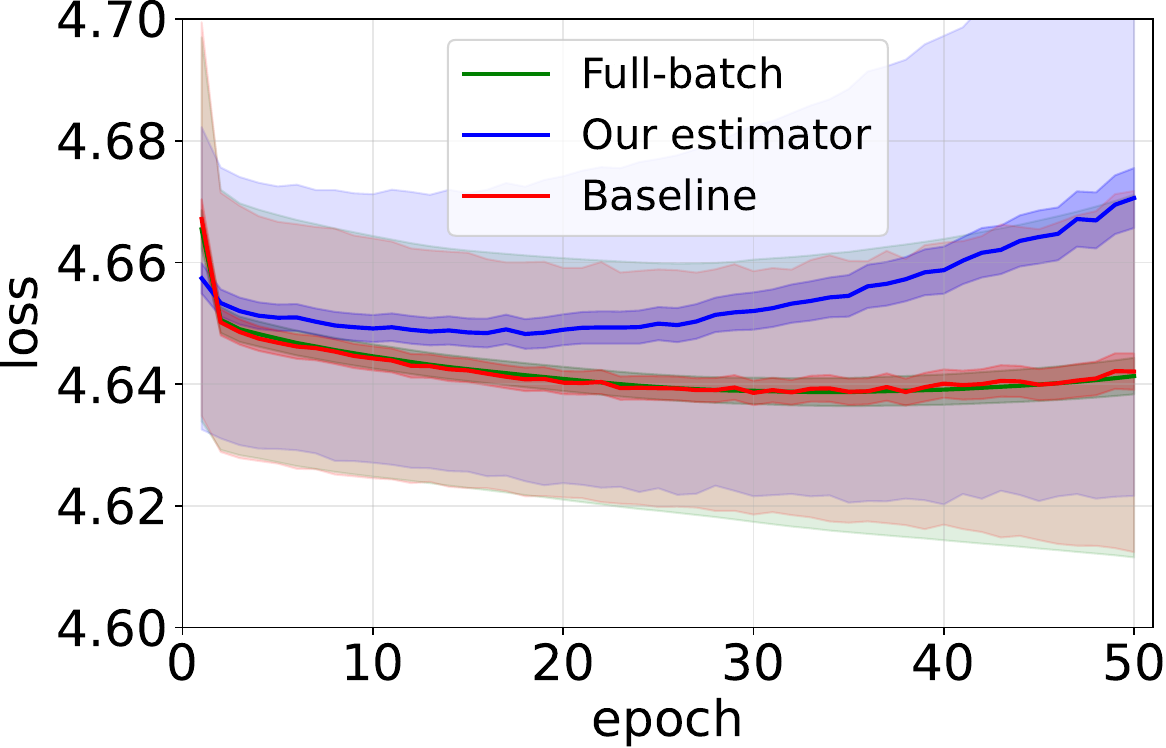} &
\includegraphics[valign=m,width=0.29\textwidth]{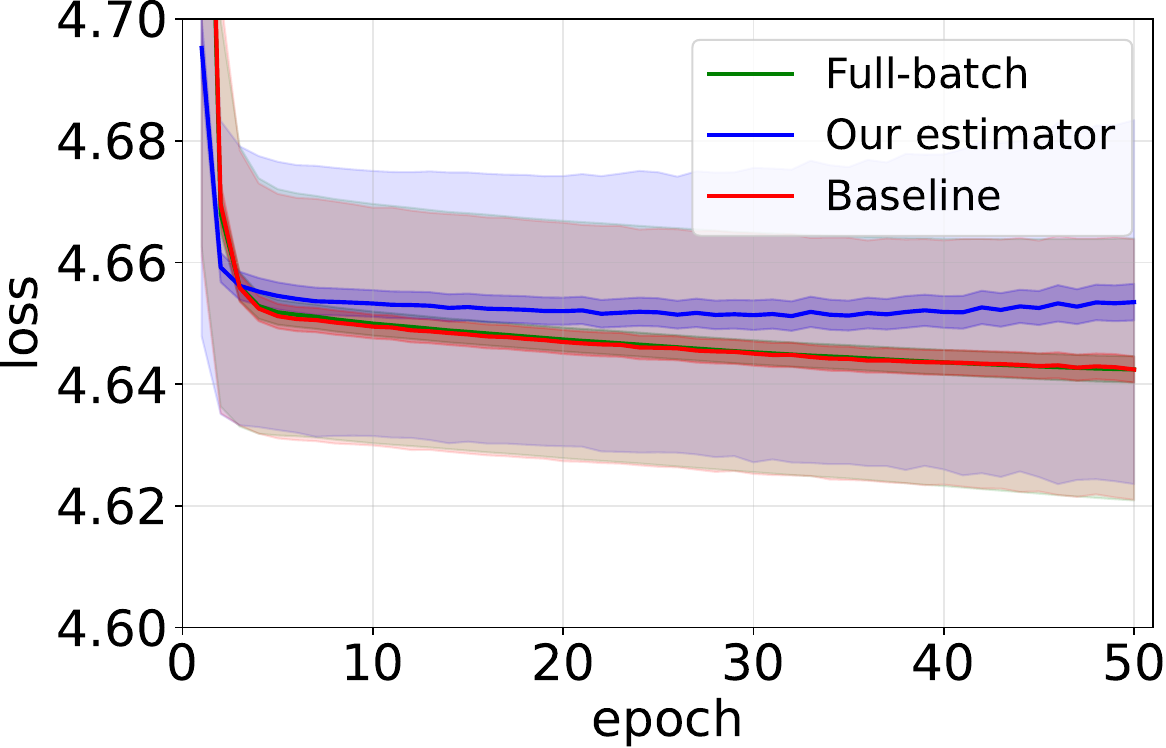} \\

\end{tabular}

\caption{The generalization performance for SGD-M optimizer with baseline (uniform mini-batch), model-assisted (our estimator) and full-batch gradients. Full-batch gradient case is listed for comparison purposes. The darker curve represents the average test loss of 400 runs, the dark shading is the 95\% confidence interval and lighter shading shows the standard deviation. Columns correspond to batch sizes and rows to datasets.}
\label{fig:sgdm_loss_grid}
\end{figure*}
\begin{figure*}[t]
\centering
\setlength{\tabcolsep}{3pt}
\renewcommand{\arraystretch}{0.8}

\begin{tabular}{>{\centering\arraybackslash}m{0.04\textwidth}
                >{\centering\arraybackslash}m{0.29\textwidth}
                >{\centering\arraybackslash}m{0.29\textwidth}
                >{\centering\arraybackslash}m{0.29\textwidth}}

& \textbf{Batch size 10} & \textbf{Batch size 50} & \textbf{Batch size 100} \\

\rotatebox[origin=c]{90}{\footnotesize\textbf{Synthetic}} &
\includegraphics[valign=m,width=0.29\textwidth]{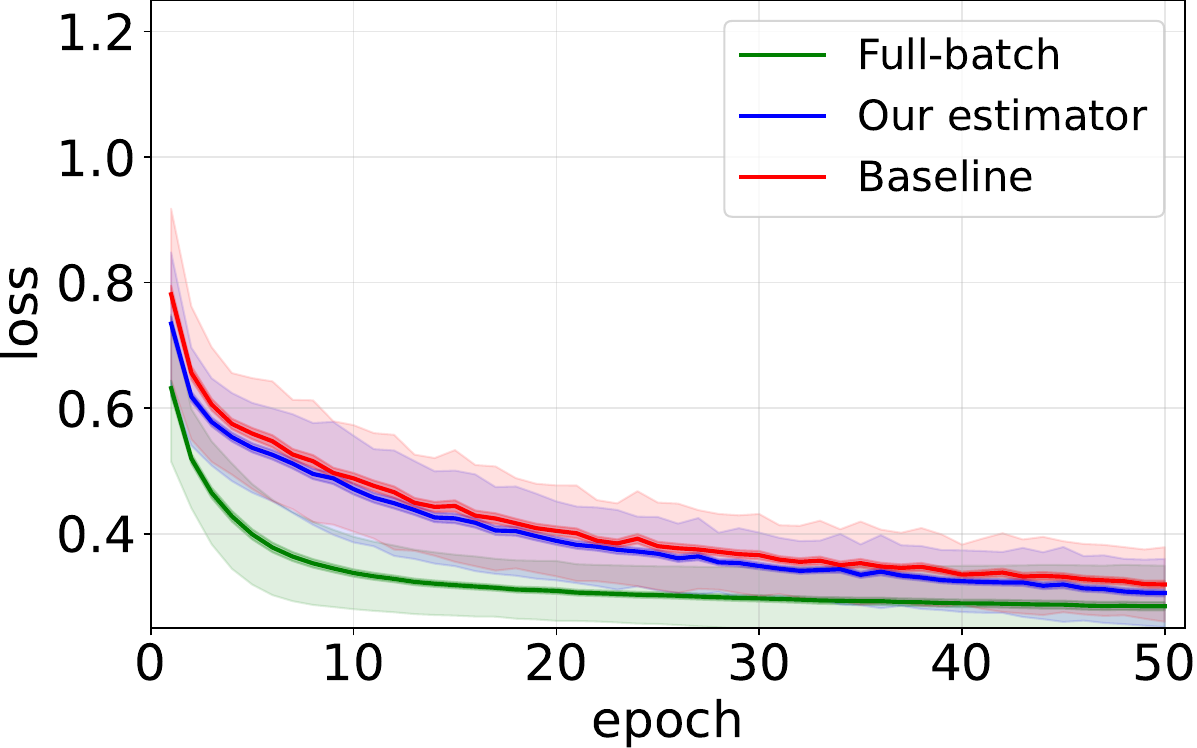} &
\includegraphics[valign=m,width=0.29\textwidth]{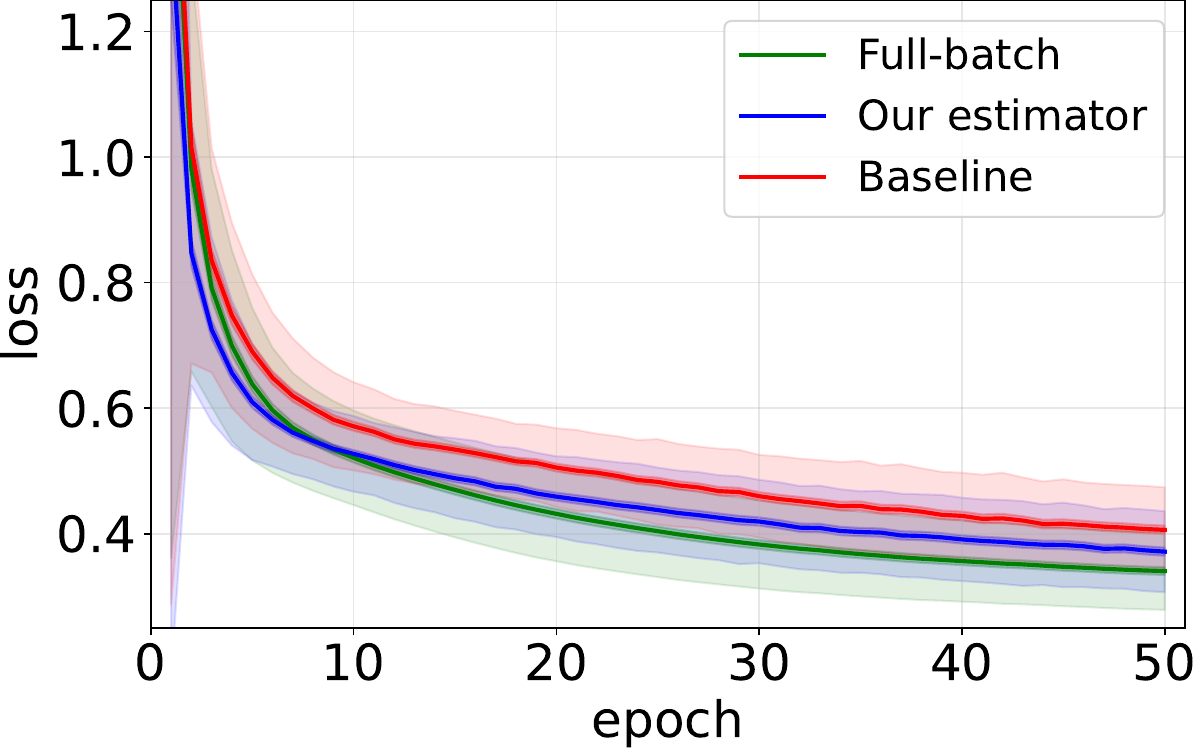} &
\includegraphics[valign=m,width=0.29\textwidth]{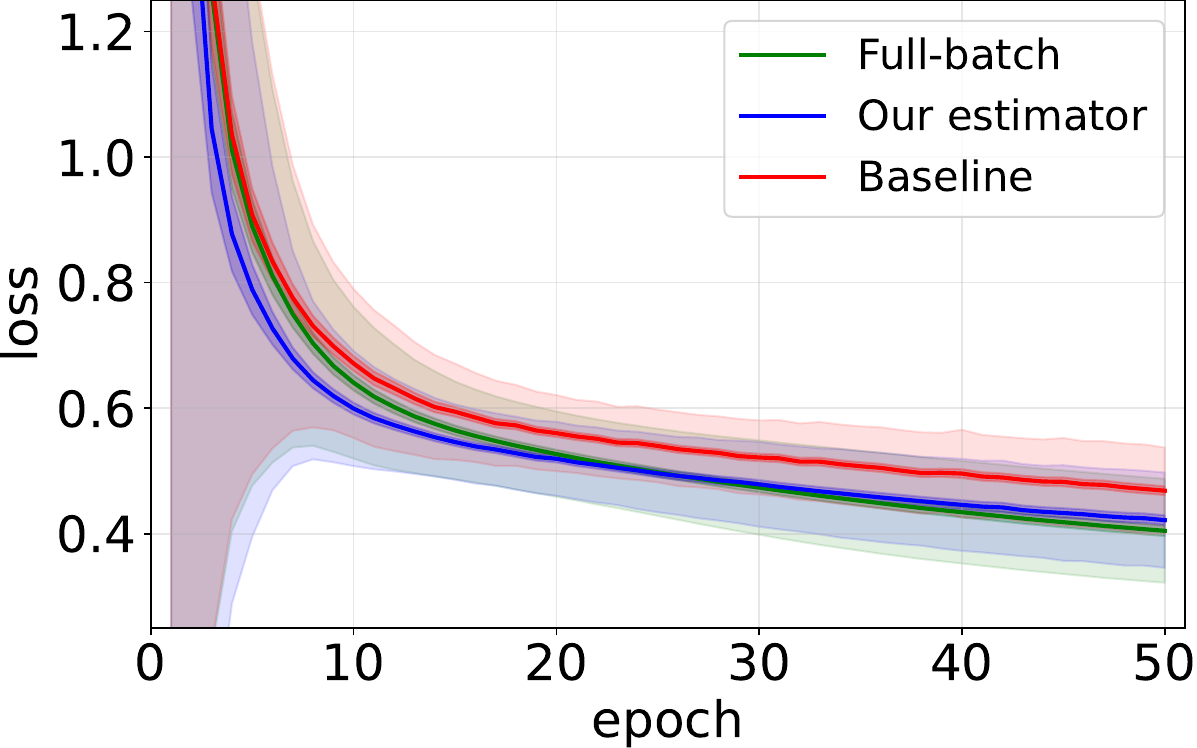} \\

\rotatebox[origin=c]{90}{\footnotesize\textbf{Airfoil self-noise}} &
\includegraphics[valign=m,width=0.29\textwidth]{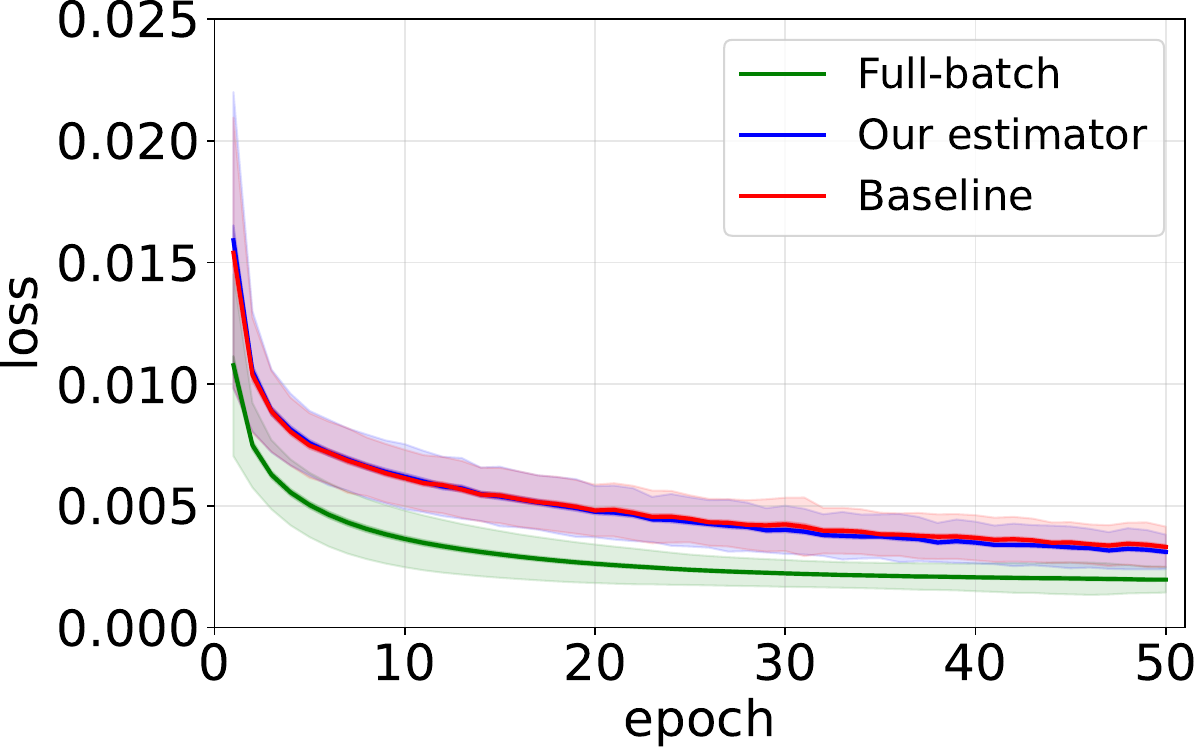} &
\includegraphics[valign=m,width=0.29\textwidth]{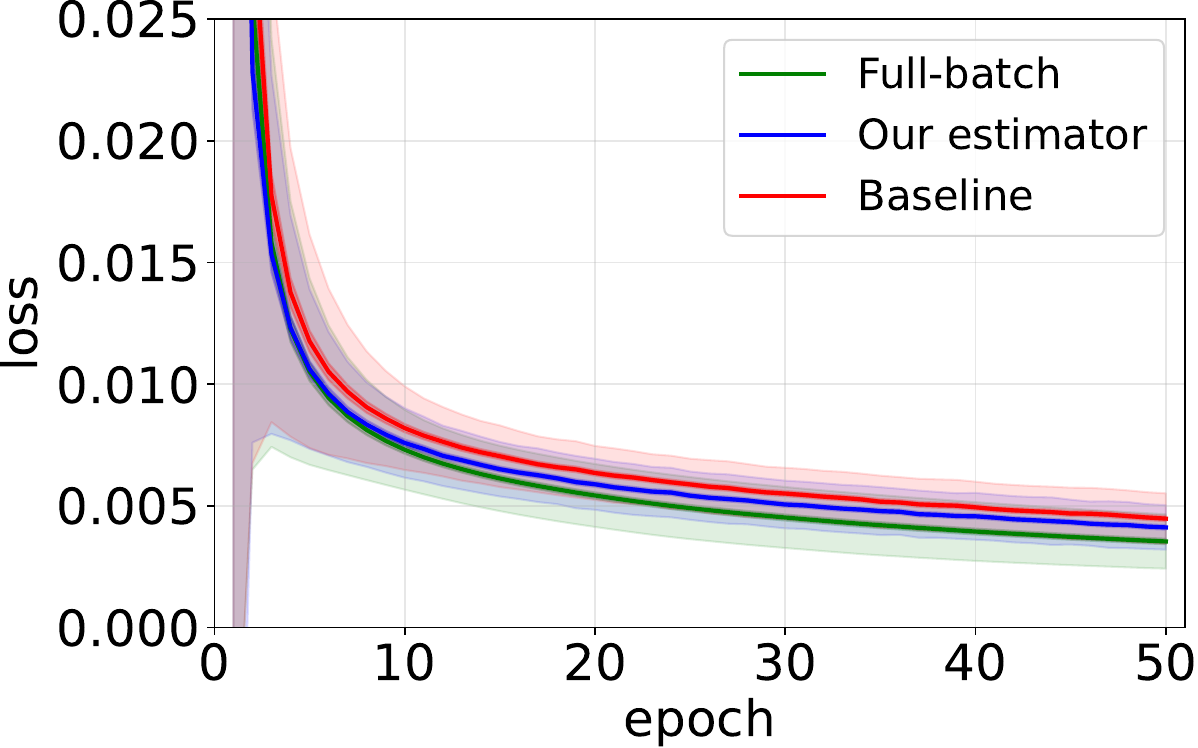} &
\includegraphics[valign=m,width=0.29\textwidth]{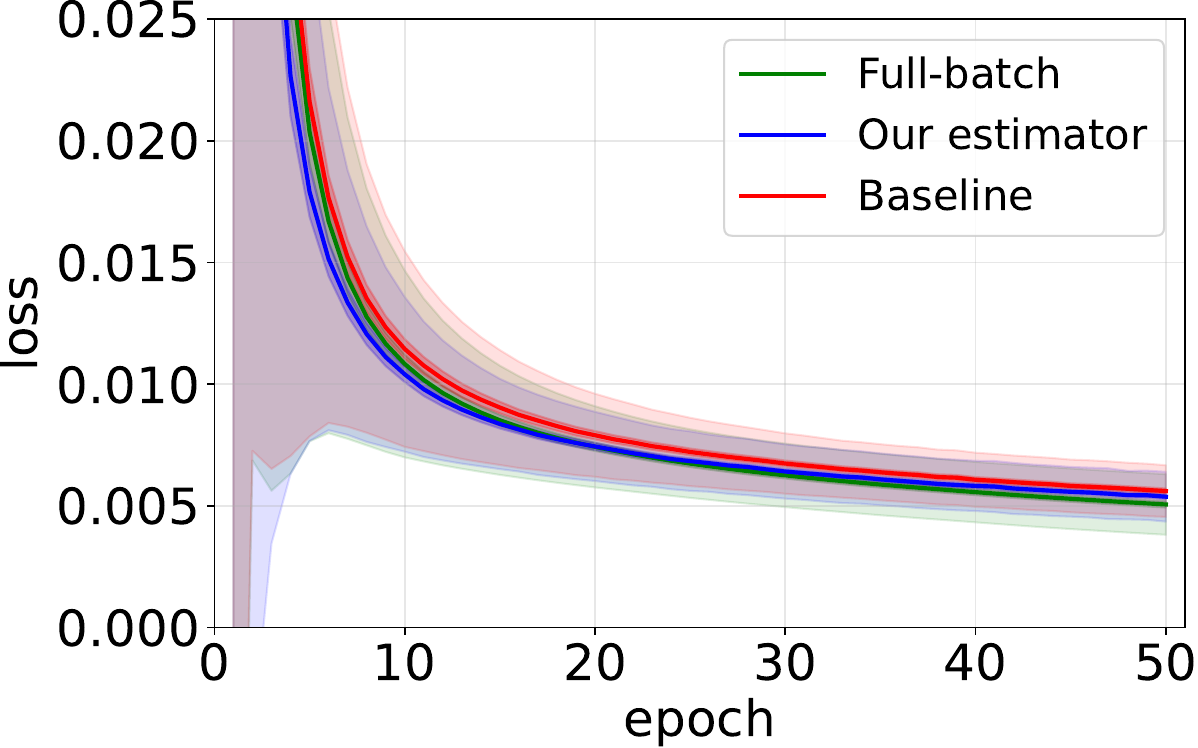} \\

\rotatebox[origin=c]{90}{\footnotesize\textbf{Appliances energy}} &
\includegraphics[valign=m,width=0.29\textwidth]{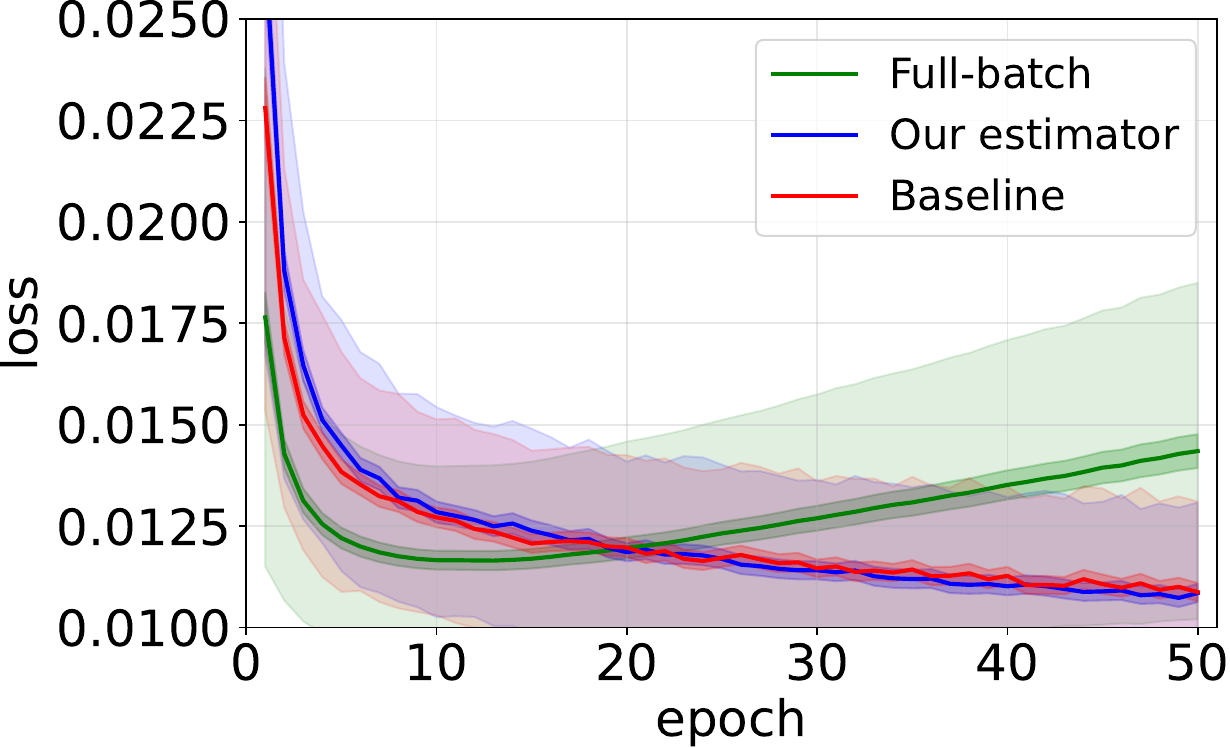} &
\includegraphics[valign=m,width=0.29\textwidth]{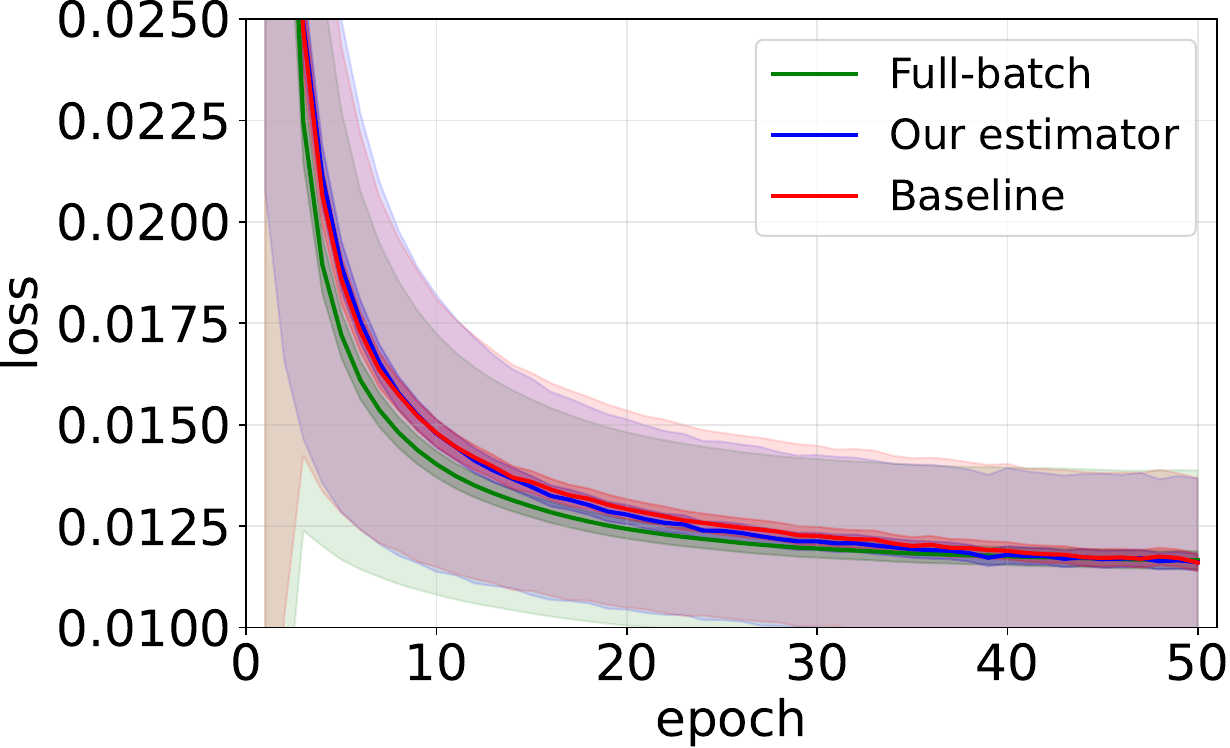} &
\includegraphics[valign=m,width=0.29\textwidth]{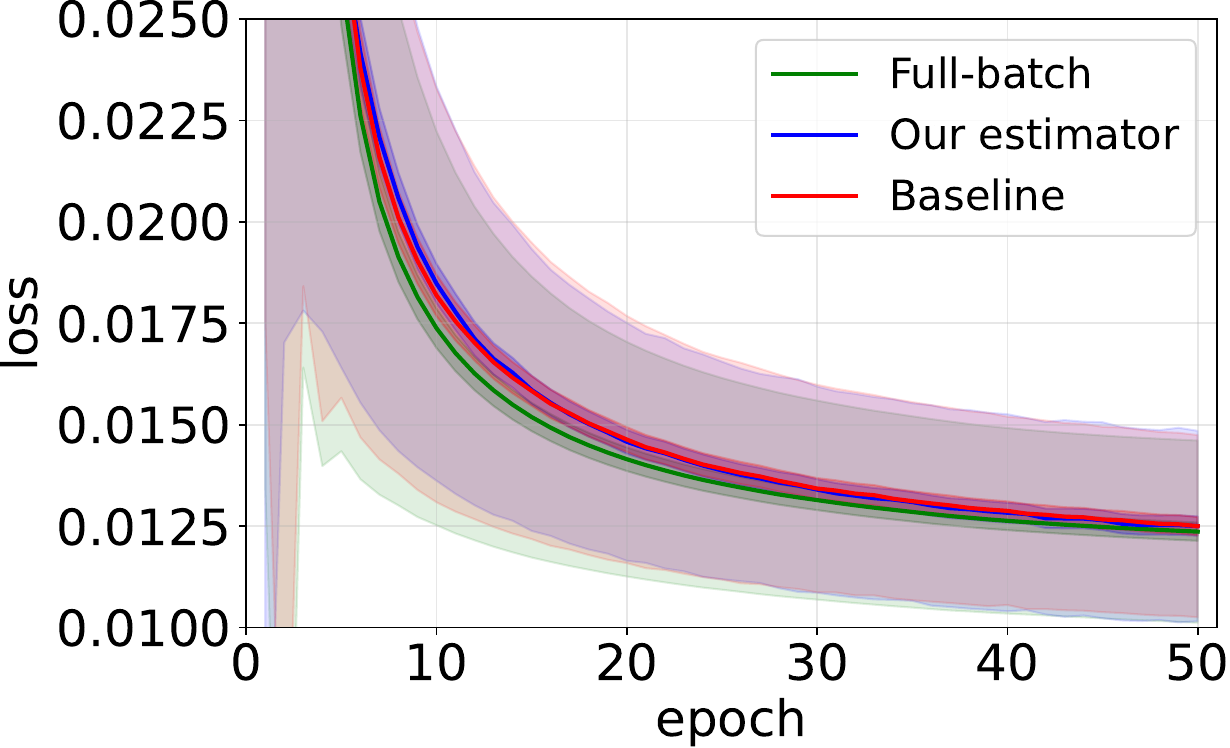} \\

\rotatebox[origin=c]{90}{\footnotesize\textbf{MNIST}} &
\includegraphics[valign=m,width=0.29\textwidth]{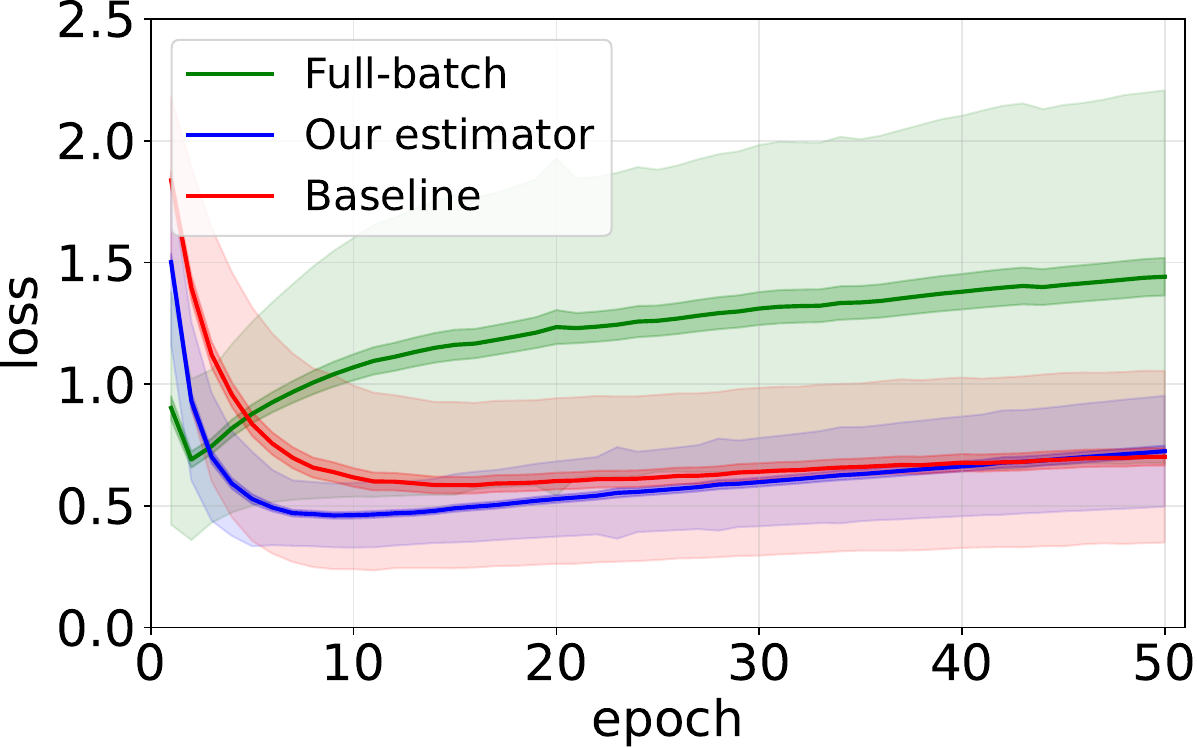} &
\includegraphics[valign=m,width=0.29\textwidth]{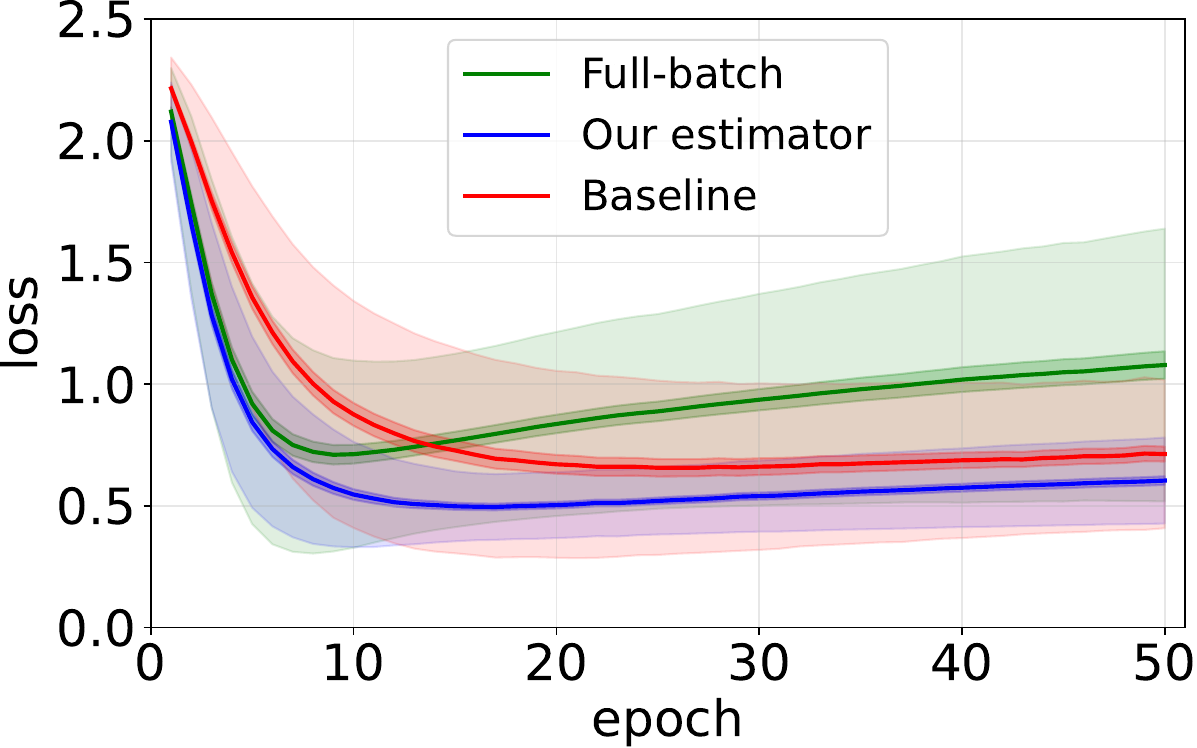} &
\includegraphics[valign=m,width=0.29\textwidth]{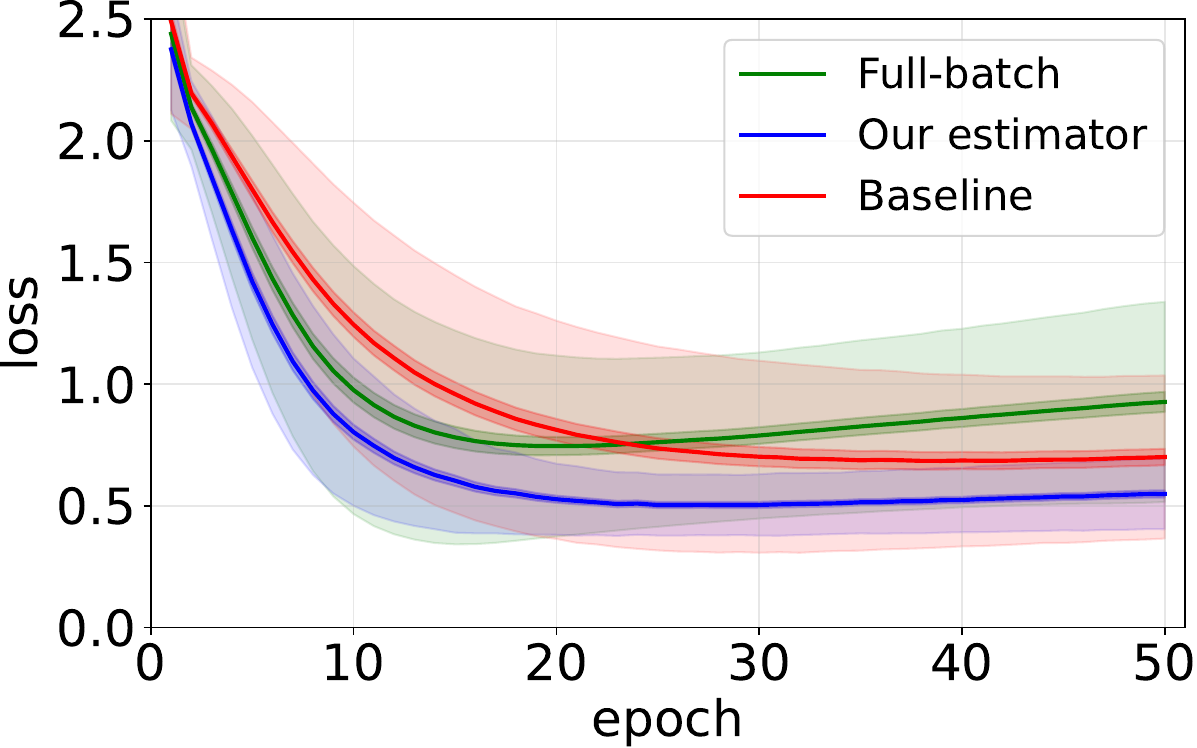} \\

\rotatebox[origin=c]{90}{\footnotesize\textbf{Fashion-MNIST}} &
\includegraphics[valign=m,width=0.29\textwidth]{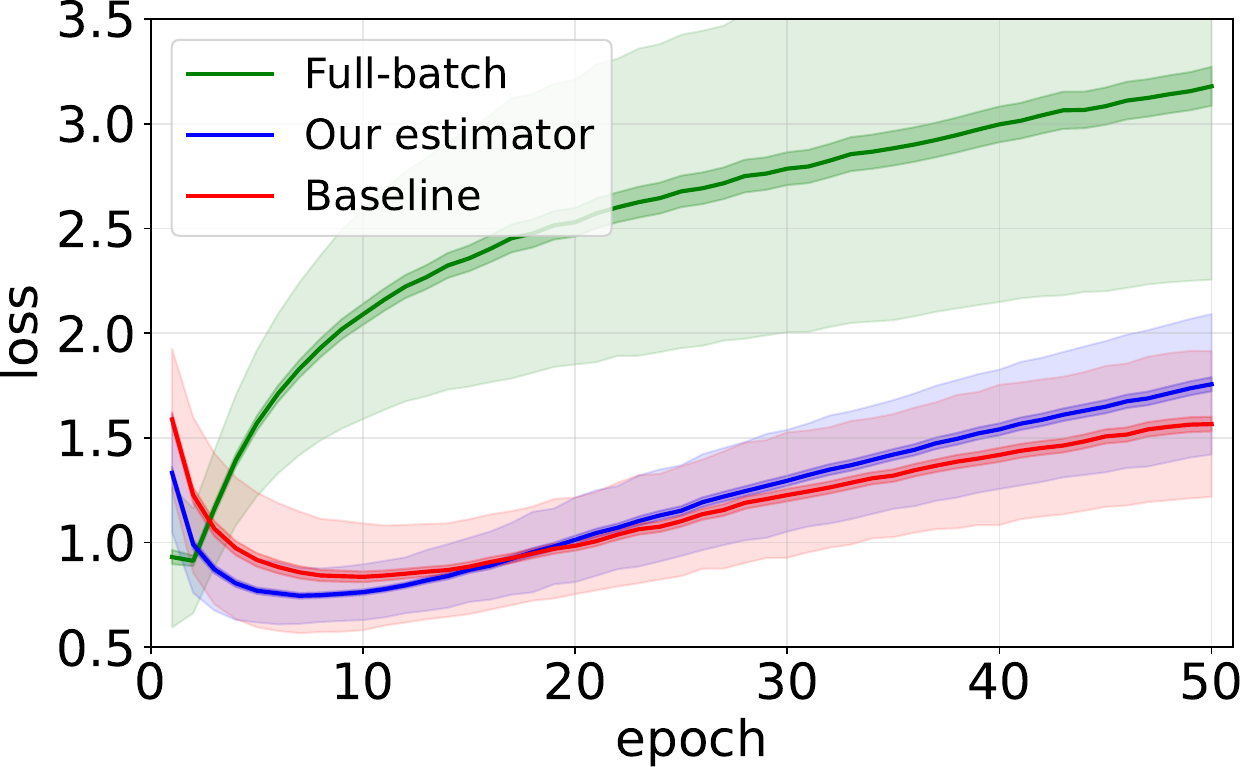} &
\includegraphics[valign=m,width=0.29\textwidth]{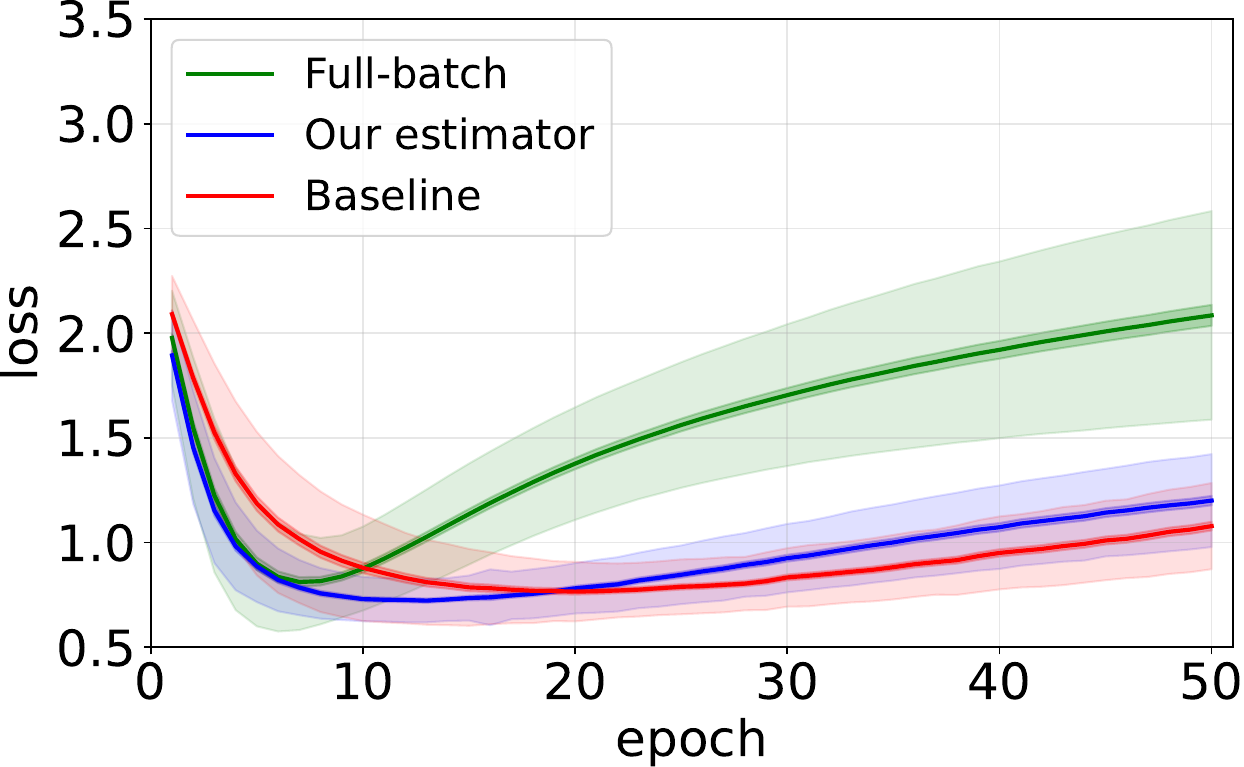} &
\includegraphics[valign=m,width=0.29\textwidth]{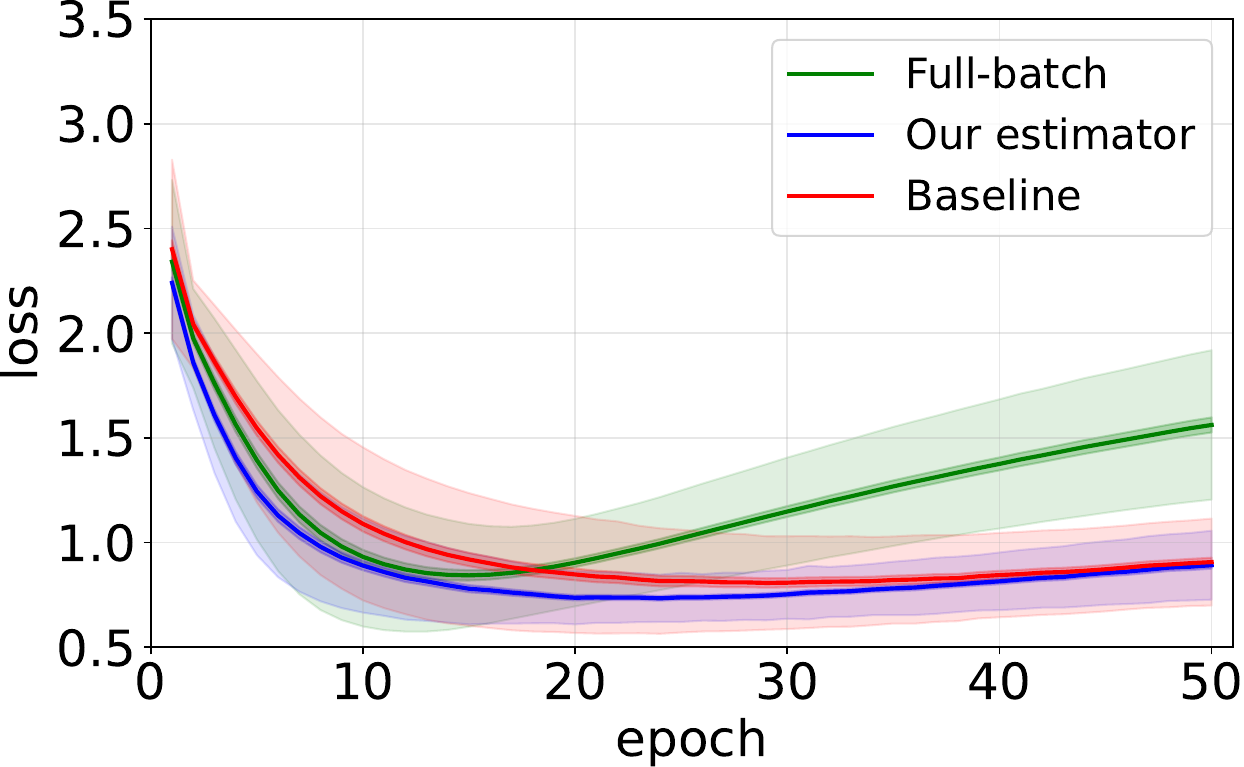} \\

\rotatebox[origin=c]{90}{\footnotesize\textbf{CIFAR-10}} &
\includegraphics[valign=m,width=0.29\textwidth]{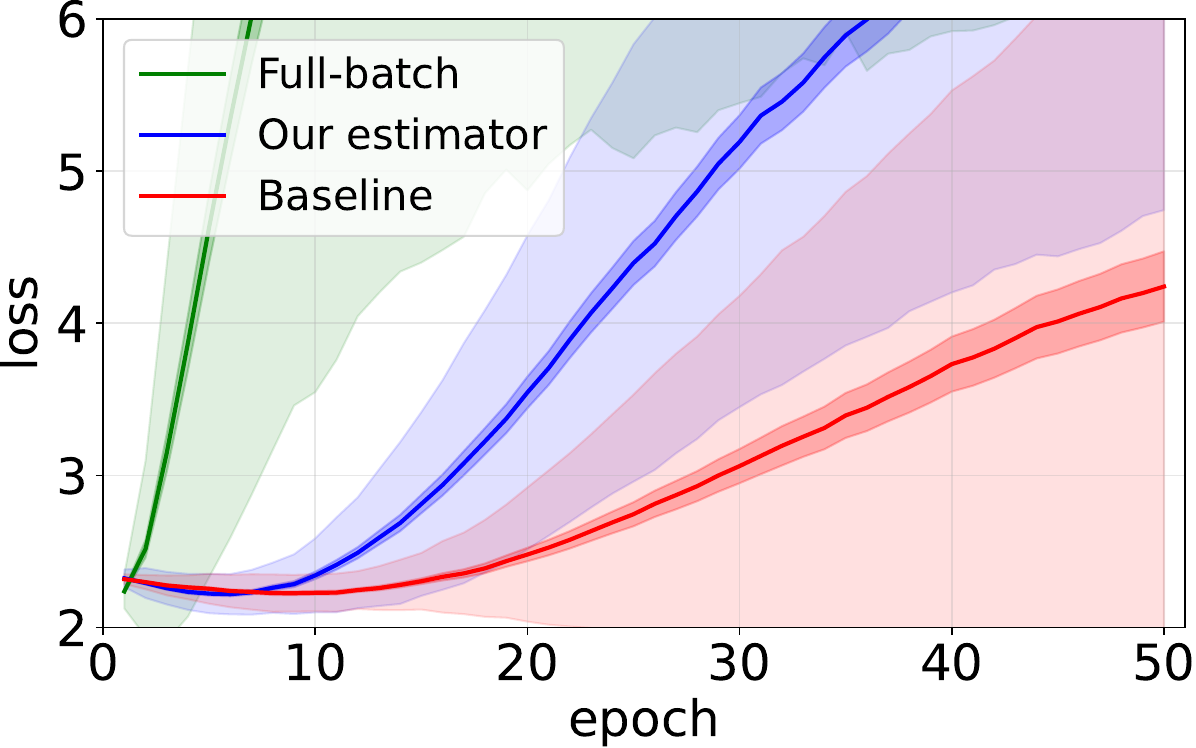} &
\includegraphics[valign=m,width=0.29\textwidth]{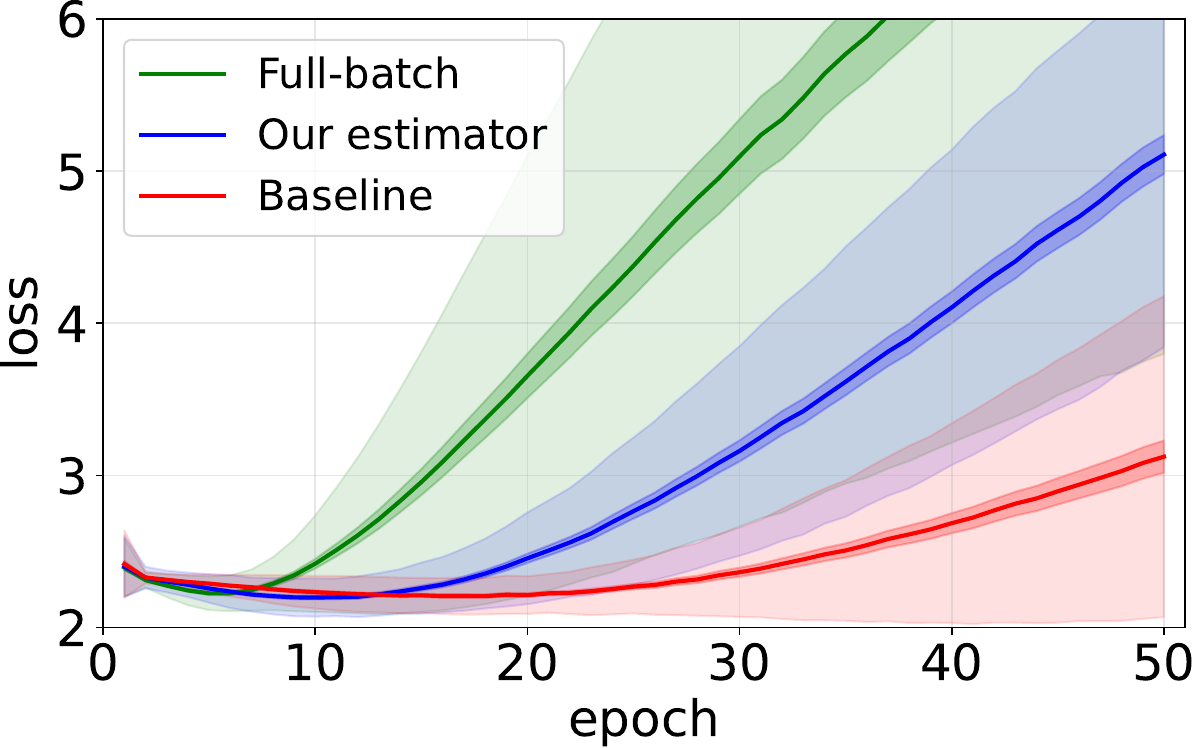} &
\includegraphics[valign=m,width=0.29\textwidth]{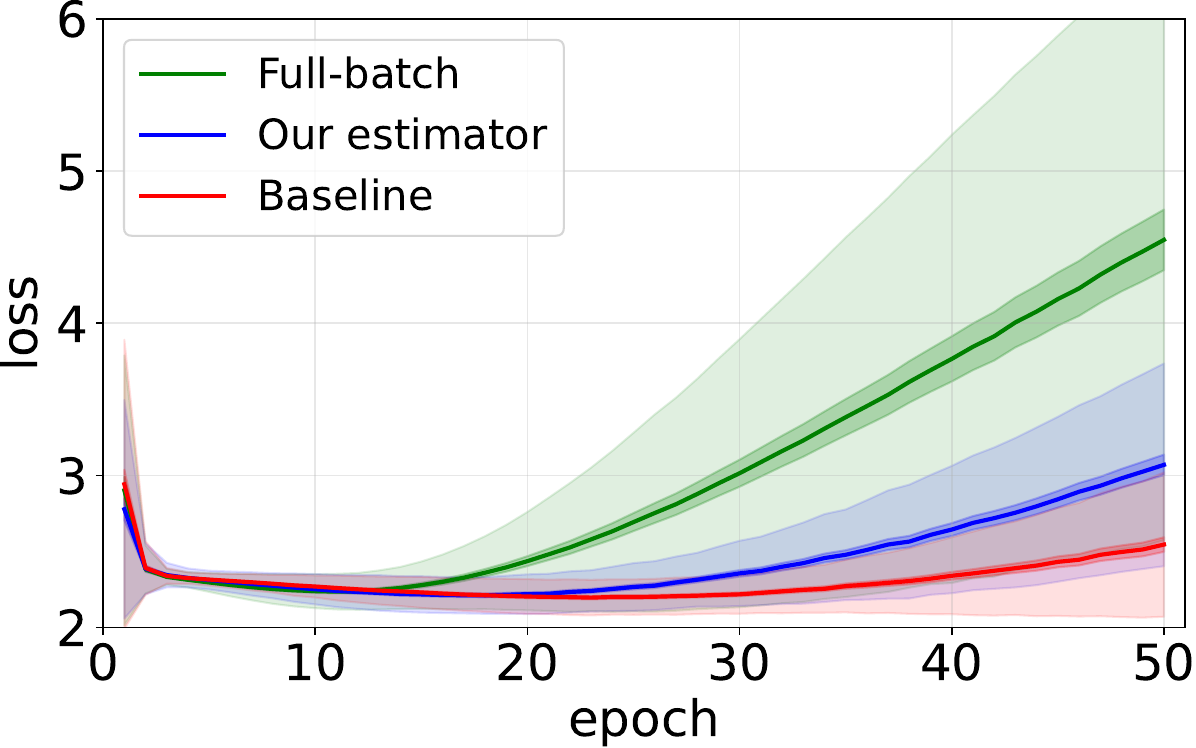} \\

\rotatebox[origin=c]{90}{\footnotesize\textbf{CIFAR-100}} &
\includegraphics[valign=m,width=0.29\textwidth]{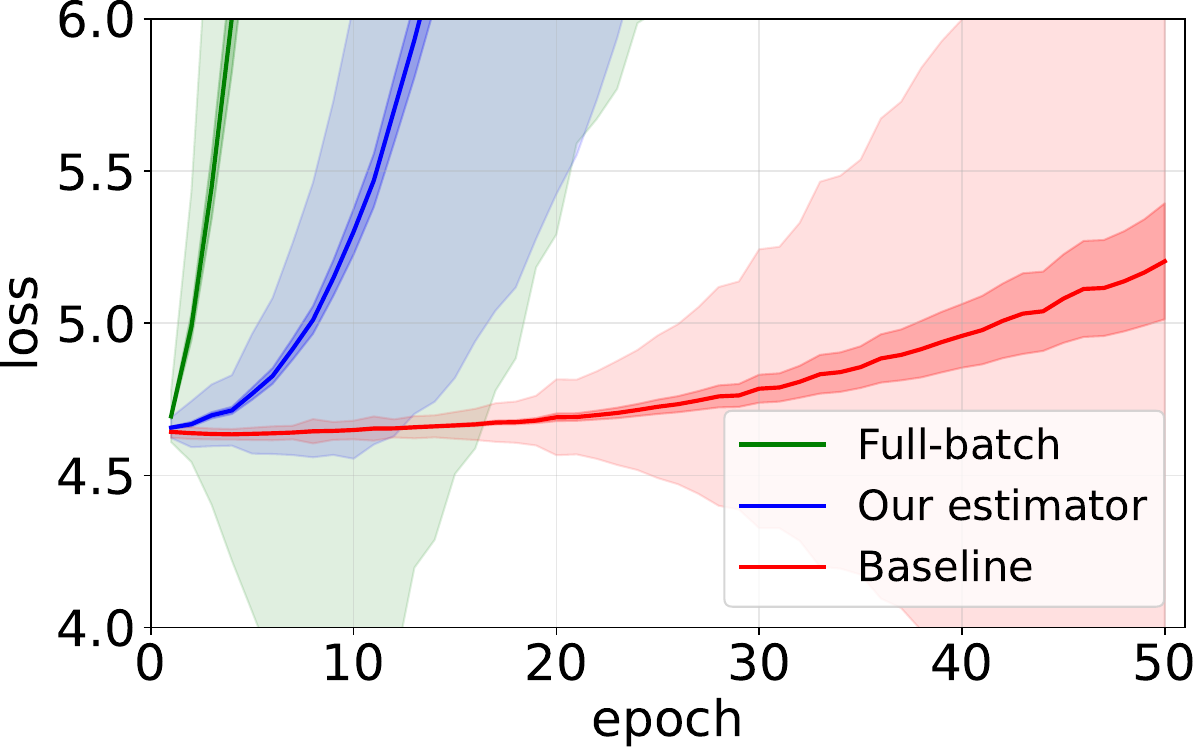} &
\includegraphics[valign=m,width=0.29\textwidth]{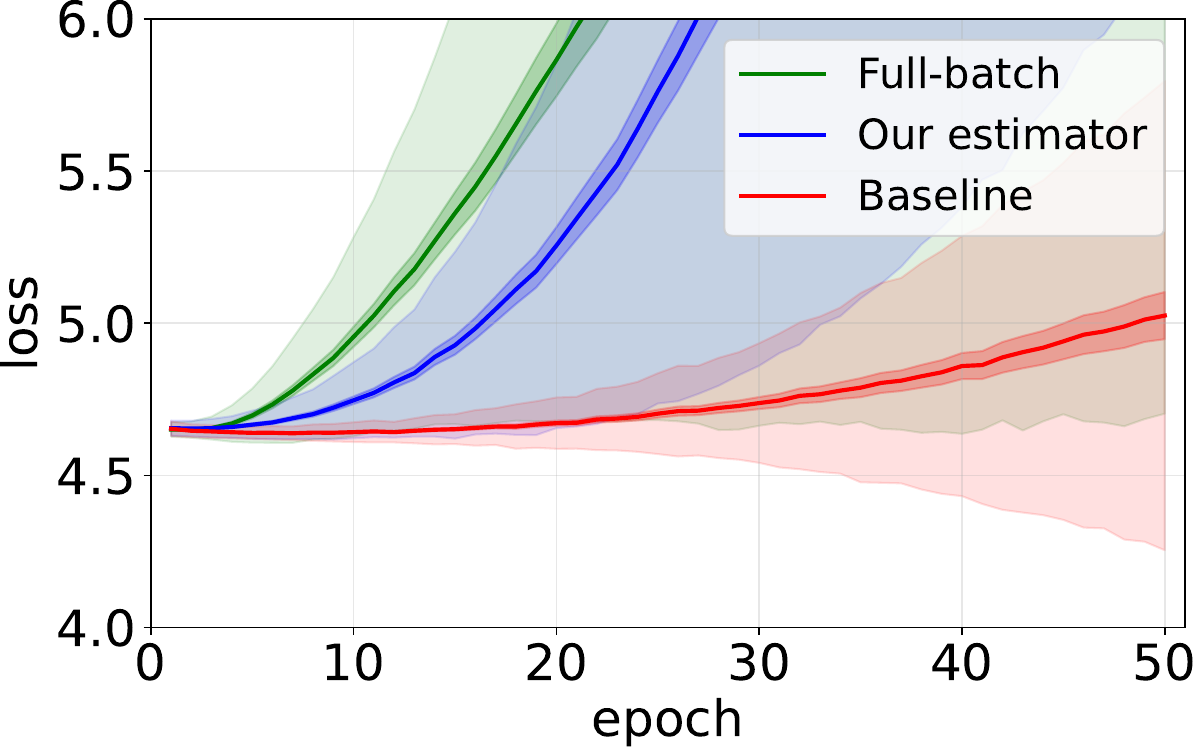} &
\includegraphics[valign=m,width=0.29\textwidth]{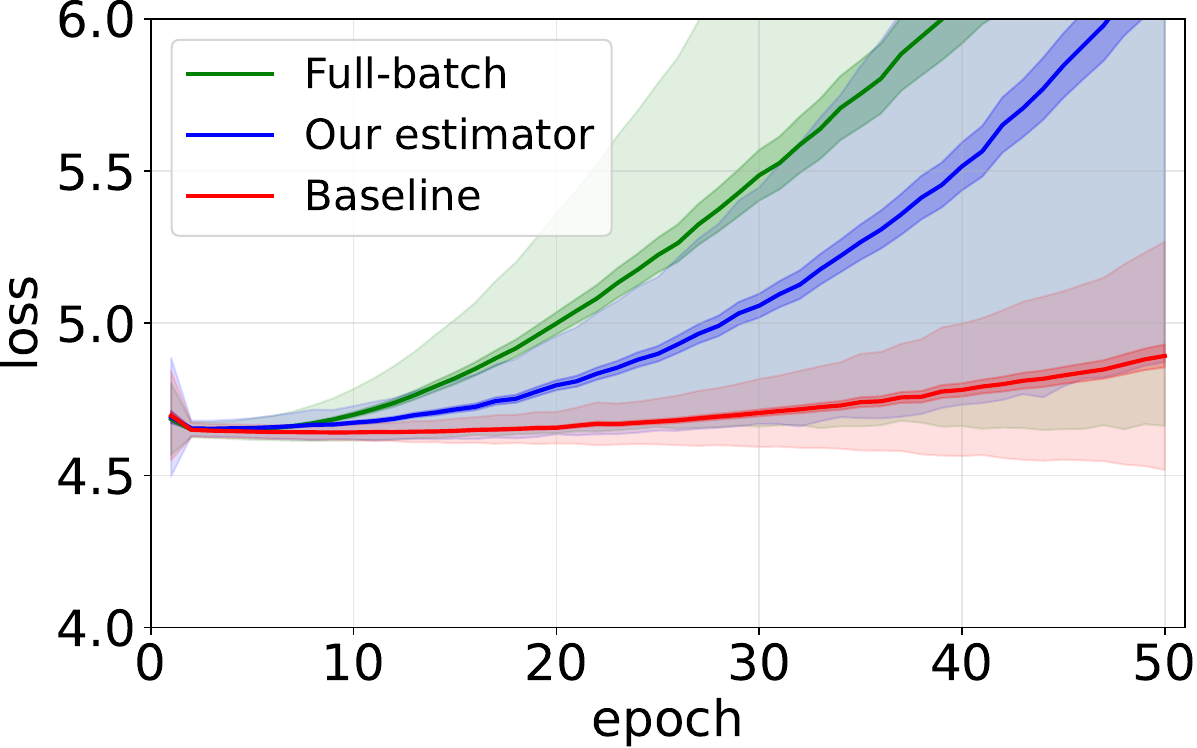} \\

\end{tabular}

\caption{The generalization performance for Adam optimizer with baseline (uniform mini-batch), model-assisted (our estimator) and full-batch gradients. Full-batch gradient case is listed for comparison purposes. The darker curve represents the average test loss of 400 runs, the dark shading is the 95\% confidence interval and lighter shading shows the standard deviation. Columns correspond to batch sizes and rows to datasets.}
\label{fig:Adam_loss_grid}
\end{figure*}

\begin{figure*}[t]
\centering
\setlength{\tabcolsep}{3pt}
\renewcommand{\arraystretch}{0.8}

\begin{tabular}{>{\centering\arraybackslash}m{0.04\textwidth}
                >{\centering\arraybackslash}m{0.29\textwidth}
                >{\centering\arraybackslash}m{0.29\textwidth}
                >{\centering\arraybackslash}m{0.29\textwidth}}

& \textbf{Batch size 10} & \textbf{Batch size 50} & \textbf{Batch size 100} \\

\rotatebox[origin=c]{90}{\footnotesize\textbf{Synthetic}} &
\includegraphics[valign=m,width=0.29\textwidth]{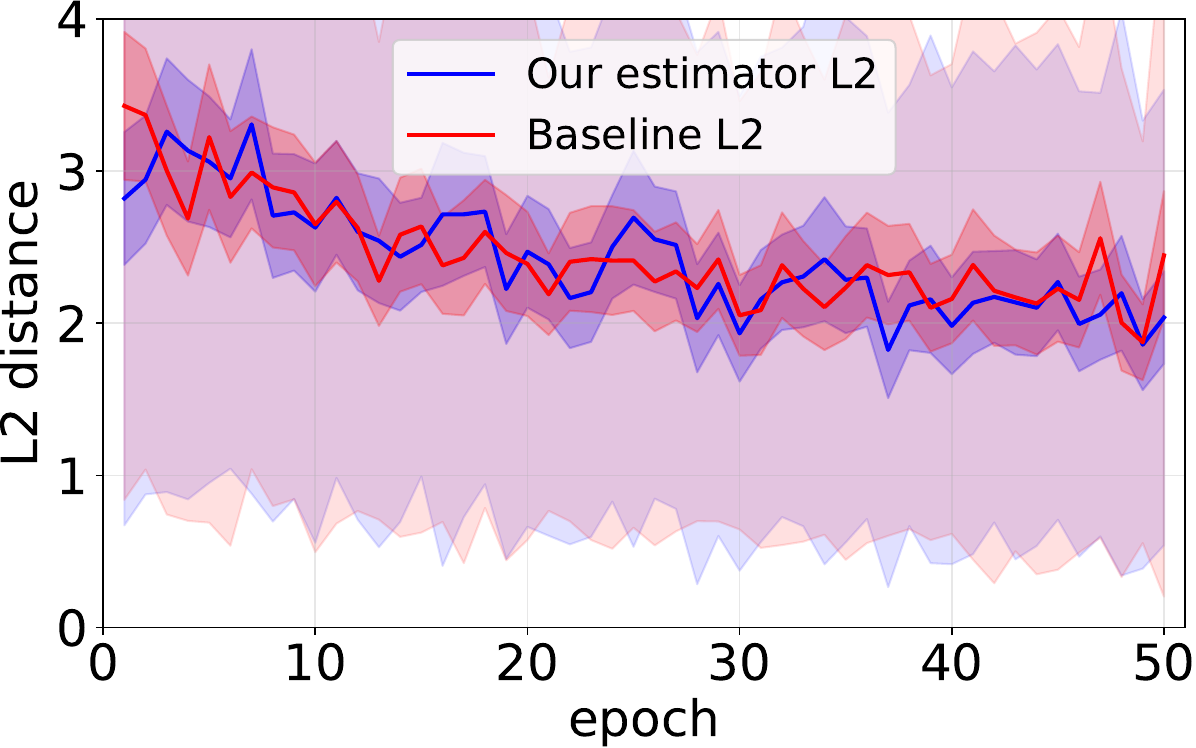} &
\includegraphics[valign=m,width=0.29\textwidth]{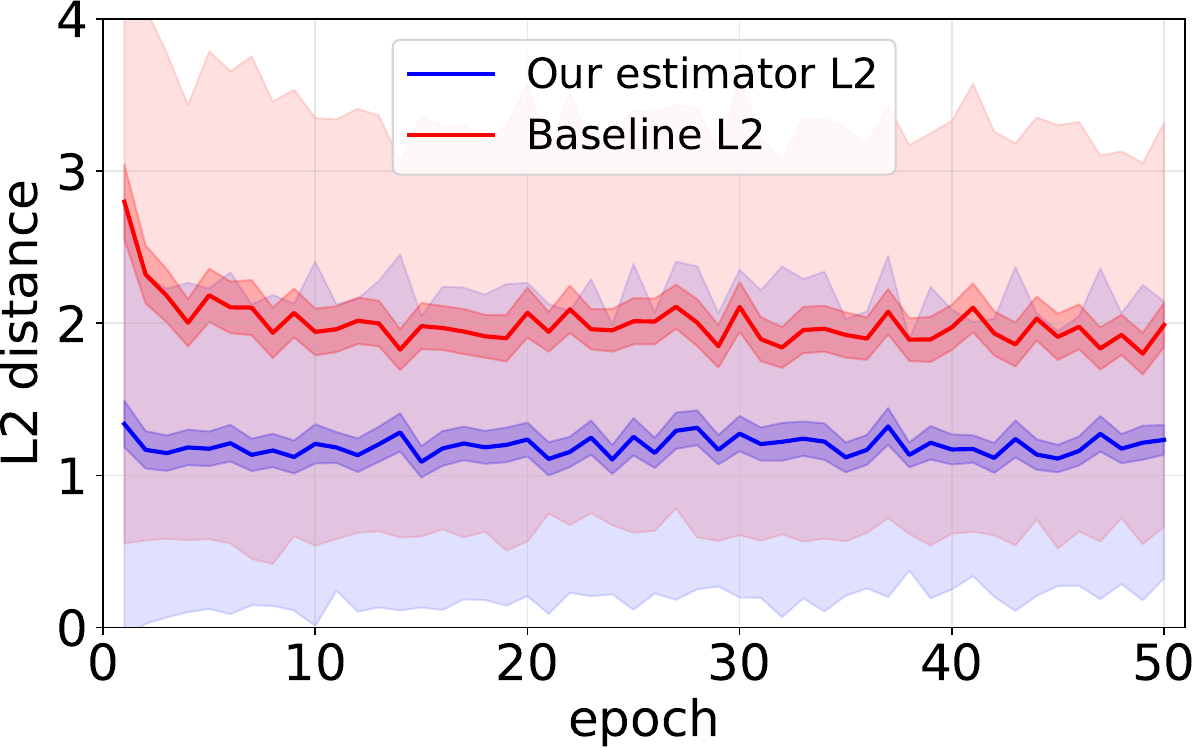} &
\includegraphics[valign=m,width=0.29\textwidth]{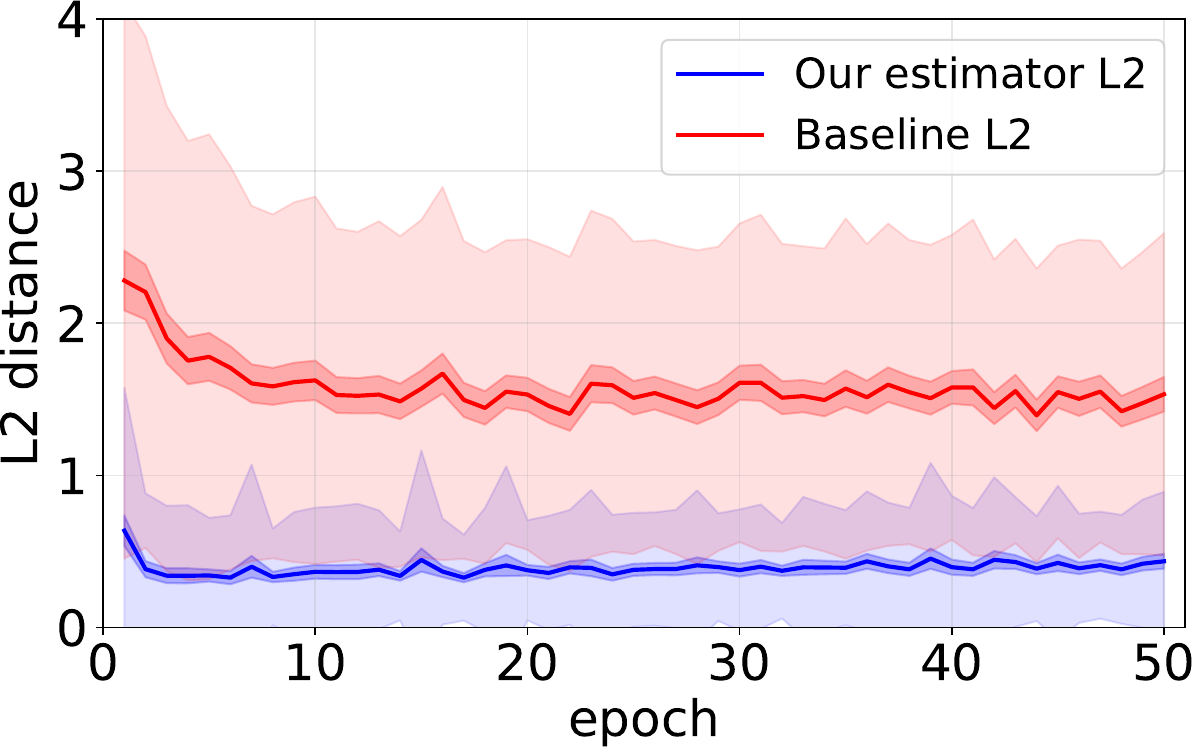} \\

\rotatebox[origin=c]{90}{\footnotesize\textbf{Airfoil self-noise}} &
\includegraphics[valign=m,width=0.29\textwidth]{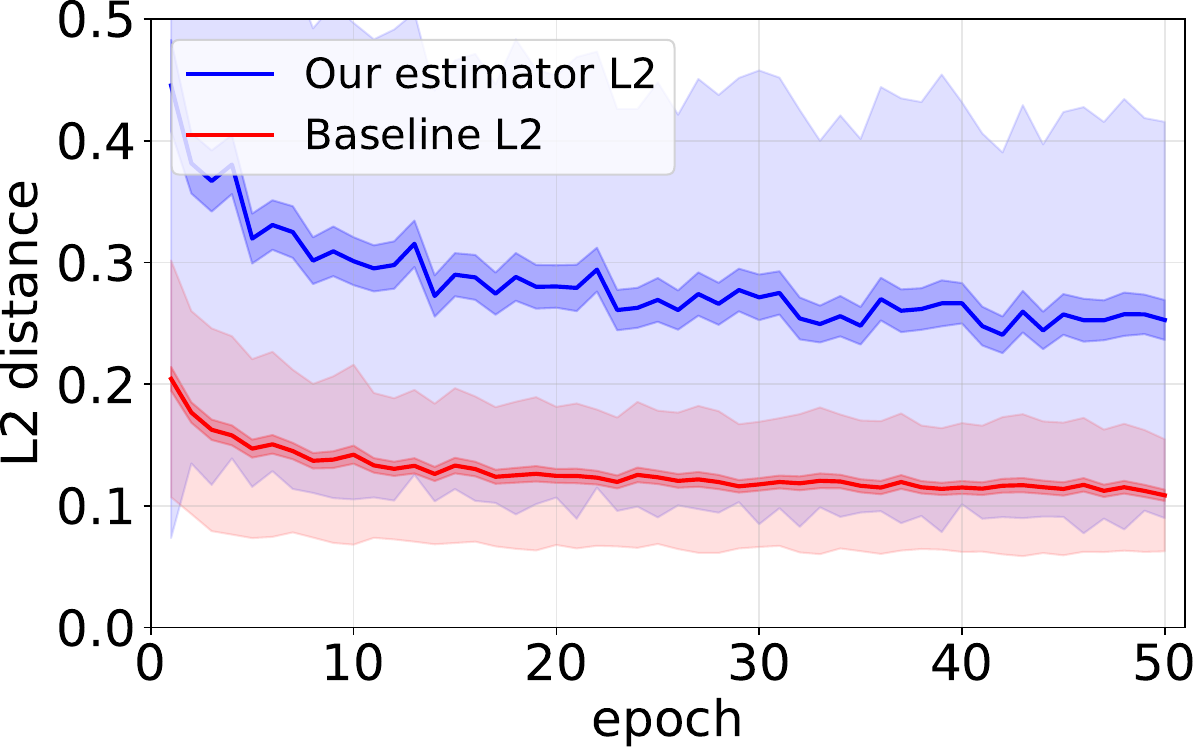} &
\includegraphics[valign=m,width=0.29\textwidth]{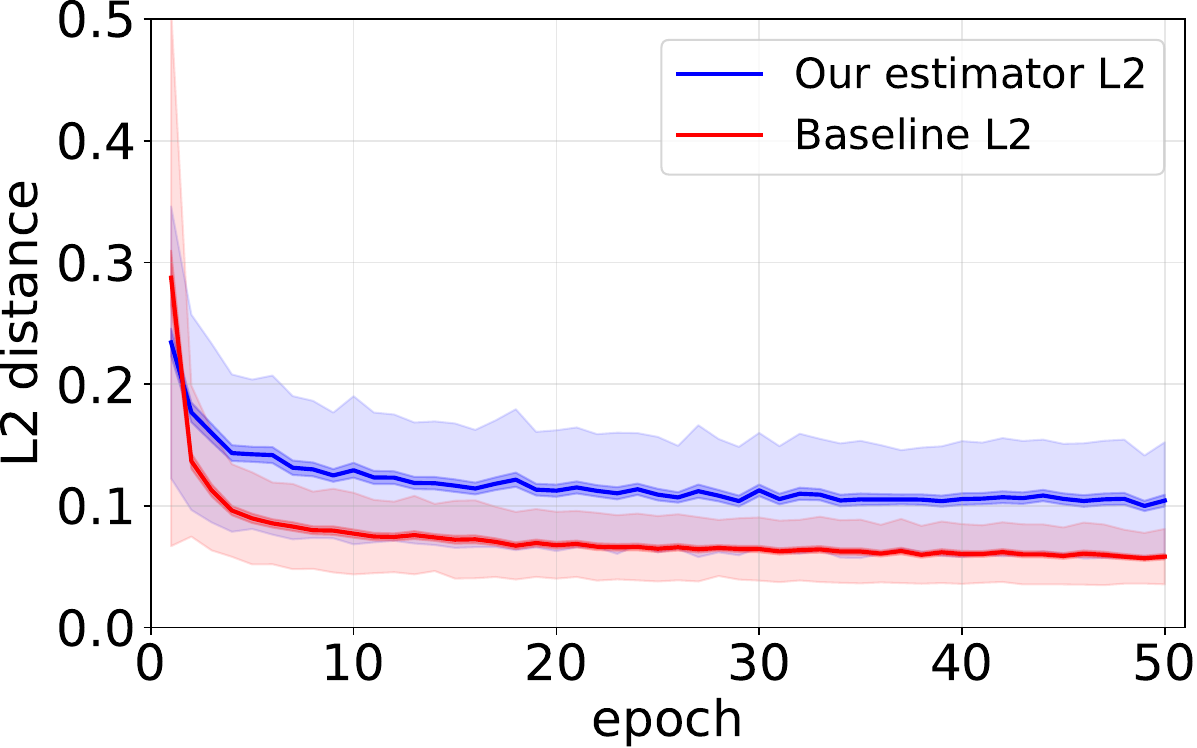} &
\includegraphics[valign=m,width=0.29\textwidth]{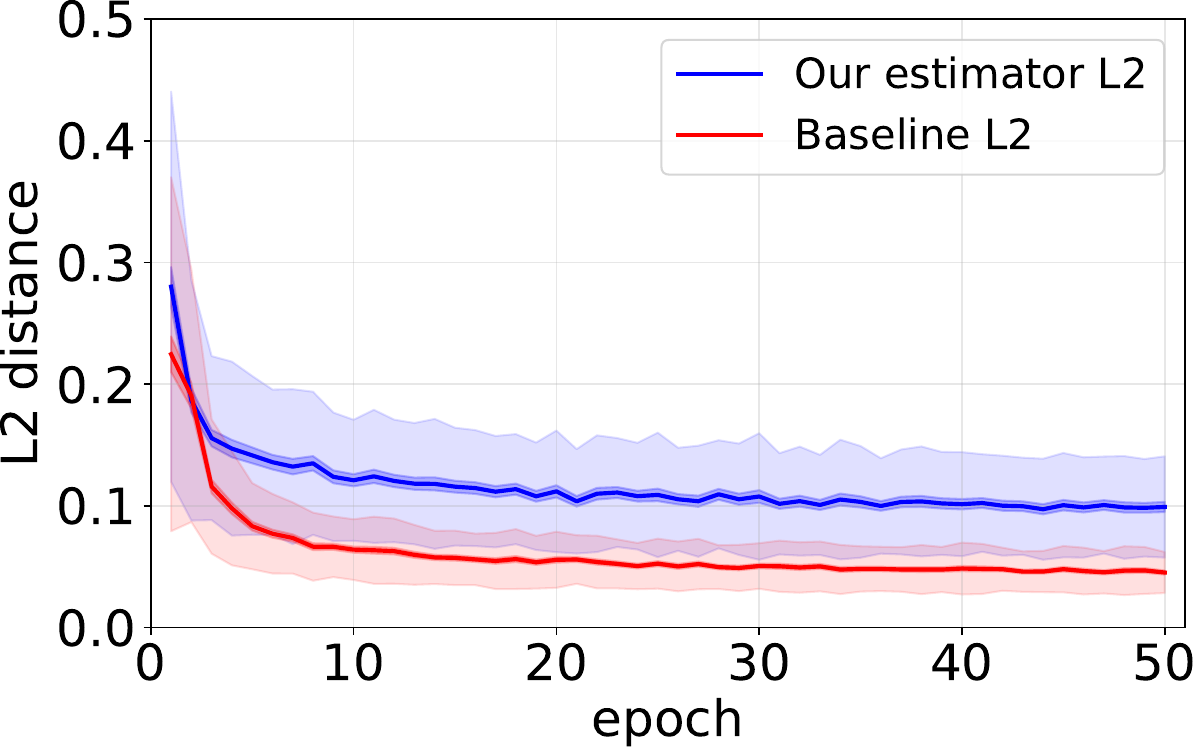} \\

\rotatebox[origin=c]{90}{\footnotesize\textbf{Appliances energy}} &
\includegraphics[valign=m,width=0.29\textwidth]{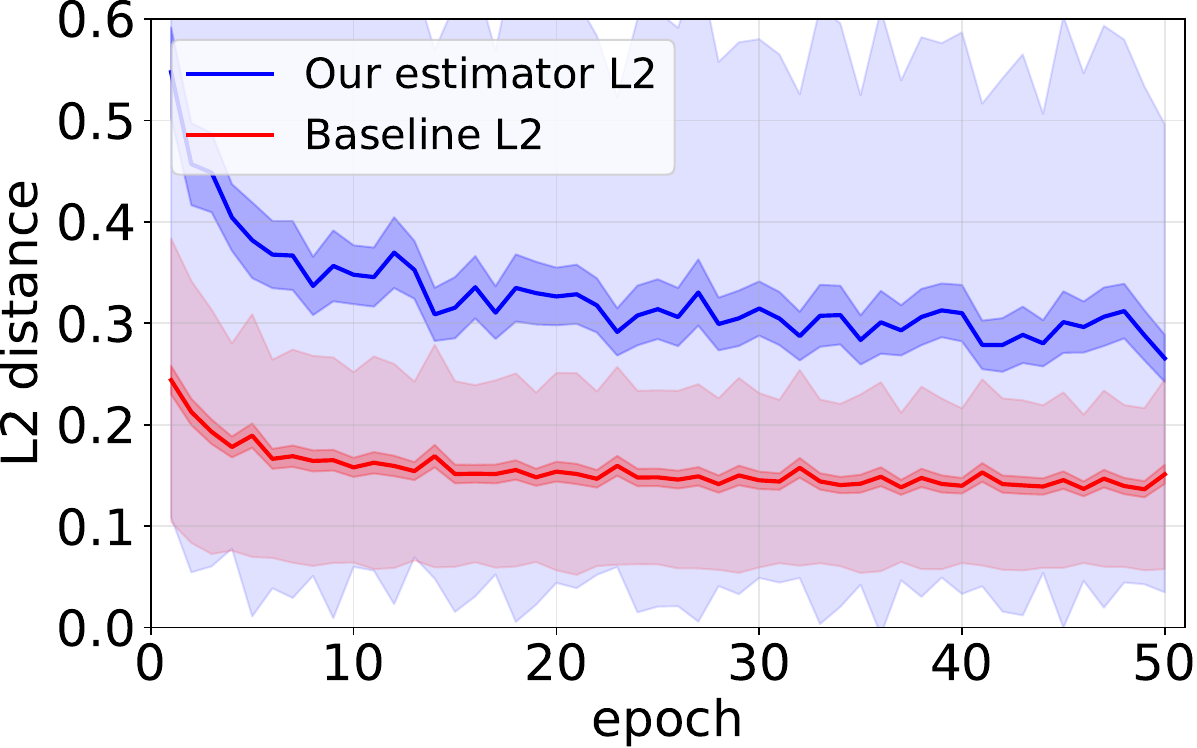} &
\includegraphics[valign=m,width=0.29\textwidth]{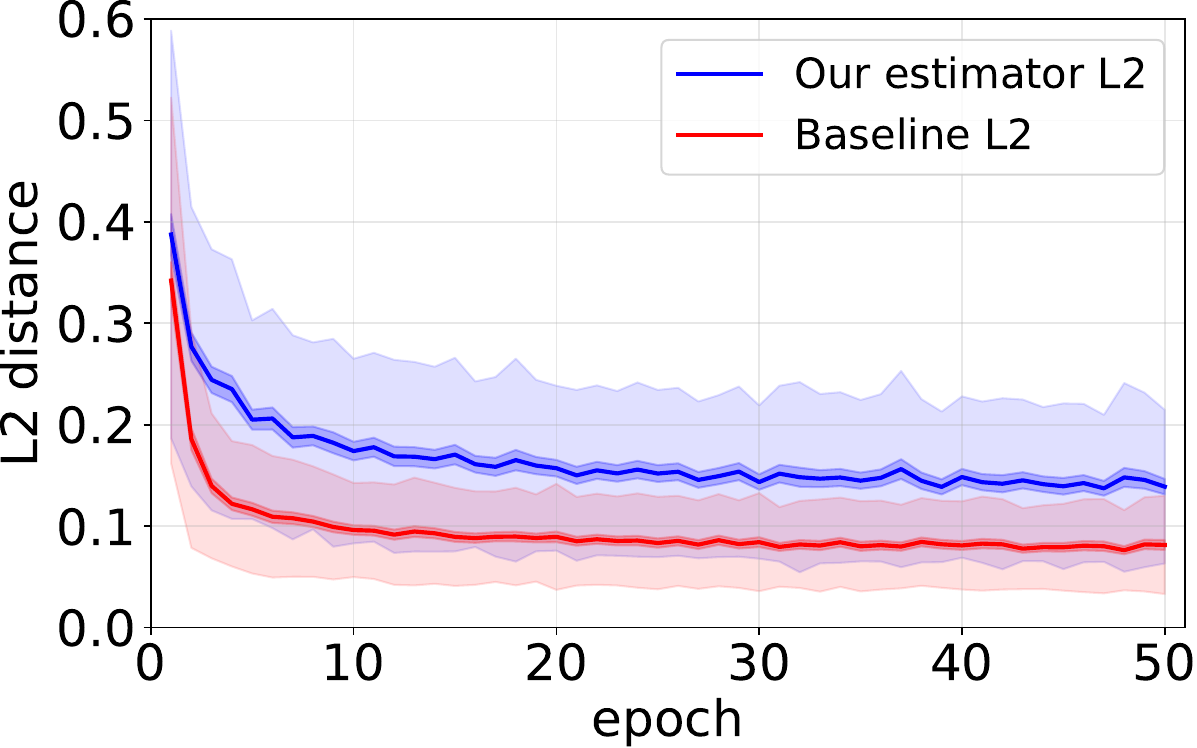} &
\includegraphics[valign=m,width=0.29\textwidth]{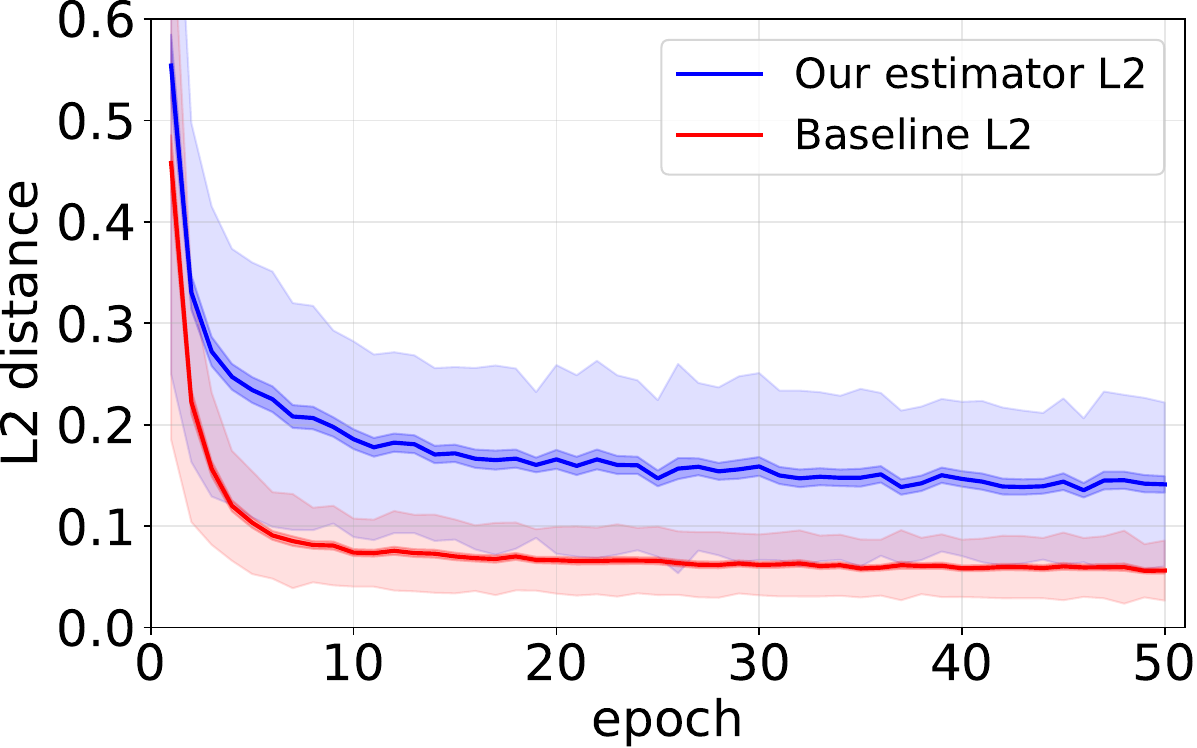} \\

\rotatebox[origin=c]{90}{\footnotesize\textbf{MNIST}} &
\includegraphics[valign=m,width=0.29\textwidth]{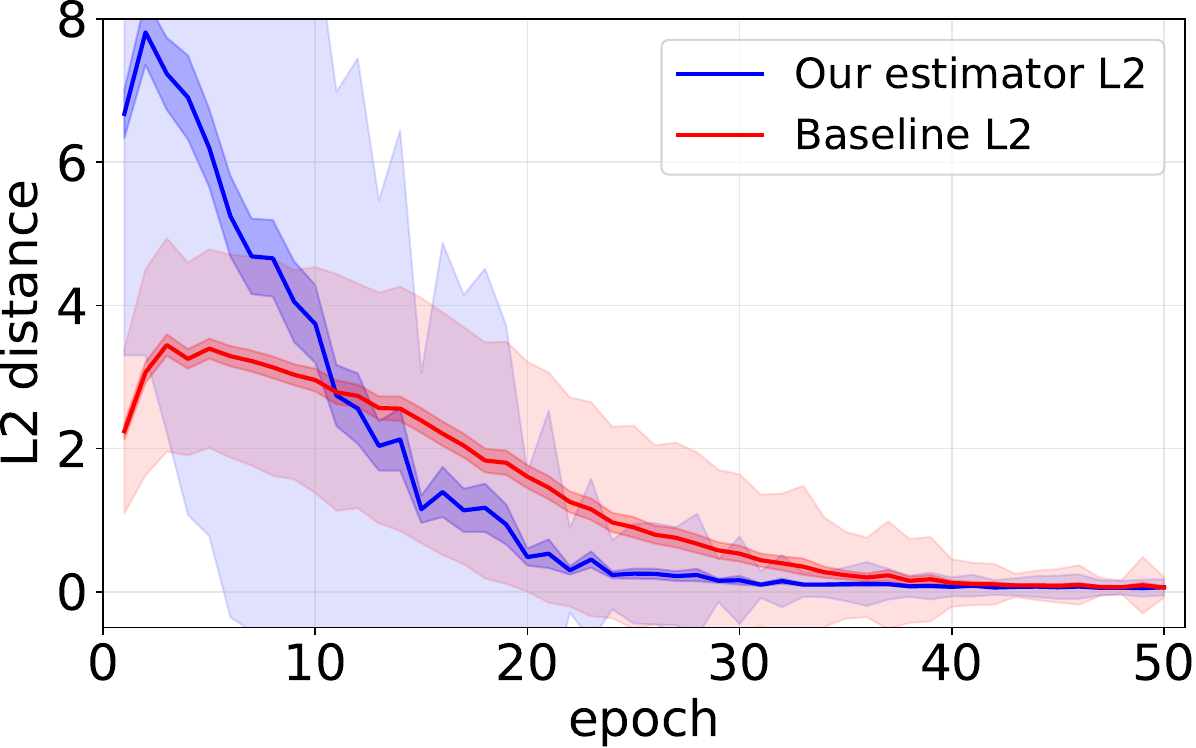} &
\includegraphics[valign=m,width=0.29\textwidth]{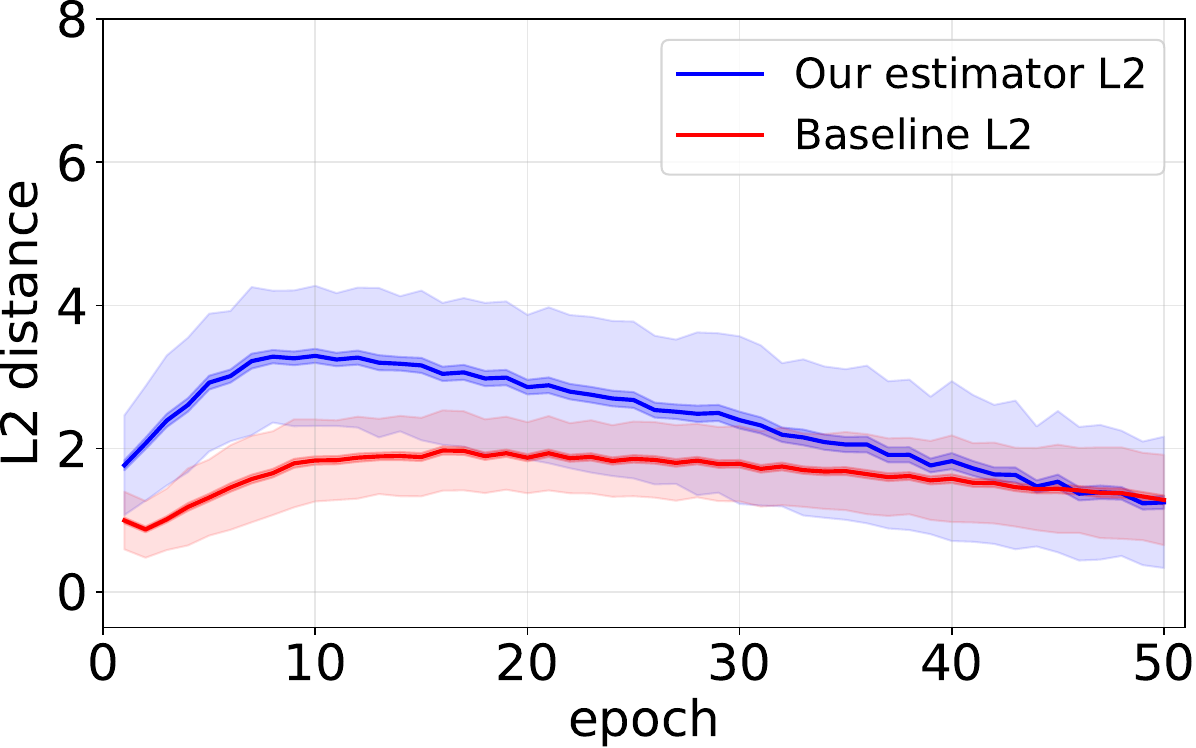} &
\includegraphics[valign=m,width=0.29\textwidth]{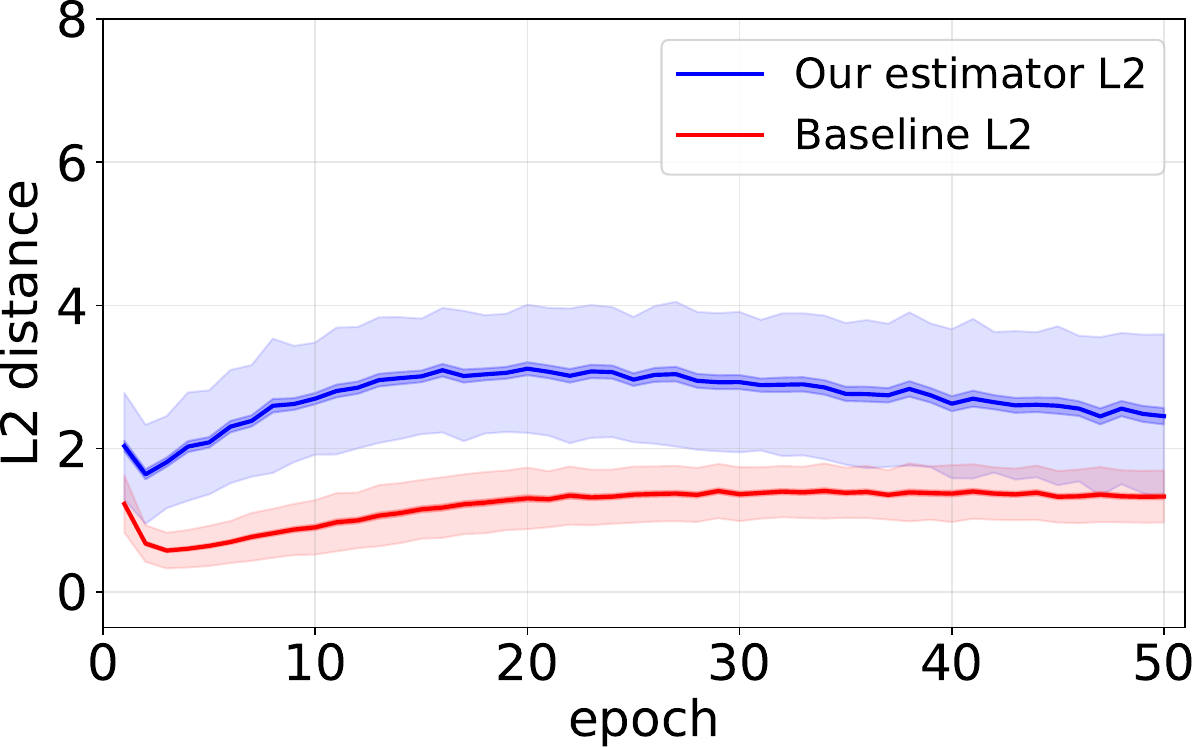} \\

\rotatebox[origin=c]{90}{\footnotesize\textbf{Fashion-MNIST}} &
\includegraphics[valign=m,width=0.29\textwidth]{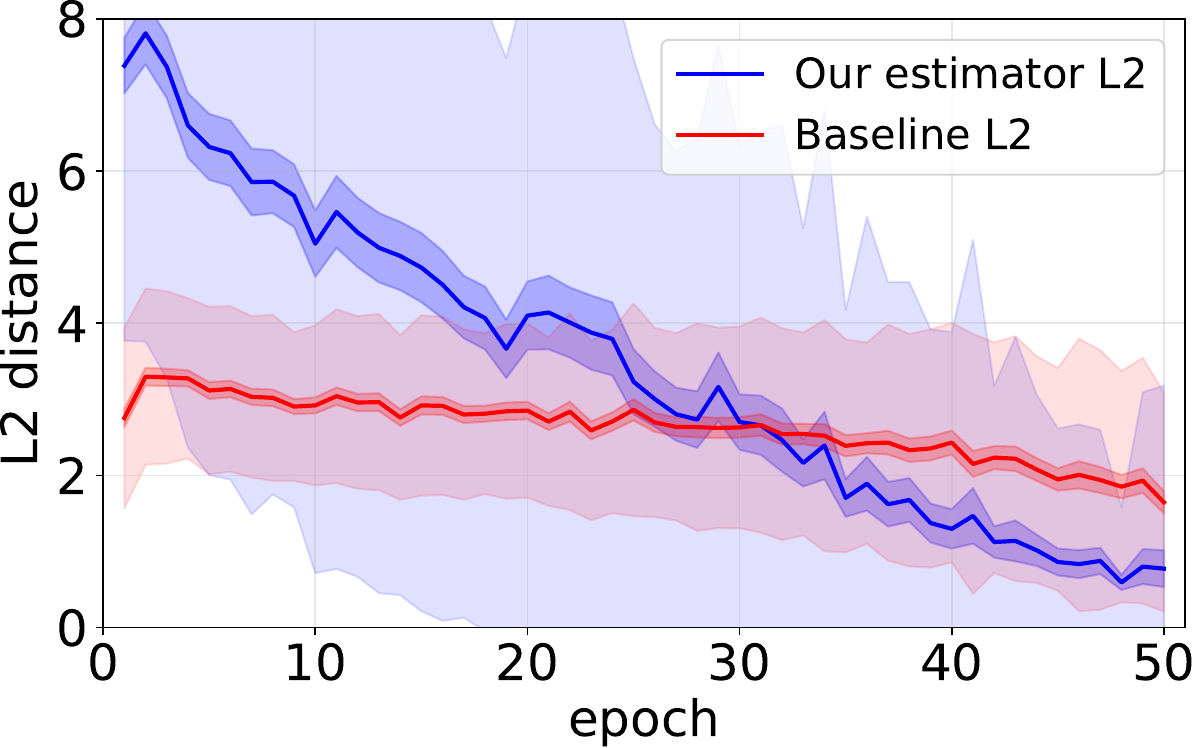} &
\includegraphics[valign=m,width=0.29\textwidth]{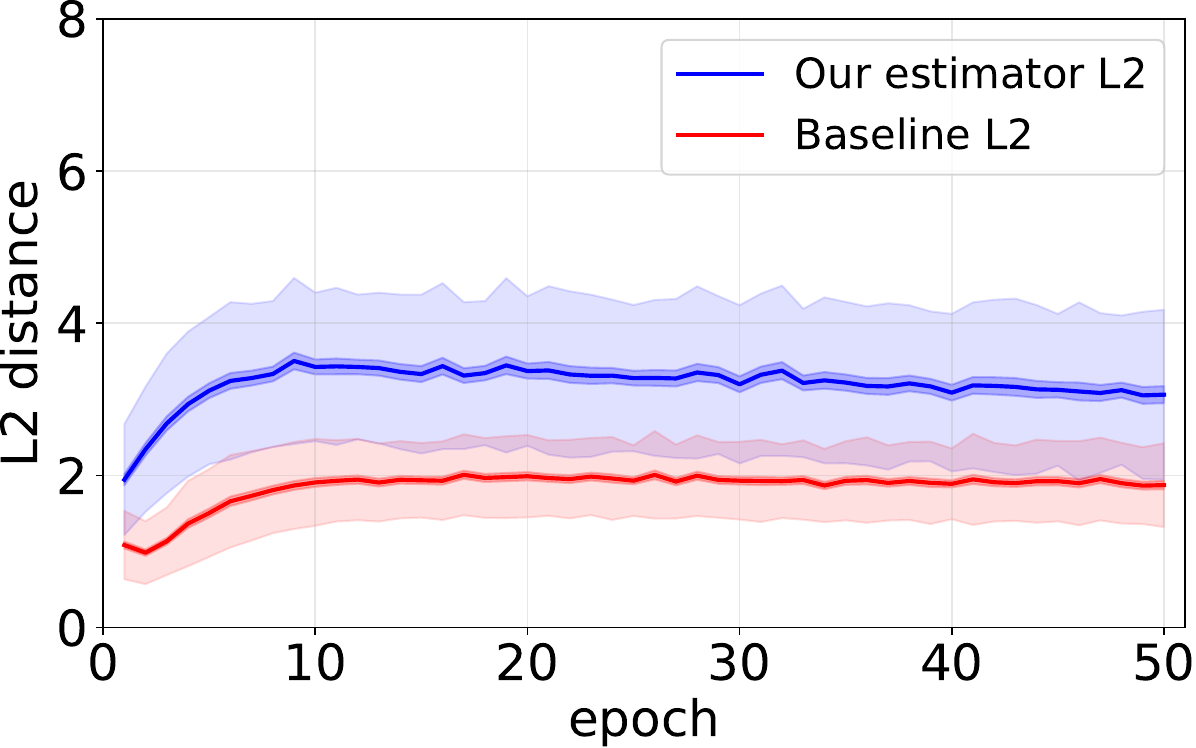} &
\includegraphics[valign=m,width=0.29\textwidth]{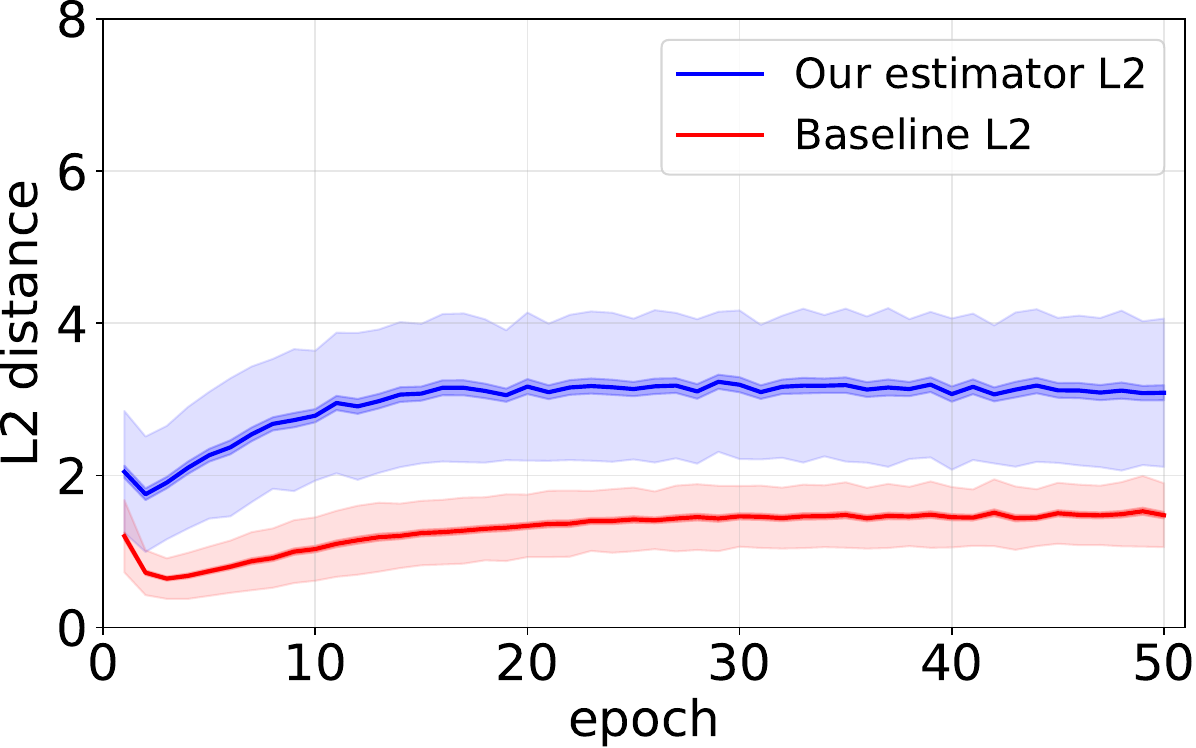} \\

\rotatebox[origin=c]{90}{\footnotesize\textbf{CIFAR-10}} &
\includegraphics[valign=m,width=0.29\textwidth]{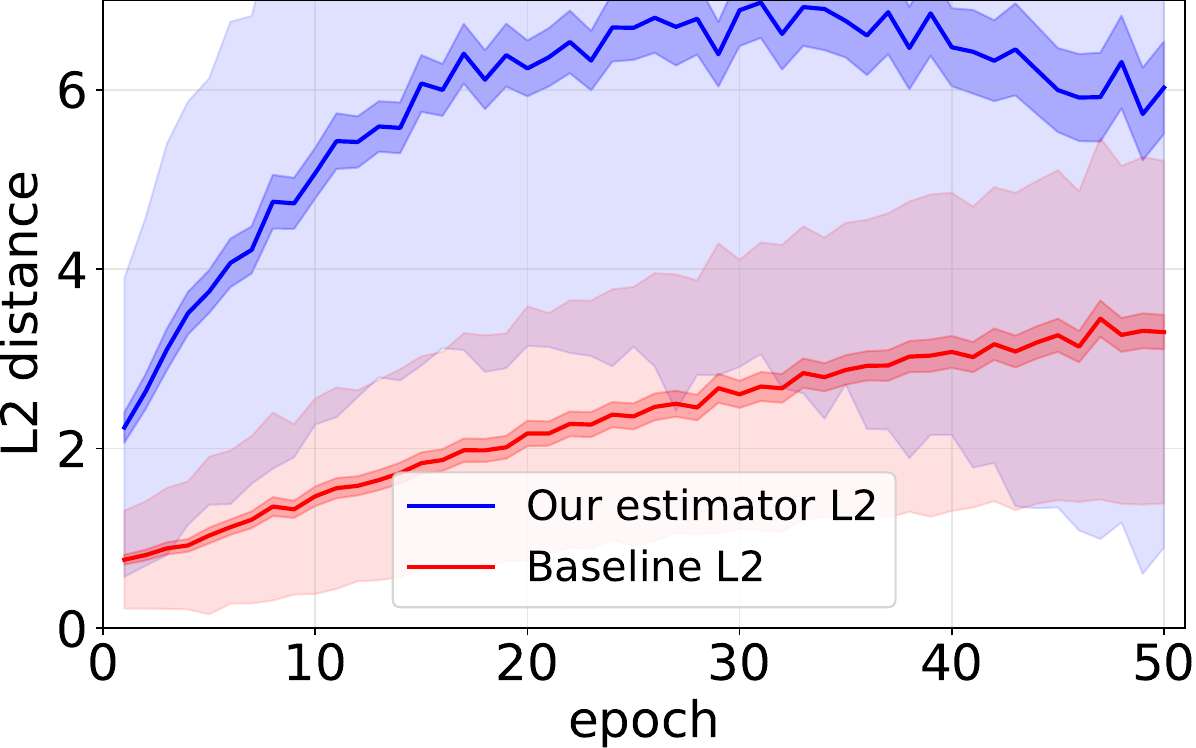} &
\includegraphics[valign=m,width=0.29\textwidth]{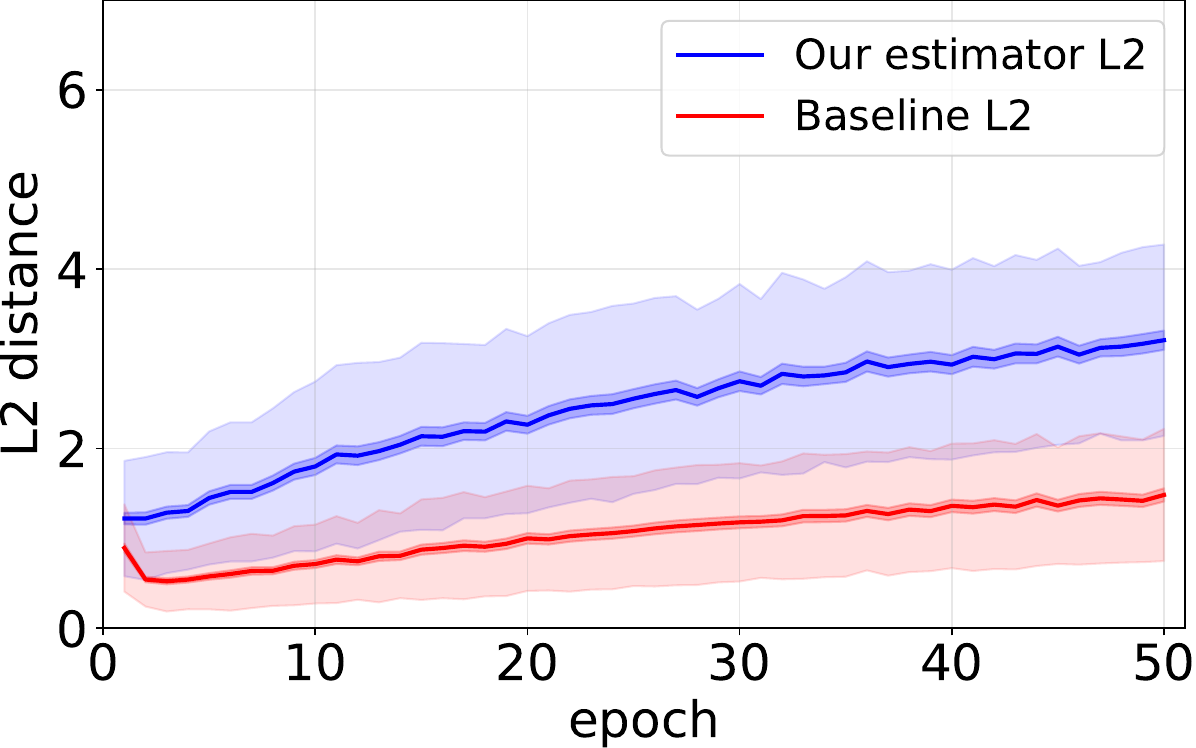} &
\includegraphics[valign=m,width=0.29\textwidth]{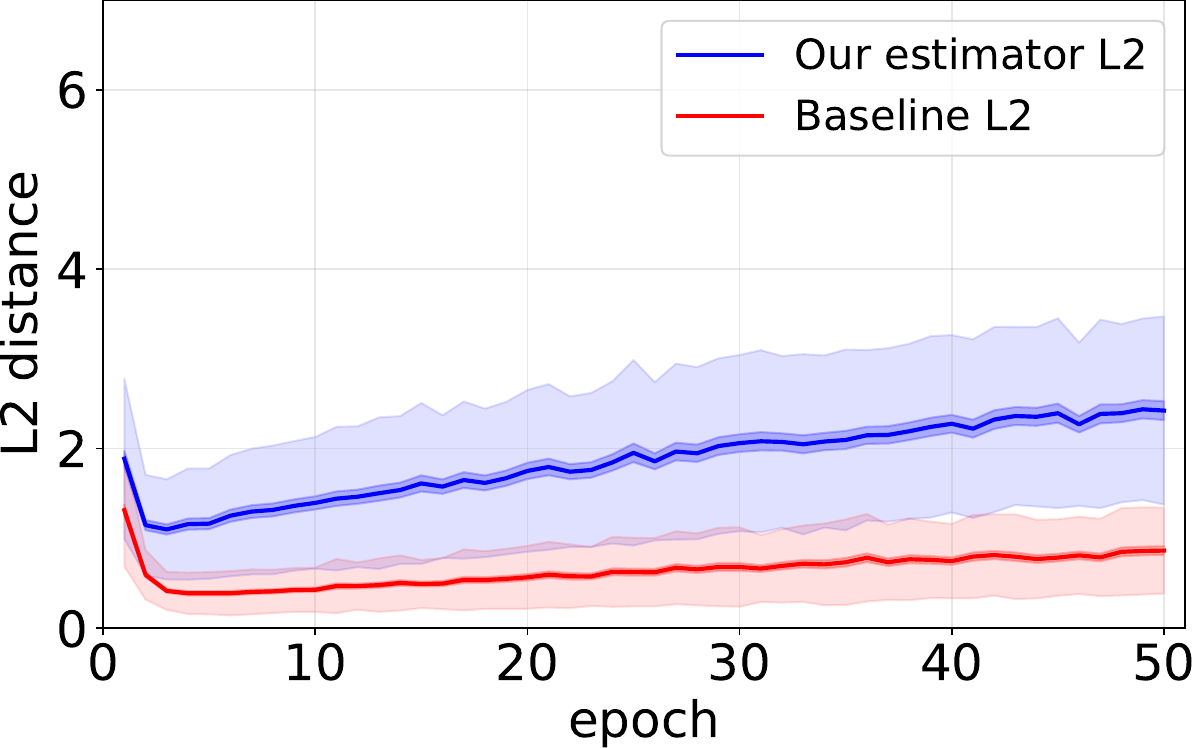} \\

\rotatebox[origin=c]{90}{\footnotesize\textbf{CIFAR-100}} &
\includegraphics[valign=m,width=0.29\textwidth]{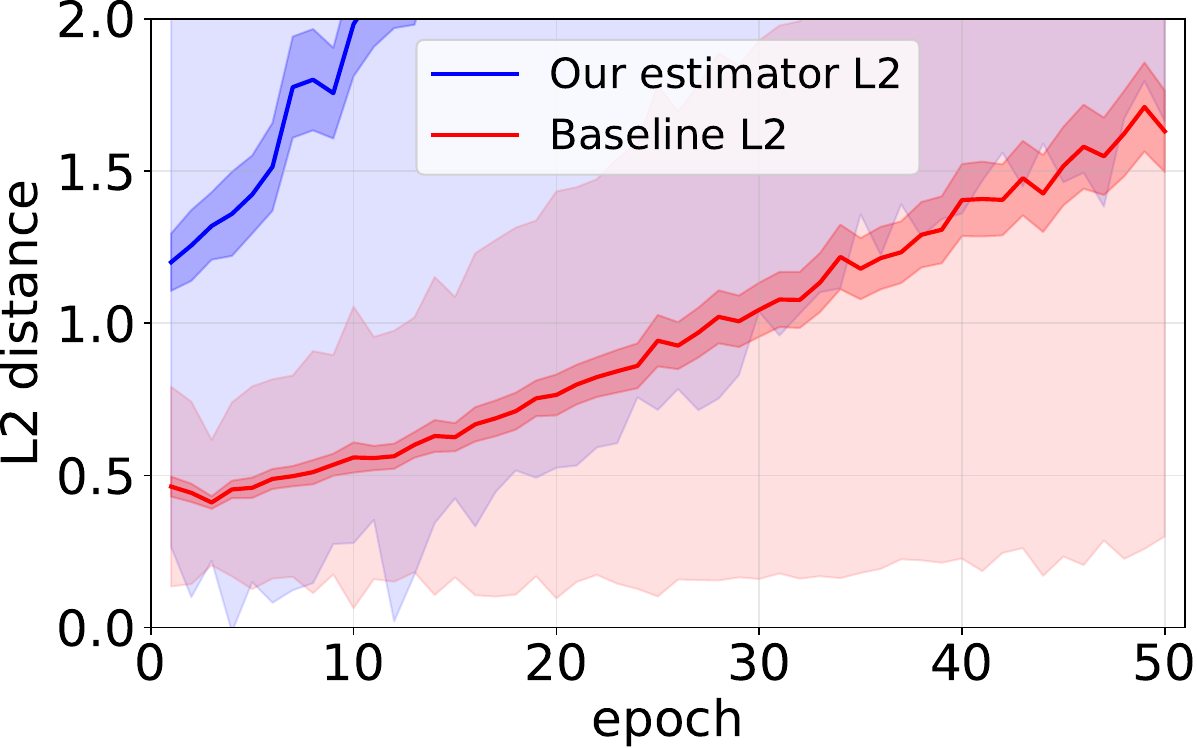} &
\includegraphics[valign=m,width=0.29\textwidth]{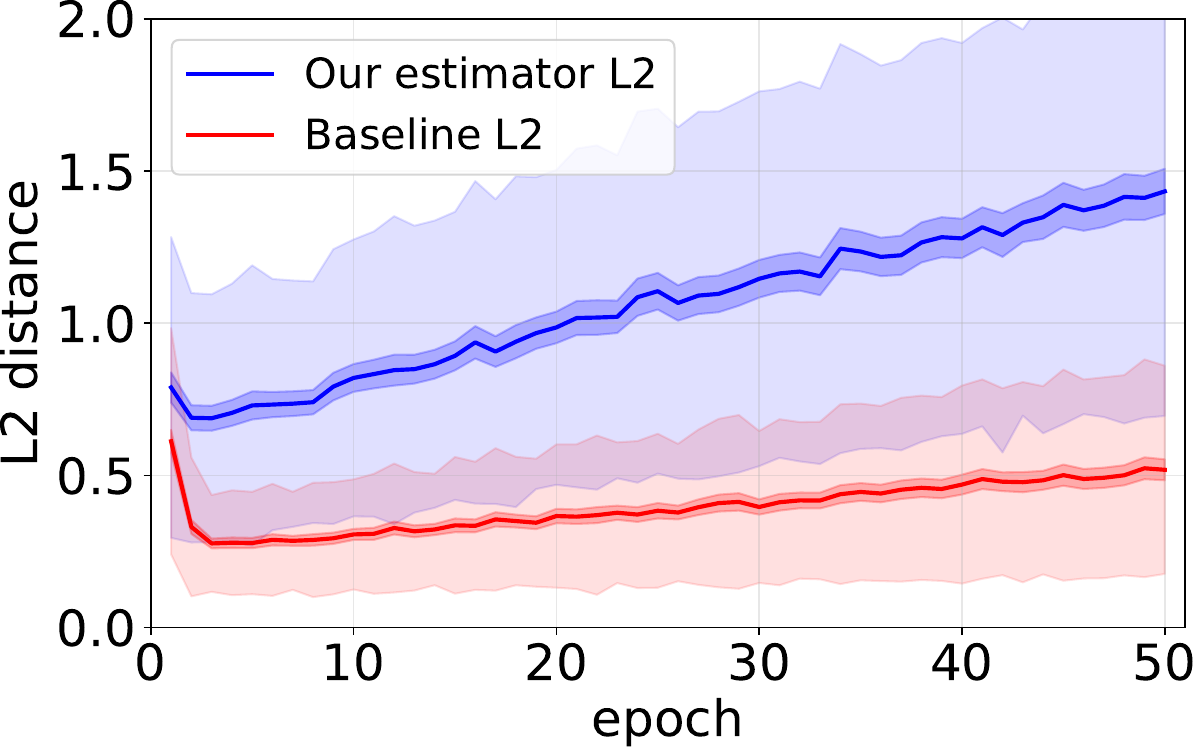} &
\includegraphics[valign=m,width=0.29\textwidth]{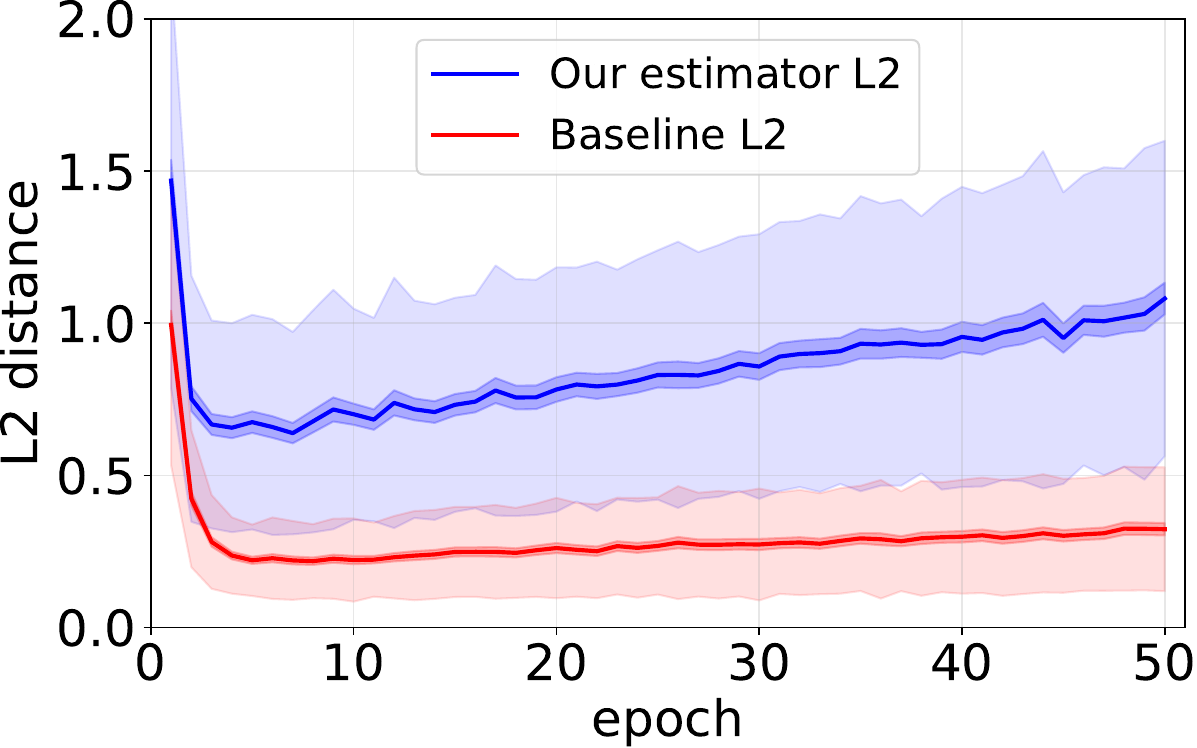} \\

\end{tabular}

\caption{L2 distance to full-batch gradient SGD-M. The darker curve represents the average of 400 runs, the dark shading is the 95\% confidence interval and lighter shading shows the standard deviation. Columns correspond to batch sizes and rows to datasets.}
\label{fig:sgd_grid}
\end{figure*}
\begin{figure*}[t]
\centering
\setlength{\tabcolsep}{3pt}
\renewcommand{\arraystretch}{0.8}

\begin{tabular}{>{\centering\arraybackslash}m{0.04\textwidth}
                >{\centering\arraybackslash}m{0.29\textwidth}
                >{\centering\arraybackslash}m{0.29\textwidth}
                >{\centering\arraybackslash}m{0.29\textwidth}}

& \textbf{Batch size 10} & \textbf{Batch size 50} & \textbf{Batch size 100} \\

\rotatebox[origin=c]{90}{\footnotesize\textbf{Synthetic}} &
\includegraphics[valign=m,width=0.29\textwidth]{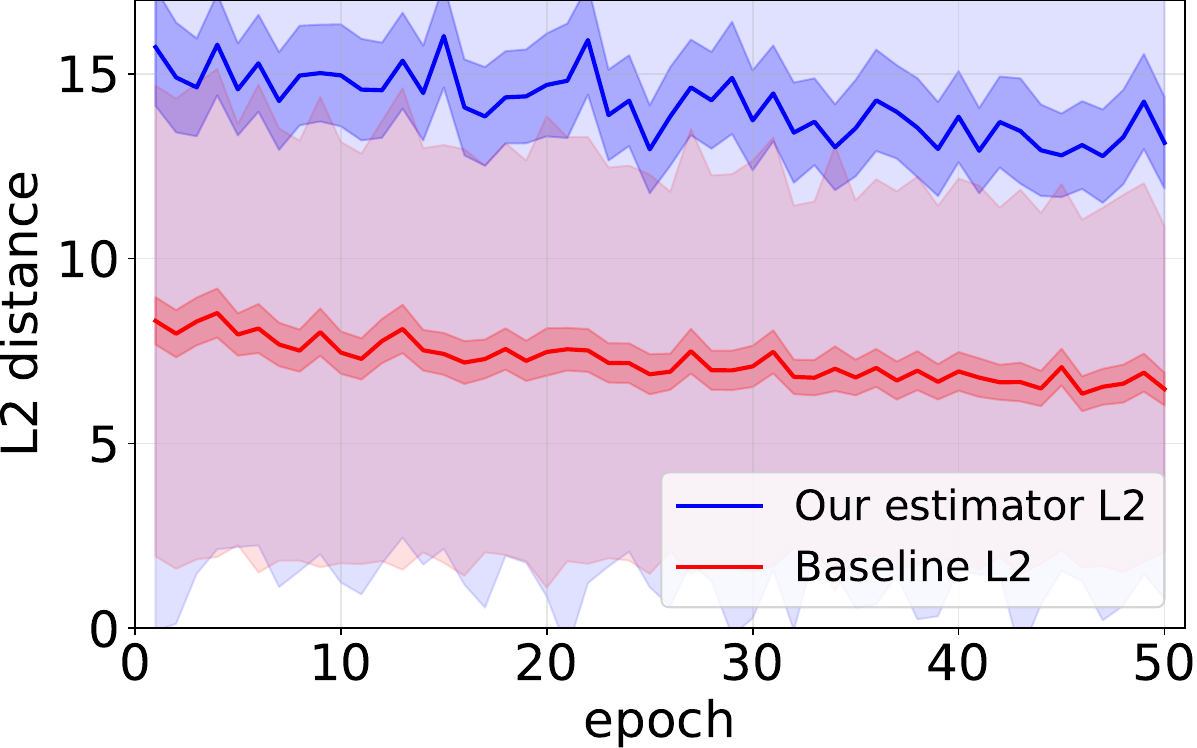} &
\includegraphics[valign=m,width=0.29\textwidth]{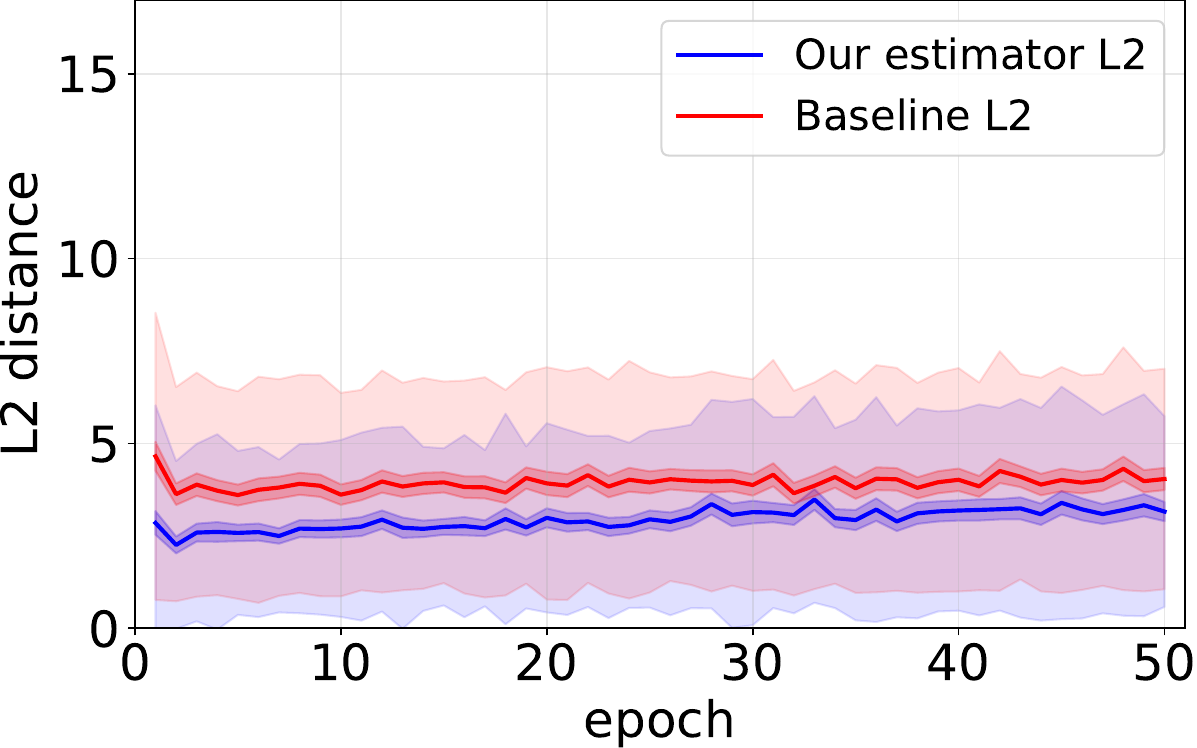} &
\includegraphics[valign=m,width=0.29\textwidth]{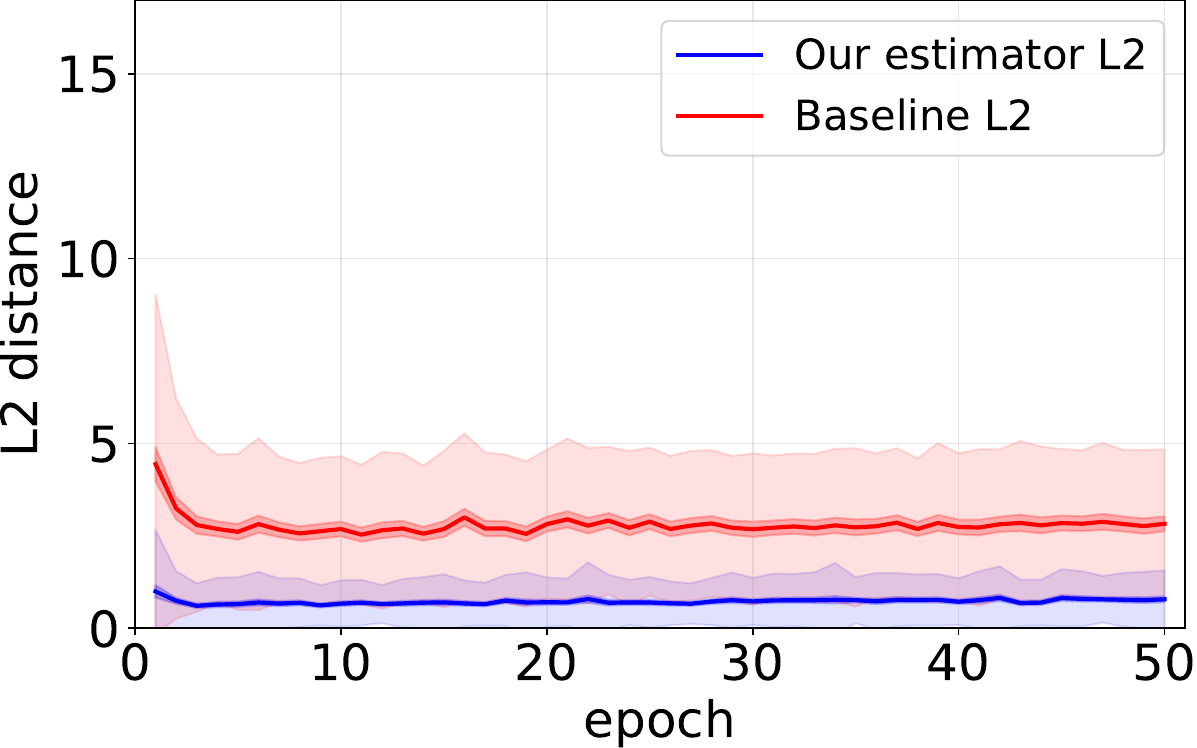} \\

\rotatebox[origin=c]{90}{\footnotesize\textbf{Airfoil self-noise}} &
\includegraphics[valign=m,width=0.29\textwidth]{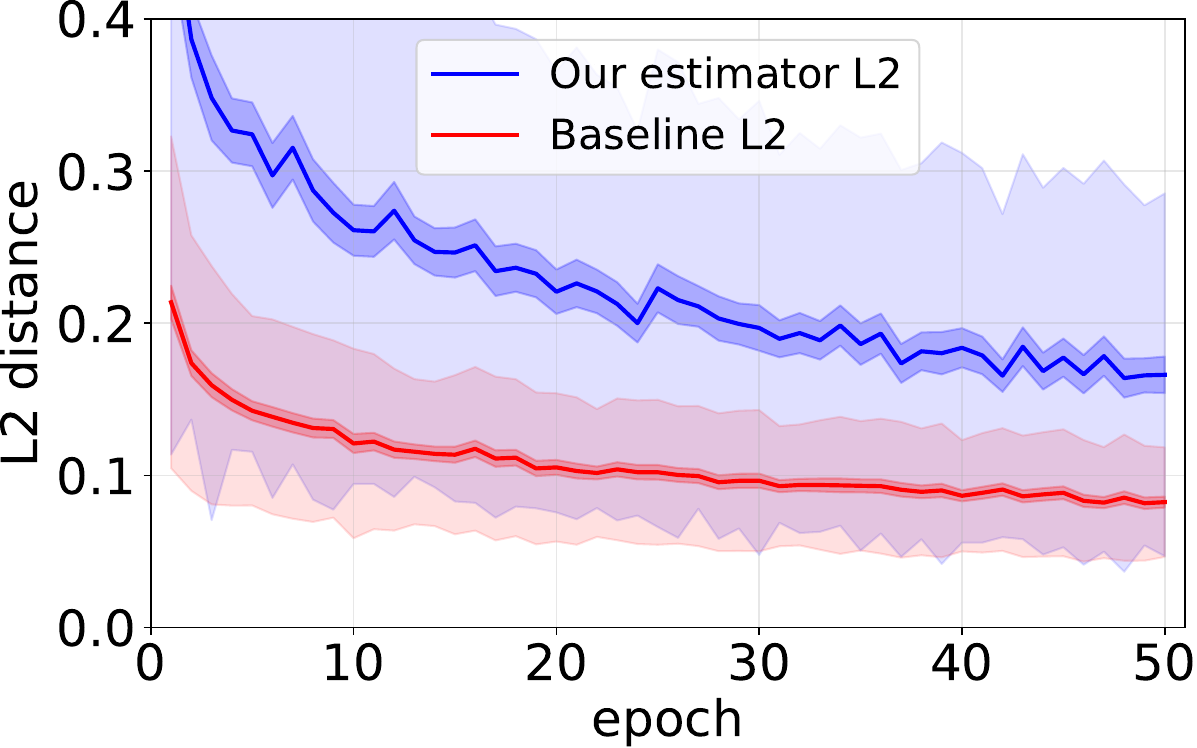} &
\includegraphics[valign=m,width=0.29\textwidth]{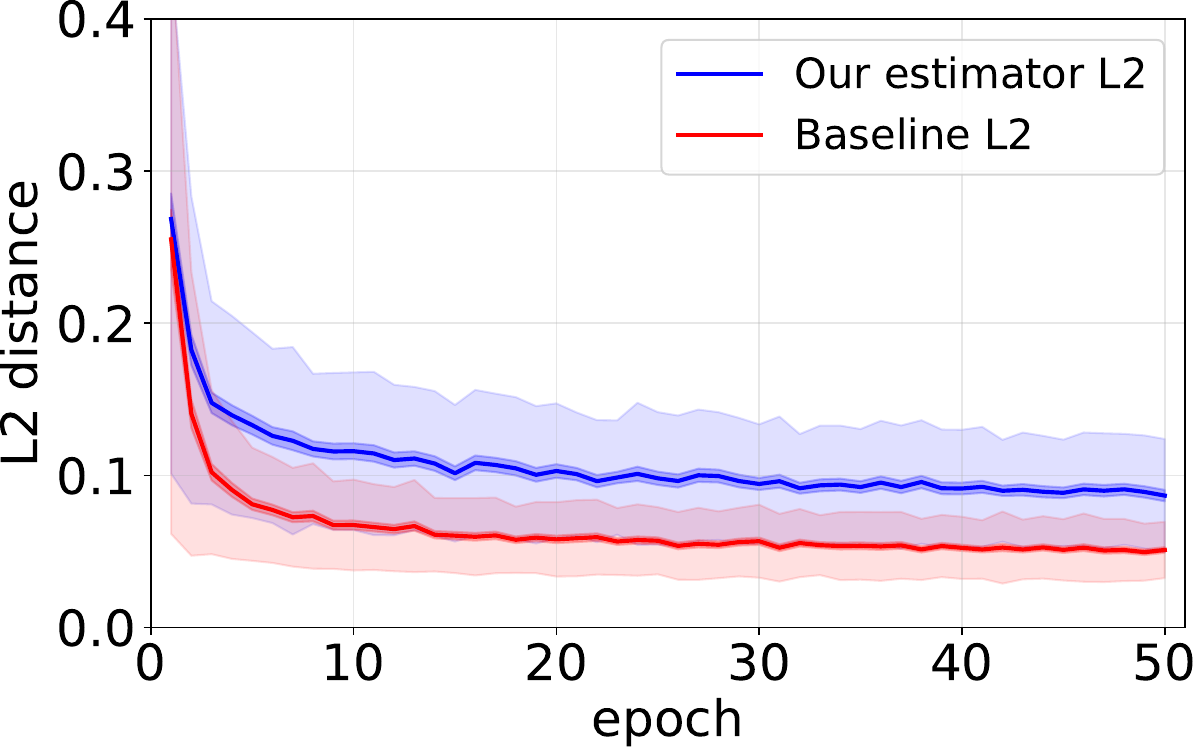} &
\includegraphics[valign=m,width=0.29\textwidth]{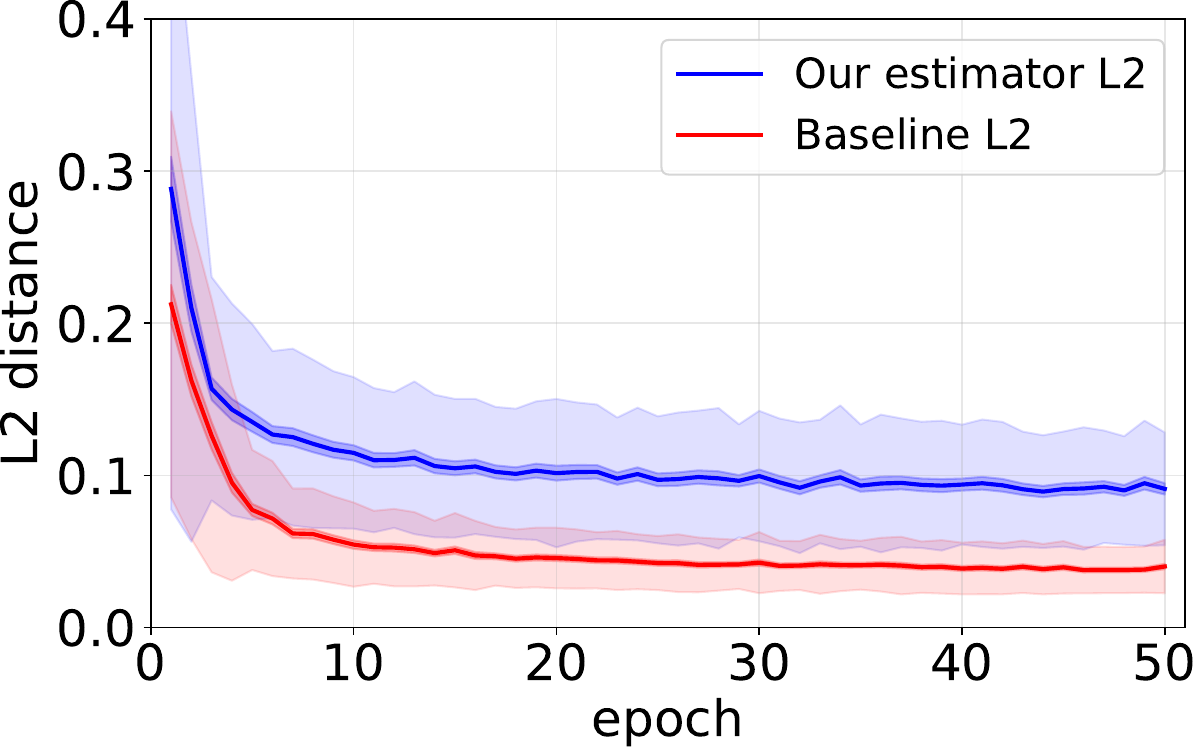} \\

\rotatebox[origin=c]{90}{\footnotesize\textbf{Appliances energy}} &
\includegraphics[valign=m,width=0.29\textwidth]{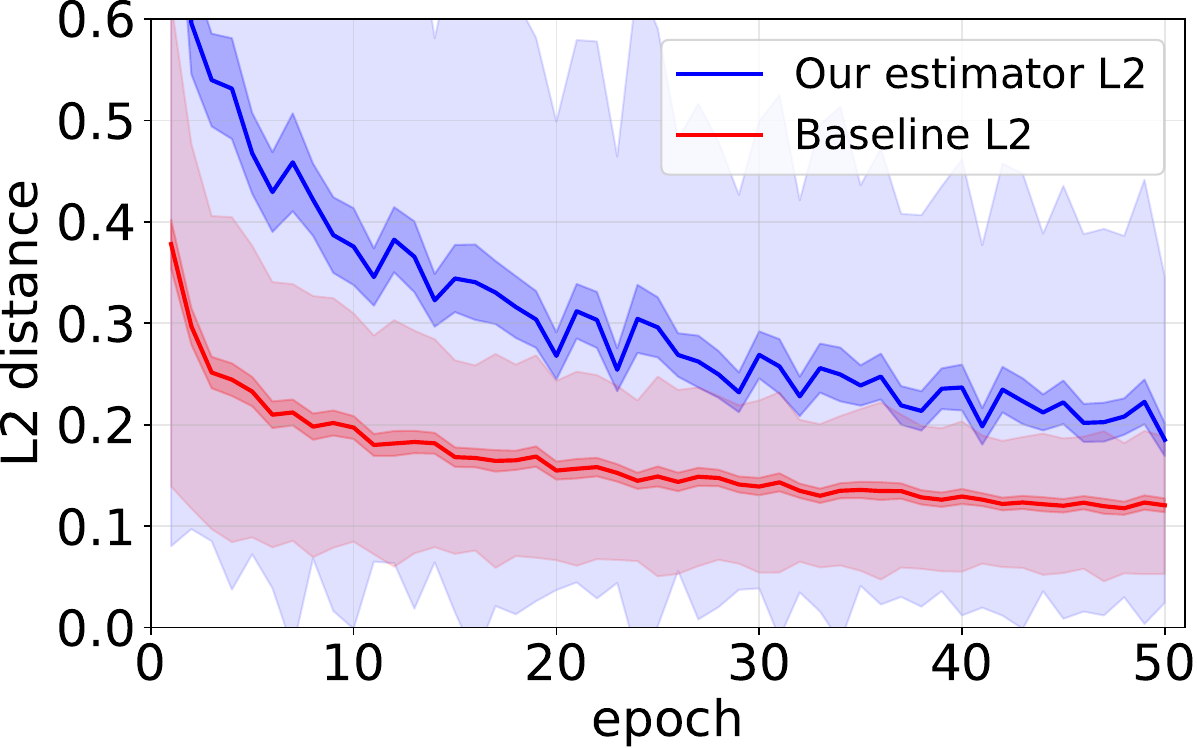} &
\includegraphics[valign=m,width=0.29\textwidth]{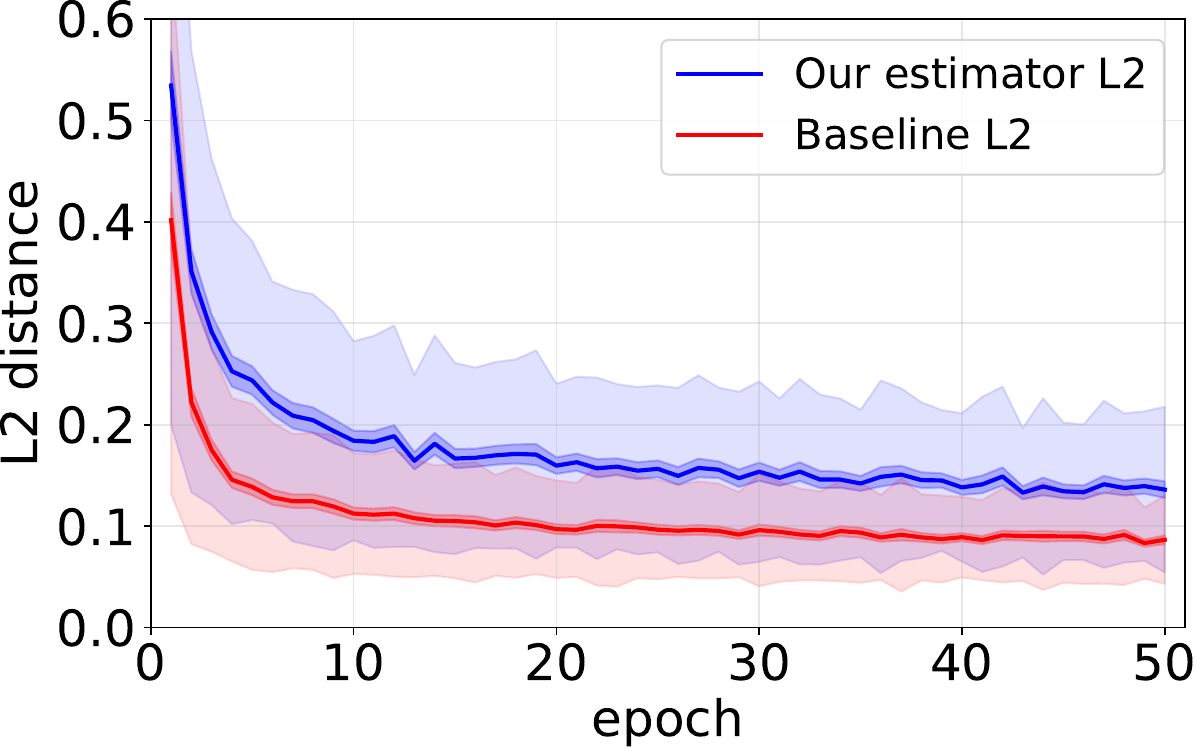} &
\includegraphics[valign=m,width=0.29\textwidth]{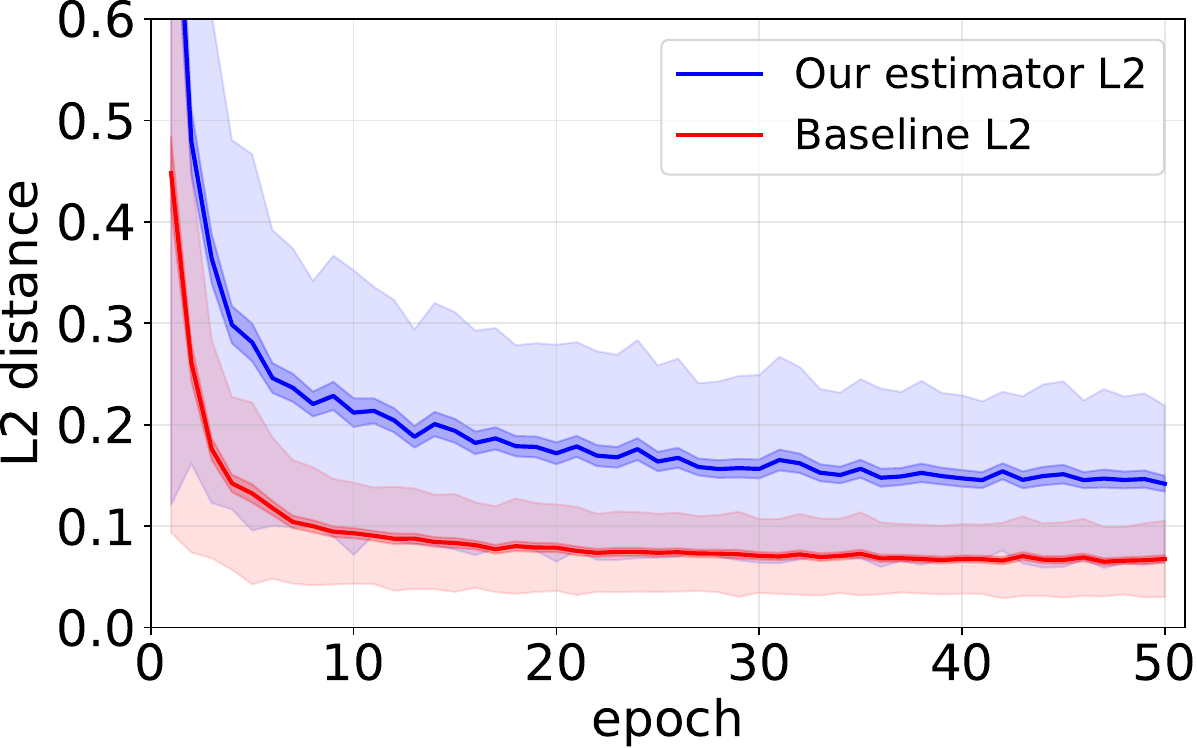} \\

\rotatebox[origin=c]{90}{\footnotesize\textbf{MNIST}} &
\includegraphics[valign=m,width=0.29\textwidth]{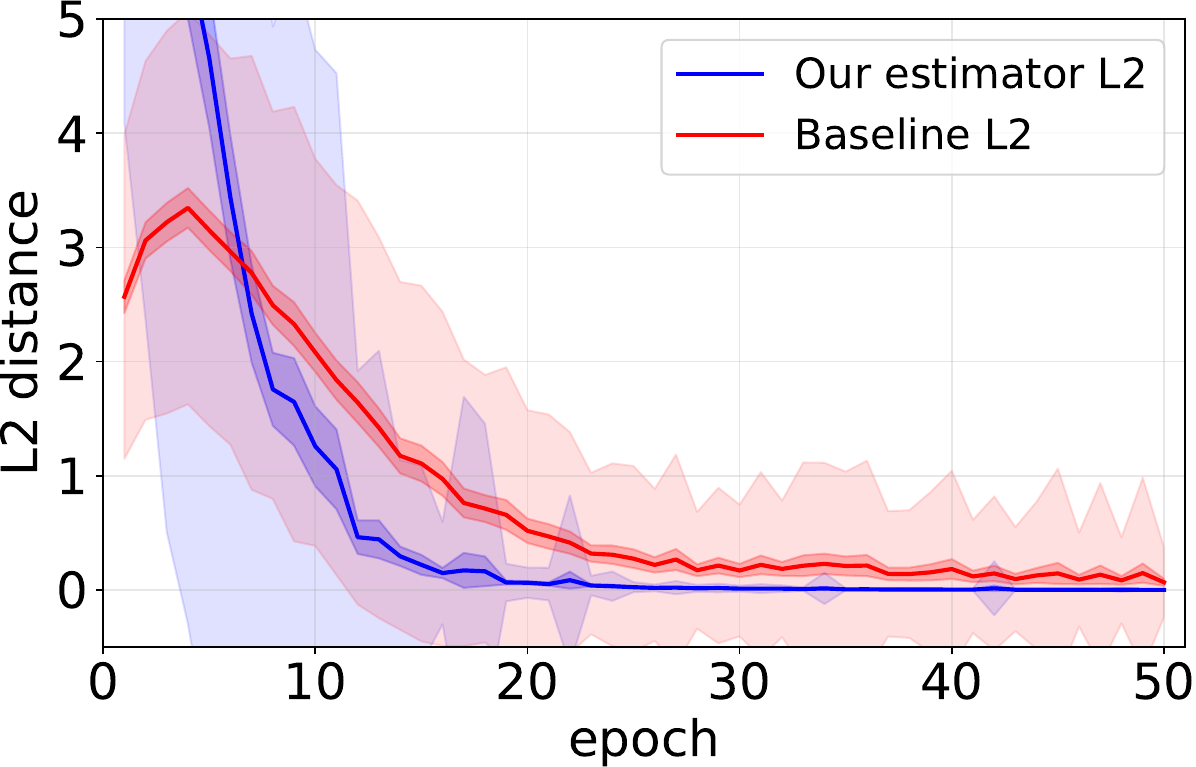} &
\includegraphics[valign=m,width=0.29\textwidth]{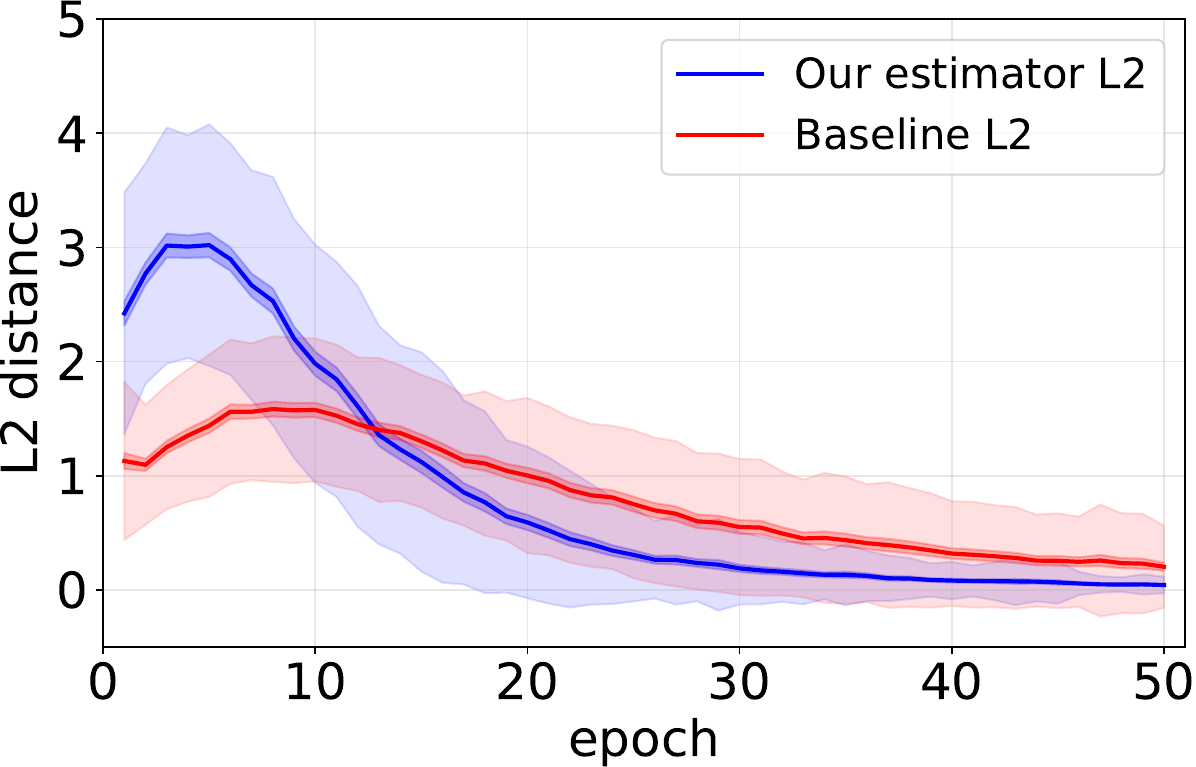} &
\includegraphics[valign=m,width=0.29\textwidth]{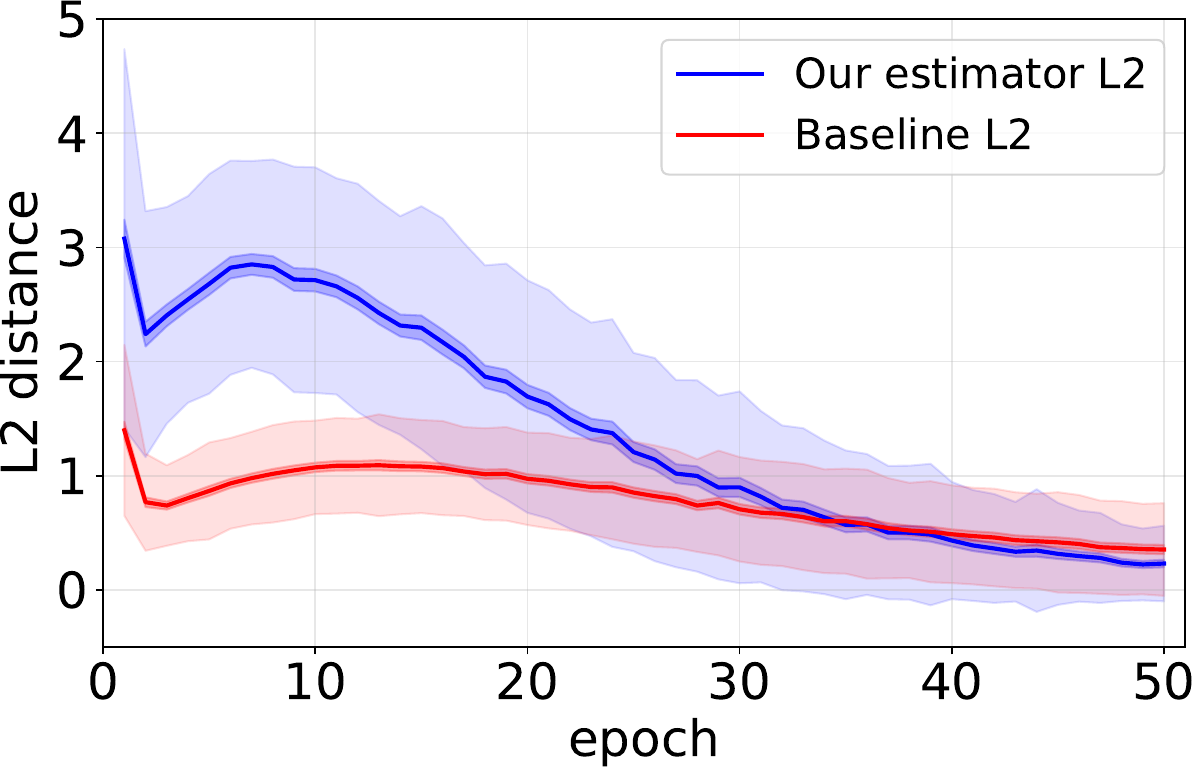} \\

\rotatebox[origin=c]{90}{\footnotesize\textbf{Fashion-MNIST}} &
\includegraphics[valign=m,width=0.29\textwidth]{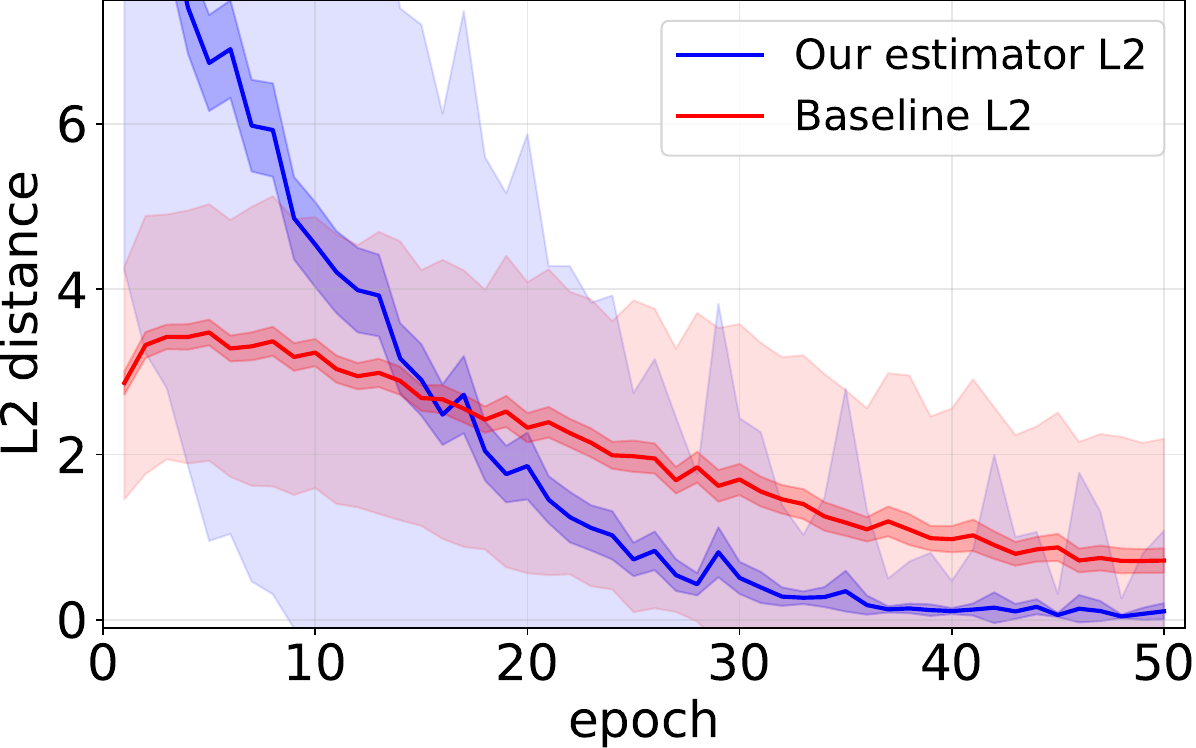} &
\includegraphics[valign=m,width=0.29\textwidth]{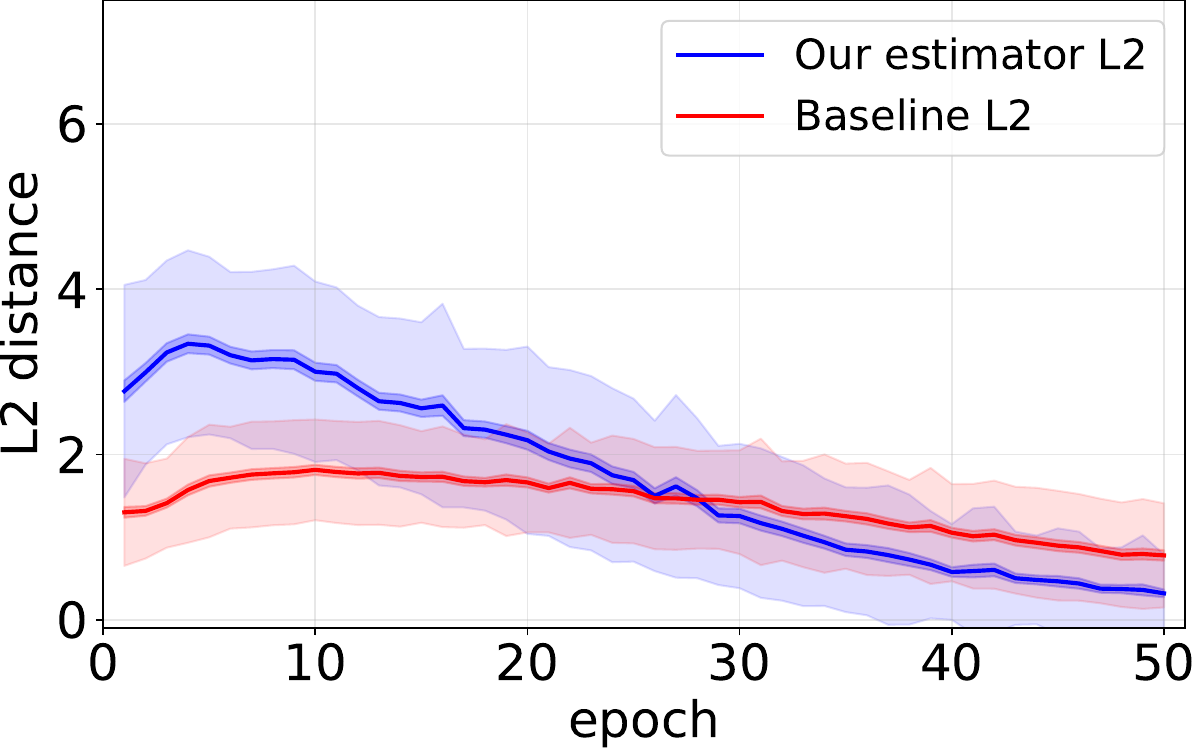} &
\includegraphics[valign=m,width=0.29\textwidth]{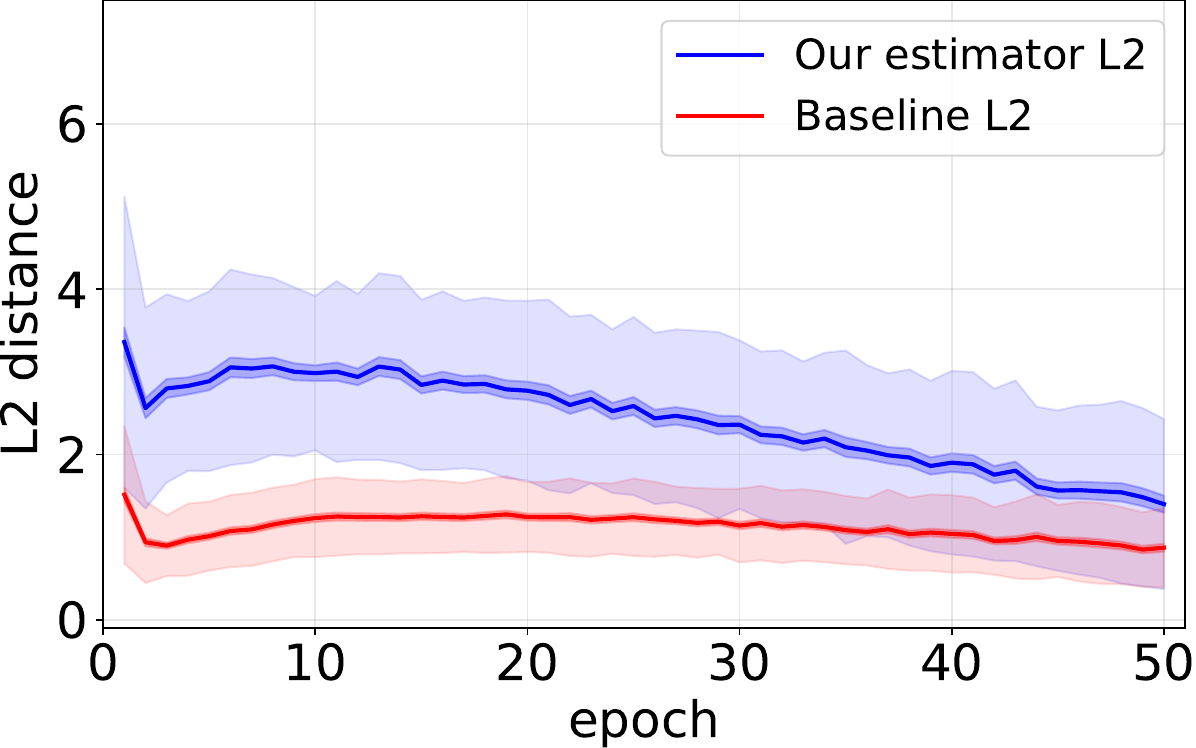} \\

\rotatebox[origin=c]{90}{\footnotesize\textbf{CIFAR-10}} &
\includegraphics[valign=m,width=0.29\textwidth]{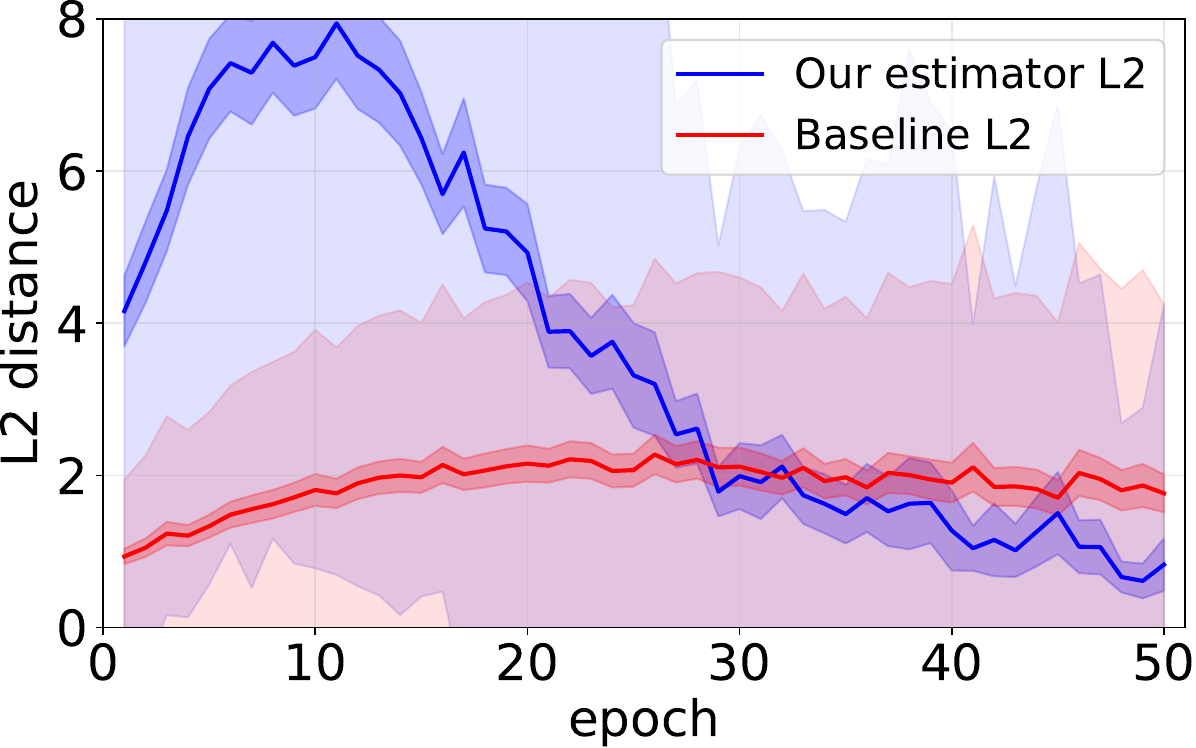} &
\includegraphics[valign=m,width=0.29\textwidth]{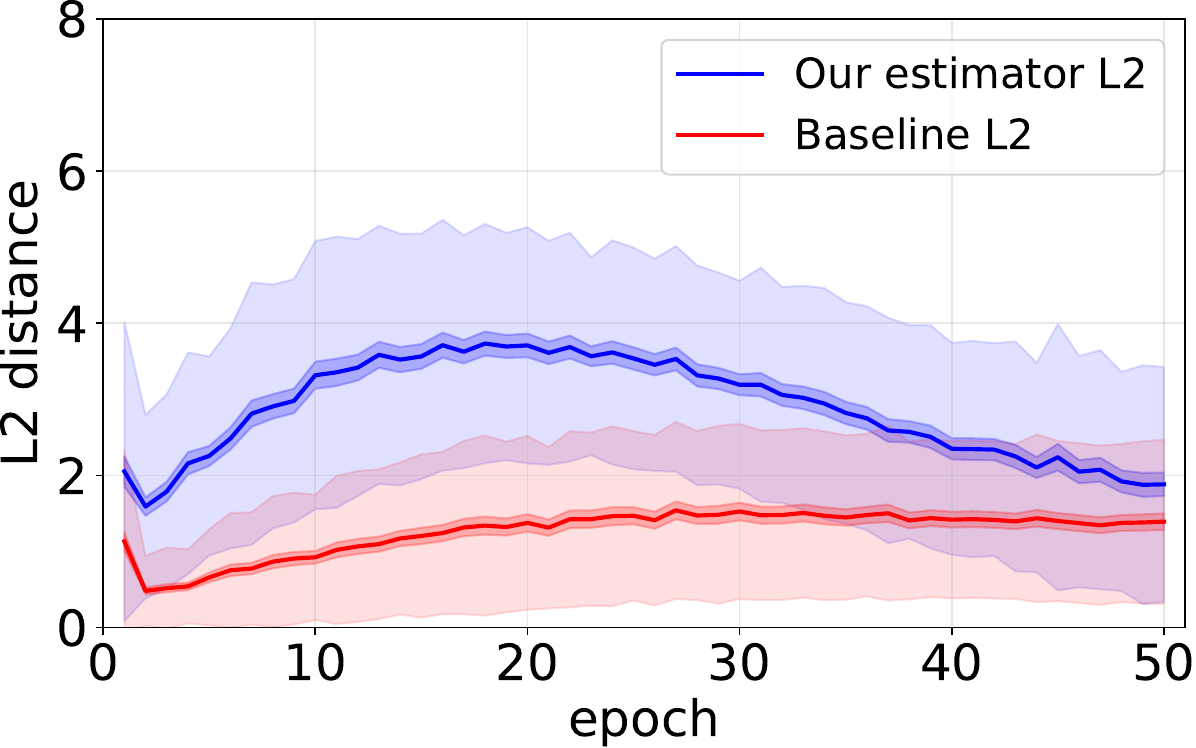} &
\includegraphics[valign=m,width=0.29\textwidth]{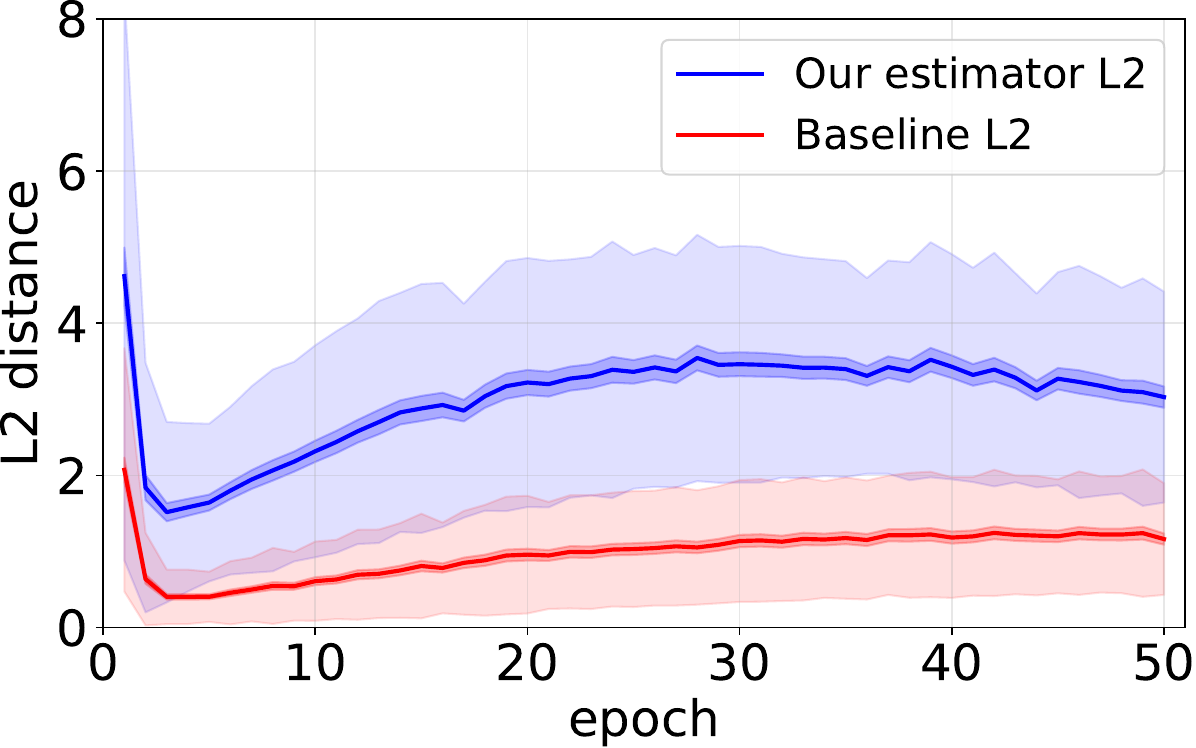} \\

\rotatebox[origin=c]{90}{\footnotesize\textbf{CIFAR-100}} &
\includegraphics[valign=m,width=0.29\textwidth]{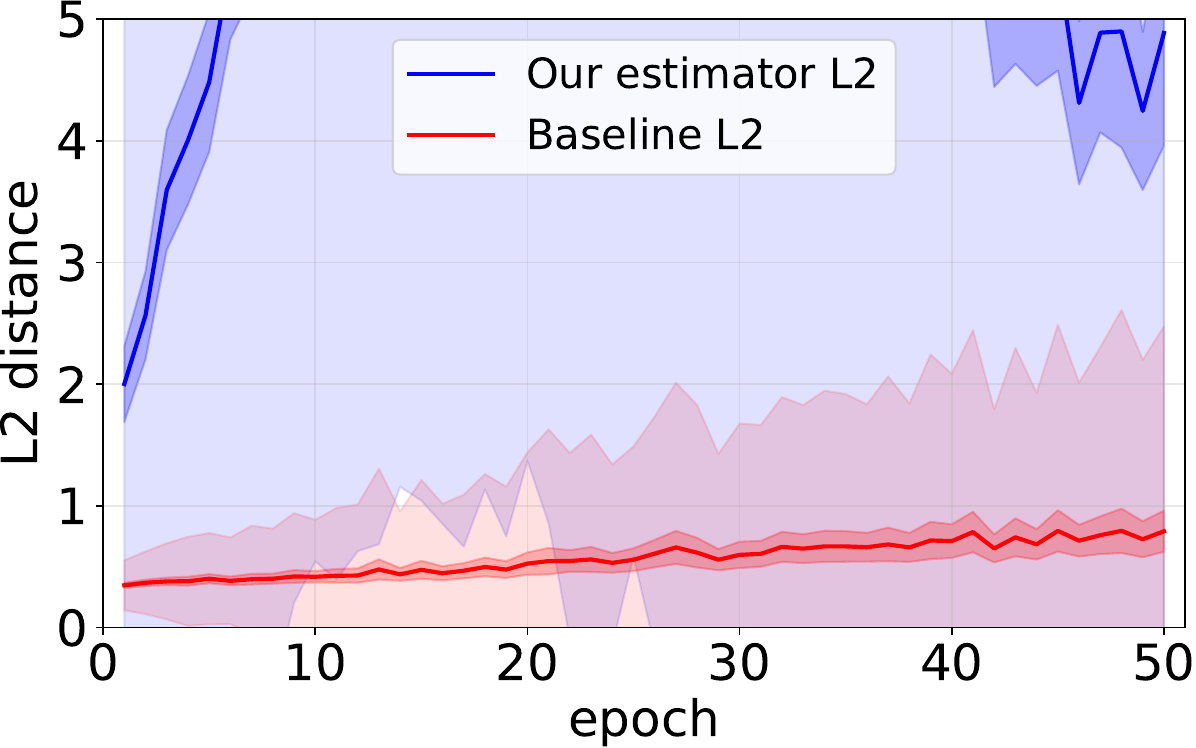} &
\includegraphics[valign=m,width=0.29\textwidth]{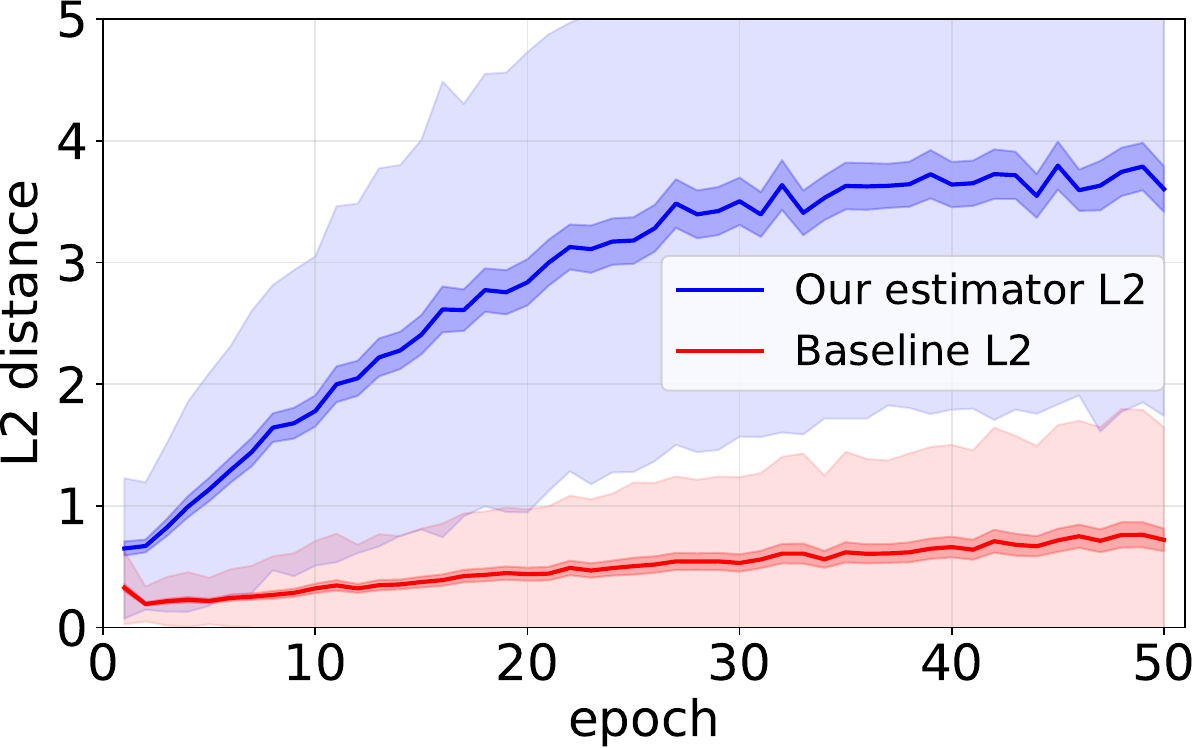} &
\includegraphics[valign=m,width=0.29\textwidth]{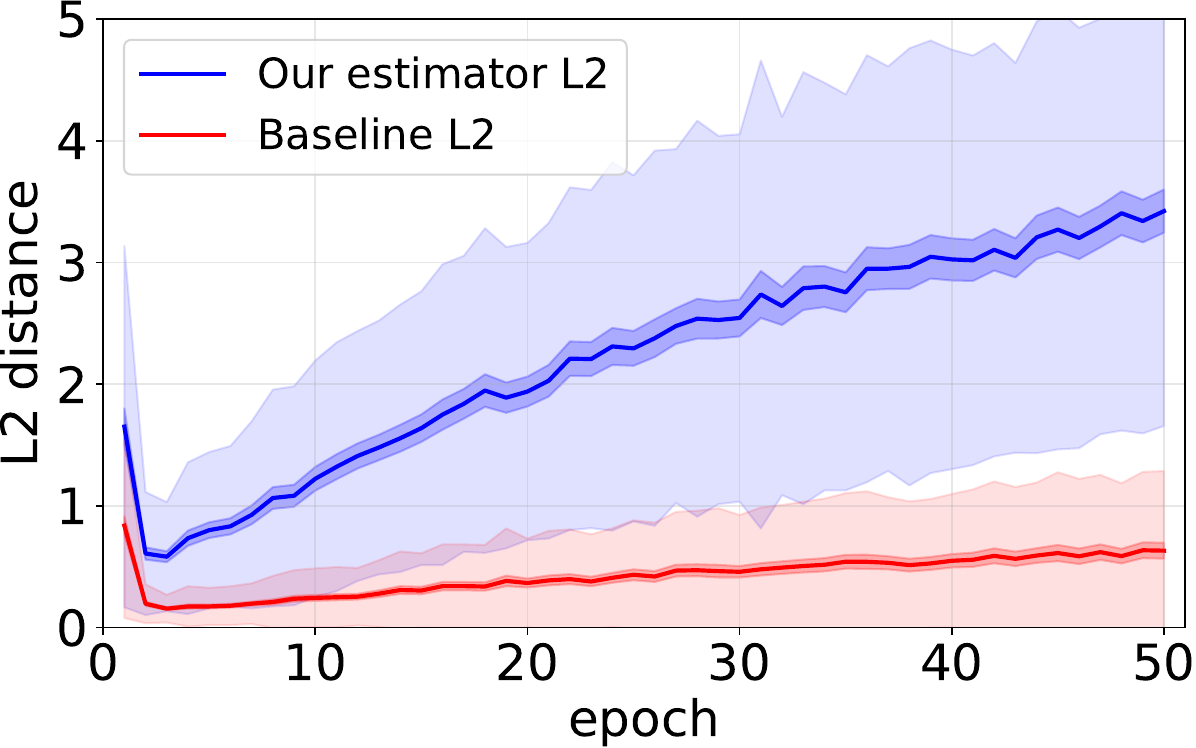} \\

\end{tabular}

\caption{L2 distance to full-batch gradient Adam. The darker curve represents the average of 400 runs, the dark shading is the 95\% confidence interval and lighter shading shows the standard deviation. Columns correspond to batch sizes and rows to datasets.}
\label{fig:Adam_l2_grid}
\end{figure*}

\subsection*{Difference estimator covariance}
\begin{equation} \label{Equation::difference_estimator_variance_proof}
    \begin{aligned}
        \mathbb{V}(\Qdiff) &= \Es{\left(\Qdiff-\sum_{k=1}^N \textbf{q}_k\right)\left(\Qdiff-\sum_{j=1}^N \textbf{q}_j\right)^T} \\
        &= \Es{\left(\sum_{i=1}^N \hat{\textbf{q}}_i + \sum_{i=1}^N I_i\frac{\textbf{q}_i-\hat{\textbf{q}}_i}{\pi_i}-\sum_{k=1}^N \textbf{q}_k\right)\left(\sum_{i=1}^N \hat{\textbf{q}}_i + \sum_{i=1}^N I_i\frac{\textbf{q}_i-\hat{\textbf{q}}_i}{\pi_i}-\sum_{j=1}^N \textbf{q}_j\right)^T} \\
        &= \mathbb{E}_{\mathcal{D}} \Bigg[ 
        \sum_{i,j}\hat{\bq}_i\hat{\bq}_j^T  + \sum_{i,j}I_j\frac{\hat{\bq}_i\bq_j^T-\hat{\bq}_i\hat{\bq}_j^T}{\pi_j} - \sum_{i,j}\hat{\bq}_i\bq_j^T \\
        &\phantom{=} +\sum_{i,j} I_i\frac{\bq_i\hat{\bq}_j^T-\hat{\bq}_i\hat{\bq}_j^T}{\pi_i} + \sum_{i,j}I_iI_j\frac{(\bq_i-\hat{\bq}_i)(\bq_j-\hat{\bq}_j)^T}{\pi_i\pi_j} - \sum_{i,j} I_i\frac{\bq_i\bq_j^T-\hat{\bq}_i\bq_j^T}{\pi_i} \\
        &\phantom{=} -
        \sum_{i,j}\bq_i\hat{\bq}_j^T  - \sum_{i,j}I_j\frac{\bq_i\bq_j^T-\bq_i\hat{\bq}_j^T}{\pi_j} + \sum_{i,j}\bq_i\bq_j^T \Bigg] \\
        &=  
        \sum_{i,j}\hat{\bq}_i\hat{\bq}_j^T  + \sum_{i,j}\hat{\bq}_i\bq_j^T-\sum_{i,j}\hat{\bq}_i\hat{\bq}_j^T - \sum_{i,j}\hat{\bq}_i\bq_j^T \\
        &\phantom{=} +\sum_{i,j} \bq_i\hat{\bq}_j^T-\sum_{i,j}\hat{\bq}_i\hat{\bq}_j^T + \sum_{i,j}\pi_{ij}\frac{(\bq_i-\hat{\bq}_i)(\bq_j-\hat{\bq}_j)^T}{\pi_i\pi_j} - \sum_{i,j} \bq_i\bq_j^T+\sum_{i,j}\hat{\bq}_i\bq_j^T \\
        &\phantom{=} -
        \sum_{i,j}\bq_i\hat{\bq}_j^T  - \sum_{i,j}\bq_i\bq_j^T+\sum_{i,j}\bq_i\hat{\bq}_j^T + \sum_{i,j}\bq_i\bq_j^T \\
        &= \sum_{i,j}\frac{\pi_{ij}-\pi_i\pi_j}{\pi_i\pi_j}\left(\bq_i-\hat{\bq}_i\right)\left(\bq_j-\hat{\bq}_j\right)^T.
    \end{aligned}
\end{equation}

\end{document}